# Preliminary Studies on Force/Motion Control of Intelligent Mechanical Systems

**by**

**Dinesh Rabindran, B.Tech (Hons)**

**Thesis**

Presented to the Faculty of the Graduate School of

The University of Texas at Austin

in Partial Fulfillment

of the Requirements

for the Degree of

**Master of Science in Engineering**

# Preliminary Studies on Force/Motion Control of Intelligent Mechanical Systems

Approved by
Supervising Committee:

Delbert Tesar

Chetan Kapoor

## Acknowledgements

I would like to thank Dr. Delbert Tesar, for his guidance and encouragement. I wish to also thank Dr. Chetan Kapoor for his useful comments and all the inspirational discussions I have had with him about my research. I would also like to express my sincere gratitude to the management and fellow graduate students at the Robotics Research Group for their support. Daniel. S. Poole, Undergraduate Research Assistant at UTRRG, is acknowledged for conducting experiments and compiling results for the section on dynamic parameter estimation of the PowerCube modules. To my dear parents I am indebted for their unconditional and unswerving support for all my endeavors.

This research was undertaken as part of the University Research Program in Robotics (URPR), sponsored by the Department of Energy (DOE) Grant No. DE-FG04-94EW37966. Their support is greatly valued.

August 2004

# Abstract

## Preliminary Studies on Force/Motion Control of Intelligent Mechanical Systems

Dinesh Rabindran, M.S.E.
The University of Texas at Austin, 2004
Supervisor: Delbert Tesar

According to a survey conducted by the Robotics Industries Association (RIA), North American robot orders increased 19% in 2003, the best year for robotics since 2000 [RIA Online]. To rationalize the relatively high investment that industrial automation systems entail, research in the field of intelligent machines should target high value functions such as fettling, die-finishing, deburring, fixtureless manufacturing [Butler and Tesar, 1992]. To achieve this goal, past work has concentrated on force control algorithms at the system level without an investigation of the feasibility of performance expansion at the actuator level.

We present pertinent literature at both system and component levels in the field of force control. A general overview of the problem we are faced with is presented together with some research areas that will help create a science base that can make significant contributions in the area. Some simple force control experiments are conducted on a modular robot testbed to study the issues involved in force control implementations at the system level and also to present the class of problems this research thread addresses.

The goal of this research work is to facilitate efficient execution of robotic contact processes using systems assembled on demand using Multi-Domain Actuators (MDA) and controlled using a model based intelligent Multi-Domain Control (MDC) scheme. The approach at UT Austin has been to maximize the number of choices within the actuator to enhance its intelligence. Drawing on this 20-year research history in electromechanical actuator architecture, in this report we propose a new concept that mixes two distinct subsystems (in motion and force domains respectively with a kinematic scaling of approximately 14:1) within the same actuator called a Force/Motion Actuator (FMA). A detailed kinematic and dynamic model of the FMA is presented. The actuator performance is evaluated with an operational specification in the motion domain using a weighted minimum prime-mover velocity norm criterion. It is shown that the design choice of 14:1 scaling between the motion and force sub-systems results in the selective flow of force- and motion- sub-system attributes to the output. We demonstrate that the velocity side of FMA contributes to the motion predominantly and the force-side contributes to external disturbance rejection. We also present the future work that would draw on this effort to establish a science base for Multi-Domain Control in the system domain.

# TABLE OF CONTENTS

## List of Figures

## List of Tables

# Chapter 1

# Introduction

*"I offer this work as the mathematical principles of philosophy, for the whole burden of philosophy seems to consist in this – from the phenomena of motions to investigate the forces of nature, and from these forces to demonstrate the other phenomena***"**

-Sir Isaac Newton (1643-1727)[1]

According to a survey conducted by the Robotics Industries Association (RIA), North American robot orders increased 19% in 2003, the best year for robotics since 2000 [RIA Online]. More recently, in the first quarter of 2004 this figure was in the vicinity of 17%. These numbers illustrate that in this time, when the economy is springing back from a slump, the manufacturing sector is striving hard to remain competitive and is investing in robotics and automation related technologies to produce better quality products more cost-effectively. In the recent past, the Butler report [Butler and Tesar, 1992] conducted a comprehensive survey of robotics application domains and concluded that our national security is closely coupled to our manufacturing capability. To rationalize the relatively high investment that industrial automation systems entail, research in the field of intelligent machines should target high value functions (fettling, die-finishing, deburring etc). The precision levels, force levels, and costs demanded by some applications is summarized in Table 1-1.

[1] Quote from the preface to *Philosophiæ Naturalis Principia Mathematica*

**Table 1-1. Summary of High-Value Functions [Butler and Tesar, 1992]**

| Application | Force Level (lbs) | Precision Level ( $\pm$in ) | Cost ($) |
|---|---|---|---|
| Drilling | 50-300 | 0.006-0.015 | 350k |
| Deriveting | 150-250 | 0.002-0.005 | 2.6m-3.5m |
| Milling and Routing | 15-25 | 0.015-0.03 | 150k-350k |

Robotic systems (or more generally intelligent machines) are frequently controlled only in the position domain where there is minimal or no information about the contact state of the robot with the environment, especially in the presence of uncertainties. A scenario where the robot interacts dynamically with its environment may be termed a contact task. Such a task necessitates a more evolved control methodology where force or tactile sensing is used to control interaction forces between the robot and its environment while tracking a pre-defined motion trajectory. This control scheme, often referred to as *Interaction Control* or *Force/Motion Control (FMC)*, enhances the capability of robots and renders them more suitable for high-value functions.

## 1.1 Purpose and Motivation for Force/Motion Control

There are several approaches to implementing robotic tasks. Position-based control, wherein the robot is commanded to track a pre-defined motion trajectory irrespective of the forces it generates due to its interaction with the environment, is the most rudimentary and inexpensive approach. For most non-contact tasks (like material handling or arc welding) this method suffices. For contact tasks with small uncertainties (like assembly operations), a common technique is to install a passive compliant interface between the robot end-effector and the tool. When uncertainties are large and the interaction forces are as important as the commanded motion trajectory, force/motion control is a typical alternative. The relative complexity of these different approaches is shown in Figure 1-1. As we move from pure motion to force/motion control, the control complexity increases while the task error decreases. For a contact task, force/motion control is appropriate.

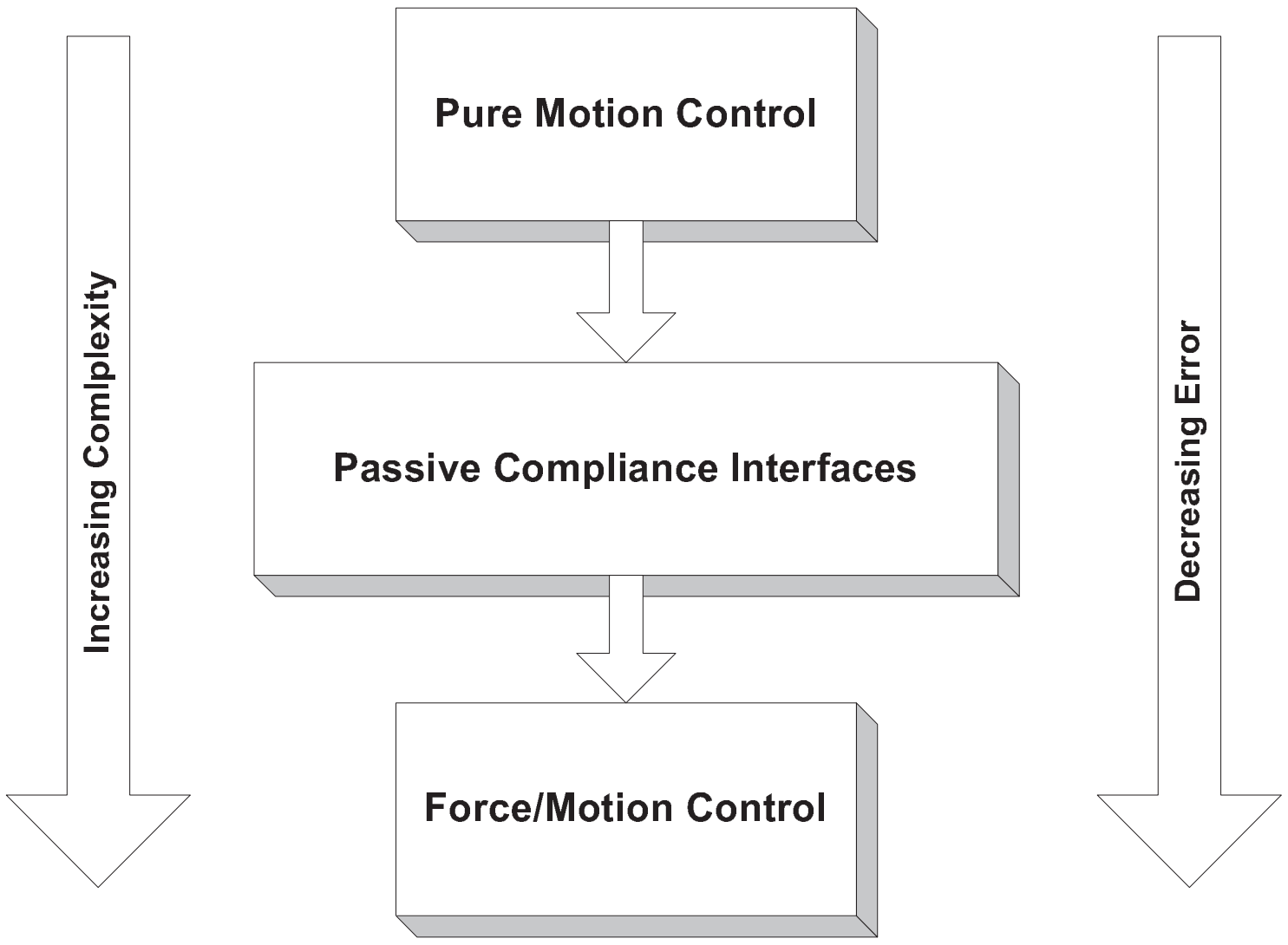

**Figure 1-1. Progression of Control Complexity on Contact Tasks**

There are myriad of application areas that lend themselves to force/motion control. Figure 1-2 shows some of these tasks in the manufacturing domain to demonstrate the motivation for this research work. These and other potential areas of relevance are described below.

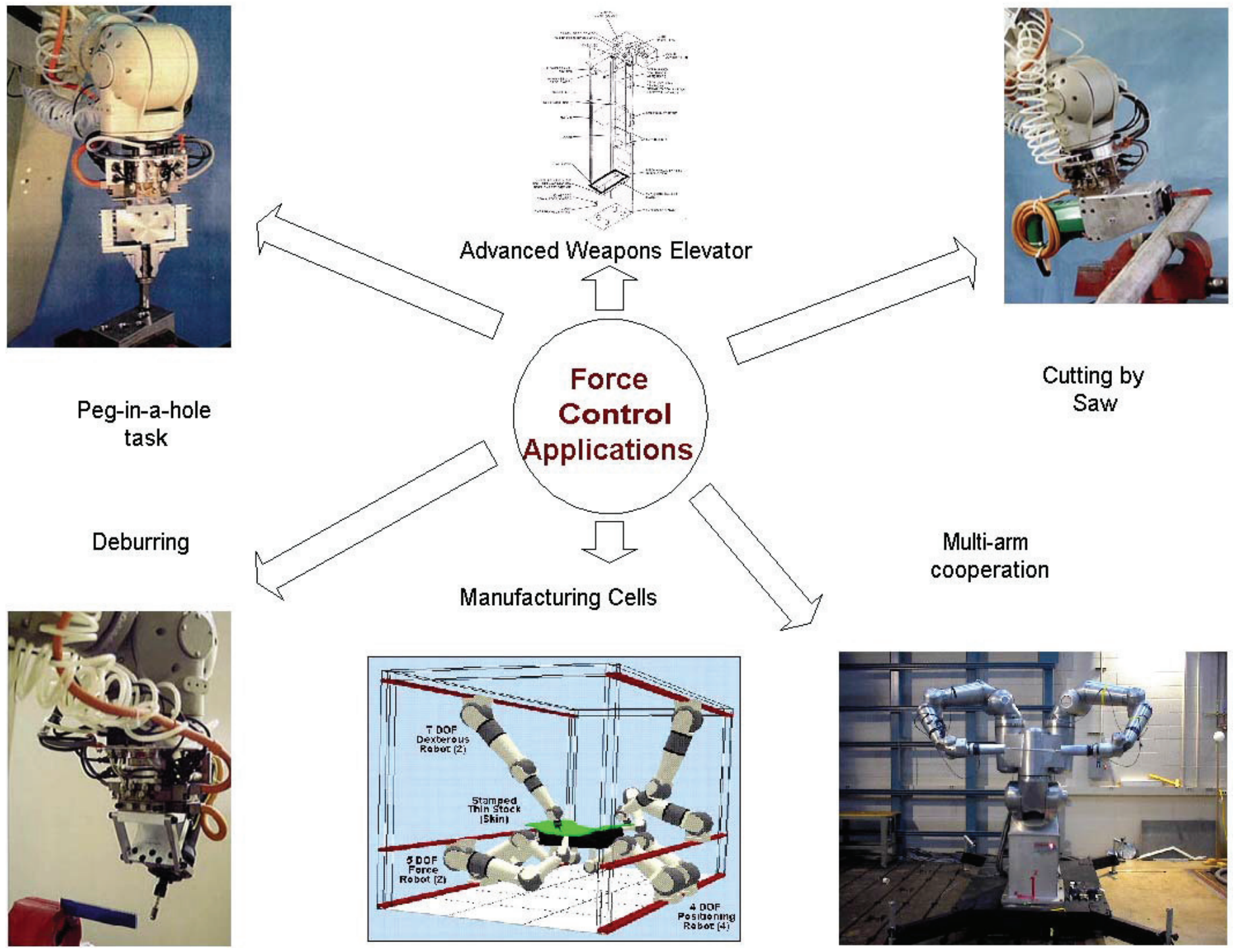

**Figure 1-2. Force/Motion Control Application Domains**

- *Deburring:* Requires a fairly constant material removal rate and relatively low precision (approximately $\pm 0.025"$).
- *Die Finishing:* Requires higher precision levels ($\pm 0.005"$) than deburring.
- *Robotic Assembly:* For a peg-in-the-hole task with large uncertainties, appropriate contact state estimation (using force/torque sensors) coupled with advanced force/motion control techniques can potentially give quicker and accurate assemblies.
- *Flexible Fixtures:* "Rigidized" jigs and fixtures are unattractive alternatives for precision manufacturing cells with the possibility of product changeover. For example, using computer-controlled manipulators as “flexible” fixtures facilitates rapid product changeover which connotes reduced costs. Precision machining requires precision of the order of $< \pm 0.015"$.

- *Microsurgery:* Tele-operated micro-surgery requires fine-motion control, tremor management and controlled interaction with tissue surfaces. Virtual fixtures, that are soft-constraints to the robot end-effector motion, are used for training novice surgeons.
- *Surface Characterization:* In the emerging area of Nano Electro-Mechanical Systems (NEMS), interaction control is used for surface characterization of material samples using an Atomic Force Microscope (AFM).

## 1.2 Fixtureless Manufacturing Cells

Fixtureless manufacturing cell is a very relevant focus area for the Force/Motion Control research thread within the Robotics Research Group at The University of Texas at Austin (UTRRG). The reliance on monolithic fixtures (jigs) is the root cause of product obsolescence. A jig may cost almost 10X the cost of a manipulator and is harder to maintain. Besides, the use of jigs hinders information flow to and from the manufacturing process (as depicted in Figure 1-3) and makes intelligent manufacturing hard to achieve.

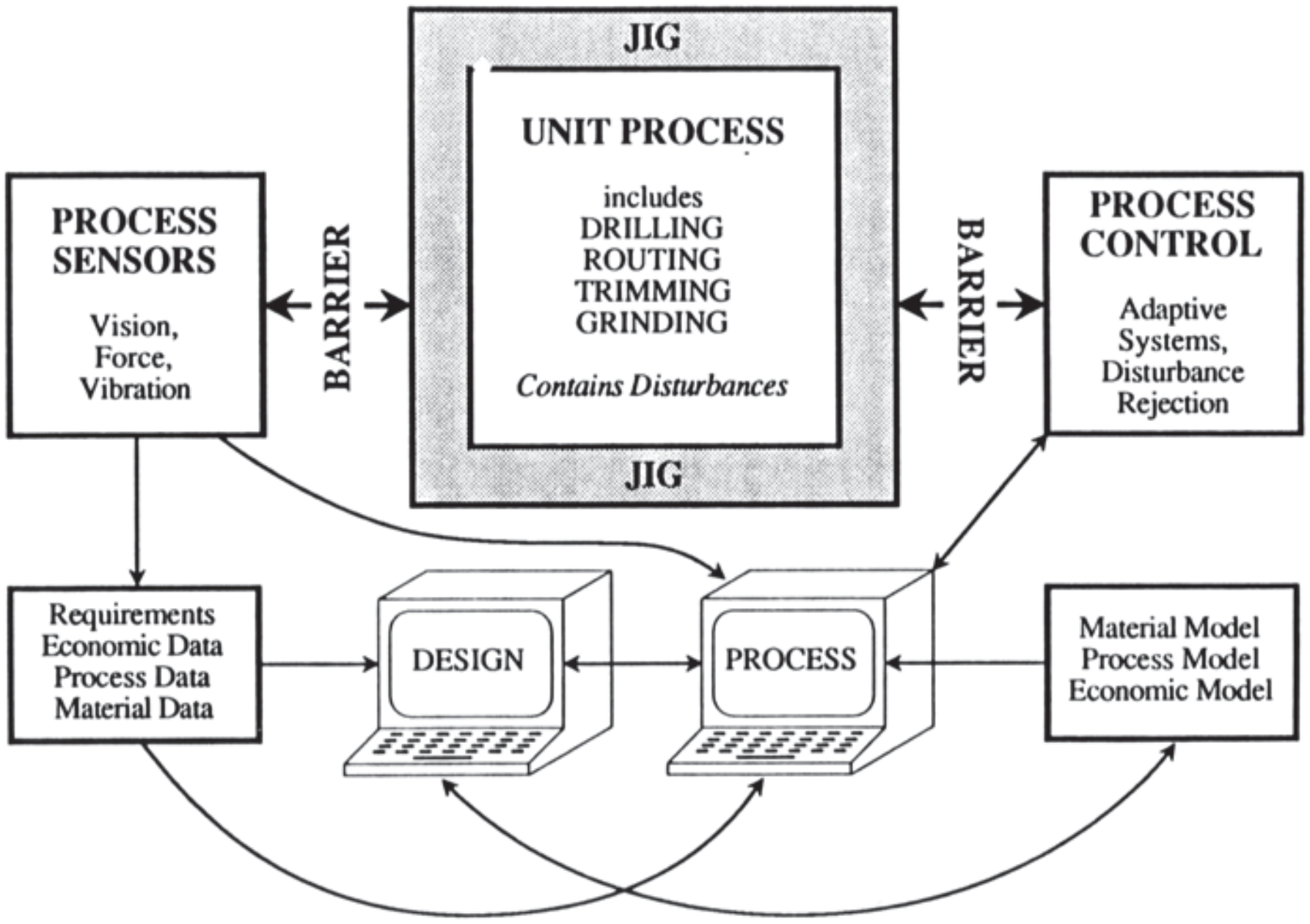

**Figure 1-3. Barrier to the Future Factory (RRG Concept of 1980)**

Automobile companies have invested heavily in the area of fixtureless manufacturing and have technologies that are becoming available. For example, the General Motors (GM) R&D center in Warren developed a reconfigurable fixture which can run 4-, 6-, and 8-cylinder engine parts on the same machine, thus enabling multiple functions, reducing change-over costs, and improving quality [Shen et al., 2003]. This agile fixture will be installed within a low-volume GM power-train plant in the first quarter of 2005.

The principal components of a fixtureless manufacturing cell (as shown in Figure 1-4) are:

- Position-controlled "dexterous" serial manipulators for precision operation. These are typically kinematically redundant manipulators of 7 or more Degrees of Freedom (DOF)

- Serial manipulators in closed configuration used for fixturing. These robots, with around 5-DOF, would be used to hold the part at hard points. This function requires rigidity. A secondary demand of precision may be placed on this sub-system.
- Additionally, low-DOF manipulator systems (approximately 4 DOF) may be used to position the part of interest to arrange it to best interface with the rest of the system.

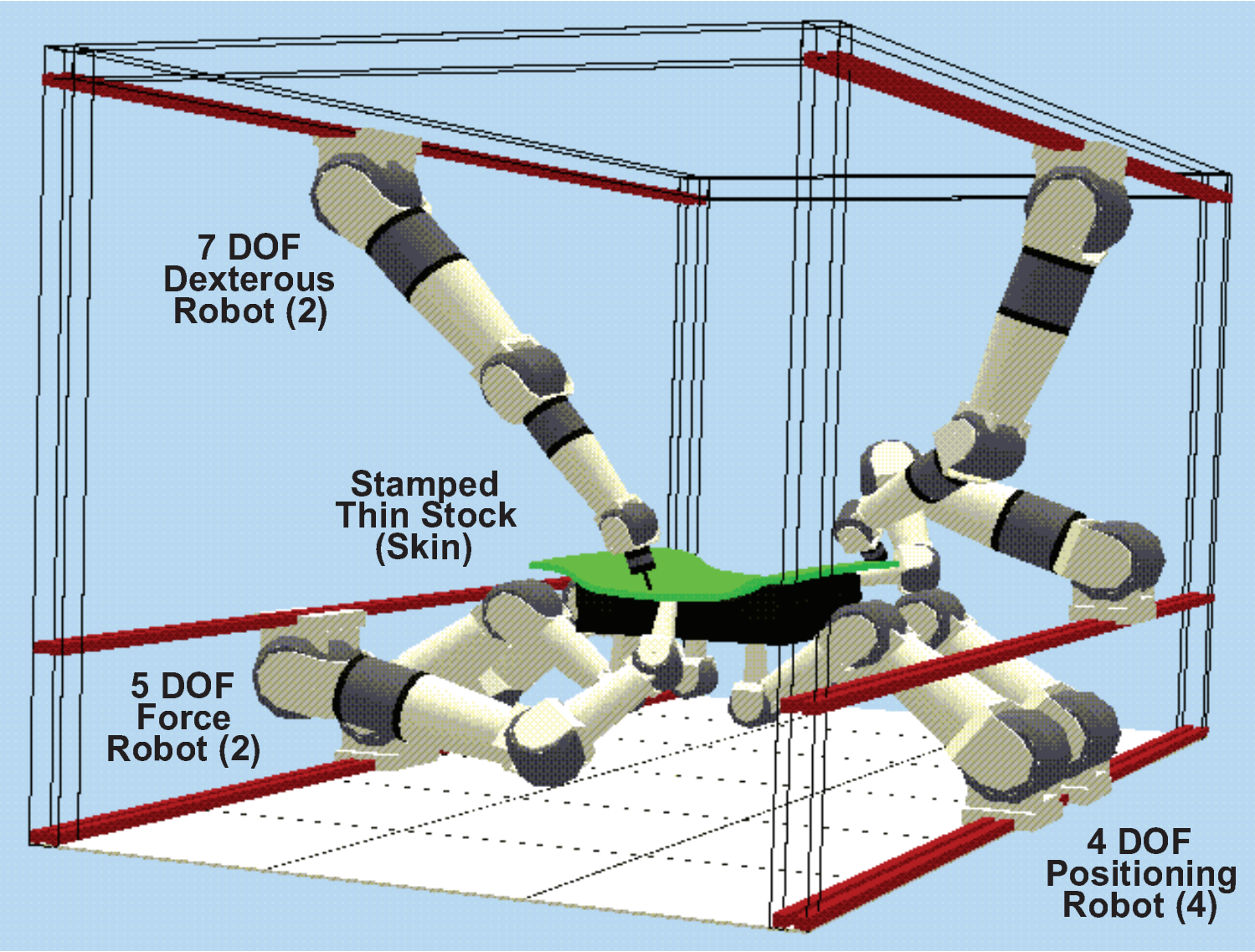

**Figure 1-4. Light Stock Machining Cell for Airframe Manufacturing (RRG Concept of 1990)**

## 1.3 Problem Overview

Having presented the broad picture and the major challenges that this research effort is targeted at, we now elucidate the more specific issue of force/motion control at the manipulator level.

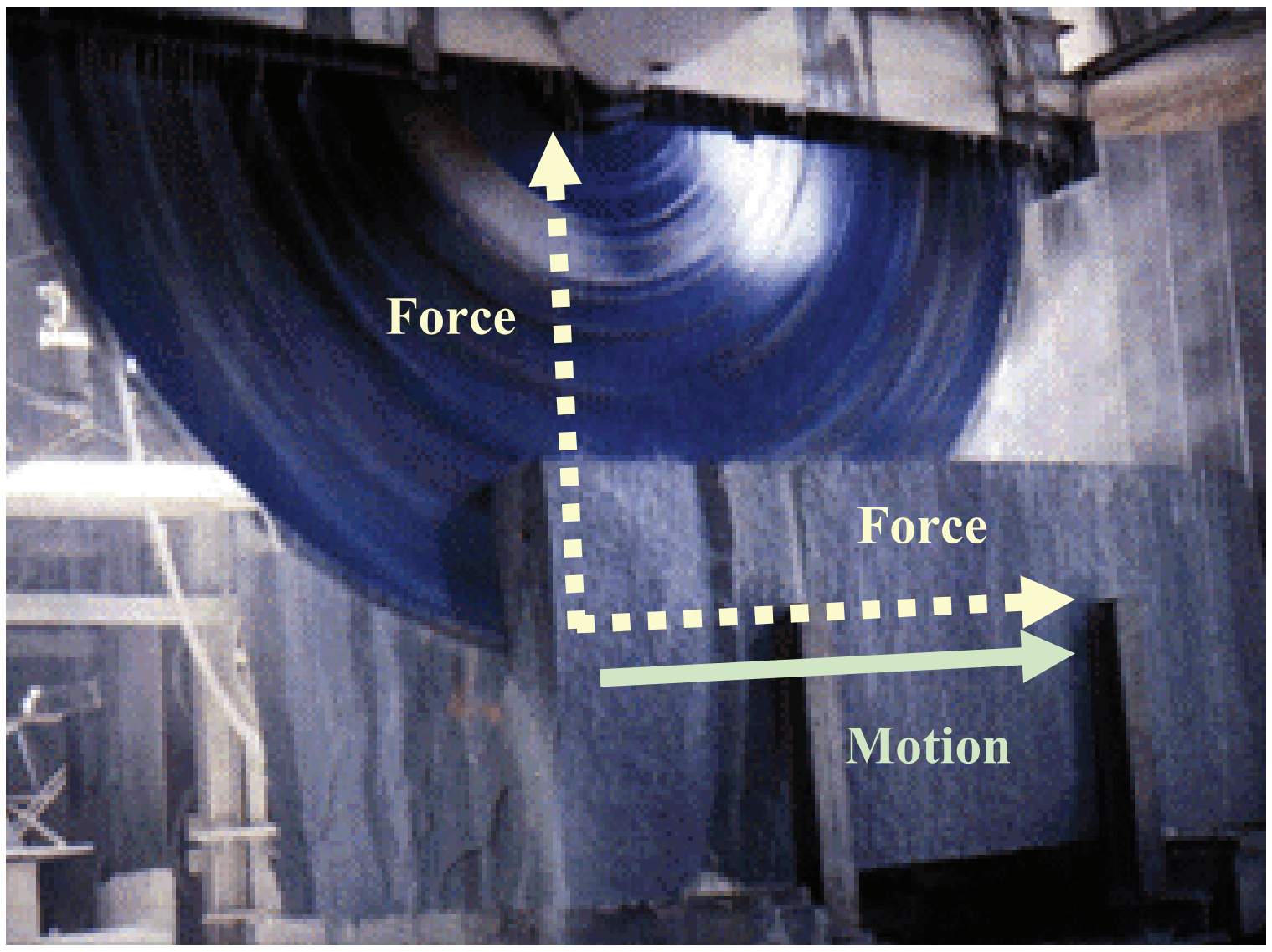

**Figure 1-5. Saw Blade in Wood**

At the manipulator level, the problem may be reduced to managing forces (both static and dynamic) and motions either in the same direction (as in most machining processes like grinding/deburring) or in mutually perpendicular directions (as in most assembly or forced-forming operations). Based on contact task-specifications, a suitable motion plan and reference force trajectory can be developed. Refer to Figure 1-5 for a schematic of an interaction task with force and motion constraints in the same direction. Subsequent to determining the motion plan and force trajectory, a control scheme is implemented depending on system parameters and sensing capability. The force/motion control template is shown in Figure 1-6.

A multitude of approaches have been developed to tackle the force/motion challenge at the system level. Most commonly, improved sensing and advanced control methods have been used to enhance force/motion capability at the output of a mechanical system. However, more recently, there is a surprisingly overdue research thrust at the component level which is justified. Since actuators are the building blocks for robots and other intelligent machines, a performance enhancement of the overall system may be achieved by enhancing the capabilities of their actuators.

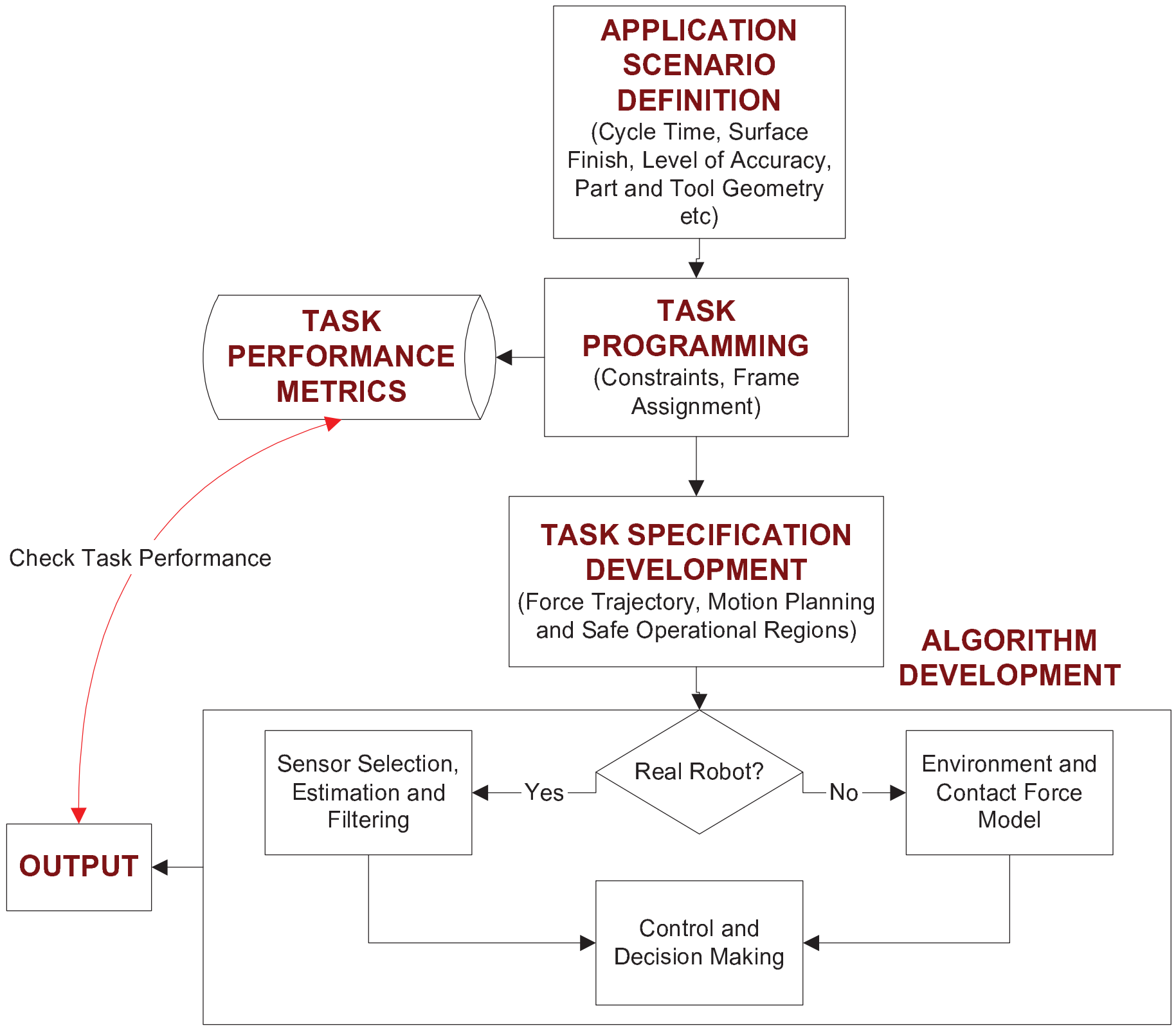

**Figure 1-6. Force/Motion Control Template**

Tesar [2003a] proposed an actuator that was hypothesized to improve the force and motion capability of a mechanical system at the component level. An investigation of this actuator and related system level issues is the focus of this report. Due to the magnitude of the challenges presented in the earlier sections of this chapter, the force/motion issue has to be examined at the level of granularity of an actuator at the outset to make the effort more manageable. Drawing on this research, a dedicated effort at the system level should be carried out to analyze the impact of such an actuator on system performance.

As a first step toward the larger goal presented in the former half of this chapter, we envisage accomplishing the following tasks:

- Develop a good understanding of literature pertinent to force/motion control both at system and component level.
- Implement force control applications on a real modular robot test-bed to identify major issues in experimental force control at the system level.
- Introduce the concept of multi-domain actuation and investigate the Force/Motion Control (FMC) paradigm at the actuator level. This entails detailed kinematic and dynamic analysis of the Force/Motion Actuator (FMA) followed by extensive simulation work.
- Based on the lessons learned this report will layout a roadmap that identifies potential areas of research for the ensuing research work.

## 1.4 Solution Approach

The concept of force/motion management using a performance map/model of the system based on as-built parameters is termed as *Process Control.* Any form interaction control methodology is supported by this all-encompassing paradigm. We propose multi-input control at the actuator level as an effective method of control within Process Control. UTRRG has developed an array of multi-input actuators, both velocity summing and force summing. On the component level we will investigate the possibility of embedding multi-domain inputs in the same actuator (Force/Motion Actuator). In such an actuator, the force and motion subsystems have a scale change of 10 to 15-to-1. The kinematic and dynamic model of the actuator, under load and external disturbances, will be developed from first principles. This model will be used for simulating the actuator to evaluate its performance in the force and motion domains.

Force control experiments using representative contact tasks will be implemented on a real robotic system. A serial modular manipulator, equipped with a 6-DOF end-of-arm force/torque sensor will be used as the test-bed. This effort will mainly focus on identifying implementation issues of force control techniques.

## 1.5 Report Outline

Chapter 1 presents the major challenges in the effort to enhance the capabilities of intelligent machines for contact tasks involving force and motion control. It discusses the main application areas of FMC. Subsequently, the chapter defines the problem addressed by this report and elucidates the solution approach followed.

Chapter 2 reviews pertinent literature in the field of FMC. The goal of this chapter is to analyze past work, internal and external to UTRRG that are relevant to the problem dealt with in this report and identify significant contributions to the field while identifying major research issues. Issues surveyed are generalized modeling, parameter identification, force control criteria, force-controlled task description, force control techniques/algorithms, component-level technologies, visual servoing, and force/motion management software among others. This chapter identifies specific research tasks based on lessons learned from the literature review.

In Chapter 3, we describe the experimental set-up used to implement force control applications on a real robotic system. The modular robot test-bed consisting of the manipulator, force/torque sensor, and associated software, is discussed herein. The chapter reports experimental activity to estimate dynamic and associated parameters of the modular robot and also describes test environments used for experimental force control in Chapter 4.

Chapter 4 presents the results of experimental force control work carried out at the UTRRG robotics lab. Beginning with verification of frame transformations for force/torque readings, it delves into implementation issues of force/motion control techniques at the system level. An objective of the effort documented in this chapter is to isolate the force/motion control issues that are of relevance to the component (actuator) level. A secondary goal of this effort is to enhance the capability of the UTRRG robotics lab.

Chapter 5 considers the force/motion control question at the level of granularity of an actuator. It describes the historical context within which this concept is proposed. The chapter presents the conceptual embodiment of the actuator and also develops a model for the system based on generalized kinematic influence coefficients. This model is then used for numerical simulations to examine the performance enhancement resulting from embedding multi-domain inputs at the actuator level. This chapter concludes with a discussion on performance criteria and performance envelopes for such multi-domain actuator systems.

In Chapter 6, we present our conclusions regarding experimental work carried out at the system level and simulation efforts at the actuator level. Based on lessons learned from these endeavors, a roadmap for the subsequent research work is laid out. It suggests further experimental evaluation of the force/motion actuator. Having investigated this issue at the component level, further research has to be done to investigate how these actuators can be configured to achieve significant improvements in the force/motion domains at the output of the system. At this juncture, all possible topologies of mechanical systems (serial, parallel and series-parallel) should be considered.

# Chapter 2

# Literature Review

In most cases, when a manipulation device performs a contact task, the force experienced by the effector has to be managed (or controlled) in conjunction with its motion. This is due to the fact that the effector is dynamically interacting with the environment. For this reason, such a control or decision-making scheme can be called Force/Motion Control (FMC) or Interaction Control. This section summarizes pertinent literature in the general area of manipulator control and focuses on the force/motion control problem. The objective here was to conduct a comprehensive literature survey of all aspects of force/motion control to identify seminal and significant works in the field, analyze their contributions, and isolate major research issues. As mechanical manipulators are inherently-coupled nonlinear devices and are representative of complex mechanical systems, a significant amount of time and effort was spent in analyzing literature pertaining to implementing force/motion control strategies on these devices performing contact tasks.

Generalized modeling of mechanical systems, dynamic parameter identification, and force control criteria are discussed at the beginning. Before delving into conventional force control approaches, a summary of work done in description of force-controlled tasks is presented. Among conventional approaches, passive and active methods are considered. A treatment of component level technologies that influence the force/motion question at the system level is given. As visual servoing is being increasingly used for force control, it has been incorporated in this review. A review of previous force control surveys and a synopsis based on them is included. The issue of contact transition when the effector moves from free space to constrained space is addressed. Apart from serial chain structures, implementations and algorithms for closed chain systems are presented for completeness. A review of force control approaches would be incomplete without the inclusion of software architectures which facilitate the implementation of these. Hence, some software libraries and architectures are recounted at the end. The chapter concludes with a summary and listing of identified research issues.

## 2.1 Modeling

Modeling of manipulator systems using Generalized Kinematic Influence Coefficients (GKIC) was developed by Thomas and Tesar [1982]. In this comprehensive analytical effort the concept of GKIC was shown to be extensible to mechanical systems of any conceivable topology. A kinematic influence coefficient may be defined, in simple terms, as the geometrical effect of the input on the output of a system. The GKIC governs the scaling of forces and motions from the input to the output of a system. A major advantage of this methodology is that it lets us examine the scaling of velocities and forces between any two generalized reference points on a system.

### 2.1.1 Generalized Manipulator Kinematics

Consider a spatial robotic manipulator with $n$-DOF and $m$-DOF input and output spaces respectively. If $m = n$, then the manipulator is said to be fully-constrained. For the conditions $m < n$ and $m > n$, the system is called kinematically redundant and over-constrained, respectively. If $m=6$, the output coordinates are the end-effector position and orientation (also collectively called the *manipulator pose*) which may be represented as follows:

$$\mathbf{u}(t)=[x \quad y \quad z \quad \phi_x \quad \phi_y \quad \phi_z]^T \tag{2-1}$$

Considering the case where $n=6$

$$\boldsymbol{\theta}(t)=[\theta_1 \quad \theta_2 \quad \theta_3 \quad \theta_4 \quad \theta_5 \quad \theta_6]^T \tag{2-2}$$

The manipulator pose, $\mathbf{u}\in\mathcal{R}^m$, is a function of the independent generalized coordinates (or joint angles), $\boldsymbol{\theta}\in\mathcal{R}^n$.

$$\mathbf{u}(t)=f(\boldsymbol{\theta}(t)) \tag{2-3}$$

Since a manipulator is a non-linear coupled system, $f(\boldsymbol{\theta}(t))$ is a time-varying non-linear function. Differentiating Eq.(2-3)

$$\frac{d}{dt}\mathbf{u}(t)=\left[\frac{\partial\mathbf{u}}{\partial\boldsymbol{\theta}}\right]\frac{d}{dt}\boldsymbol{\theta}(t) \tag{2-4}$$

$$\dot{\mathbf{u}}=[G_\theta^u]\dot{\boldsymbol{\theta}} \tag{2-5}$$

$[G_\theta^u]$ is a function purely of the manipulator geometry and configuration and signifies the variation of the system output function (or manipulator pose) w.r.t the input (or joint angles). Any term of this matrix is called a first-order influence coefficient or *g-function.* In other terms, $[G_\theta^u]$ is the manipulator Jacobian and is expanded (for a 6-DOF spatial manipulator) in Eq.(2-9). $[G_\theta^u]$ is used for the analysis of differential motion and statics.

$$\boldsymbol{\tau}=\left[G_\theta^u\right]^T\mathbf{F} \tag{2-6}$$

where $\boldsymbol{\tau}\in\mathcal{R}^n$ is the vector of equivalent joint torques and $\mathbf{F}\in\mathcal{R}^m$ is the vector of external forces and moments exerted on the End-Effector (EEF) by the environment. These end-effector forces and moments may be collectively called a *wrench.*

$$\boldsymbol{\tau} = \begin{bmatrix} \tau_1 & \tau_2 & \tau_3 & \tau_4 & \tau_5 & \tau_6 \end{bmatrix}^T \quad \textbf{(2-7)}$$

$$\mathbf{F} = \begin{bmatrix} F_x & F_y & F_z & M_x & M_y & M_z \end{bmatrix}^T \quad \textbf{(2-8)}$$

$$[G_\theta^u] = \begin{bmatrix} \frac{\partial x}{\partial \theta_1} & \frac{\partial x}{\partial \theta_2} & \frac{\partial x}{\partial \theta_3} & \frac{\partial x}{\partial \theta_4} & \frac{\partial x}{\partial \theta_5} & \frac{\partial x}{\partial \theta_6} \\ \frac{\partial y}{\partial \theta_1} & \frac{\partial y}{\partial \theta_2} & \frac{\partial y}{\partial \theta_3} & \frac{\partial y}{\partial \theta_4} & \frac{\partial y}{\partial \theta_5} & \frac{\partial y}{\partial \theta_6} \\ \frac{\partial z}{\partial \theta_1} & \frac{\partial z}{\partial \theta_2} & \frac{\partial z}{\partial \theta_3} & \frac{\partial z}{\partial \theta_4} & \frac{\partial z}{\partial \theta_5} & \frac{\partial z}{\partial \theta_6} \\ \frac{\partial \phi_x}{\partial \theta_1} & \frac{\partial \phi_x}{\partial \theta_2} & \frac{\partial \phi_x}{\partial \theta_3} & \frac{\partial \phi_x}{\partial \theta_4} & \frac{\partial \phi_x}{\partial \theta_5} & \frac{\partial \phi_x}{\partial \theta_6} \\ \frac{\partial \phi_y}{\partial \theta_1} & \frac{\partial \phi_y}{\partial \theta_2} & \frac{\partial \phi_y}{\partial \theta_3} & \frac{\partial \phi_y}{\partial \theta_4} & \frac{\partial \phi_y}{\partial \theta_5} & \frac{\partial \phi_y}{\partial \theta_6} \\ \frac{\partial \phi_z}{\partial \theta_1} & \frac{\partial \phi_z}{\partial \theta_2} & \frac{\partial \phi_z}{\partial \theta_3} & \frac{\partial \phi_z}{\partial \theta_4} & \frac{\partial \phi_z}{\partial \theta_5} & \frac{\partial \phi_z}{\partial \theta_6} \end{bmatrix} \quad \textbf{(2-9)}$$

Differentiating Eq.(2-4) w.r.t time gives us the second-derivative of the output function which, in the above case, is the end-effector acceleration.

$$\ddot{\mathbf{u}} = \left( \frac{\partial \mathbf{u}}{\partial \boldsymbol{\theta}} \right) \ddot{\boldsymbol{\theta}} + \dot{\boldsymbol{\theta}}^T \left( \frac{\partial^2 \mathbf{u}}{\partial \boldsymbol{\theta}^2} \right) \dot{\boldsymbol{\theta}} \quad \textbf{(2-10)}$$

$$\ddot{\mathbf{u}} = \left[ G_\theta^u \right] \ddot{\boldsymbol{\theta}} + \dot{\boldsymbol{\theta}}^T \left[ H_{\theta\theta}^u \right] \dot{\boldsymbol{\theta}} \quad \textbf{(2-11)}$$

From Eq.(2-11), we observe that $\left[ G_\theta^u \right]$ maps the joint accelerations to the end-effector acceleration. The additional term $\left[ H_{\theta\theta}^u \right]$ is an outcome of Coriolis and centripetal effects. $\left[ H_{\theta\theta}^u \right]$ is a Hessian array and, like $\left[ G_\theta^u \right]$, is also a function only of the manipulator geometry and configuration. Any term of this Hessian is called a second-order influence coefficient or *h-function.*

The above analysis develops the expressions for end-effector velocities and accelerations. However, it is fairly straightforward to develop the *g*- and *h-functions* for any fixed point on the manipulator w.r.t to the system inputs. For a generalized and exhaustive treatment of GKICs, refer to Thomas and Tesar [1982].

### 2.1.2 Generalized Manipulator Dynamics

Having introduced the concepts of first- and second-order kinematic influence coefficients, we now present the dynamic analysis of a spatial robotic manipulator based on them. A generalized dynamic model, as described in this section, is essential for advanced control and design of manipulators. The derivation below is from Thomas and Tesar [1982] which uses the Lagrange formulation to develop the controlling equations of motion.

Consider a link *jk* on a spatial manipulator with a Center Of Mass (COM) at the point *C*. The total kinetic energy of the manipulator system may be given as in Eq.(2-12). From Eq.(2-12), the effective inertia matrix $\left[I^*_{\phi\phi}\right]$ may be represented as in Eq.(2-13).

$$\begin{aligned} K.E &= \sum_{j=1}^{N}\left\{M_{jk}\dot{\boldsymbol{\varphi}}^T\left[{}^jG^c_\phi\right]^T\left[{}^jG^c_\phi\right]\dot{\boldsymbol{\varphi}} + \dot{\boldsymbol{\varphi}}^T\left[G^{jk}_\phi\right]^T\left[\Pi^{jk}\right]\left[G^{jk}_\phi\right]\dot{\boldsymbol{\varphi}}\right\} \\ &= \frac{1}{2}\dot{\boldsymbol{\varphi}}^T\left[I^*_{\phi\phi}\right]\dot{\boldsymbol{\varphi}} \end{aligned} \qquad \textbf{(2-12)}$$

$$\left[I^*_{\phi\phi}\right] = \sum_{j=1}^{N}\left\{M_{jk}\left[{}^jG^c_\phi\right]^T\left[{}^jG^c_\phi\right] + \left[G^{jk}_\phi\right]^T\left[\Pi^{jk}\right]\left[G^{jk}_\phi\right]\right\} \qquad \textbf{(2-13)}$$

where $M_{jk}$ is the mass of the link, $\Pi^{jk}$ is the global inertia of the link, $\dot{\boldsymbol{\varphi}}$ is the vector of joint velocities, $N$ is the number of Degrees Of Freedom (DOF) of the mechanism, $\left[{}^{j}G_{\phi}^{c}\right]$ and $\left[G_{\phi}^{jk}\right]$ are the *g*-functions from the input to the COM, *C*, of the link *jk*, and the distal end of the link *jk* respectively.

The centripetal and Coriolis effects are encapsulated within the inertia power modeling matrix $\left[P_{\phi\phi\phi}^{*}\right]$ that is given in Eq.(2-14).

$$\left[P_{\phi\phi\phi}^{*}\right]=\sum_{j=1}^{N}\left\{\begin{array}{l}M_{jk}\left[{}^{j}H_{\phi}^{c}\right]^{T}\left[{}^{j}G_{\phi}^{c}\right]+\left[H_{\phi}^{jk}\right]^{T}\left[\Pi^{jk}\right]\left[G_{\phi}^{jk}\right]\\+\left[G_{\phi}^{jk}\right]^{T}\left[\Pi^{jk}\right]\left(\left[G_{\phi}^{jk}\right]\times\left[G_{\phi}^{jk}\right]\right)\end{array}\right\} \tag{2-14}$$

$\left[I_{\phi\phi}^{*}\right]$ and $\left[P_{\phi\phi\phi}^{*}\right]$ are functions purely of the topology, configuration and mass-distribution of the manipulator. Including gravity loads and external force/torque loads, we may develop the equation for equivalent joint torque as below.

$$\tau=\left[I_{\phi\phi}^{*}\right]\ddot{\phi}+\dot{\phi}^{T}\left[P_{\phi\phi\phi}^{*}\right]\dot{\phi}+\sum_{j=1}^{N}\left[{}^{j}G_{\phi}^{c}\right]^{T}{}^{j}L_{g}+\left[{}^{j}G_{\phi}^{e}\right]^{T}L_{e} \tag{2-15}$$

where ${}^{j}L_{g}$ is the gravity load due to the *j*-th link and $L_{e}$ is the external force/torque load. $\left[{}^{j}G_{\phi}^{e}\right]$ is the *g*-function from the input to the point of application of the load on the *j*-th link.

The complete research on GKICs and dynamic/compliance analysis based on them, spans over almost two decades and four seminal papers. Benedict and Tesar [1978] was the first such published work which laid out the analytics for GKICs for N-DOF planar systems. Thomas and Tesar [1982] developed the result for a serially linked open-chain spatial manipulator. Freeman and Tesar [1988] extended these results and showed how they worked for closed-chain mechanisms as well. This work also dealt with transfer of GKIC representations between various sets of generalized coordinates. Hernandez and Tesar [1996] added compliance modeling to this already comprehensive analytical framework. Also,

Yi and Kim [2000] have extended this framework to omni-directional mobile robot systems.

There has been no treatment of joint friction in this body of analytics. In the present framework described above, friction coefficients have to be chosen by the operator based on experience. A theoretical investigation of friction and its consequent addition to the dynamic model based on GKIC would be a timely effort now.

### 2.1.3 Dynamic Parameter Identification

Developing a mathematical model to describe the kinematics and dynamics is a first step toward understanding a mechanical system. This effort is futile if we do not have fairly accurate estimates of system parameters. This requirement calls for very precise metrology and efficient parametric identification techniques. Behi and Tesar [1991] identified some of the local dynamic parameters (mass, stiffness and damping coefficients) of a Cincinnati Milacron T3-776 using experimental modal analysis. This experimental study was thorough in presenting the mathematical model, designing the experiment and verifying the experimentally determined parameters.

### 2.1.4 Force Control Criteria

Yi, Walker, Tesar and Freeman [1991] studied the effect of robot configuration on force control stability during contact tasks. This may be explained by considering a human finger in contact with a rough surface as shown in Figure 2-1.

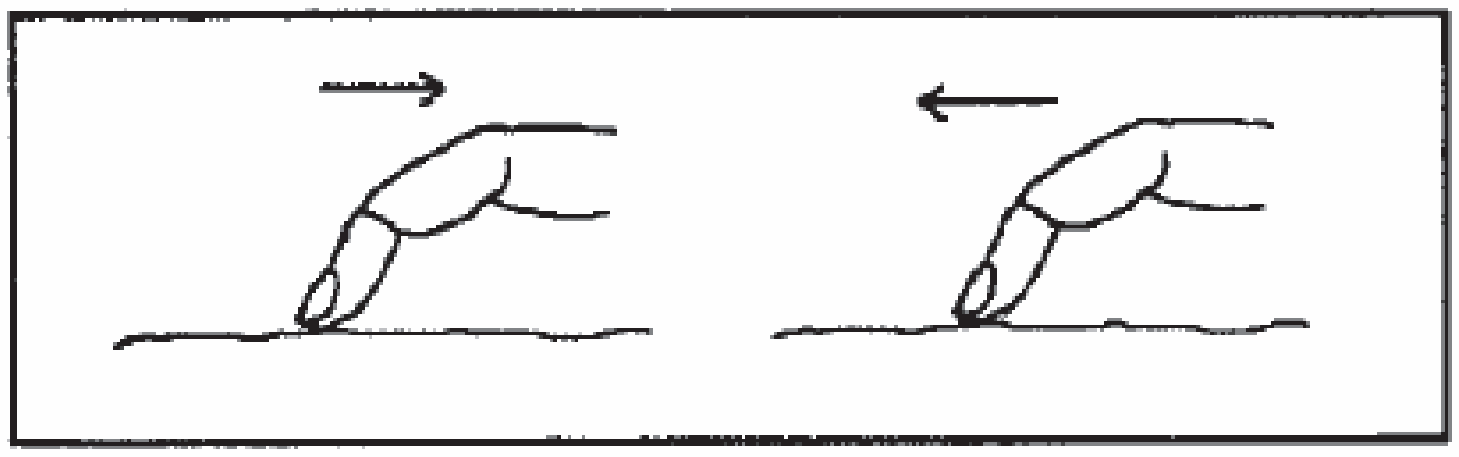

**Case (A)**

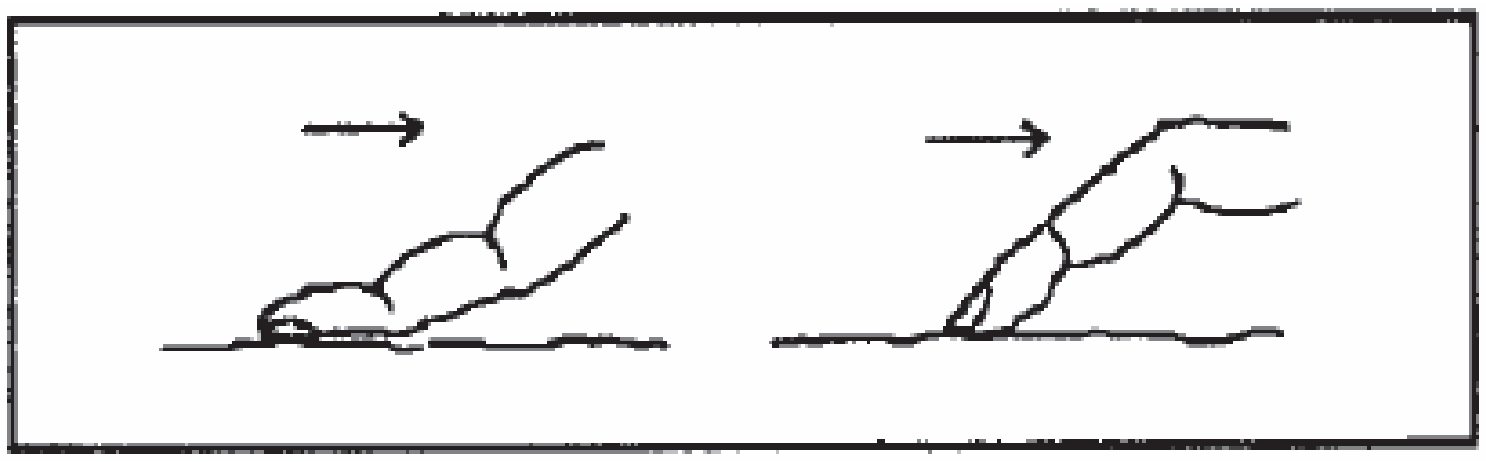

**Case (B)**

**Figure 2-1. Geometric Stability in Force Control [Yi et al., 1991]**

In Case (A), the inward movement is more stable than the outward motion since the latter case produces unwanted chattering. Similarly, in Case (B), the "elbow-down" configuration results in lower structural impacts. Force control criteria may be used to make these configuration choices that optimize secondary objectives. To this effect, an antagonistic stiffness matrix may be used as a decision-making tool. For a detailed derivation of antagonistic stiffness, please see Yi et al. [1991]. This paper suggests impact minimization as a force control criterion which may be used to resolve kinematic redundancy.

## 2.2 Formalisms for Force-Controlled Task Programming

Prior to implementing force/motion control algorithms, it is imperative to establish a framework to describe the task at hand. This description entails generating the constraints on the manipulator based on the task geometry.

### 2.2.1 Compliance Frame Formalism (CFF)

Mason [1981] proposed a general model, called the *Compliance Frame Formalism (CFF)*, for compliant motion. According to Mason, compliant motion is executed when the manipulator is not in free space and is constrained by the task geometry. He regards "pure motion control" and "pure force control" as duals of each other. This is illustrated in this paper by considering two complementary scenarios – (a) a manipulator tip buried in an immobile rigid body (b) a manipulator tip in free space. In the former case, the manipulator end-effector possesses force-freedom but not position-freedom and in the latter case, vice-versa. Compliant motion is a scenario between these extreme cases. In CFF, Mason proposes a task configuration space (called a *C-Surface*) which has a position-constraint in the normal direction and a force-constraint along the tangent to the *C-Surface*. This paper is seminal because it was used for Hybrid Force/Position Control which grew to become one of the main threads in force control research. The contribution of this work was CFF, which makes the description of force-controlled tasks very simple and intuitive.

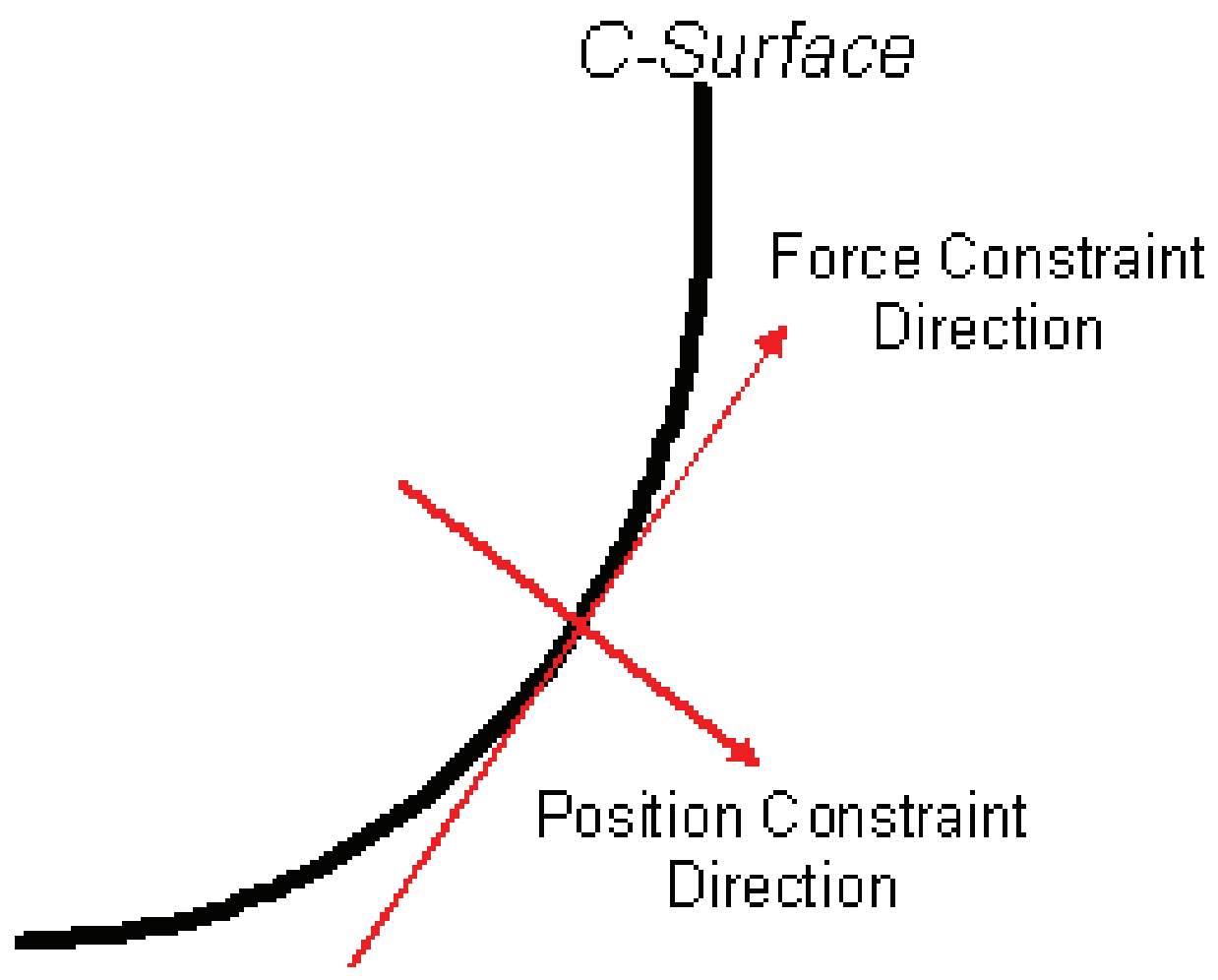

**Figure 2-2: Graphic illustrating the C-Surface Concept [Mason, 1981]**

### 2.2.2 Task Frame Formalism (TFF)

Bruyninckx and De Schutter [1996] introduced *'Task Frame Formalism' (TFF)*, the roots of which can be traced back to CFF [Mason, 1981], as a way of specifying force controlled actions. This effort is probably third generation work that drew on the original work within the same research group at Katholieke Universiteit Leuven by De Schutter and Van Brussel [1988] and Van Brussel [1976]. In this exceptional work [Bruyninckx et al., 1996] they have examined basic force controlled actions and their corresponding kinetostatic constraints. The objective of Bruyninckx et al. was to break down a force-controlled task into sub-tasks which can be described as a 'relative motion while trying to maintain a desired contact state'. The Task Frame (TF) they define for constrained motion can be partitioned into mutually exclusive twist (or position-controlled) and wrench (or force-controlled) sub-spaces. The dimensions of these sub-spaces add up to 6. TFF defines the force/motion restraints that follow from task geometry as *natural constraints* and those that follow from the prescribed reference trajectories as the *artificial constraints.* Bruyninckx and De Schutter exemplify TFF through a 'peg-in-a-hole' example shown here in Figure 2-3.

In the 'peg-in-a-hole' scenario shown in Figure 2-3, the TF is time-invariant with its origin on the axis of the hole. The wrench space consists of translations along and rotations about the X and Y axes and the twist space includes translation along and rotation about the Z-axis, which coincides with the axis of the hole. Note here that the twist space is 4-D and the wrench space is 2-D, both of them adding up to 6-D that is the dimension of the task space.

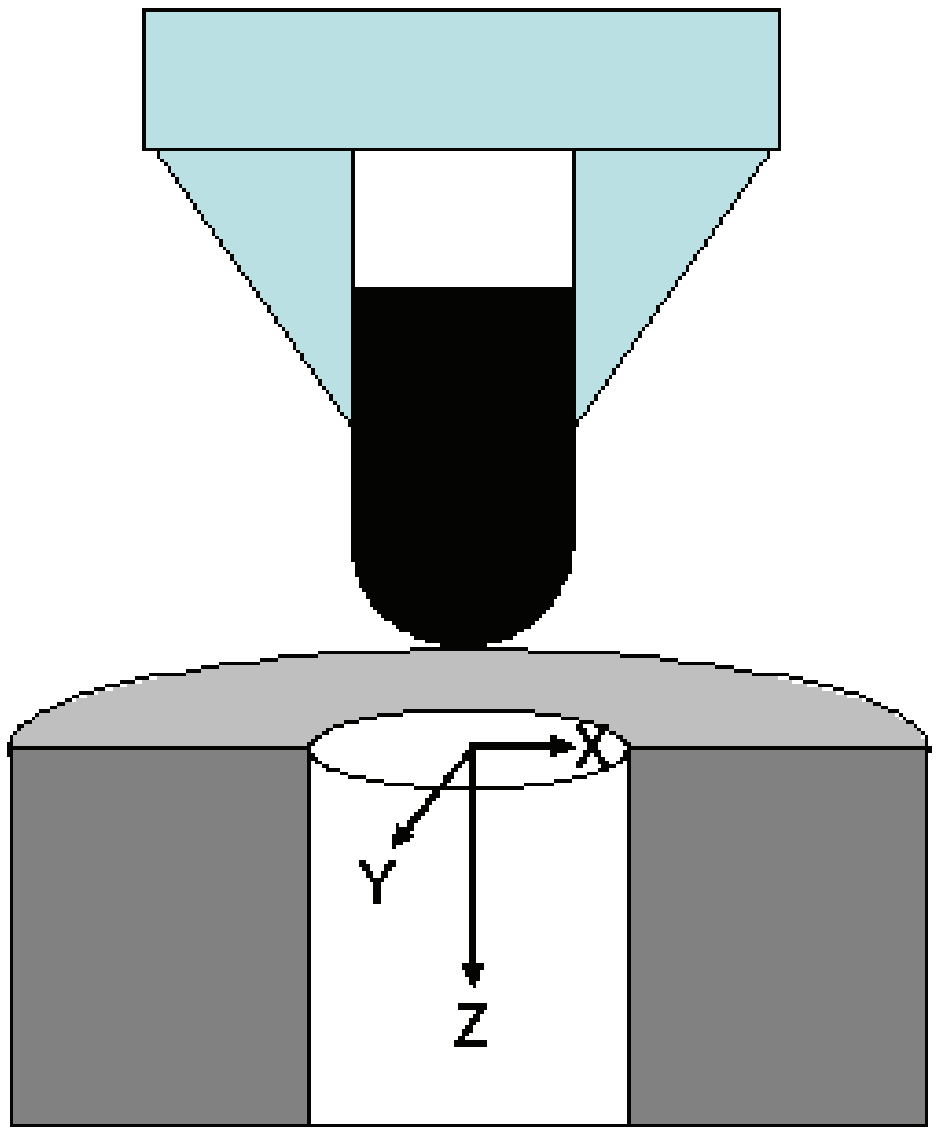

**Figure 2-3: Task Frame Assignment for a Peg-in-a-Hole Scenario**

The advantage of TFF is that the contact situation description does not change with time even if the TF does. Combining the CFF and TFF concepts, we could also say that the TF can be regarded as a rigid body trying to track the C-Surface. In a way, the TF is to task-planning as the Frenet frame is to motion planning [Wu and Jou, 1989]. Bruyninckx and De Schutter give many other examples of force-controlled actions represented by TFF. They also present a counter example of incompatible seam following for which TFF fails.

The work where the task frame was initially discussed was carried out by De Schutter and Van Brussel [1988]. This work is divided into two parts. In Part I, they describe the step-by-step functional specification of compliant manipulation as a significant step before implementing force control. In Part II, they suggest control methodologies which use the task frame specifications. They considered contour tracking and assembly as representative cases for their formalism. In this study they came up with a method to do "compliant motion planning" which includes a motion plan (frequently in end-effector space) and a trajectory for the task frame. They also developed database structures for these frames, in various cases, for programmatically implementing them. Both the frame specification and control algorithm were demonstrated on an experimental test-bed using a Cincinnati Milacron –T3 robot. The results of this research demonstration, for a peg-in-a-hole problem, show that they were capable of achieving successful insertions for peg diameters of $(30 \pm 0.1)mm$ for orientation errors up to $3^{\circ}$. The framework used by De Schutter et al. [1988] points to the future in compliant control during contact tasks.

A drawback of both CFF and TFF is that they assume the force-controlled and position-controlled directions to be in mutually exclusive directions (based on what is called the *reciprocity principle* [Bruyninckx et al., 1996]). These are simplifying assumptions about a real world where contact friction, stiction and object compliance are existent. The TFF relies on the robustness of the control algorithm to react to such aberrations from the idealized model. Functionally specifying compliant motion considering these real world issues is a major research step required in future.

## 2.3 Force Control Approaches

Having recounted work done in task description and a programming framework for implementing control algorithms, we now present and analyze some conventional force control approaches. The approaches have been divided into active and passive methods depending on the existence, in the control loop, of force sensor information and programmable compliance at the end-of-arm.

### 2.3.1 Passive Force Control

A *Passive Approach* does not utilize sensor information to monitor task-errors in order to provide compensatory action through a control system. Instead, it uses a mechanical compliant element which is designed to compensate for variations from desired performance. In a sense, this is a pure feed forward approach. There have been several successful implementations of force controlled contact task-execution using such devices, like the *Remote Center Compliance (RCC)* [Whitney, 1982] (Refer Figure 2-4). Also, this is a low-cost and quick-response approach. However, it is unsuitable for tasks where the uncertainties are large. This approach is hard to generalize. Since the compliances in different directions are fixed and hard-coded mechanically, they cannot be changed from task to task. From a safety perspective the passive approach lacks efficacy due to the absence of sensory data.

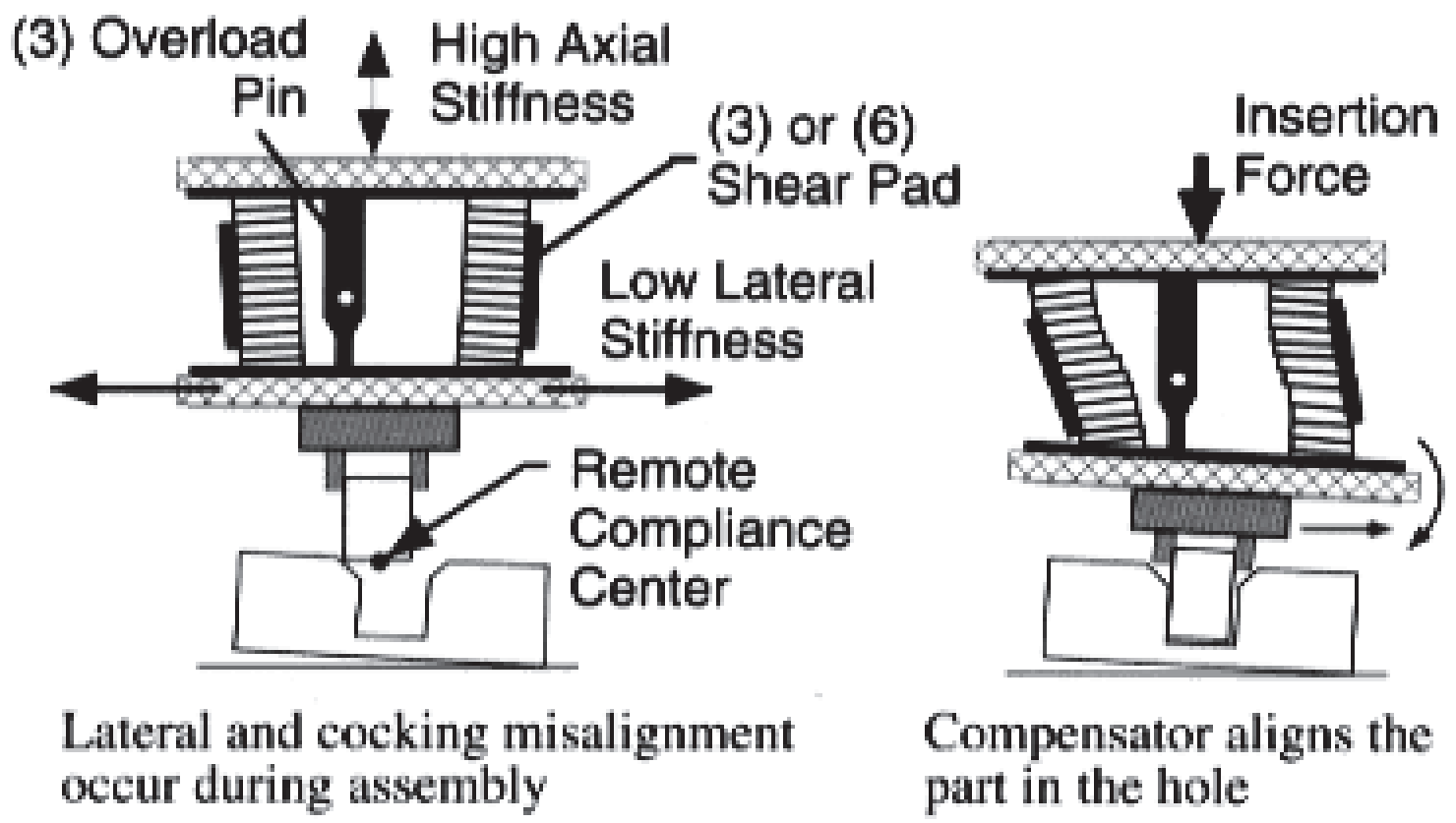

**Figure 2-4: A Remote Center Compliance Interface [From ATI Inc. Website]**

### 2.3.2 Active Force Control

*Active Force Control* methods are those in which the contact forces between the end-effector and the environment are fed back to the Decision Making System (DMS) and an interaction control strategy is used to determine further motion commands. Such a control method may also be characterized by the existence of a power source in the system. The advantage of active control over passive methods is that the interaction between the end-effector and the environment is soft-coded and thus manageable. Although this flexibility exists, the desired interaction is still limited by inherent properties of the system (inertia, energy dissipating elements, etc.) and the cycle times of the control system itself.

Force sensing is a principal component in an active control scheme and the quality of the sensor can directly influence the complexity required from the control methodology. Van Brussel at al. [1985] discussed the design of multi-axis force sensing and the effect of force sensor characteristics on force control systems. This paper is based on a sound survey of literature on tactile and force/torque sensors. Simons and Van Brussel [1980, 1985] have thoroughly examined active force control for assembly operations. They also experimentally implemented [Van Brussel et al., 1982] an intelligent force control scheme, based on active adaptable compliance, for an industrial manipulator (Cincinnati Milactron-T3).

#### 2.3.2.1 Hybrid Position/Force Control

The Hybrid Position/Force Control paradigm was proposed by Craig and Raibert [1979]. This method partitions the manipulator's task space into mutually orthogonal force-controlled and position-controlled subspaces, just like in the Compliance Frame Formalism [Mason, 1981], using a switching matrix. In this strategy, there are two control loops operating concurrently, viz., the position-control loop for the position-controlled task sub-space and the force-control loop for the force-controlled task subspace. The dimensions of these subspaces add up to the output degrees of freedom, which in the most general case is 6 for a fully constrained spatial manipulator. The input to the system is a union of the input commands from the position and force control loops. For example, consider a 3-DOF planar manipulator executing a 3-DOF force-controlled task wherein the contact force is to be regulated at a reference value along one direction (X-direction) and the motion has to be controlled in the remaining two mutually independent directions (Y- and Θ- directions). Although force and position sensors provide information in all 3-DOF in the output space, the force information in the X-direction and position feedback in the Y- and Θ- directions alone are used for decision-making. A union of command signals from these mutually exclusive loops is then sent to the manipulator servos. These two loops are separated in the analytics by means of a diagonal switching matrix. Raibert and Craig implemented their paradigm on a JPL Scheinman manipulator in two degrees of freedom.

An and Hollerbach [1987a] proved hybrid force/position control to be erroneous due to kinematically induced instabilities. They used the MIT serial direct drive arm to test the kinematic stability under three different force control paradigms, viz., hybrid control, resolved acceleration control and stiffness control. They found that hybrid control resulted in kinematically induced instabilities because of the switching matrix and the other two paradigms did not incite unstable behavior. An alternative formulation was suggested by Fischer and Mujtaba [1992] to eliminate these deficiencies. Since the hybrid control method uses the Jacobian inverse to map motion errors from Cartesian space to joint space, it is susceptible to all the consequent drawbacks such as scaling issues [Schwartz et al., 2003].

The hybrid twist-wrench control was proposed by Lipkin and Duffy [1988] which is invariant with respect to units and representation spaces. In this conceptual paper they disapprove of "orthogonality" as a term used to explain the relationship between differential motion and reaction forces at the contact point in an interaction task. They propose an alternative concept called "kinestatics" to elicit the dual relationship between the above mentioned physical quantities. They use screw theory concepts and suggest that the twist $T$ (instantaneous motion) and wrench $w$ (static forces and torques) are "reciprocal", i.e, $T^T w = 0$, or that the instantaneous work at contact vanishes. Using Euclidean and projective geometry they developed an invariant form of hybrid control. Note that the wrench represents only the static forces. Inertial and frictional forces are not considered in this formulation.

#### 2.3.2.2 Compliance-Based Methods

In the previous section, the hybrid position/force control paradigm was introduced. In this control technique, force and motion are explicitly and separately controlled in mutually exclusive directions. In this section, we will describe force control paradigms that do not control force or motion explicitly but model the contact between a manipulator and its environment as a pre-defined dynamic response. These may be called as *Compliance-Based Methods*. The different kinds of compliance based force control techniques differ from each other in the manner they define the dynamic response of the manipulator during contact with an environment of arbitrary stiffness.

Salisbury [1980] proposed the simplest form of compliance control which is called *Active Stiffness Control*. In this method the response between the manipulator and the environment is defined by a user-defined 6x6 stiffness matrix which quantifies the apparent stiffness of the end-effector to execute a contact task. The difference between the actual position and the nominal position (reference) is mapped to the virtual spring force using this 6x6 stiffness matrix. The stiffness matrix is generally designed to be diagonal to minimize coupling among the different task space directions. In effect the end-effector can be programmed dynamically to behave like a RCC. The method was demonstrated experimentally for a sprinkler assembly operation. The advantage of this paradigm is that the apparent stiffness can be varied depending on the task-requirements. A major drawback is the quantification and characterization of the apparent end-effector stiffness based on the task constraints.

A major research thread in force control literature is commonly known as Impedance Control. This was proposed by Neville Hogan at MIT. Hogan [1985] models the interaction between a manipulator and its environment as a second order dynamic response. It suggests that pure position/force control is inadequate and hence attempts to control the dynamic relationship between them in the task-space. In simple terms, impedance control introduces a *spring-mass-damper* system as a virtual interface between the robot end-effector and the environment. The objective, in impedance control, is to render the robot relatively stiff or compliant by changing the parameters of this interface to obtain a dynamic response that is suitable for the task at hand. In impedance terminology this desired response is also known as the *target impedance*. Hogan's [1985] first publication on this paradigm drew from bond graph modeling theory and suggested that in an interaction task the manipulator should behave as "impedance" and the environment should act as "admittance".

Hogan's conventional impedance formulation related the contact force vector to the end-effector motion error vector through a second-order ordinary differential equation. Variants of this basic equation were later proposed. Chan and Liaw [1997] related end-effector motion and force errors which facilitated motion- and force-tracking. Impedance Control implementation usually requires dynamic control capability. Reynolds et al. [1993] suggested a position biasing and used it to implement saw-cutting at the Savannah River Site (SRS). This method accounted only for feed-rate as a cause for saw-binding. Position-based impedance control assumes perfect position-tracking of the servos. This assumption is not warranted because more often than not full state feedback is necessary to ascertain good tracking performance. Pelletier and Doyon [1994] describe this limitation of position-based impedance control in detail. Bae et al. [1990] extend the basic impedance formulation to include robustness and thus compensate for modeling errors and environmental disturbances.

By imposing dynamic constraints on the end-effector in task space, impedance control claims to achieve motion and force tracking. However, this claim is underachieved because there is always a trade-off between motion tracking and force tracking in a given direction of control. This occurs since force and motion in one direction are related by a physical law and are not completely independent controllable entities. The foremost stumbling block in implementing impedance control is the ambiguous nature of determining the user-defined desired impedance parameters (or the target impedance), like the spring constant, damping and inertia of the virtual interface. If these parameters are not chosen judiciously, then the algorithm can also end up inflicting harm. Due to this drawback, the transient force response in the system can have unacceptable overshoots.

Impedance control almost always requires a dynamic model of the system (if the servos are unreliable with regard to position tracking), yet a fairly accurate dynamic model is frequently unavailable. Another shortcoming of the impedance paradigm is that the overall performance of the control system is still limited by the properties of the "real system" (manipulator, tooling and environment) like inertia, friction, compliance etc, irrespective of the target response desired from the "virtual interface".

Impedance control defines the interaction between a manipulator and its environment such that the manipulator is an impedance. In other words, the environment is seen as a source of force disturbances to which the manipulator has to react. Admittance control defines the interaction as in impedance control with the difference that the environment is considered to act as an impedance or source of motion disturbances. In a way, both these paradigms are conceptually similar except for differences in causality. *Accommodation Control,* which is one of the earlier damping methods, was first proposed by Whitney [1977]. In admittance control, the end-effector forces are mapped to differential motions in the task-space. An experimental implementation of admittance control on an industrial manipulator was done by Glosser and Newman [1994].

#### 2.3.2.3 Merits and Demerits of Active and Passive Methods

In this section we summarize some of the advantages and disadvantages of active and passive control paradigms (Table 2-1).

**Table 2-1: Comparison of Active and Passive Force Control**

| | Active Force Control | Passive Force Control |
|---|---|---|
| **Flexibility** | Good | Bad |
| **Mechanical Design** | Less Complicated | More Complicated |
| **Stability** | Stability to be Ensured | Stable by Default |

### 2.3.3 Force Control Using Passive Programmable Devices

There is a method that is midway between passive and active control methodologies. In this approach, compliant motion is established by means of passive programmable mechanical devices. Passivity of the system renders it stable, because energy is never added to the system. Concurrently, its programmability yields flexibility.

Asada and Kakumoto [1988] address the design and analysis of a dynamic Remote Center Compliance (RCC) hand for high speed assembly tasks. In this work, Asada and Kakumoto draw attention to the failure of the quasi-static assumption in RCC [Whitney, 1982] during high-speed assembly. As an alternative solution, they consider an insertion process wherein the inertia forces are predominant in comparison to the static spring forces. Based on this model, they presented the design procedure for a dynamic RCC. This was then tested experimentally for a peg-in-hole task with chamfer angles of $45^o$ and $60^o$ and a clearance of 20μm. They achieved successful insertions in approximately 120ms. Their method, however, failed when the initial misalignment between the peg and the hole was significant. A similar approach was taken by Hudgens and Tesar [1991] in the case of a micromanipulator. The work of Simons and Van Brussel [1980] [1985] are notable in the area of active robotic assembly. Cutkosky [1985] developed a programmable compliant wrist whose center of compliance could be modified dynamically.

Goswami and Peshkin [1991] suggest a parallel hydraulic redundant wrist at the end of the manipulator for high-bandwidth force controlled insertions. Since the whole arm does not have to move for accommodating task errors, higher speeds of operation can be achieved for fine compliant compensatory motions during an assembly task. The topology of this wrist can be changed infinitesimally about a nominal point by adjusting the variable conductance constrictions in the hydraulic network. This flexibility can be utilized to achieve a range of responses for the wrist to end-effector forces. The end-effector forces are mapped to the end-effector velocities by means of a 6x6 matrix called the *Accommodation Matrix.* For a RCC, this accommodation matrix is diagonal and thus is said to have a *Center-of-Accommodation*. Goswami and Peshkin [1993] show that with a programmable mechanical device, the range of achievable accommodation matrices (with or without a center-of-accommodation) is larger.

### 2.3.4 Force Control through Visual Servoing

To implement the aforementioned active force control algorithms on a real robotic system, it is customary to have a wrist mounted Force/Torque (F/T) sensor to measure the contact force experienced by the end-effector while interacting with an environment of arbitrary stiffness. The F/T sensor gives highly localized information at the interface between the manipulator and the environment. Some researchers have endeavored to use visual servoing in conjunction with F/T sensing. The advantage in doing so is that these two forms of sensing complement each other and can result in commendable improvements in performance. This section summarizes research efforts that use visual servoing for force controlled tasks.

Ishikawa et al. [1990] proposed the use of both force and vision sensors for error-correction during an assembly task. They used a CCD camera mounted on the end of a planar 3-DOF manipulator to carry out a bolt-tightening operation with compliant motion. However the sampling time for image processing they could use was relatively slow (500 ms). Nelson et al. [1995] at the Robotics Institute in Carnegie Mellon University devised three different strategies for combining force and vision information, viz., (a) traded control, where the contact surface is approached under visual servoing while the contact task is executed using force control, (b) hybrid control, where visual servoing and force servoing are used simultaneously but in mutually orthogonal directions, and (c) shared control, where visual and force servoing are used in the same direction for compliant motion. Nelson et al. also describe the specific task scenarios where these strategies are applicable. One advantage they demonstrated, of using vision and force sensing, was the ability to achieve faster approach velocities for the end-effector with minimal impact forces on contact. Baeten and De Schutter [2002] [Baeten et al., 2003] consider a planar contour tracking problem using

“eye-in-hand” approach. They mounted a camera on the end-effector so that it would detect corners/high curvature points in the contour in real-time. At the same time the force sensor on the end-effector would provide real-time contact force information. These two sensory inputs were then used to modify velocities in Cartesian space. They compare results from two experiments, one using only force sensing to negotiate a right angle corner and another using hybrid vision/force sensing. They achieved almost 5x improvements in allowable tool-tip velocities (50 mm/s) using the latter method. However, their relatively simple corner-detection algorithm resulted in a reduction in force-levels below the reference at high-curvature points.

Although visual-servoing assists force-control to a great extent, the issue of sensor fusion in multi-sensory control is still very complex. Also the range of tasks that can be handled by hybrid vision/force control is limited because many simplifying assumptions about the task have to be made to make the control problem tractable.

## 2.4 Survey Literature on Force Control

This section lists noteworthy surveys and pertinent monographs on force control. Whitney’s [1987] survey is one of the widely cited ones, wherein he describes the algorithmic evolution of various paradigms. He identifies some major drawbacks of force control before the 1990s which include lack of advanced estimation and filtering techniques for force/torque data and lack of good decision making strategies among others. Whitney acknowledges that his focus on the force control strategy (the means by which a high-level command is generated based on sensory information) itself is insufficient in this paper. Zeng and Hemami [1997] present a broad review of fundamental force control approaches with a strategy focus. Zeng and Hemami classify fundamental force control methods on the basis of the sensory feedback used (position or force or both) and the variables modified (Position or Velocity or Contact Force). They also analyzed and compared strategies in the task and joint spaces. In addition to the fundamental approaches they address advanced control methods like learning control and robust control. It is astonishing that this survey reveals the same drawbacks of force control as the ones exposed by Whitney [1987]. This indicates that a concerted research effort in sensing/estimation and algorithms is vital. This survey would have been more thorough had it included some experimental comparisons of the approaches they consider. Siciliano and Villani [1999a] conducted an experimental comparison between various force control approaches using an industrial 6-DOF robot (Comau SMART-3 S). This paper serves as a good compendium of various force control algorithms. Yoshikawa [2000] examines two major research threads in force control, viz., Hybrid Control and Impedance Control. In a discussion about related research topics, he addresses the question of contact transition control and the use of visual servoing for controlled contact tasks, which are normally skirted in such surveys. Although this survey is well-written, it does not conclude effectively because it does not present a list of research issues that have to be addressed in the field.

Three monographs written on force control are notable and thus are included in this survey. Gorinevsky's [1997] recent effort on the subject is unique in that it presents an extensive discussion on analysis and design of force sensors. He concludes by addressing potential application areas like surface machining and part mating. Most of the force control research conducted in the PRISMA lab at the University of Naples under the guidance of Siciliano [1999b] was published as a monograph. This includes rigorous analyses of fundamental force control approaches with experimental results. However Siciliano et al. present experiments conducted by their research group alone. Natale's [2003] publication is the most recent monograph on force control. In this work, titled *"Interaction Control of Robot Manipulators -- Six Degree of Freedom Tasks"*, Natale elicits the importance of both force and motion management for contact task execution. In addition to serial chain systems, he presents results of force control experiments conducted on closed-chain systems as well.

### 2.4.1 Synopsis of Force Control Surveys

- The bandwidth in active control is limited by the control system and the mechanical properties of the mechanical system
- Estimation/Filtering of force/torque data is still a primary concern. Issues like multi-sensor fusion and robot state estimation have to be addressed
- Non-collocation of sensor and actuator: If the sensor and actuator are separated structurally by other components (mainly transmissions), then the attainable control bandwidth is limited because of the resonant frequency of the system
- Some primary requirements for good force control are low robot inertia, frictionless and zero backlash transmissions

## 2.5 Developments in Component Technologies

Any investigation of force/motion control should be taken to the grass-root level of the mechanical system, i.e., the component level. The importance of actuators in determining the performance in force and motion at the end-of-arm (or output of the system) cannot be stressed more. We could say that a mechanical system is only as good or bad as its component actuators. There are limits to performance enhancements using only control techniques.

Having described some fundamental force control approaches at the system level, we will now present a summary of developments in actuator research that affect the force/motion control problem at the system level. This section summarizes literature, both internal and external to the Robotics Research Group at the University of Texas at Austin (UTRRG), in multiple-input actuation. Layered control, which is one such actuation scheme, is achieved when the output of the system is controlled at various scales by different sets of inputs. The output of a mechanical system may be in two domains, namely, force and motion. The different inputs in a multi-input actuator could potentially be in the same domain (position or force) or in different domains, as will be discussed in this section.

### 2.5.1 Internal Literature on Process Control and Intelligent Automation

Intelligence in decision-making is facilitated by the existence of mutually independent redundant resources in the system. The concept of duality in intelligent actuation was used for *fault tolerance* by Tesar [EMAA, 2003a] when two prime-movers were embedded in the same actuator to facilitate rapid reconfiguration during failure or performance degradation. These actuators could either be force summing, wherein the output force (or torque) is the parallel sum of those generated at each stage, or velocity summing, wherein the output velocity (or motion) is the serial sum of those generated at each stage. Further, the control command for both the stages in these dual actuators should be in the same domain, either force or velocity (motion). If the serially configured actuators influence the outputs at different scales (say 100:1), then the concept of multiple actuation can be extended to *layered control* [Tesar, 1999a]. For example, consider a micromanipulator appended at the output of a macromanipulator. The output of the system is a union of the outputs of the micro- and macro-subsystems. However, the gross motions at the output are influenced by the macromanipulator and the relatively small motions are caused by the micromanipulator for disturbance rejection. This concept is called *Control-In-The-Small (CITS)* and was patented by Tesar [1985]. Tesar [1999b] showed, in the CHAMP proposal, that CITS empowers us to control the output at various scales of motion. The macro-subsystem is highly coupled and nonlinear. Nonetheless, linearization of the micro-subsystem is warranted due to its relatively small range of motion. Design and analysis of such a micromanipulator was done by Hudgens and Tesar [1991].

Intelligent actuation could, alternatively, be configured in a mixed arrangement [Tesar, 2003a]. Tesar [2003b] proposed an embodiment for such an actuator (*Force/Motion*) where one of the two parallel stages controls the output force and the other controls motion. In addition to being in parallel these stages should have a scale change of 10 to 15-to-1. If this method of allocation of resources is ascertained and supported by an open architecture software framework to manage resource allocation based on performance, then it will lead to an expansion of choices at the output of the system. Investigating the promise of this proposition is an objective of this report.

In all the above instances of multiple embedded inputs in intelligent actuation, performance criteria pertaining to each subsystem and those pertaining to the combination of these subsystems are very critical. A significant amount of effort is required to investigate this mixing of criteria.

Having explained the component-level issues regarding dual actuation, we will now put this is in the broader perspective of RRG’s vision for *Intelligent Automation* at the system level (Figure 2-5). To realize modular production systems, there ought to be a systematic manner in which resource allocation is achieved. For this, a residual is generated between the model-predictive and real-time process states and used for performance based decision-making.

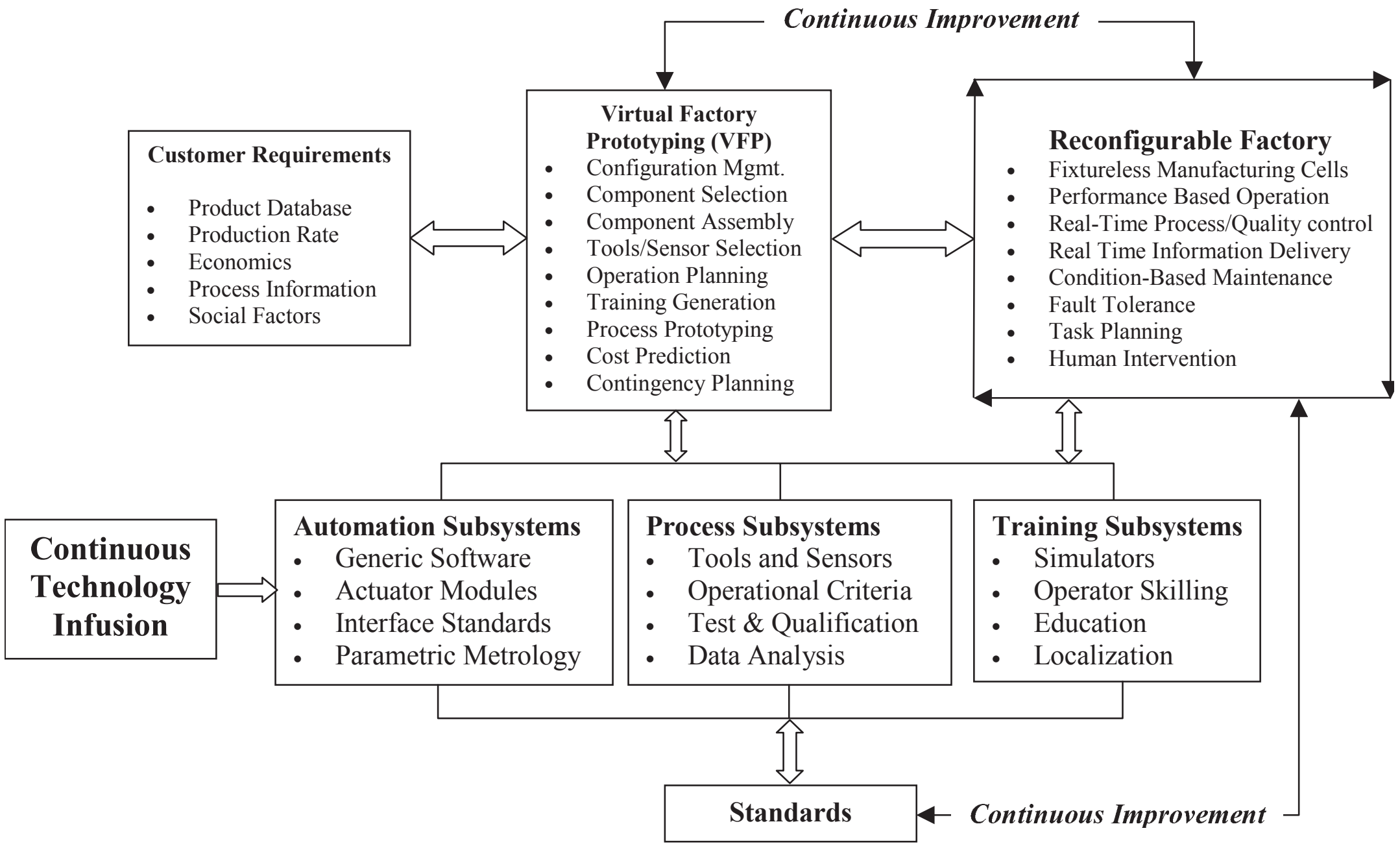

**Figure 2-5: Open Architecture Reconfigurable Manufacturing [Tesar and Kapoor, 1999a]**

The principal component of Intelligent Automation is *Real-Time Process Control and Monitoring*. Process control is a novel digital way of controlling all functions of the system. To place the concept of process control in perspective, consider a modular robot system. If this system is to execute a surface-finishing task, then the process involves parameters that completely define the task and thus need to be monitored in real-time. These parameters, for example, could be the tolerance, the Material Removal Rate (MRR), the force-levels etc. Adherence to or deviation from the references for these process parameters is manifested through real-time data from a sensor-suite (actual process state). To facilitate decision-making and optimal resource allocation, we will need a residual between this actual state and a model-predictive state. The model-predictive state should be based on the ‘as-built’ system as well as process parameters. This requires a parametric description of the process also. By means of extensive simulations, the concept of Force/Motion Control (FMC) will be described more deeply in Chapter 5.

### 2.5.2 External Literature

Eppinger and Seering [1987, 1992] have investigated the force control problem and identified the major dynamic problems which limit control bandwidth. These works propose good performance metrics for actuator design for force/motion control. We herein discuss three actuator design efforts that have worked based on these papers by Eppinger and Seering.

Morell and Salisbury [1996] at MIT proposed a Parallel Coupled Micro-Macro Actuator (PaCMMA). PaCMMA (Figure 2-6) consists of a micro-actuator and macro-actuator coupled in parallel using a compliant transmission. It bears similarity with the Control-In-The-Small (CITS) concept patented by Tesar [1985].

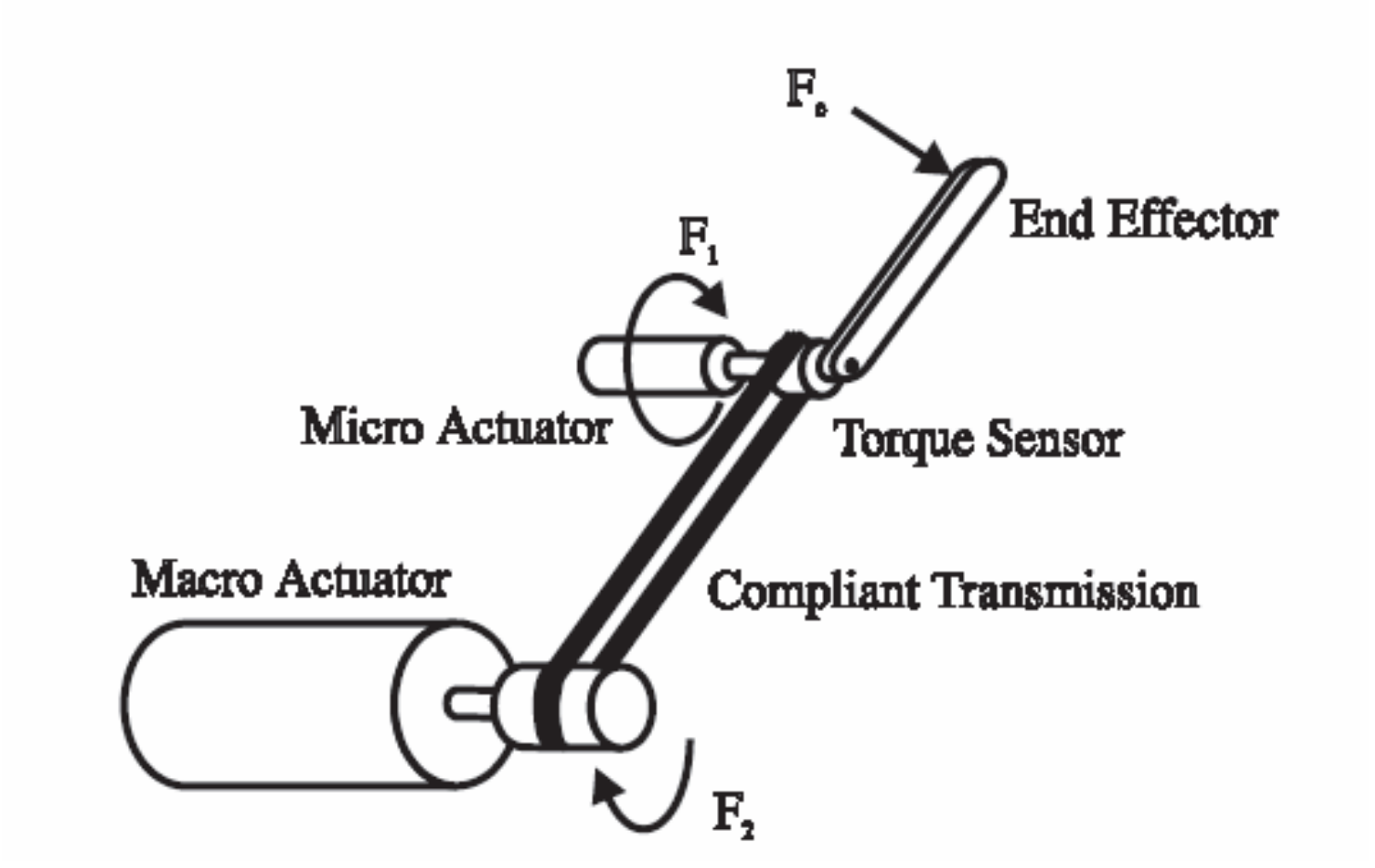

**Figure 2-6: Parallel Coupled Micro-Macro Actuator [Morell and Salisbury, 1996]**

The micro-actuator is direct-drive and thus can be controlled at a higher bandwidth. The compliant transmission was introduced to improve the dynamic range of forces obtainable at the end-of-arm. Morell compared PaCMMA against other single actuator systems based on force control bandwidth, position bandwidth, dynamic range, impact force, force distortion, force performance space and backdriveability as performance metrics. However, Morell uses classical linear control approach which fails to incorporate some nonlinear effects like backlash and nonlinear friction. PaCMMA is conceptually comparable to the Force/Motion Actuator (FMA) suggested by Tesar [2003b]. The difference between the two actuators is that the former actuator's subsystems are commanded in the same domain (Force) and the latter one's are commanded in two different domains (Force and Motion).

Using a stiff interface between an actuator and its load is frequently considered a rule of thumb for robot design. However, Pratt and Williamson [1995] proposed Series Elastic Actuators (SEA). In this design an elastic mechanical interface is purposefully placed between the actuator and the load. They claimed that this would reduce the impact during unexpected contacts. This concept was carried through to a working prototype on which a PID control algorithm was implemented. The disadvantage of this design is that the compliant element reduces the achievable control bandwidth. Also, the impact loads on the actuator can excite resonant modes of the flexible element seriously limiting the disturbance rejection capability.

In a more recent effort, Zinn, Khatib and Roth [2004] attempted to strike a compromise between control performance and safety by proposing the Distributed Macro-Mini ($DM^2$) actuation approach. Stiff actuation structures are good for control bandwidth as is required for force control, but they have high effective inertia which is not good for human-centered robot systems. Compliant structures, like the SEA, are good for safety as they are more forgiving, but their performance is limited by an inferior control bandwidth. In the $DM^2$ actuation approach, the torque generation is split into separate low and high frequency actuators whose torques sum in parallel. Simultaneously, it supports distributed actuation so that low frequency actuators are available at the base and high frequency ones are located at the joints. Low frequency actuators are used for low bandwidth trajectory tracking and high bandwidth ones are used merely for disturbance rejection. This work is in prototype-development stage. A proof of concept was recently demonstrated at ICRA 2004. This research thread needs to be actively monitored in the near future since it attempts to produce a universal actuation design that strikes a compromise between various design trade-offs for force/motion control.

## 2.6 Closed Chain Systems

This section focuses on the implementation of force control on parallel and series-parallel systems. Due to their topology, such systems are characterized by the presence of multiple force paths and are, thus, stiffer compared to serial chain systems. Parallel structures can also be more precise with less moving mass because deformations, backlash, errors and load are all distributed [Tesar and Butler, 1989].

Object manipulation using a dual arm robot system is very complex due to arm-interaction, variation in object mechanical properties and dynamics. Rackers and Tesar [1996] implemented a coordinated control system on the Robotics Research Corp's dual arm system K/B 2017. The work was well focused on experimental demonstration of a force/position control algorithm on a series-parallel system for multiple task scenarios, using performance criteria to monitor operational attributes. Considering processor-speed and other hardware limitations, they achieved reference forces and torques with an average lag-time of 2 seconds and 20 seconds respectively. Cox and Tesar [1992] suggest a hierarchical control architecture for dual arm robots. This comprehensive work characterizes typical operations carried out by cooperating manipulators based on the constraints on the end-effector wrenches (6-D vectors of forces and moments). The modes of operation considered in this research were *Lift, Squeeze, Insert, Twist, Transport* and *Bend*.

Walking Machines constitute a class of parallel mechanical systems which consist of articulated legs (typically six in number). Some of the earliest efforts in walking devices were carried out by Bessonov in Russia. He worked on gait analysis [Bessonov and Umnov, 1973], stability of walking systems and steering mechanisms [Bessonov and Umnov, 1979] among other issues. McGhee and Waldron [1985] built an Adaptive Suspension Vehicle (ASV) under a DARPA project at Ohio State University. This project was certainly a significant improvement over the state-of-the-art. Terrain adaptive vehicles were hardly implemented before this [Waldron, 1986]. In this device, there was a set of 18 actuators to control the body on the suspension in 6-DOF output space. This redundancy posed great challenges for control and coordination. Static indeterminacy and structural rigidity were major issues which challenged the force/motion management software that controlled the system. The speeds achievable by such systems are, however, modest (approximately 8 mph) compared to other types of locomotion devices. Despite massive efforts, walking machine technology is still an unattractive option for land locomotion because the performance is not commensurate with the prohibitive computational investment and mechanical complexity that it entails. The energy transfer in the system due to the gait is a major issue which needs attention. The weight of such walking machines is a limiting factor. Furthermore, the mathematical model used for these systems is simplified to a great extent to make the control problem tractable. For example, the ASV model neglects the interaction force, which causes slip at the legs, in the horizontal plane.

Multi-fingered grasping is analogous to the walking machine challenge. The issue in grasping is object manipulation by multiple parallel articulated digits. Kumar and Waldron [1988] studied force distribution in systems with multiple frictional contact points between dynamically coordinated mechanisms. Multi-fingered grasp is such a problem. In this analysis, the contact forces are partitioned into an equilibrating force field (or particular solution) and an interaction force field (or homogenous solution). Physical interpretation of the solutions is presented based on Screw System Theory. Hester, Cetin, Kapoor and Tesar [1999] proposed a criteria-based approach to grasp synthesis. This work divides grasp synthesis into a preliminary grasp, attained through grasp deconstruction and use of finger-level criteria, followed by an optimal grasp configuration that is achieved using hand-level criteria. A locally optimal grasp was achieved in a computationally efficient manner and applied to the NASA-JSC Robonaut Hand [Lovchik and Diftler, 1999].

## 2.7 Contact Transition Control

Smooth transition from free space motion to constrained space motion (and vice-versa) is essential to the successful execution of a contact task. Smooth contact transition should be construed as minimal rate of change of force level. As correctly identified by Whitney [1987] and Qian and De Schutter [1992], the presence of compliance and damping contributes toward stabilizing contact transition. This is because they assist in dissipating impact energy. Khatib and Burdick [1986] suggested impact transition control with pure velocity damping which acts for the first 100 ms after contact is established. Akella, Seigwart and Cutkosky [1991] achieve actively-controlled damping through the use of electro-rheological ‘soft’ fingertips. Hyde and Cutkosky [1993] proposed input shaping as a feed-forward alternative. They identified the dominant modes during impact and suppressed it in the input command, thus pre-shaping it. An and Hollerbach [1987b] studied the force responses and associated dynamic stability issues of force control implemented on multi-link manipulators with revolute joints. They present results of experiments carried out on the MIT serial link direct drive arm.

The simplicity of proportional force control makes it an attractive option. However it could cause an undesirable bouncing effect in case of stiff contact. Qian and De Schutter [1992] add a high-gain nonlinear damping term, depending on the force derivative, in the control equation for impact. It should be, however, ensured in this algorithm that the force derivative signal is free of noise. Xu, Han, Tso and Wang [2000] use joint acceleration and velocity feedback to implement switching control on a single-link robot. This control method requires the determination of environment parameters such as stiffness and damping factor, which vary noticeably from one contact task to another.

## 2.8 Force Control Software

For force control implementations, a generalized operational software framework is essential. It has to address most issues associated with force control such as sensing, decision making algorithm(s), force/motion performance criteria and contact force models for implementations on simulation platforms. Besides these basic requirements, it should also support other vital robot control components such as kinematics, dynamics, motion planning, obstacle avoidance etc.

Kapoor and Tesar [1996] developed Operational Software Components for Advanced Robotics (OSCAR). OSCAR [OSCAR Online Reference] is a framework which does not conform to any specific control architecture (hierarchical or distributed). Nevertheless, hierarchical control is often used. OSCAR provides the building blocks for all layers of hierarchical control [Kapoor and Tesar, 1998]. Figure 2-7 shows the three principal layers of hierarchical control. The topmost layer consists of the human-machine interface. The middle one is the operational software layer which consists of generalized kinematics [Kapoor and Tesar, 1999] and dynamics, performance criteria [Kapoor et al., 1999], obstacle avoidance, motion planning etc. The third and the bottom most layer is the device interface layer which talks to servo controllers, different end-effector tools, and sensors. Real-time issues are addressed in the device interface layer. This hierarchical relationship, however, need not be rigidly followed. For example, the human machine interface could circumvent the control components and talk directly to the device interface, if need be.

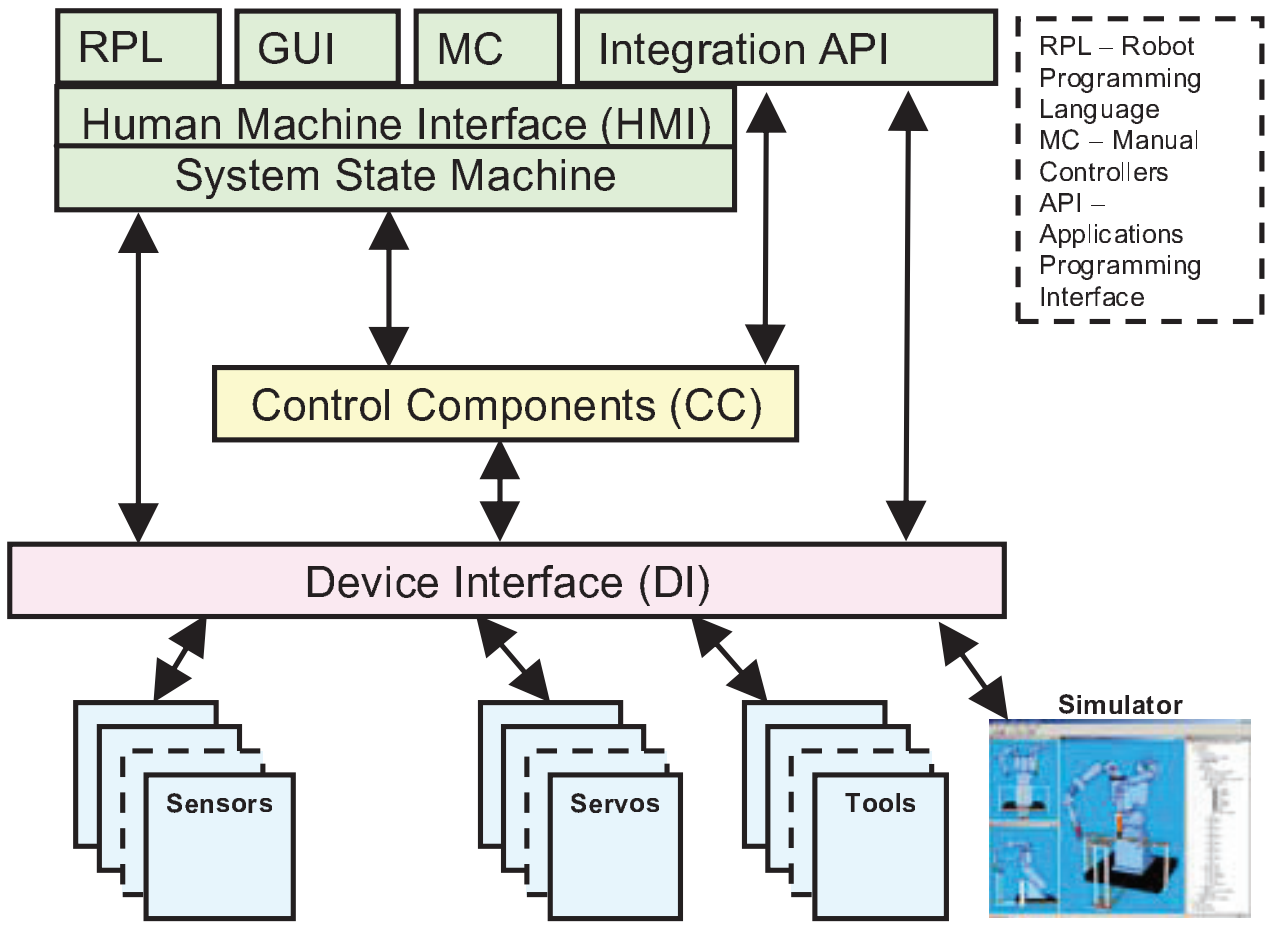

**Figure 2-7: Hierarchical Control Implemented with OSCAR**

In addition, OSCAR also contains support domains. Support includes a math domain, which contains all the mathematical constructs necessary for modeling and control of manipulators. Besides it has other support components like error-handling, data I/O, communications etc. (Table 2-2). In this report, OSCAR is extended to include a general control systems component which is demonstrated through force control implementations.

**Table 2-2: OSCAR Domains**

| Operational Domains | Support Domains |
|---|---|
| Generalized Kinematics | Math |
| Generalized Dynamics | IODevices |
| Performance Criteria | Communications |
| Obstacle Avoidance | Data I/O |
| Device Interface | Error Handling |
| Motion Planning | |

Koeppe, Heindl and Natale developed a Force Control Library (FCL) [FCL Webpage, 1999] at the Institute of Robotics and Mechatronics at Deutsches Zentrum fur Luft-und Raumfahrt (DLR), Germany. FCL uses an object-oriented framework and consists of modules for algorithms, force sensor computations and automatic force control synthesis. However, this library is not generalized because it addresses only the inner motion/outer force control algorithm. Natale et al. [2000] presented a step-by-step design procedure for force control of industrial robots. This automatic synthesis procedure works only for force regulation problems using the inner motion/outer force control algorithm and is thus not generic.

Open Robot Control Software (OROCOS) was developed by Herman Bruyninckx [OROCOS Online] at the Katholieke Universiteit (KU), Leuven. OROCOS is released under an open source software license and was funded as part of the Information Society Technologies (IST) projects of the European

Union. OROCOS consists of a real-time motion control kernel, CORBA based communication primitives for robotics environments and task execution sequencing component. Recently an effort is being made by Gadeyne [2002] to append a Bayesian Filtering Library (BFL) to OROCOS to facilitate robot state estimation. By Bayesian estimation here they attempt the recognition of the robot's contact state, object recognition (part-mating) or compliant motion sequencing, all in the presence of environmental uncertainties. The BFL can be used for autonomous compliant motion (or hybrid force/vision control) [Bruyninckx et al., 2000] among other applications.

## 2.9 Chapter Summary

This section outlines the force control literature survey reported in this chapter. A comprehensive survey of most fields within force control was conducted. The main objective of this survey was to identify the key issues in the field and identify works that are relevant to the work documented in this report. We list the key contributions to force control in various areas in Table 2-3 and also rank them according to their relevance to this report.

**Table 2-3: Compendium of Significant Contributions Related to Force Control**

| Area | Contribution | Credit(s) | Institution(s) | Year(s) |
|---|---|---|---|---|
| **Task Programming** | Task Frame Formalism (TFF)** | Hendrik Van Brussel, Joris De Schutter Bruyninckx | KU Leuven | 1980-1996 |
| | Compliance Frame Formalism (CFF)* | Matthew Mason | AI Lab, MIT | 1981 |
| **Operational Software** | OSCAR** | Kapoor and Tesar | UTRRG | 1996 |
| | OROCOS** | Bruyninckx | KU Leuven | 2001 |
| **Component Technology** | Multi-Input Actuators, Layered Control, CITS and EMAA*** | Tesar | UTRRG | 1978-2003 |
| | Parallel Coupled Micro-Macro Actuator** (PaCMMA) | Morell and Salisbury | MIT | 1996 |
| | Distributed Macro-Mini ($DM^2$) Actuation** | Zinn, Khatib and Roth | Stanford | 2000-2004 |
| **Algorithms** | Hybrid Control Research Thread* | Raibert and Craig | MIT | 1981 |
| | | Lipkin and Duffy | UFL | 1988 |
| | Compliant Control** | Hendrik Van Brussel | KU Leuven | 1980 |

*** Most relevant to this report

** Intermediate relevance to this report

*Some relevance to this report

### 2.9.1 Lessons Learned

- Component technologies which affect the force and motion control question at the system level have to be investigated
    - The PaCMMA [Morell, 1996] and $DM^2$ [Zinn, Khatib, and Roth, 2004] are representative literature sources of tested prototypes that can help this investigation
    - The concept of multi-domain actuation [Tesar 1999b] [Tesar 2003a] [Tesar 2003b] has to be studied at the actuator level as an initial step. Multi-Domain Input Actuators (MDIA) mix force and motion sub-systems in series or parallel arrangements to facilitate different applications.
        - Mixing of force and motion control capability using the Force/Motion Actuator (FMA) is a special case of the MDIA paradigm, proposed by Tesar [2003b]. A research effort is required to study the force/motion question at the actuator level first and then extend these results to the system level
        - Mixing of constituent force and/or motion subsystems and their coupling is a relevant research issue which needs to be explored
        - Performance criteria and performance maps of the force/motion actuator have to studied
- We need emphasis on Model-Based Process Control. Stability is considered as an important performance metric in most control efforts. Although stability is essential for implementing control, it is not sufficient. Control should be interpreted under the broader perspective of Criteria based Decision-Making.

- Although there have been numerous surveys on force control, there are not many studies of the effect of actuator characteristics on the overall force control performance. To this end, force control experiments on a real robot have to be conducted to study these effects.
- Task frame formalism, due to Bruyninckx and De Schutter, is an intuitive method to represent force-controlled tasks. This certainly assists in describing the contact task constraints in a generic manner
- Force/Motion Management software is still a weak area. Addressing this issue and adding necessary components to OSCAR [Kapoor, 1996] will facilitate on-demand implementation of force control strategies on intelligent mechanical systems
- There is ambiguity in determining the control parameters for Impedance Control and other Compliant Control paradigms (such as Active Stiffness Control and Admittance Control). This is a major limitation which makes them an unattractive option for practical implementation.

# Chapter 3

# Force Control Testbed Development

This chapter describes the test bed used for the experimental force control implementations done as part of this research work in the Robotics Research Group at the University of Texas at Austin (UTRRG). It covers all the physical elements of the set-up together with a description of the software platform used to develop these force control applications.

## 3.1 Test Bed for Experimental Force Control

The components necessary for demonstration of force control paradigms on a real robotic system are described in this section. UTRRG is equipped with a modular robot test-bed for experimental force control implementations. The components of this test bed are listed below and described in following sections.

- Modular Manipulator Components
- Multi-axis Force/Torque (F/T) Sensing System
- Peripheral Tooling

### 3.1.1 Modular Manipulator Components

A modular manipulator can be assembled using a set of standardized modules from Amtec GmbH, Germany [Amtec Website]. The actuator modules, called *PowerCube*$^{TM}$ modules, come with standardized interfaces that facilitate reconfiguration. These cube-shaped actuator modules [Figure 3-1] contain the prime-mover, a harmonic drive for transmission, a brake and an embedded controller.

The PowerCube™ modules can be controlled in position, velocity and current modes and use the Controller Area Network (CAN) bus for communication[2]. These are available in various standardized sizes based on different torque capacities. The modules available in the UTRRG metrology lab are PR 110 and PR 090, which are standardized rotary modules, PW 090, which is a standardized 2-DOF wrist module, and the PG 090, which is a parallel-jaw standardized gripper module. The modules come with associated software interfaces by means of which they can be controlled in position, velocity, and current modes.

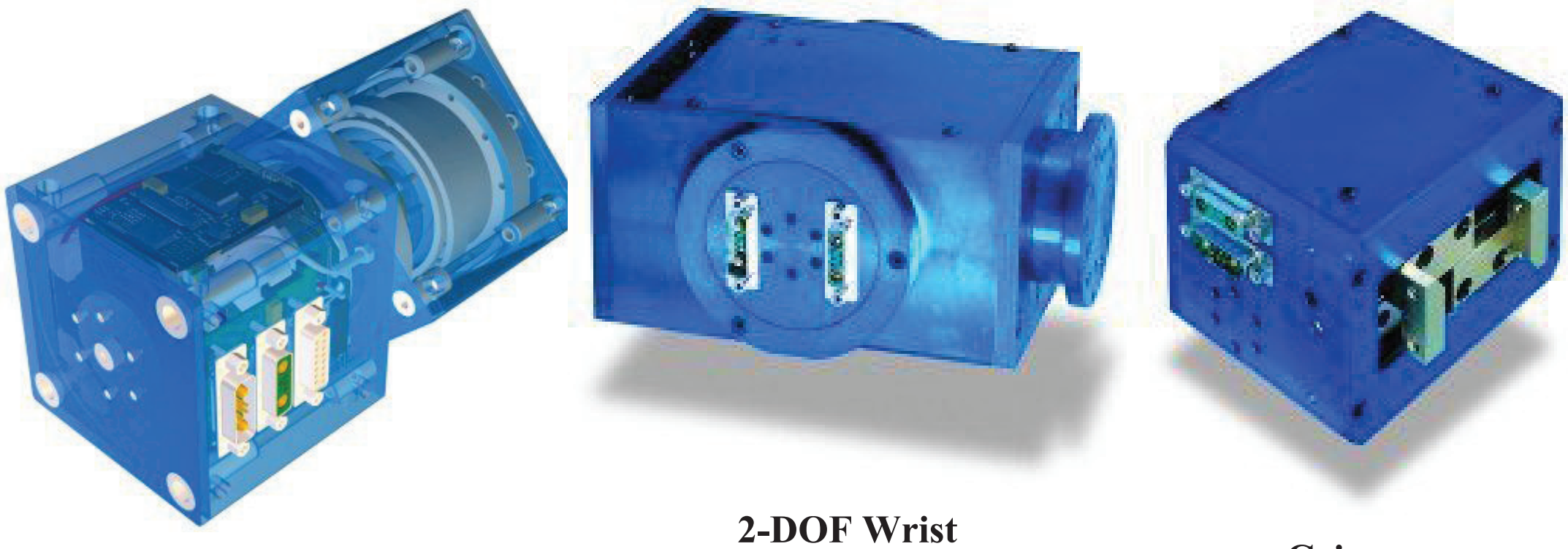

**Rotary Module** **2-DOF Wrist** **Gripper**

**Figure 3-1. PowerCube™ Actuator Modules [Amtec Website]**

At UT Austin, the PowerCube modules have been used for demonstrating glovebox automation and Feature Oriented Programming (FOP) techniques for integration of modular manipulator control software [Modular Small Automation Website].

[2]Host PC specifications are: Intel P4, 2.2 GHz, 523 MB RAM, Windows 2000 OS

For the purpose of force control demonstrations, the modules were assembled into a standard 6-DOF industrial robot [Figure 3-2] configuration[3] with a Roll-Pitch-Roll (RPR) spherical wrist. The regional structure consists of two PR 110 modules used for the base and shoulder joints, and a PR 090 module for the elbow joint. The wrist is composed of a PR 090 roll joint and a 2-DOF PW 090 wrist module. A 6-DOF ATI force/torque sensor is attached between the tool and the manipulator tool-flange. One goal of this work is to initiate the determination of parameters necessary to implement dynamic control on the PowerCube™ manipulator.

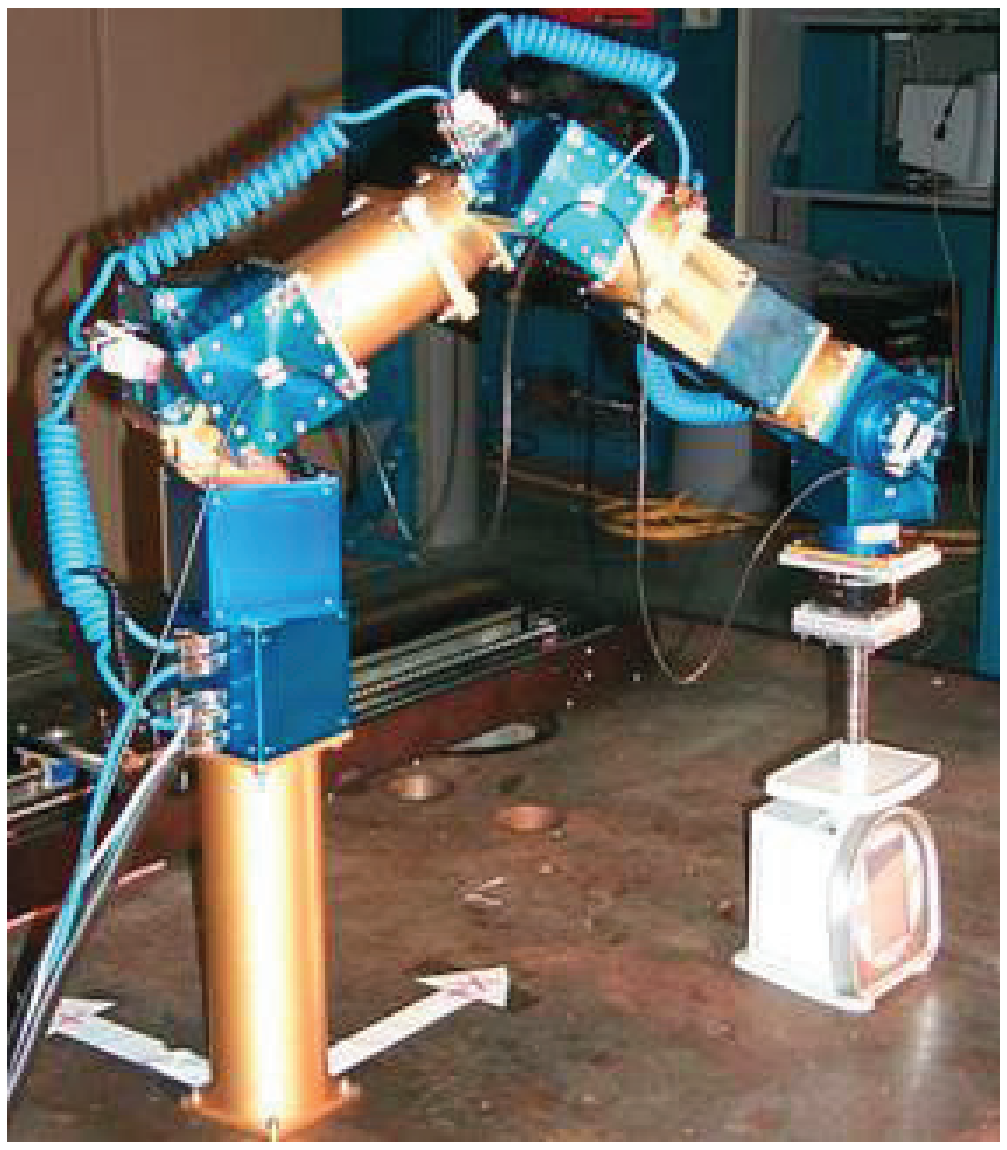

**Figure 3-2. PowerCube™ Modular Manipulator**

[3] Kinematic parameters of the robot are presented in the Appendix

The experiments reported in Chapter 4 are based on position-based force control which uses the manipulator kinematic model given in Eq.(3-1).

$$\dot{\mathbf{u}} = [G_\theta^u]\dot{\boldsymbol{\theta}} \tag{3-1}$$

where $\mathbf{u}(t)$ and $\boldsymbol{\theta}(t)$ are respectively the manipulator configuration represented in EEF and joint spaces.

$$\mathbf{u}(t) = [x \quad y \quad z \quad \phi_x \quad \phi_y \quad \phi_z]^T \tag{3-2}$$

$$\boldsymbol{\theta}(t) = [\theta_1 \quad \theta_2 \quad \theta_3 \quad \theta_4 \quad \theta_5 \quad \theta_6]^T \tag{3-3}$$

#### 3.1.1.1 Determination of Current-Torque Mapping

The PowerCube™ actuator modules, by design, do not support torque control. However they may be indirectly controlled in torque mode by varying the current. To this effect, it was necessary to experimentally determine the mapping between the current applied and torque exerted for every module over the whole operating range of the module. Torque control capability facilitates dynamic control of the modules based on a dynamic model of the manipulator. Without torque-control capability, we would have to resort to position-based control and thus rely on the actuator's own joint-controller that is usually proprietary.

##### 3.1.1.1.1 Experimental Procedure

To determine the current-torque mapping of a given module, the following procedure was followed. The actuator module was mounted on a rigid structure and a known weight was hung from the end of a bar, with known length and mass, attached to the module output. The bar was held in the vertical position so the weight initially acts along the axis of the bar. A specific current was applied so the weight moves quasi-statically to settle down at a specific angle with the vertical. Noting this angle from the PowerCube demo interface, the torque it is applying on the module is calculated analytically. Thus the output torque is mapped to the current applied.

### 3.1.1.1.2 Results

The results of the current-torque mapping are presented in Figure 3-3 and Figure 3-4. The trend followed is linear to a great extent, except that there is a dead zone corresponding to no output torque for a non-zero current. The square of the coefficient of correlation for a linear least squares fit between the current and torque was 0.9806 and 0.9948 respectively for the PR 110 and PR 090 rotary actuators.

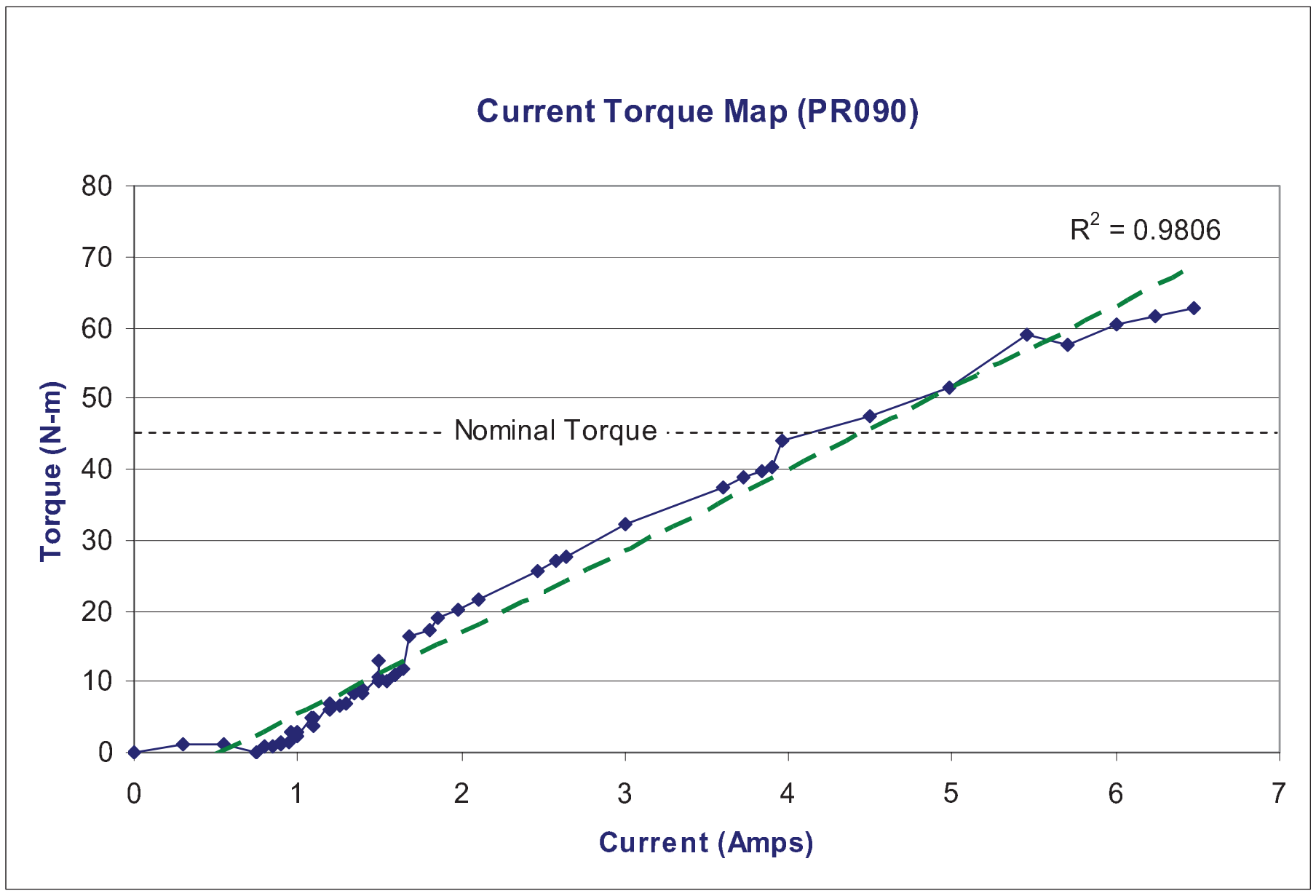

**Figure 3-3. Current-Torque Mapping for PR 090**

### 3.1.1.1.3 Observations

It was important not to apply more than the nominal torque to a module in a quasi-static case. This is because the maximum allowable current (or torque) for the static motor is approximately only one-third of that for the rotating motor. A dead band, corresponding to no torque-production for a given non-zero current, is observed in the ranges 0-0.8 Amps and 0-0.55 Amps for PR 110 and PR 090 modules respectively.

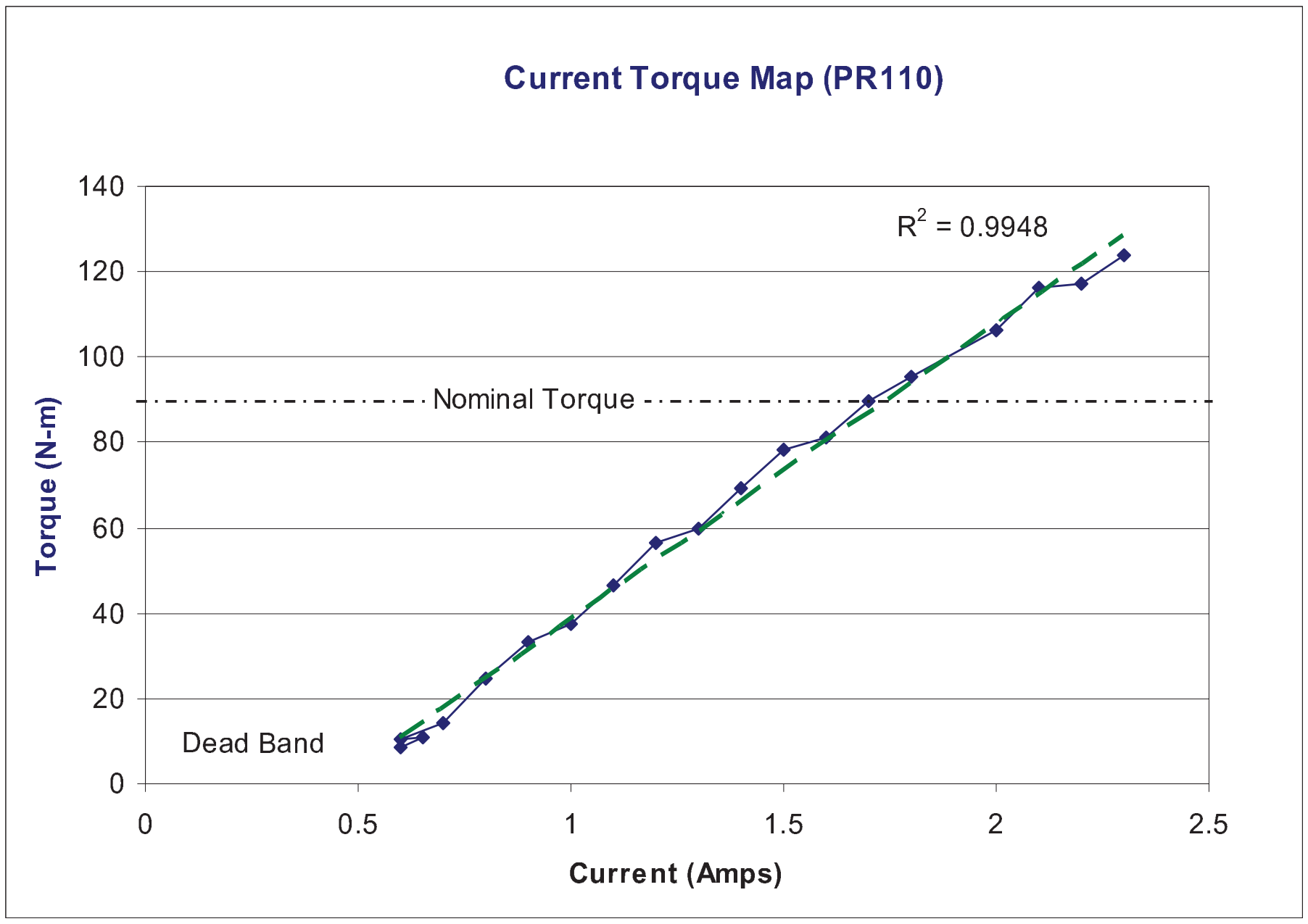

**Figure 3-4. Current-Torque Mapping for PR 110**

The possible reasons for this could be:

- Friction in the harmonic drive transmission. This is a function of the actuator load and velocity.
- Some current is used for the magnetic holding brake and the controller electronics and is therefore not used for torque production.
- The effect of module temperature on the current-torque mapping was neglected in this experiment. However, in future experiments to determine this mapping, temperature should be considered.

#### 3.1.1.1.4 Sources of Error

This section lists the possible sources of error in the experimental procedure used to determine the current-torque mapping for the different actuator modules.

- The additional unaccounted dynamic effects due to the load motion could cause the torque readings to be in excess of the actual reading
- The variation in the current-torque mapping from a linear trend is probably due to the actuator transmission friction

### 3.1.1.2 Determination of Dynamic Parameters

The dynamic parameters, like link and module inertias and Centers of Mass (COM), are necessary for dynamic control. In this section, we describe the experiments conducted to determine these parameters.

#### 3.1.1.2.1 Experimental Procedure

The experimental setup (See Figure 3-5) consisted of a trifilar pendulum [Crede, 1948], the parameters of which are listed in the appendix. It is assumed that the COM of the machine component (whose inertia is to be determined) is known prior to conducting the trifilar pendulum experiment. The COM of the PowerCube modules is not listed in the catalog. Consequently, the location of the COM of the modules had to be determined initially. This was done using a knife edge test in the three principal axes of the module to determine the balancing point. To determine its inertia, a module was positioned on the bottom plate of the trifilar pendulum such that its COM coincides with the center of the plate. This bottom plate was disturbed infinitesimally ($< 5^{\circ}$) from its equilibrium position and allowed to oscillate about the normal to the plate passing through the center of mass of the compound system (plate and module). The period of oscillation was measured using a stopwatch. Knowing the period of oscillation and the parameters of the set-up, the moment of inertia of the module about the axis of rotation can be calculated. The experiment was repeated multiple times for all modules and a mean of the readings was taken.

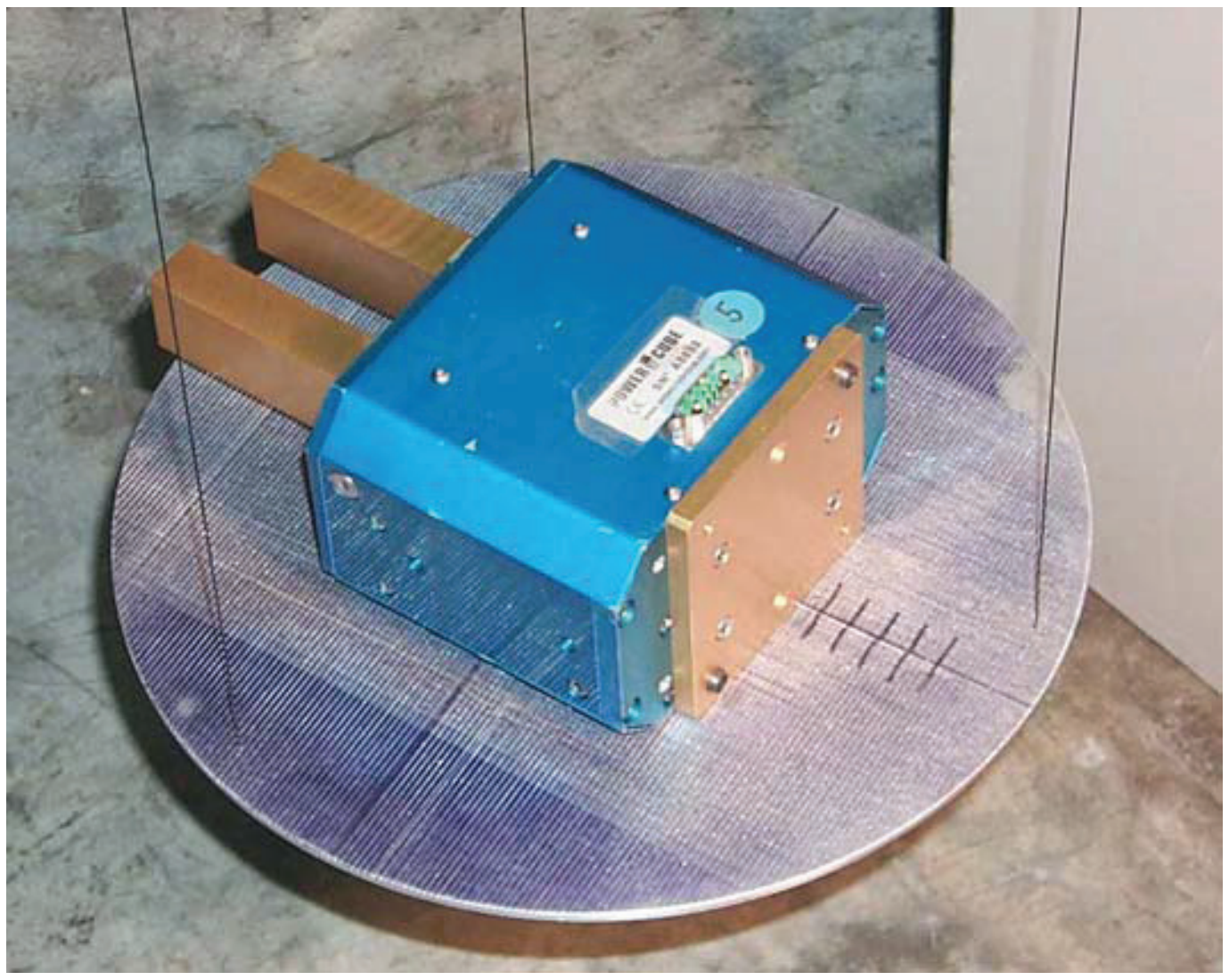

**Figure 3-5. The Gripper Module Under Test On the Trifilar Pendulum**

#### 3.1.1.2.2 Observations

The experimentally determined inertia values of objects with standard shapes (cylinder, plate, tube) were compared with their theoretically determined inertias. An error was noticed that was a function of the ratio of module-inertia to plate-inertia. As this ratio decreased, the error between theoretical and experimental values increased. Noticing this trend in error, a least squares curve fit was introduced based on the readings from the test sample of objects with standard shapes. The error was correlated with the inertia ratio mentioned above. The experimentally determined inertia parameters of the PowerCube modules are listed in the appendix. A plot of the error trend, based on the above mentioned experiments with standard test samples, is also included in the appendix.

#### 3.1.1.2.3 Sources of Error

The following are some possible sources of error in the above mentioned experimental procedure.

- *Leveling Inaccuracy:* The bottom plate should be set-up such that the gravity vector is normal to it.
- *Scale Inaccuracy in Mass and Length Measurements*
- *Variations in Wire Tension:* A fishing line was used for the three cables between the bottom and top plates in the trifilar pendulum. Its extension under the weight of the loaded machine component might introduce errors in the readings.
- *COM Placement:* The COM of the test part needs to coincide with the COM of the bottom plate. Inscribed graduations (at intervals of 1mm) on the plate were used to achieve this centering. However, without precise equipment it is hard to attain perfect centering.
- *Lateral Oscillations of the Plate:* Throughout the experiment the pendulum should rotate only about the normal to the plate passing through the compound COM of the system. Lateral oscillations introduce additional dynamics which are not accounted for in the dynamic model used for this experimental procedure.
- *Large Amplitude of Oscillation:* The dynamic model for this pendulum is linearized about the equilibrium position. Hence it is necessary to disturb the plate with an angle not more than 5 degrees for $\frac{\sin\theta}{\theta} \cong 1$ to be true.

- *Number of Significant Digits on the Period of Oscillation:* The measured moment of inertia is proportional to the square of the period of oscillation as shown in Eq.(3-4).

$$J \propto T^2 \tag{3-4}$$

$$\Delta J \propto 2\Delta T \tag{3-5}$$

Applying finite differences on either side of Eq.(3-4), we observe that the inaccuracy in the derived quantity (J) is twice the inaccuracy in the measured quantity (T). Only one decimal place was used for the period of oscillation due to limiting experimental equipment. The accuracy of the inertia readings can be improved by two times as that of the time period.

### 3.1.2 Force/Torque (F/T) Sensing System

The UTRRG force control test-bed is equipped with a Force/Torque (F/T) sensor system from Assurance Technologies Inc (Refer Figure 3-6). This sensor measures all six components of forces and torques ($F_x$, $F_y$, $F_z$, $T_x$, $T_y$, $T_z$) on the end-effector. The system consists of a transducer, shielded high-flex cable, a stand-alone F/T controller and associated software.

The transducer has axial, transverse and torsion load capacities of 300 lbs, 150 lbs and 600 lb-in respectively. The maximum data transfer rate attainable with the control system is 600 Hz [ATI Inc. Website]. The accuracy (including manipulator vibration) during operation of the robot is 0.25 lbs with filtering. The F/T sensor controller outputs force/torque data through the RS-232 port at baud rates varying from 1200 to 38400. The communication rate of 38400 baud was used for our force control experiments. The stiffness matrix for the F/T sensor is given in Eq.(3-6). Note that the unit for translational stiffness is lb/in and that for torsion stiffness is in-lb/rad.

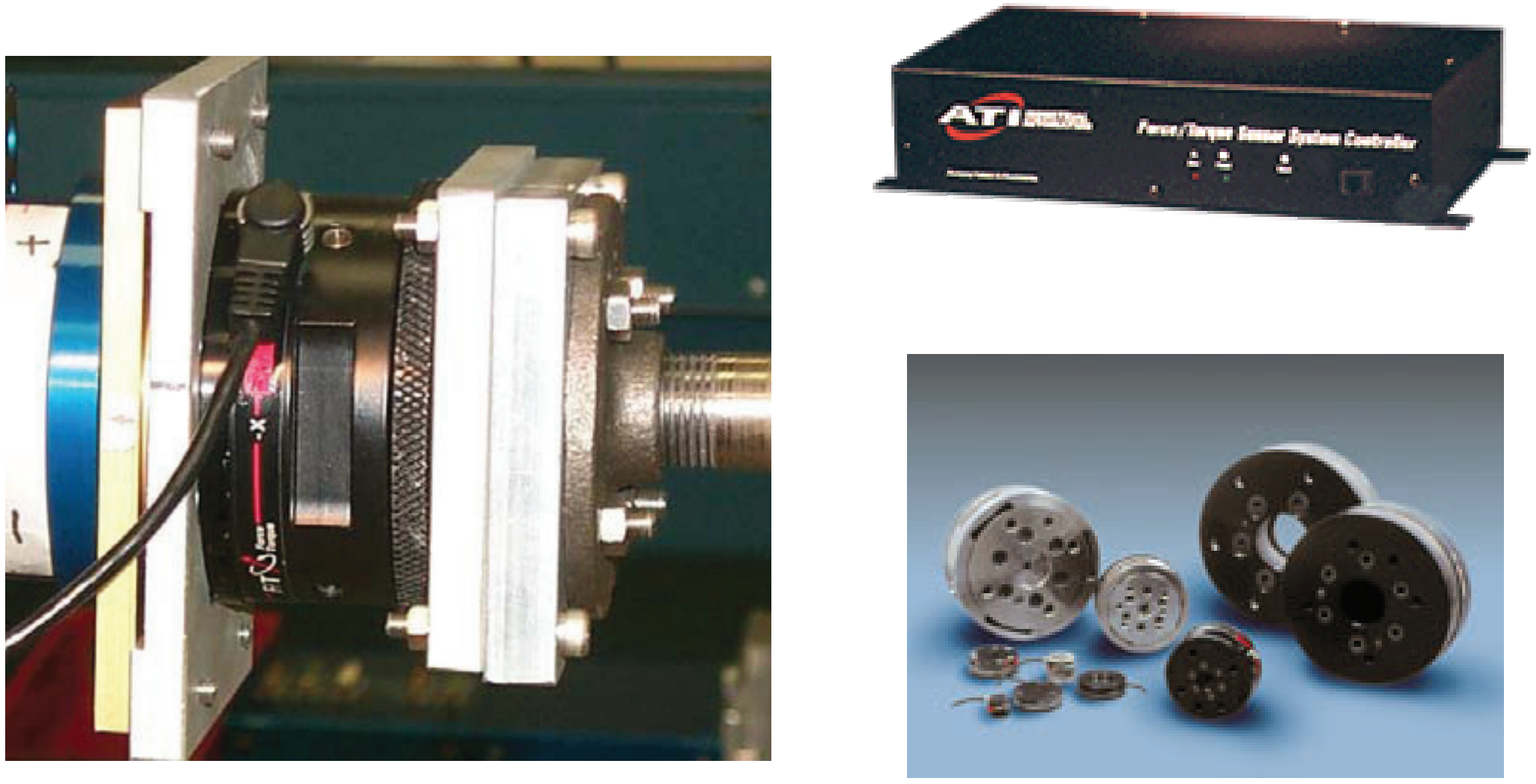

**Figure 3-6. Peripheral Force/Torque Sensor System (ATI Gamma FT 30/100)**

$$\begin{bmatrix} F_x \\ F_y \\ F_z \\ \tau_x \\ \tau_y \\ \tau_z \end{bmatrix} = \begin{bmatrix} 50 & 0 & 0 & 0 & 0 & 0 \\ 0 & 50 & 0 & 0 & 0 & 0 \\ 0 & 0 & 100 & 0 & 0 & 0 \\ 0 & 0 & 0 & 90 & 0 & 0 \\ 0 & 0 & 0 & 0 & 90 & 0 \\ 0 & 0 & 0 & 0 & 0 & 140 \end{bmatrix} \text{x}\, 10^3 \begin{bmatrix} \Delta x \\ \Delta y \\ \Delta z \\ \Delta \phi_x \\ \Delta \phi_y \\ \Delta \phi_z \end{bmatrix} \quad \textbf{(3-6)}$$

#### 3.1.2.1 Frame Transformations

*End-point Force Control* (EFC) may be defined as an active force control scheme wherein force/torque feedback is obtained from an F/T sensor appended to the end-of-arm of the serial chain between the last link and the tool. In order to implement EFC, the F/T signals should be interpreted in a convenient frame of reference. A frequent choice for such a reference frame is parallel to the robot world frame and coincident with the wrist-center. Let us denote this frame by $\mathcal{B}$ and call the tool frame and sensor's local frame $\mathcal{T}$ and $\mathcal{S}$ respectively.

The vector containing forces and moments will be referred to as a *wrench* ($\mathbf{W} = \begin{bmatrix} F_x & F_y & F_z & \tau_x & \tau_y & \tau_z \end{bmatrix}^T$).

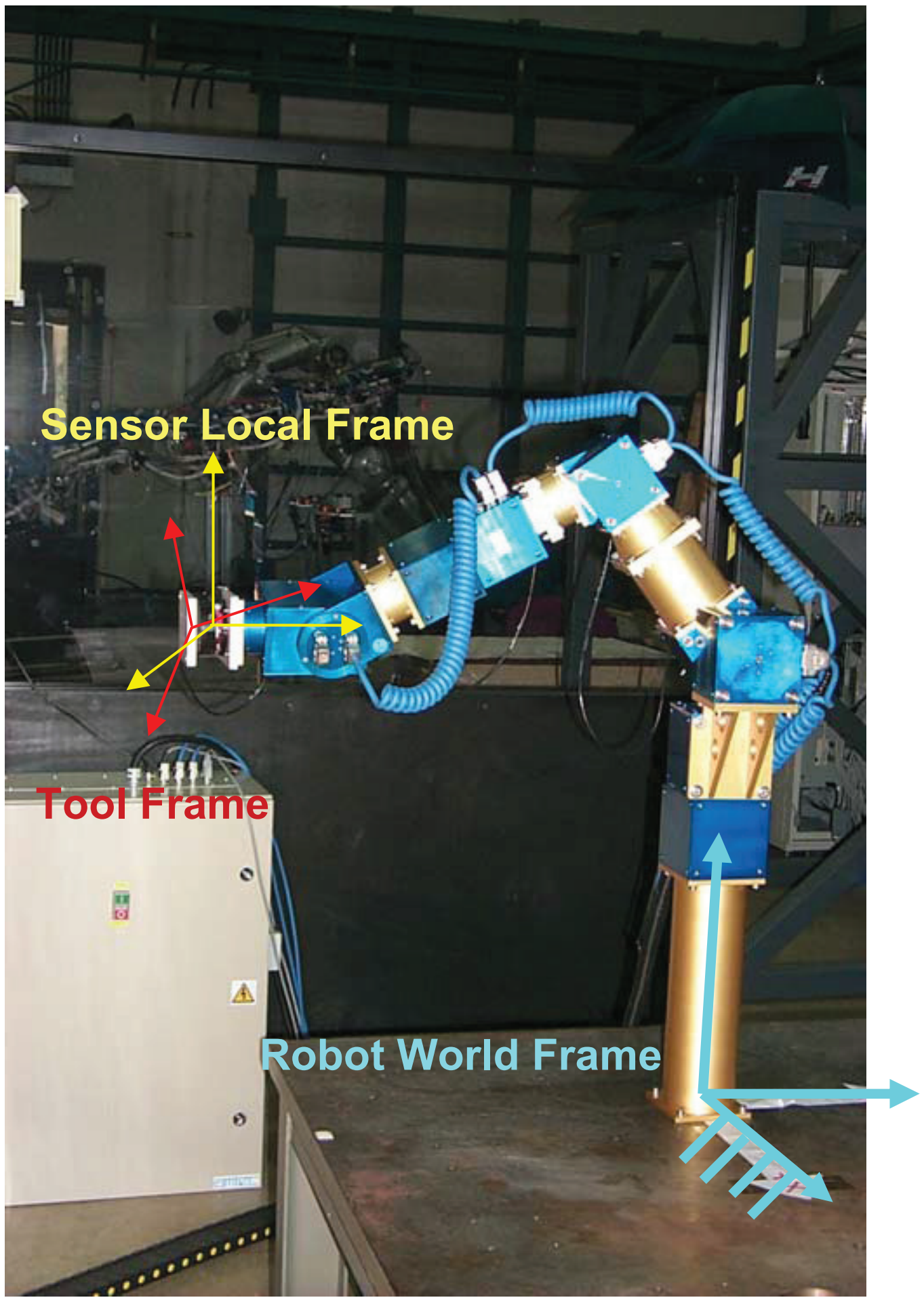

**Figure 3-7. Modular Robot Test Bed**

The end-effector wrench is transformed from $\mathcal{S}$ to $\mathcal{B}$ using Eq.(3-7).

$$\mathbf{W}_{\mathcal{B}} = \mathbf{X}_{\mathcal{T}}^{\mathcal{B}} \mathbf{X}_{\mathcal{S}}^{\mathcal{T}} \mathbf{W}_{\mathcal{S}} \tag{3-7}$$

where $\mathbf{X}_{\mathcal{F}_1}^{\mathcal{F}_0}$ denotes the spatial transformation matrix from frame $\mathcal{F}_1$ to $\mathcal{F}_0$.

$$\mathbf{X}_{\mathcal{F}_1}^{\mathcal{F}_0} = \begin{bmatrix} \mathbf{R}_{\mathcal{F}_1}^{\mathcal{F}_0} & \mathbf{0} \\ \mathbf{p}_1^0 \times \mathbf{R}_{\mathcal{F}_1}^{\mathcal{F}_0} & \mathbf{R}_{\mathcal{F}_1}^{\mathcal{F}_0} \end{bmatrix} \tag{3-8}$$

$\mathbf{R}_{\mathcal{F}_1}^{\mathcal{F}_0}$ is the rotation matrix from $\mathcal{F}_1$ to $\mathcal{F}_0$ and $\mathbf{p}_1^0$ is the position vector of the origin of $\mathcal{F}_1$ in $\mathcal{F}_0$.

#### 3.1.2.2 F/T Sensor Calibration

An F/T sensor has to be calibrated when it is freshly installed on a manipulator. The information that needs to be determined to begin implementing force control applications on the manipulator is described in this section.

The spatial force transformation matrix is important information required to usefully interpret the F/T sensor signal. In Eq.(3-7), the transformation $\mathbf{X}_{\mathcal{T}}^{\mathcal{B}}$ is time-varying and has to be determined based on the manipulator pose at any instant. However the transformation $\mathbf{X}_{S}^{\mathcal{T}}$ is static and can be determined based on the orientation of the sensor's local frame w.r.t the tool frame. This relative orientation and the static transformation are given in the appendix.

Ideally for a force control application, the forces and torques measured by the F/T sensor should be zero when there is no payload attached. To satisfy this condition, the weight of a peripheral tool should be subtracted every time an F/T reading is taken so that these represent the active forces and torques applied by the environment. This tare weight correction is called a *bias* and is deliberately applied to compensate for the dynamic properties of the tool. A mere subtraction of the tool weight suffices to bias a force component. However, the dynamic offsetting of torques is more difficult since it is dependent on the configuration of the robot. For all the force control applications reported here, the bias is calculated automatically by averaging the tool weight over approximately 100 readings and subtracting this value from the F/T reading every time.

It is necessary to have an idea of the noise content in the F/T signal. The company specification on the signal/noise ratio for the F/T 30/100 sensor is 75. This means that the noise level of this sensor is approximately 1.33% (1/75). This is a good enough number for real-time force control. However manipulator jitters, human interactions, and vibrations may contribute to the noise content. Hence a 16 point moving average filter was implemented for all force control applications.

#### 3.1.2.3 F/T Sensor Software

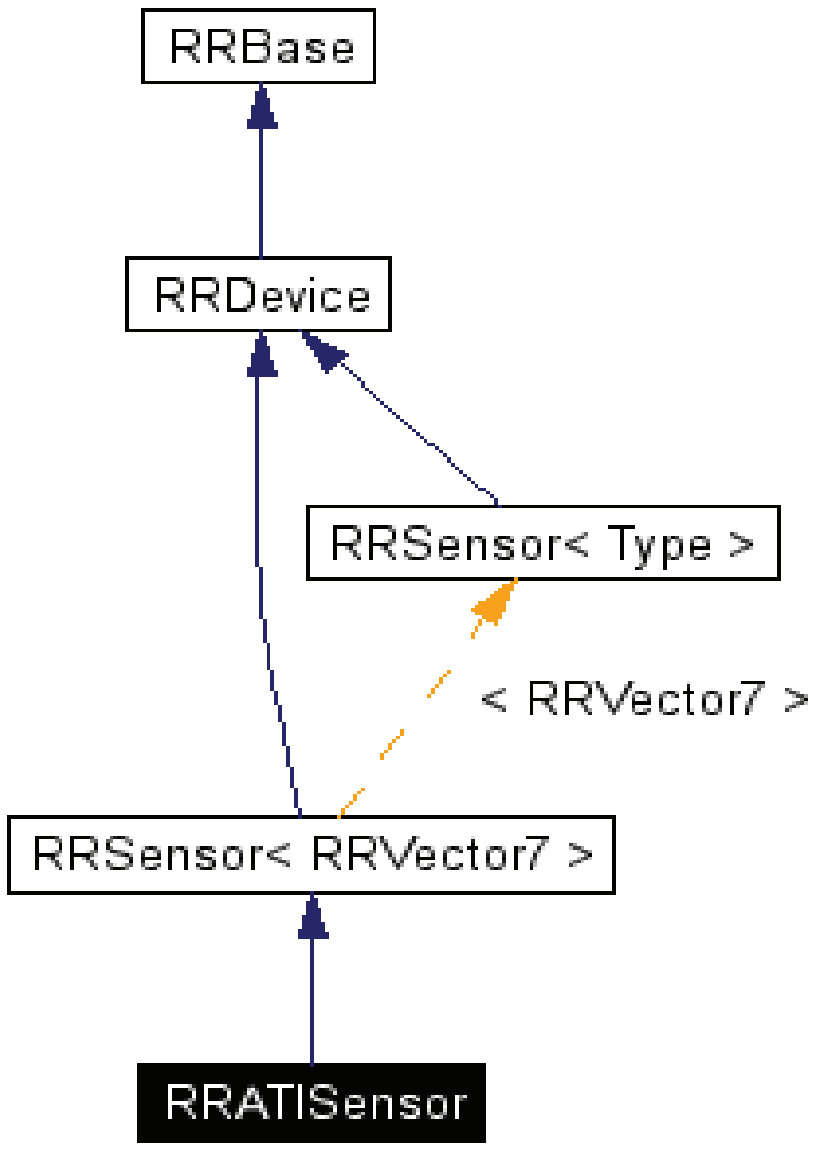

**Figure 3-8. Sensor Class Inheritance Diagram**

Operational Software Components for Advanced Robotics (OSCAR) [Kapoor and Tesar, 1996] is an object-oriented generalized software framework to develop manipulator control applications. The OSCAR class used for applications associated with the force/torque sensor is *RRATISensor*.

This class derives from the abstract class *RRSensor<Type>* under the Device library [OSCAR Online Documentation]. Figure 3-8 shows the inheritance[4] diagram for this class in which we show the root class from which the *RRSensor* class is derived. Table 3-1 shows the F/T sensor class interface.

**Table 3-1. Force/Torque Sensor Class Interface**

| Method | Description |
|---|---|
| ***RRATISensor(SerialPortNo, OSCARError);*** | Constructor |
| ***Initialize();*** | Opens the serial communication interface and sets its parameters. A call to this method is mandatory |
| ***Read(ForceVector);*** | Reads 6-axis forces from the sensors and runs error checks |
| ***SetSpatialXform(FrameTransformation Matrix);*** | Facilitates user-definition of the data member corresponding to the frame transformation matrix. By default it is set to identity rotation and zero displacement |
| ***GetSpatialXform(FrameTransformation Matrix);*** | Queries and returns the frame transformation matrix |
| ***SetBias(Bias Vector);*** | Sets the bias for the force components. By default it is set to zero vector. The use of this method is optional |
| ***GetBias(BiasVector);*** | Queries and returns the bias vector |
| ***ApplyTransform(True or False);*** | If the argument is 'true', the force readings from Read (ForceVector) are in the frame defined by the frame transformation matrix. Else (by default), it is in the sensor's local frame. |
| ***Close();*** | Closes the serial communication interface |

[4] Inheritance diagram is a graphical representation following a tree structure that illustrates the relation between root classes and derived classes.

### 3.1.3 Test Environments and Contact Models

Two sample surfaces were used as representative stiff and compliant contact environments for the force control applications. The compliant surface was a weighing scale and the stiff surface was a vibration damper pad made of cork with hard rubber lining. These are shown in Figure 3-9 (a) and (b) respectively. The stiffness of the compliant surface was measured to be approximately 25 lbs/in and that of the F/T sensor (in the local Z-direction) is $100 \times 10^3$ lbs/in.

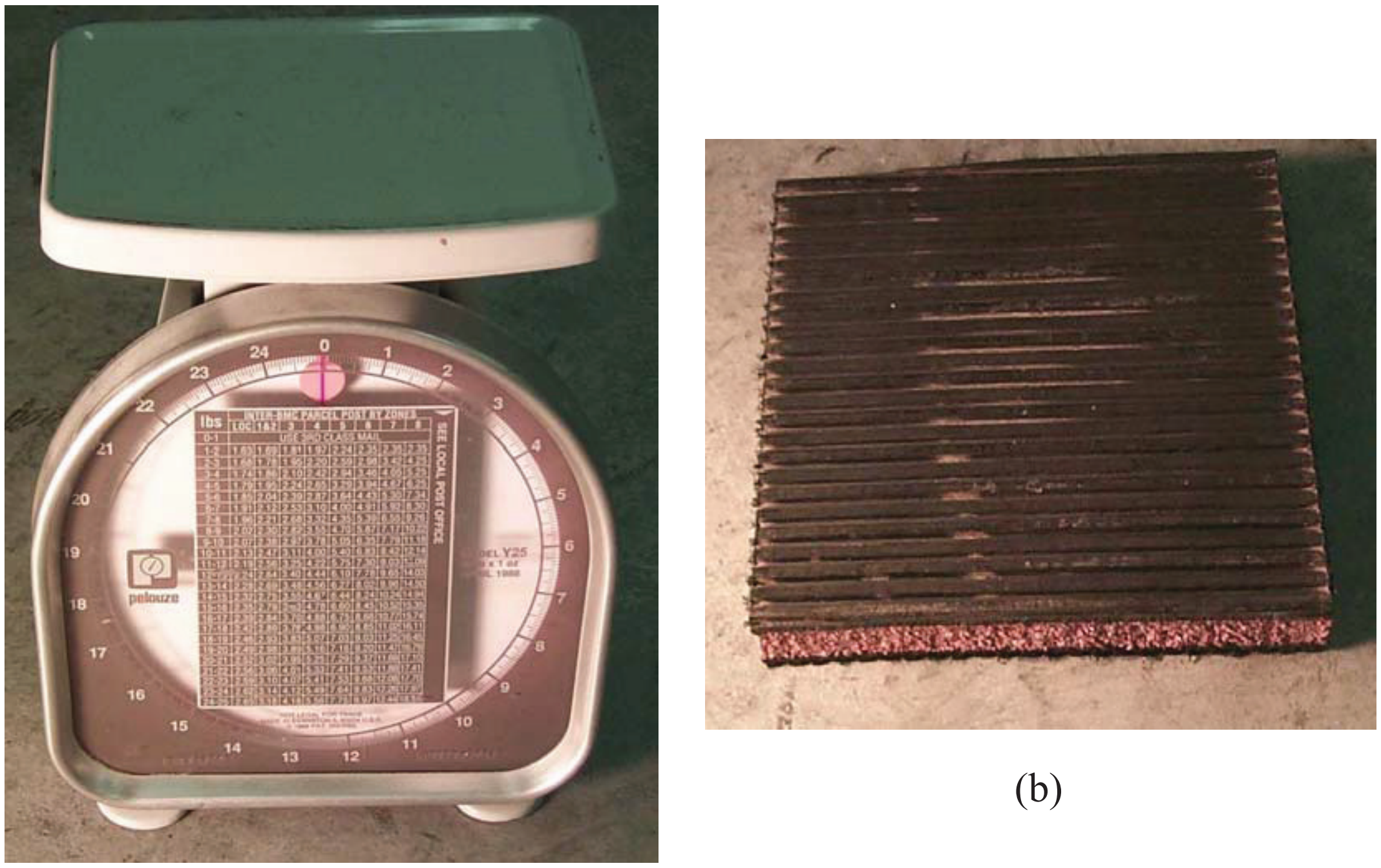

(a) (b)

**Figure 3-9. Test Environments Used for Force Control**

Figure 3-10 shows the simplest interaction model between the manipulator EEF and the test environment. This is a stiffness model where the stiffness of the sensor, tool, and environment are summed in series to obtain the net interaction stiffness as shown in Eq.(3-9).

$$\frac{1}{K_{eff}} = \left( \frac{1}{K_{sensor}} + \frac{1}{K_{tool}} + \frac{1}{K_{env}} \right) \tag{3-9}$$

A relatively compliant surface was used for most force control implementations to ensure safe contact transitions without large impulses. The contact environment is modeled as an elastically compliant plane and the contact is assumed to be a frictionless point contact.

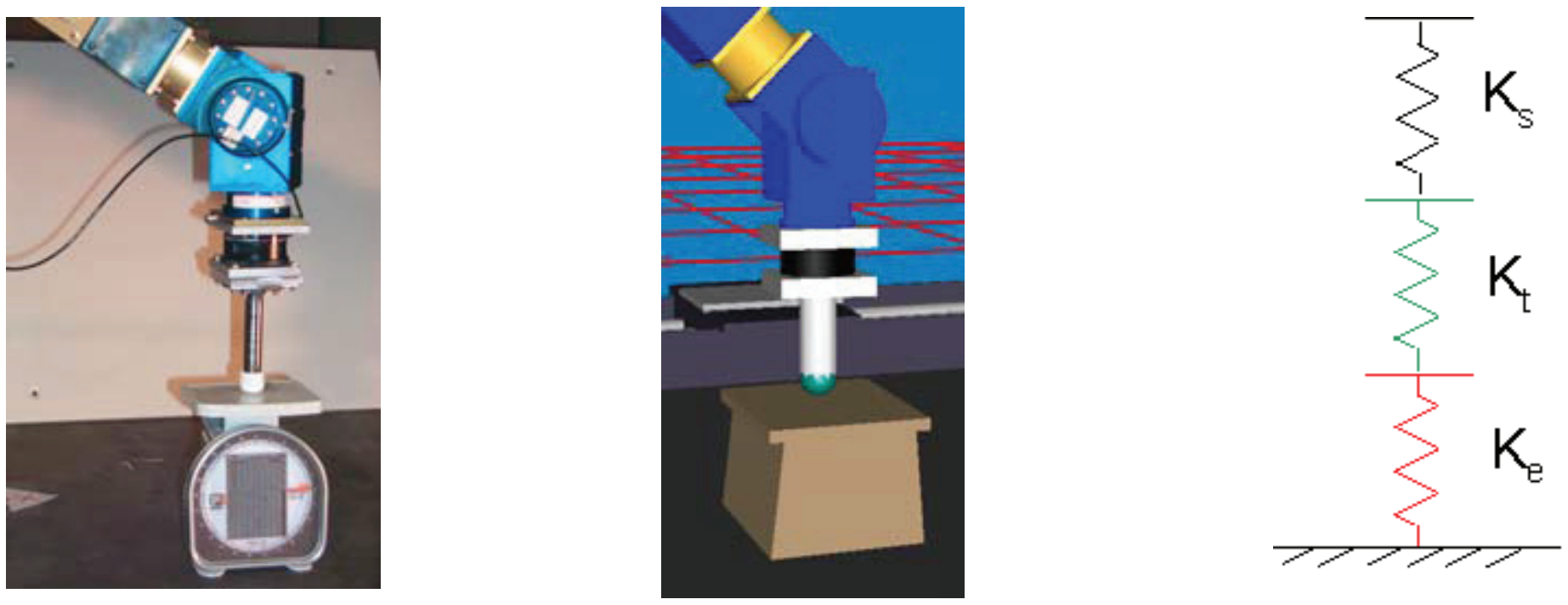

(a) (b) (c)

**Figure 3-10. Contact Model for Test Surfaces**

## 3.2 Software Platform: OSCAR

The software platform used for application development was Operational Software Components for Advanced Robotics (OSCAR) developed by Kapoor and Tesar [1996] at UT Austin. This section describes the components within OSCAR that facilitate the development of control applications for manipulators. Figure 3-11 shows the association diagram at the sub-system level for OSCAR. The control systems domain extends OSCAR to include control algorithms.

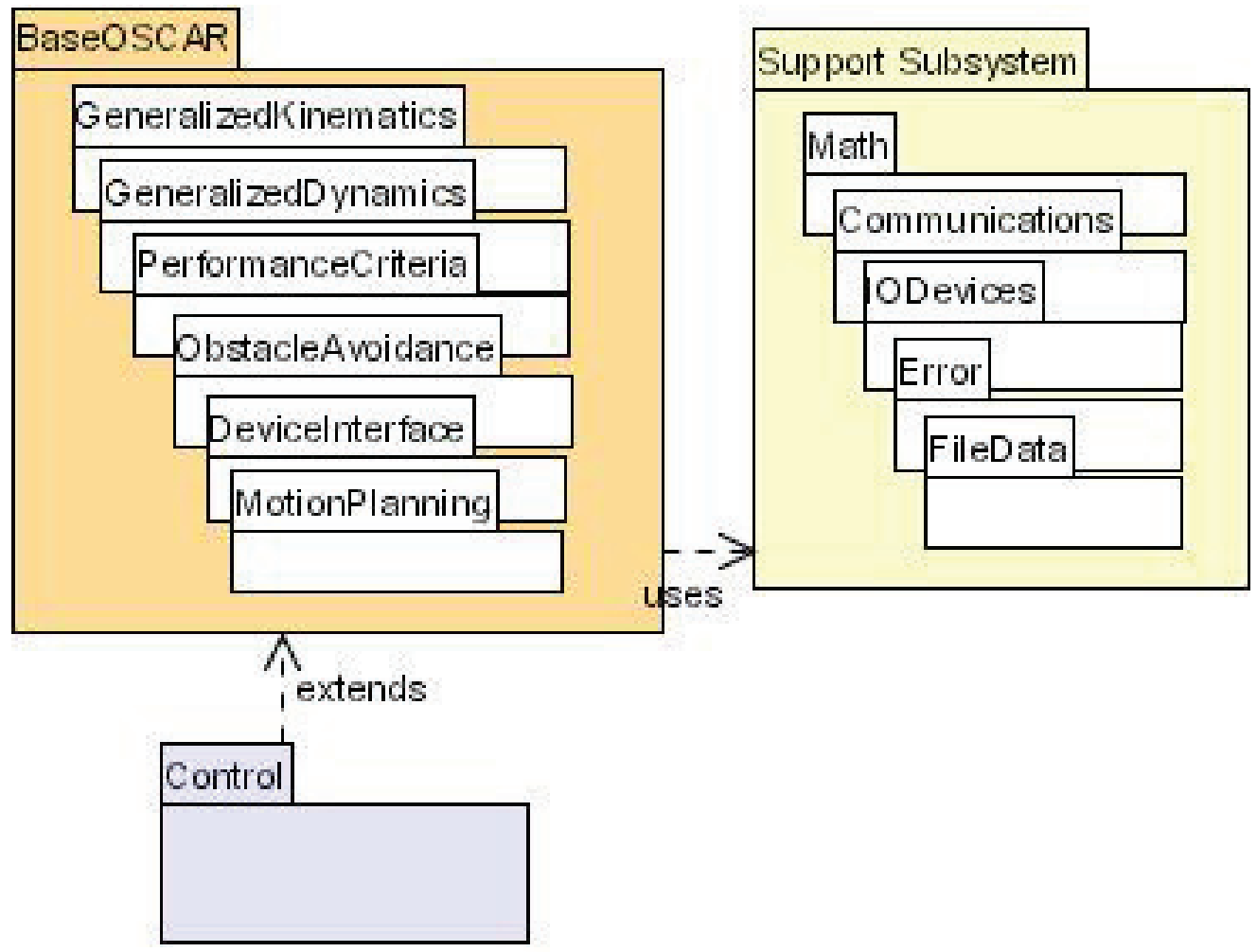

**Figure 3-11. Subsystem Level Association Diagram for OSCAR**

The initial architecture developed for control system abstraction is shown in Figure 3-12. The shaded blocks are abstract classes. *RRBase* is a parent class from which most domains are derived.

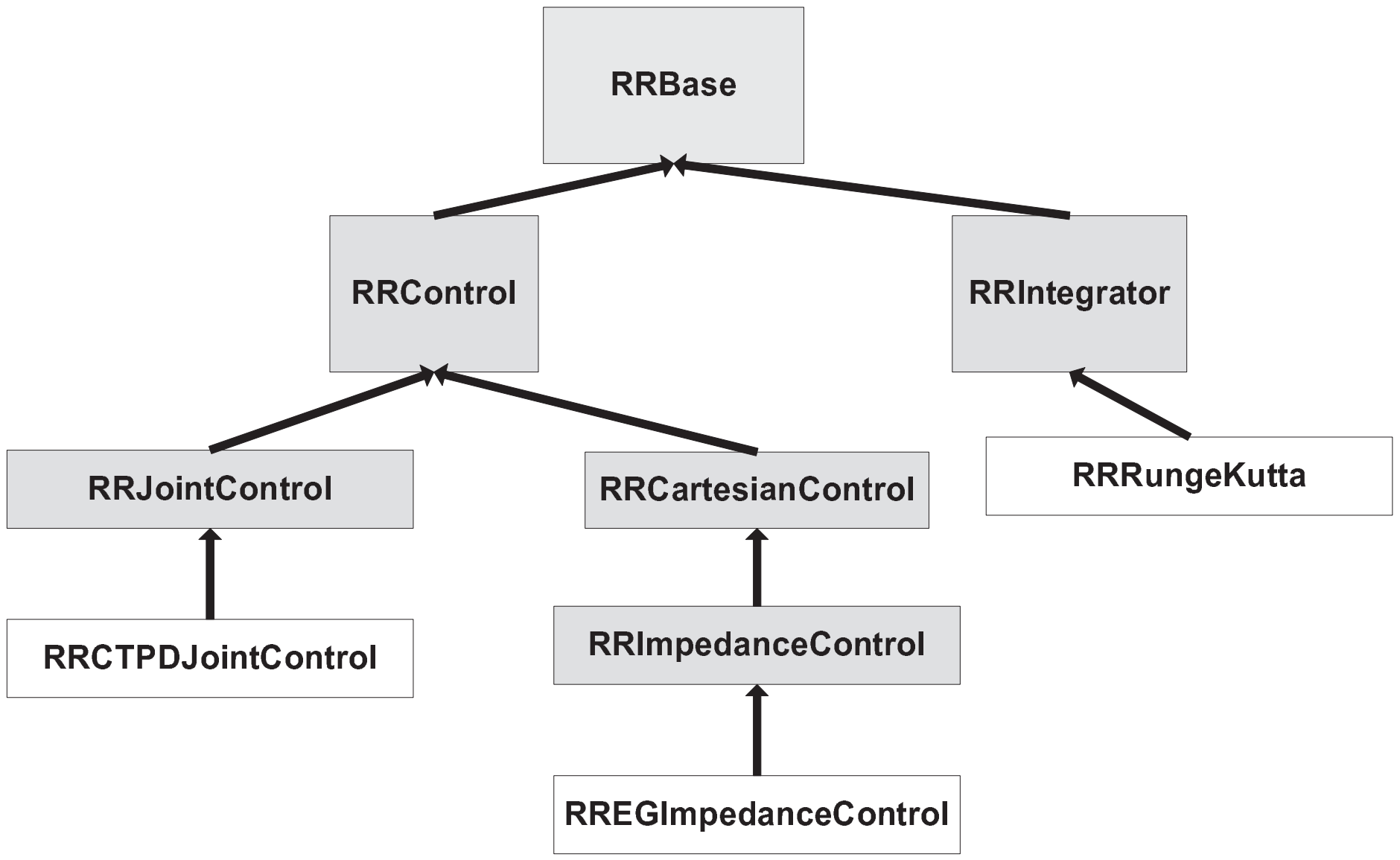

**Figure 3-12. Inheritance Diagram for the Control Systems Class**

As the architecture shows, *RRControl* is the root class within the control systems domain from which *RRJointControl* and *RRCartesianControl* are derived. These latter classes, as their names suggest, implement functionality for control algorithms in two of the configuration spaces of a robot, viz. joint space and Cartesian space. These abstract classes declare the basic functionality required for the respective control algorithms. Specific control laws that drive from these classes, like *RRCTPDJointControl* (that abstracts computed-torque proportional derivative control in the joint space and is derived from *RRJointControl*) and *RREGImpedanceControl* (that abstracts extended generalized impedance control and is derived from *RRImpedanceControl*), implement these functions.

```
#ifndef CTPDJointControl_hpp
#define CTPDJointControl_hpp

#include "JointControl.h"

class RRCTPDJointControl : public RRJointControl
{
public:
  RRCTPDJointControl(unsigned int dof, RRIDStandAlone* dyn);
  virtual ~RRCTPDJointControl();

  virtual bool SetGains(const RRVectorArray& gains);
  void SetSamplingTime(double dt);
  virtual bool GetControlTorques(RRVector& torques);

protected:
  RRMatrix Kp, Kv, Ki;
  double  samplingTime;
  RRVector integral;
  RRVector *force;
};

#endif //CTPDJointControl_hpp
```

**Figure 3-13. Example Control Class: Computed Torque PD Control**

Figure 3-13 illustrates the implementation of computed torque PD control. A Runge-Kutta 4$^{th}$ order integrator was implemented in the class *RRRungeKutta*. This is used in simulating system dynamics (forward dynamics).

## 3.3 Chapter Summary

This chapter described the force control testbed at UTRRG. Main components of this testbed are the PowerCube modular manipulator, Force/Torque (F/T sensor from Assurance Technologies Inc., representative rigid and compliant contact surfaces, and the software platform (OSCAR). Detailed documentation of the experimental work undertaken to determine the dynamic parameters of the manipulator modules is presented. The control systems domain within OSCAR is explained with illustrative examples.

# Chapter 4
# Force Control Experiments

To identify some issues in implementing force control strategies on a real robotic system, experimental force control was necessary as part of this research work. This was also required to identify the influence of some actuator parameters on system-level force control performance. The experimental activities documented in this chapter were simple implementations that raised important questions about the main subject of investigation of this report, viz., multi-domain inputs and their effect on system performance. Apart from that, these contributed to the capability of the UTRRG robotics lab. In this chapter we consider two scenarios, one where the modular manipulator equipped with a peripheral F/T sensor is used as an input device and another where it is used as an active system performing a force-controlled task on an environment with unknown stiffness. The testbed used to conduct these experiments was described in Chapter 3.

## 4.1 Position Based Force Control

The experiments described in this chapter are based on end-point, position-based, active force control. In the interest of clarity terms are defined here. *Position-based force control* does not involve the dynamic model of the system. It commands an accommodative motion based on the sensed forces and torques. *End-point force control* is based on a peripheral force/torque sensor appended to the system at its output (before the tool, for a serially-linked manipulator). *Active force control*, as opposed to passive force control, does not utilize any mechanical compliance device[5] at the end-of-arm. The decision making in this scheme is based only on force/torque sensing and user-defined control scheme.

## 4.2 Experimental Verification of Wrench Transformations

In this section we describe the applications developed to ascertain the frame transformations for the end-effector wrench ($\mathbf{W} = \begin{bmatrix} F_x & F_y & F_z & \tau_x & \tau_y & \tau_z \end{bmatrix}^T$). The approach followed was to first transform the wrench from the F/T sensor's local frame to a frame parallel to the robot world frame ($\mathcal{W}$) but coincident with the tool frame ($\mathcal{T}$) as in Eq.(4-1). Such a reference frame is denoted by $\mathcal{B}$ here.

$$\mathbf{W}_{\mathcal{B}} = \mathbf{X}_{\mathcal{T}}^{\mathcal{B}} \mathbf{X}_{\mathcal{S}}^{\mathcal{T}} \mathbf{W}_{\mathcal{S}} \qquad \textbf{(4-1)}$$

Please note that $\mathbf{X}_{\mathcal{S}}^{\mathcal{T}}$ is a static transformation dependent on the relative orientation of the tool frame w.r.t the local sensor frame but $\mathbf{X}_{\mathcal{T}}^{\mathcal{B}}$ is dynamic changing dependent on the configuration of the robot. Also note that the form of this dynamic transformation is $\mathbf{X}_{\mathcal{T}}^{\mathcal{B}} = \begin{bmatrix} \mathbf{R}_{\mathcal{T}}^{\mathcal{B}} & \mathbf{0} \\ \mathbf{0} & \mathbf{R}_{\mathcal{T}}^{\mathcal{B}} \end{bmatrix}$ since $\mathbf{p}_{\mathcal{T}}^{\mathcal{B}}$ is a null-vector due to coincidence of these frames. This transformation matrix ($\mathbf{X}_{\mathcal{T}}^{\mathcal{B}}$) is determined by extracting the rotation matrix ($\mathbf{R}_{\mathcal{T}}^{\mathcal{B}}$) corresponding to the EEF pose at any instant.

---

[5] A *Mechanical Compliance Device* is a passive forgiving element (appended at the end-effector) that is designed to compensate for variations in desired task performance due to positioning errors. An example is the Remote Center Compliance (RCC) [Whitney, 1982]

Subsequently, a proportional accommodative displacement, $\mathbf{\Delta u}_{\mathcal{W}} \in \mathcal{R}^{6}$, in end-effector space is commanded based on the perceived end-effector wrench as shown in Eq.(4-2).

$$\mathbf{\Delta u}_{\mathcal{W}} = \mathbf{K}\mathbf{W}_{\mathcal{B}} \quad \textbf{(4-2)}$$

$$\mathbf{\Delta u}_{\mathcal{W}} = \mathbf{K}\left[\mathbf{X}_{\mathcal{T}}^{\mathcal{B}}\mathbf{X}_{\mathcal{S}}^{\mathcal{T}}\mathbf{W}_{\mathcal{S}}\right] \quad \textbf{(4-3)}$$

where $\mathbf{K} \in \mathcal{R}^{6x6}$ is a positive-definite diagonal linear gain matrix. Using the control scheme in Eq.(4-3), the end-effector response to an external wrench is similar to that of an inertial mass with damping. Higher the gain in a specific direction, the less is the simulated inertial mass and damping in that direction and vice-versa.

This control scheme was implemented on the 6-DOF PowerCube manipulator with force feedback from the ATI F/T sensor. The results are summarized in Figure 4-1, Figure 4-2, Figure 4-3, and Figure 4-4. A diagonal linear gain matrix $\mathbf{K} = diag(3.5, 3.5, 5.25, 0.0, 0.0, 0.0)$ was used to command position modifications based on external forces according to Eq.(4-3). The unit of every element of $\mathbf{K}$ matrix is mm/lbf. For the purpose of verifying the spatial force transformations, the EEF was moved along an approximate figure of eight trajectory as shown in Figure 4-1. The total time taken to complete the trajectory was 13 seconds. The results presented here are in the X-Z plane of the manipulator EEF space. Figure 4-2 shows the damping behavior of the EEF in response to external forces in X and Z directions. Since the damping gain in Z-direction is 1.5 times that in the X-direction, the slope of the force vs. velocity curve is less in the former case. A dead band of 0.25 lbs, which corresponds to the minimum force input required to initiate motion, is purposefully introduced on all axes. This was done to prevent jittering of the EEF in response to F/T sensor noise. In Figure 4-3 and  Figure 4-4, the inertial response of the EEF is presented.

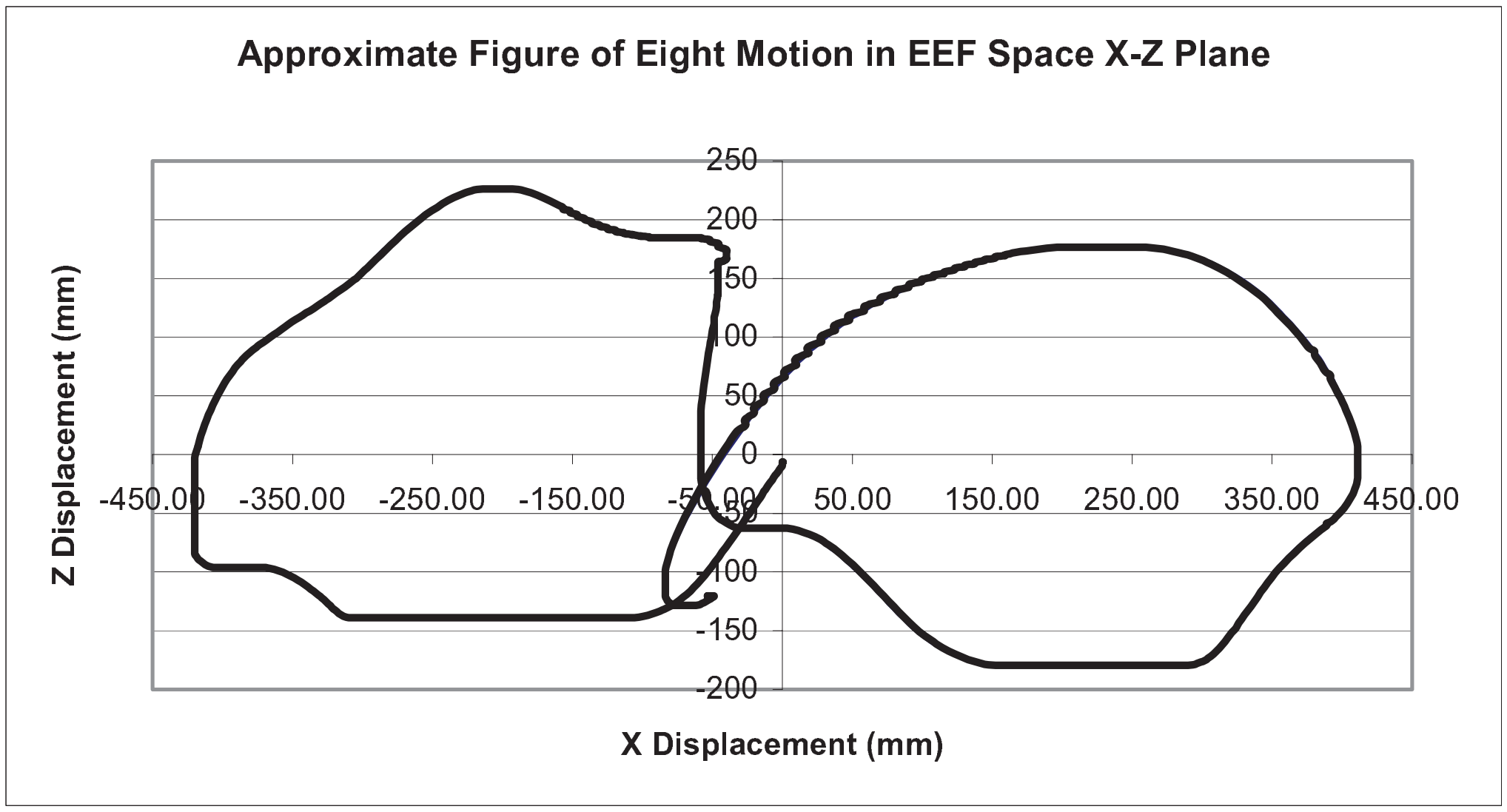

**Figure 4-1. Test Motion Trajectory in EEF Space**

Due to manipulator vibration and tremors due to operation, the acceleration signal, derived from the velocity signal, contained undesirable noise. Hence, a 16 point average trend line was fit to this data to study the variation of acceleration w.r.t force input. The acceleration signal approximately follows the trend of the force signal (analogous to an inertial element) but with a negative sign. This is because the F/T sensor measures reactive rather than active forces/torques. In other words, the F/T sensor measures forces/torques exerted on it by the environment rather than the forces/torques it exerts on the environment.

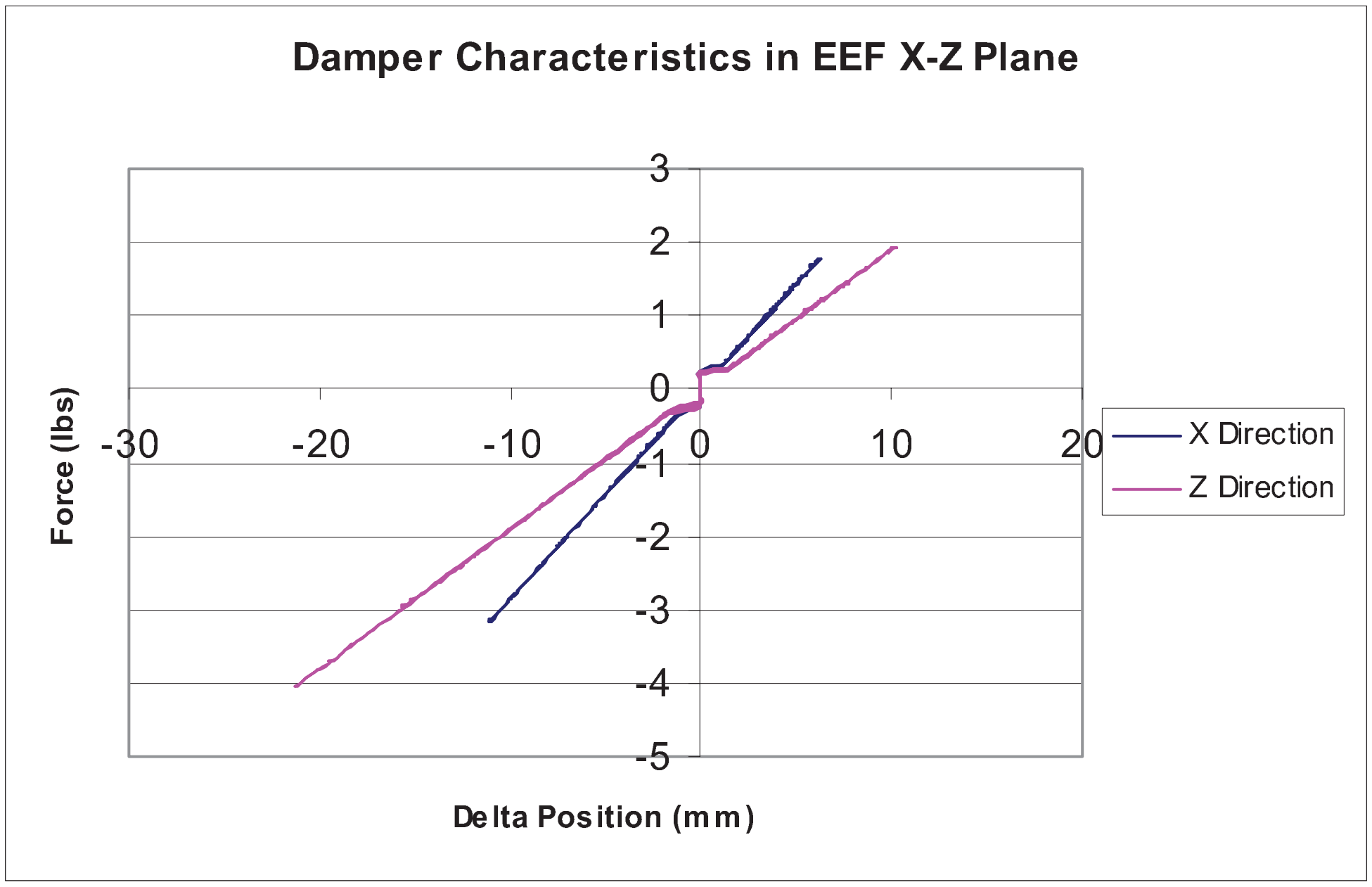

**Figure 4-2. Damping Behavior in EEF Space**

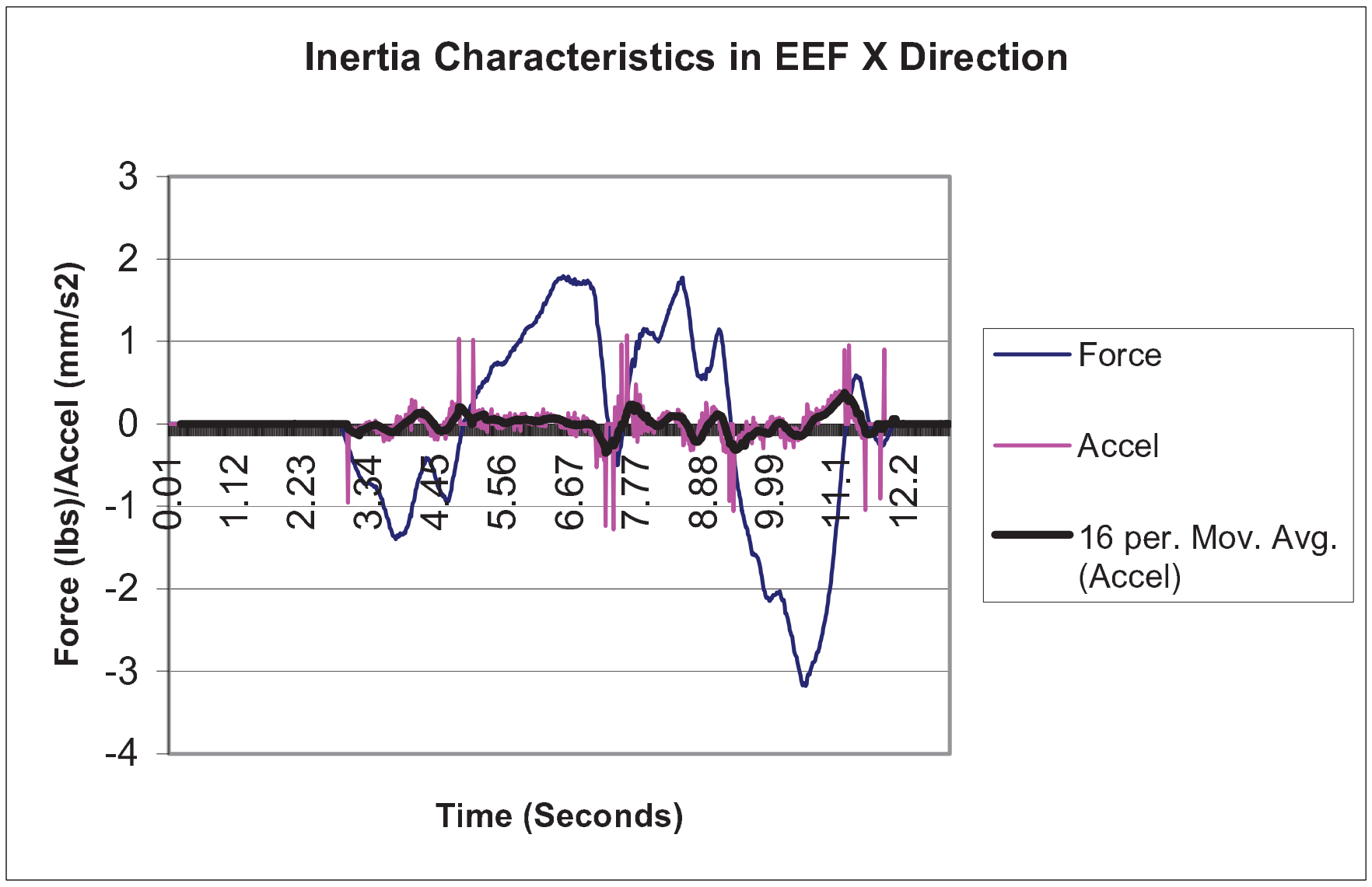

**Figure 4-3. X-Direction Inertial Response**

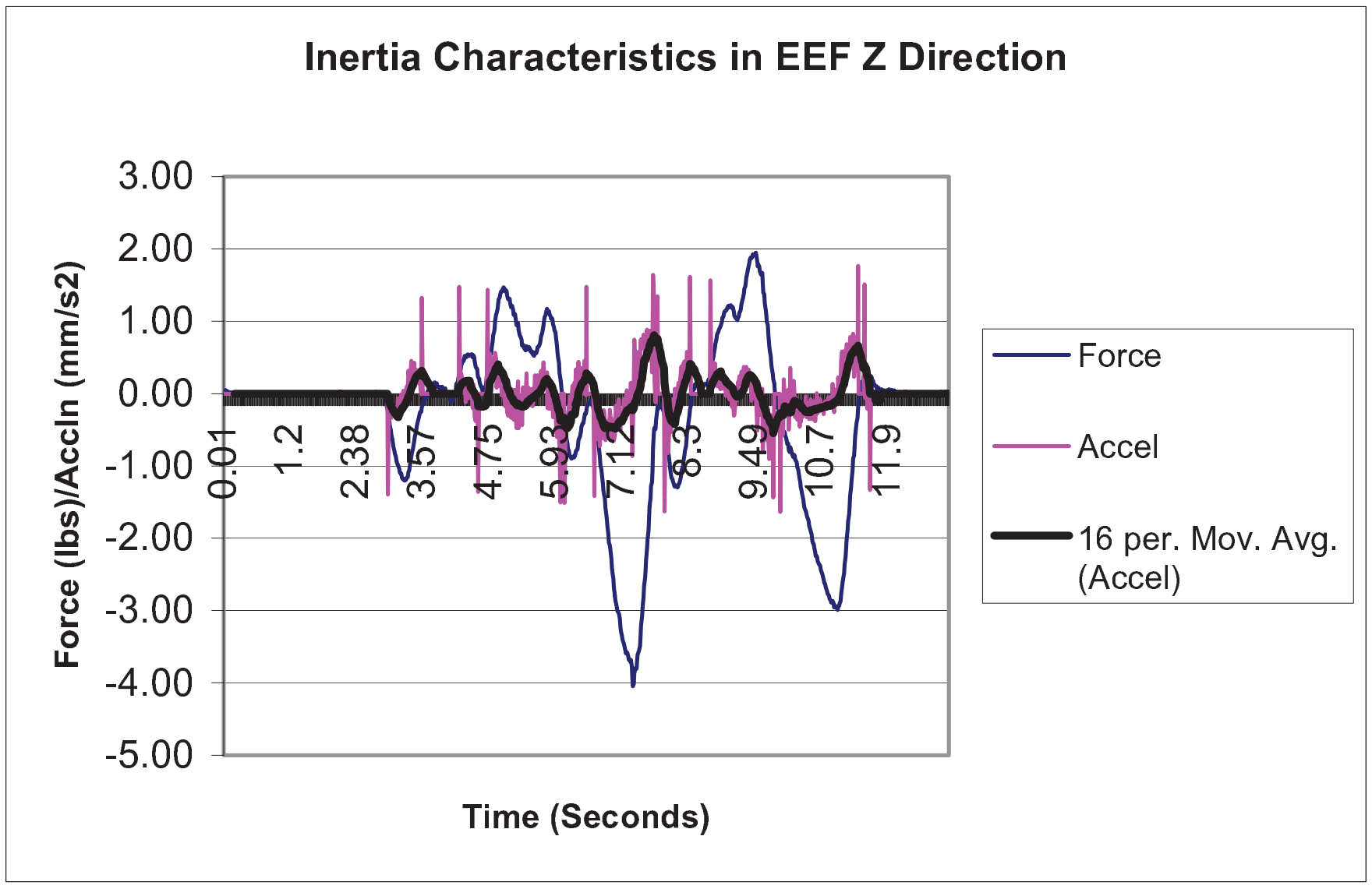

**Figure 4-4. Z-Direction Inertial Response**

A variant of the above application was developed to simulate a virtual spring at the EEF. In this case, the accommodative positional displacement in the EEF space was commanded according to Eq.(4-4).

$$\Delta \mathbf{u}_{\mathcal{W}} = \mathbf{K}\left[\Delta \mathbf{W}_{\mathcal{B}}\right] \tag{4-4}$$

$$\Delta \mathbf{u}_{\mathcal{W}} = \mathbf{K}\left[\mathbf{X}_{\mathcal{T}}^{\mathcal{B}}\mathbf{X}_{\mathcal{S}}^{\mathcal{T}}\right]\left[\Delta \mathbf{W}_{\mathcal{S}}\right] \tag{4-5}$$

The results of this implementation from the PowerCube robot testbed in the EEF space X-Z plane are presented in Figure 4-5 and Figure 4-6.

In this demonstration the EEF was displaced in the X-Z plane along random directions. The compliance matrix used for this application was $\mathbf{K} = diag(10.0, 3.0, 7.0, 0.0, 0.0, 0.0)$. The unit used for the members of the $\mathbf{K}$ matrix was mm/lbf. Figure 4-5 shows the linear spring behavior of the EEF. Note that like in the virtual inertia experiment a dead band of 0.25 lbs is used in this application too. The equilibrium pose was chosen to be $\mathbf{u}^* = \begin{bmatrix} 0 & -944 & -7 & 0^c & \dfrac{\pi^c}{2} & \dfrac{\pi^c}{2} \end{bmatrix}^T$. The orientation is represented using fixed Euler angle notation. The position is measured in millimeters. Figure 4-6 represents the phase portrait of the EEF under the influence of restoring force due to controlling equation Eq.(4-4). Notice that the EEF settles down to equilibrium after it is disturbed by an external force.

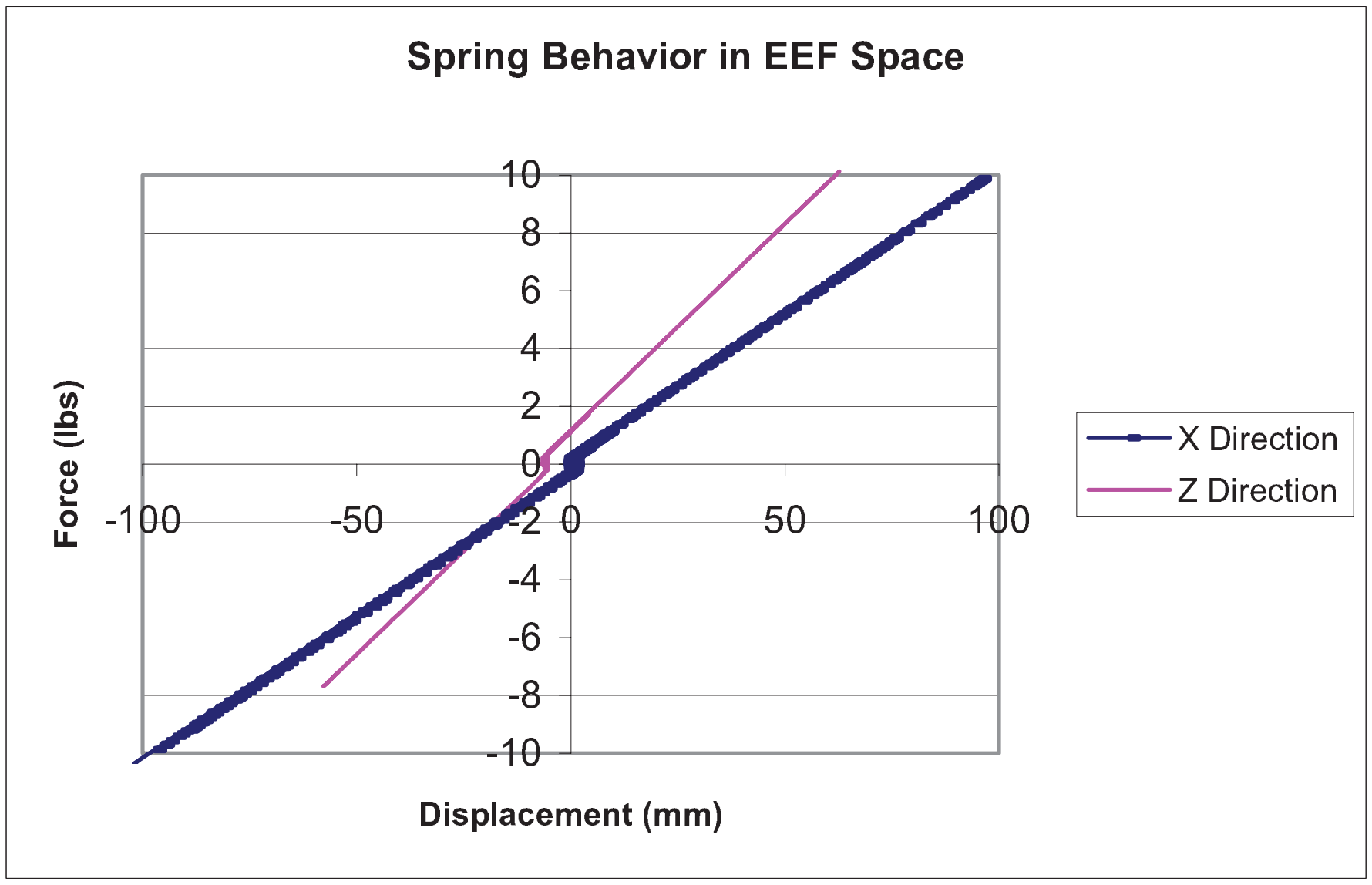

**Figure 4-5. Linear Spring Behavior of the EEF**

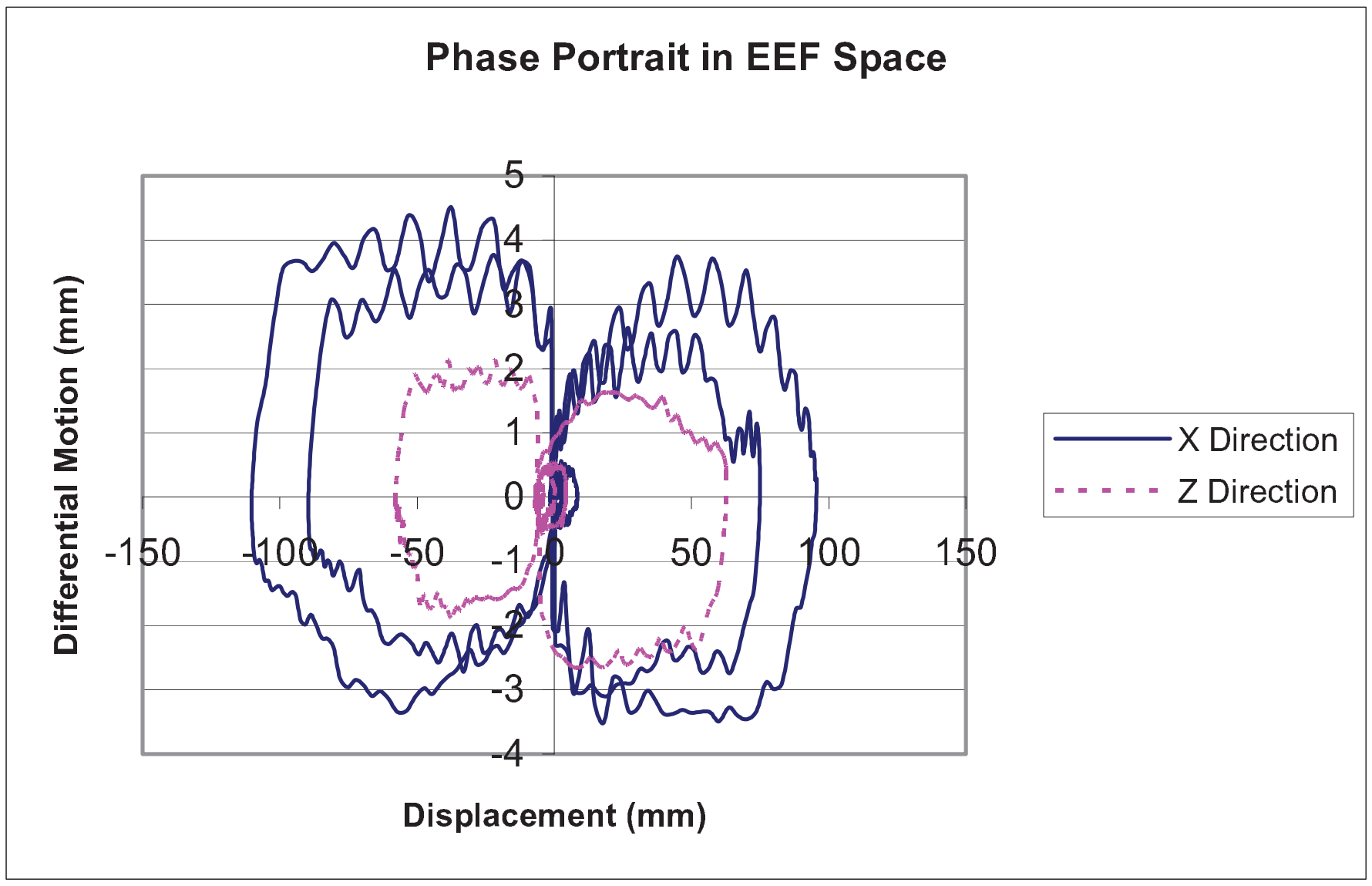

**Figure 4-6. Phase Portrait of EEF for Spring Response**

### 4.2.1 Application Domains

The experiments described in the Section-4.2 achieved the following objectives.

- Ability to command the EEF to execute an accommodative displacement in the direction of an applied external wrench confirms the transformation of the spatial force in the F/T sensor's local frame to the robot world frame.
- Spring, inertia and damper are basic mechanical elements that can describe the dynamics of a mechanical system. Ability to simulate these behaviors at the EEF enables the user to command a specific dynamic response at the EEF based on an application and environment.

As the tool is distal to the F/T sensor in a serial chain such as the PowerCube manipulator, the weight of the tool and the torque generated by it on the sensor have to be dynamically compensated for in a force-controlled application so that the sensed forces and torques are "true" external forces/torques experienced by the manipulator and are not modified due to the inertial characteristics of the tooling itself. This dynamic compensation is easier to achieve for forces than for torques. For this reason, please note that the above demonstrations have used only the EEF force information and disregarded the moments. Consequently, the orientation of the EEF is constant.

In the following sub-sections (4.2.1.1, 4.2.1.2, and 4.2.1.3) we describe some potential applications of the demonstrations reported above.

#### 4.2.1.1 Human Augmentation Technology (HAT)

*Human Augmentation Technology (HAT)* is motivated by the need for assistive devices that can alleviate the burden on a human operator while performing strenuous or hazardous tasks. A typical assistive mechanism achieves this objective by scaling (or transforming) the input motion or force thus significantly reducing the necessary effort or mental commitment from the user. An exoskeleton is a representative example of a HAT device.

The control scheme used for the virtual inertia demonstration may be used to transform input effort to an amplified differential motion so that the "perceived inertia" of the load is considerably less than its "actual inertia". This transformation may be accomplished using either a linear or a non-linear gain depending on the application.

#### 4.2.1.2 Virtual Fixtures

*Virtual Fixtures* (in the context of a robotic manipulator's workspace) may be defined as artificially generated constraints on the EEF motion. By restricting the EEF motion to a desirable sub-space of the actual reachable workspace, the precision requirement on the operator is reduced. For example, consider a surgical training system intended to guide a novice surgeon through a complicated trajectory with stringent precision requirements. If the motion of the input device is restricted to this desired trajectory, the trainee eventually develops the skill to follow it [Abbott, Hager, and Okamura, 2003]. Subsequently, the motion constraints may be gradually withdrawn so that the trainee acquires the requisite skill incrementally.

A virtual fixture plane was implemented on the Powercube manipulator. The procedure followed was to use the virtual inertia (or assisted motion) application and constrain the motion of the EEF on a plane irrespective of the direction of application of force. To define a desirable plane, $\Pi$, the user has the choice of identifying three points on this plane. Let us consider these points in 3D translational subspace, $\mathrm{T}$, of the EEF space are $\mathbf{P}_i = \begin{bmatrix} x_i & y_i & z_i \end{bmatrix}^T \ni i \in \{1,2,3\}$. Using these points on $\Pi$ we form two unit vectors, $\hat{\mathbf{u}}_1 = \dfrac{\mathbf{P}_2 - \mathbf{P}_1}{\|\mathbf{P}_2 - \mathbf{P}_1\|}$ and $\hat{\mathbf{u}}_2 = \dfrac{\mathbf{P}_3 - \mathbf{P}_1}{\|\mathbf{P}_3 - \mathbf{P}_1\|}$, that span this virtual fixture plane. We then stack these unit delta vectors in a matrix $\mathbf{A} \in \mathcal{R}^{3x2}$ as given below.

$$\mathbf{A} = \begin{bmatrix} \hat{\mathbf{u}}_1 & \hat{\mathbf{u}}_2 \end{bmatrix} \tag{4-6}$$

Using the $\mathbf{A}$ matrix, it is possible to develop a projection matrix $\mathbf{\Omega} \in \mathcal{R}^{3x3}$ that maps any force vector $\mathbf{F_T} \in \mathrm{T}$ to a projected force vector $\mathbf{F}_\Pi \in \Pi$.

$$\boldsymbol{\Omega} = \mathbf{A}\left[\mathbf{A}^T\mathbf{A}\right]^{-1}\mathbf{A}^T \qquad \textbf{(4-7)}$$

$$\mathbf{F}_{\Pi} = \left[\boldsymbol{\Omega}\right]\mathbf{F}_T \qquad \textbf{(4-8)}$$

The results from our virtual plane fixture experiments on the force control testbed are summarized in Figure 4-7 and Figure 4-8. Figure 4-7 shows the data points (EEF position coordinates) traced by the robot when subjected to the plane constraint. The virtual fixture is also shown in this graphic as meshed grid. Figure 4-8 displays the applied and projected forces that constrain the motion of the EEF to the virtual plane $\Pi$. This implementation helped in showing that a hard constraint may be placed on the EEF to facilitate guidance along a virtual fixture.

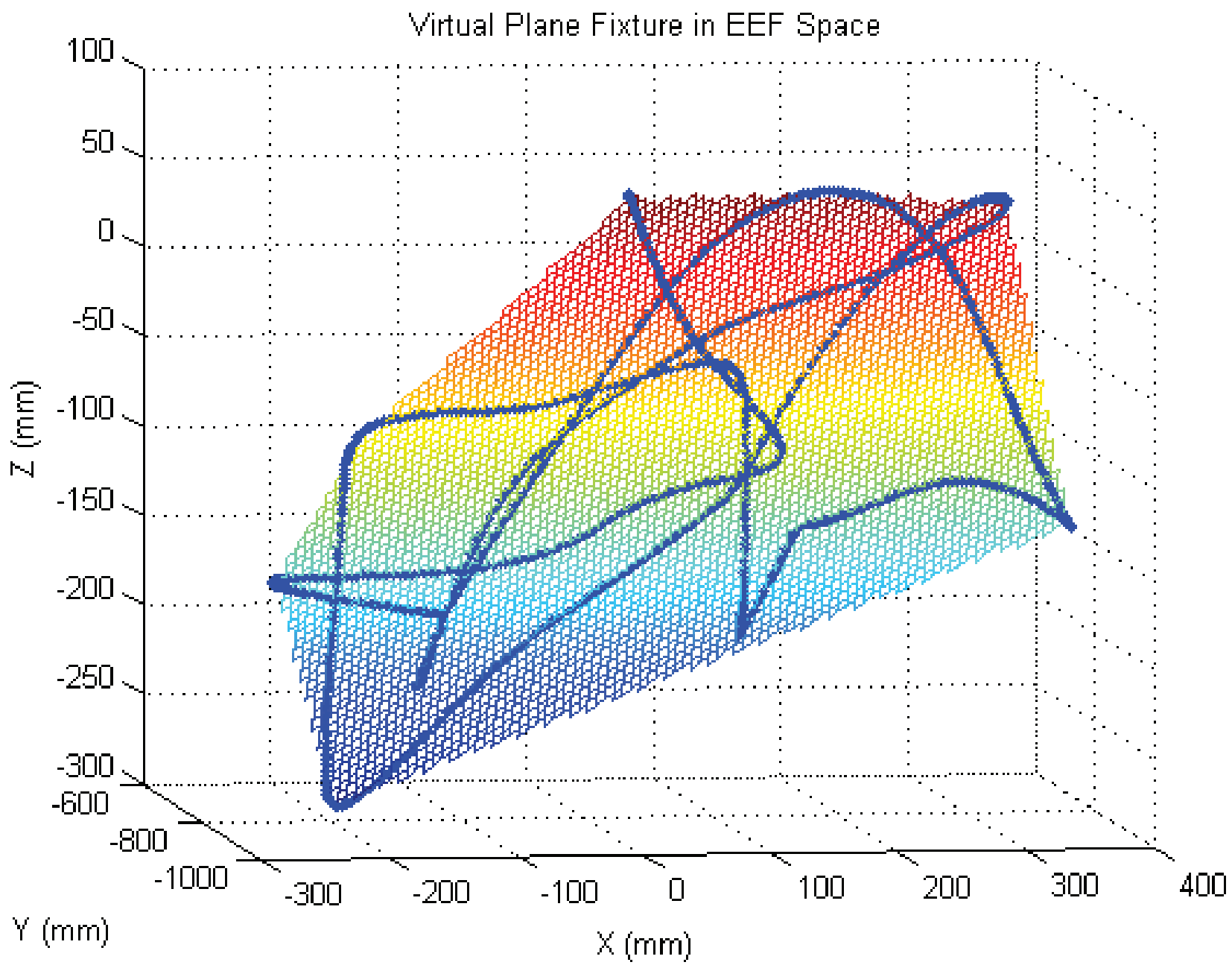

**Figure 4-7. Virtual Plane Fixture**

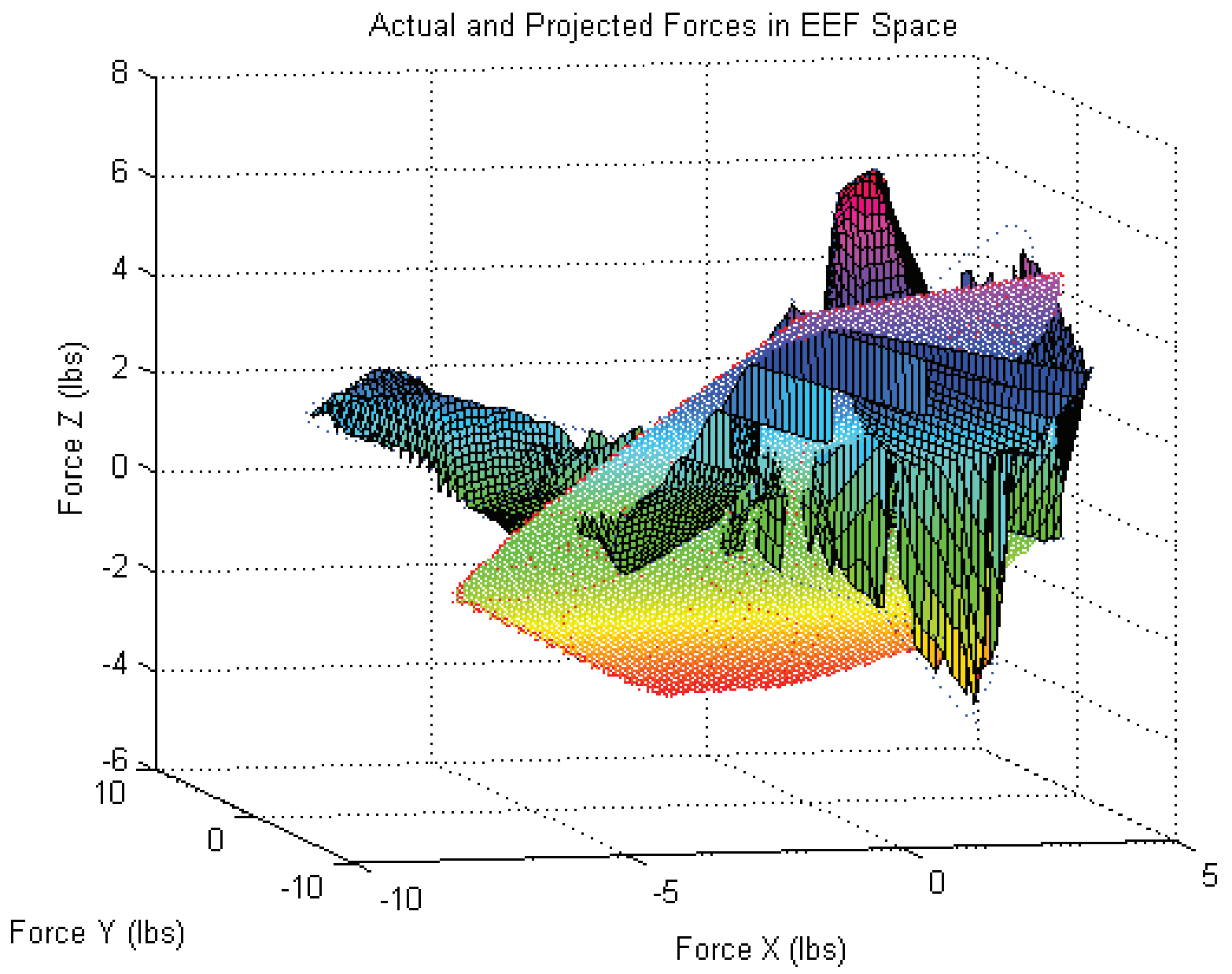

**Figure 4-8. Display of Applied Forces**

### 4.2.1.3 Teleoperation Input Device

The above demonstrations show that complete control on the motion and force (in other words *dynamic response*) at the EEF is achievable through appropriate control schemes based on position and force sensing. This capability could potentially be exploited to develop an input device for teleoperation (manual controller) with manageable dynamic response and geometry. This flexibility is a very desirable attribute for a manual controller. This thought is still under investigation and has not been put into practice as yet at UTRRG. Some foreseeable limiting factors to this proposition are listed below.

- Force/Torque Sensor Signal/Noise Ratio and Sensitivity
- Manipulator Vibrations and Tremors
- Achievable Control System Bandwidth

- Excessive Effective Inertia of the Manipulator: This is due to high gear ratios at the actuator. This is undesirable for *Human-Centric Robotics* [Zinn, Khatib and, Roth, 2004] where there is a close interaction between humans and robots.

Whee-kuk Kim and Tesar [1990] designed, analyzed, and implemented a 6-DOF force-reflecting spherical manipulator. In this research effort, the manual controller is similar to a Stewart platform with actuators located at the base to minimize dynamic effects. Kim and Tesar [1990] proposed an innovative concept called *power steering* and conducted experiments in force-feedback based teleoperation using a 3-DOF shoulder system. This was successful in that it compensated in real-time for the most important non-linear properties of the manual controller (mass, friction etc.) in order to make the device more transparent (by 90%) in force-feedback to the operator. The force gain was also adjustable so that the returning signal was much cleaner than it would have been otherwise. It may now be possible to design a cost-effective, sensor-based manual controller using appropriate torque dense actuators with very little stiction. Then power-steering would become feasible and easily implemented using OSCAR [Kapoor and Tesar, 1996].

## 4.3 Pure Force Control

In Section 4.2, we described some applications in which the force-controlled manipulator was used as a receptive device wherein the robot perceives an external force input and responds according to a user-defined control scheme. With this section, we initiate a discussion on specific control algorithms used for force-controlled tasks where the robot operates on a passive environment and transforms it, through dynamic interaction, in a manner dictated by the task requirements.

*Pure Force Control* may be defined as a control scheme where the error in force tracking is used as an important metric for decision-making. Force/torque sensory information is mandatory for the implementation of this scheme. Whitney [1977] proposed manipulator fine motion control using force feedback control and Van Brussel [1976] has worked on using such a technique for compliant assembly operations. Consider $\mathbf{W}_{\mathbf{ref}}(t) \in \mathcal{R}^6$ and $\mathbf{W}(t) \in \mathcal{R}^6$ to be the reference and actual wrenches at the end-effector for a given force-controlled application. The force tracking error $\mathbf{e}_{\mathbf{f}}(t) \in \mathcal{R}^6$ may be defined as:

$$\mathbf{e}_{\mathbf{f}} = \mathbf{W}_{\mathbf{ref}} - \mathbf{W} \tag{4-9}$$

Based on this error signal, a position-based control scheme may be developed as shown in Eq.(4-10).

$$\dot{\boldsymbol{\theta}} = \left[\mathbf{G}_{\boldsymbol{\theta}}^{\mathbf{u}}\right]^{-1} (\mathbf{K}_{\mathbf{P}}\mathbf{e}_{\mathbf{f}} + \mathbf{K}_{\mathbf{V}}\dot{\mathbf{e}}_{\mathbf{f}} + \mathbf{K}_{\mathbf{I}}\int_{t_i}^{t_f} \mathbf{e}_{\mathbf{f}}dt) \tag{4-10}$$

where $\left[\mathbf{G}_{\boldsymbol{\theta}}^{\mathbf{u}}\right]$ is the manipulator Jacobian, and $\mathbf{K}_{\mathbf{P}} \in \mathcal{R}^{6x6}$, $\mathbf{K}_{\mathbf{V}} \in \mathcal{R}^{6x6}$ and $\mathbf{K}_{\mathbf{I}} \in \mathcal{R}^{6x6}$ are the diagonal positive definite proportional, derivative and integral gain matrices. Note that this approach assumes linearity and is simplistic. However this method was chosen since the application was implemented only in 1-DOF and the approach velocity used was moderate.

The control scheme given in Eq.(4-10) was implemented in 1-DOF on the PowerCube modular manipulator using a given end-effector approach. The objective was to move the end-effector in the robot world z-direction so as to regulate a given force on a compliant surface of unknown stiffness. The surface of a weighing scale was used as a representative compliant surface. A compliant contact surface was used to reduce impulse excitation on contact and thus ensure safety. The values of proportional, derivative and integral gains chosen in the force-controlled direction (robot world z-direction) were 0.1, 0.1 and 0.01 respectively. The control bandwidth used was 15 Hz. This frequency was arrived at based on tuning and was observed to work best for the application.

### 4.3.1 Pure Force Control Results

The task was to regulate a force of 5 lbs on a compliant environment using an approach velocity of 2.25 mm/s. The results of this experiment are presented in Figure 4-9, Figure 4-10, and Figure 4-11. In Figure 4-9 we can see that the desired target force value is achieved in approximately 1.3 seconds. Contact with the environment occurs 4 seconds after the task execution begins.

Figure 4-10 plots the derivative of the force signal and shows the shaded contact zone in the time-history. The impulse experienced by the manipulator EEF at the instant of contact ($t = 4\,\text{s}$) is $|F.\Delta t| = (0.5\,\text{N})(0.75\,\text{s}) = 0.375\,\text{N-s}$.

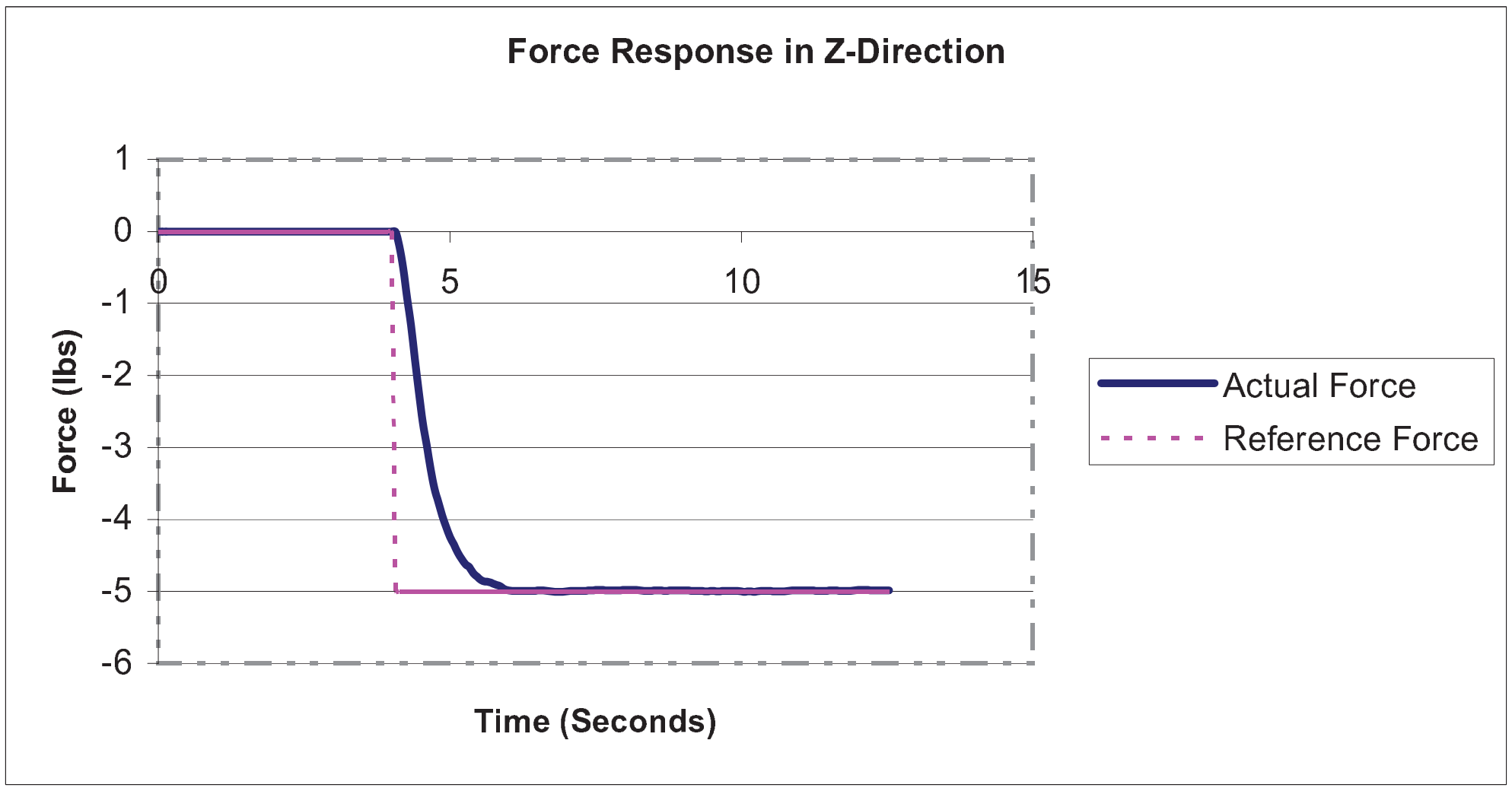

**Figure 4-9. Force Response in Pure PID Force Control**

This value is very sensitive to the approach velocity and surface compliance. Figure 4-11 shows the differential motion in the robot world Z-direction. The time-period of the control system is 0.067 seconds if the execution frequency is 15 Hz. Hence the approach velocity is $\left(\frac{0.15}{0.0667}\right) \approx 2.25 \frac{\text{mm}}{\text{s}}$.

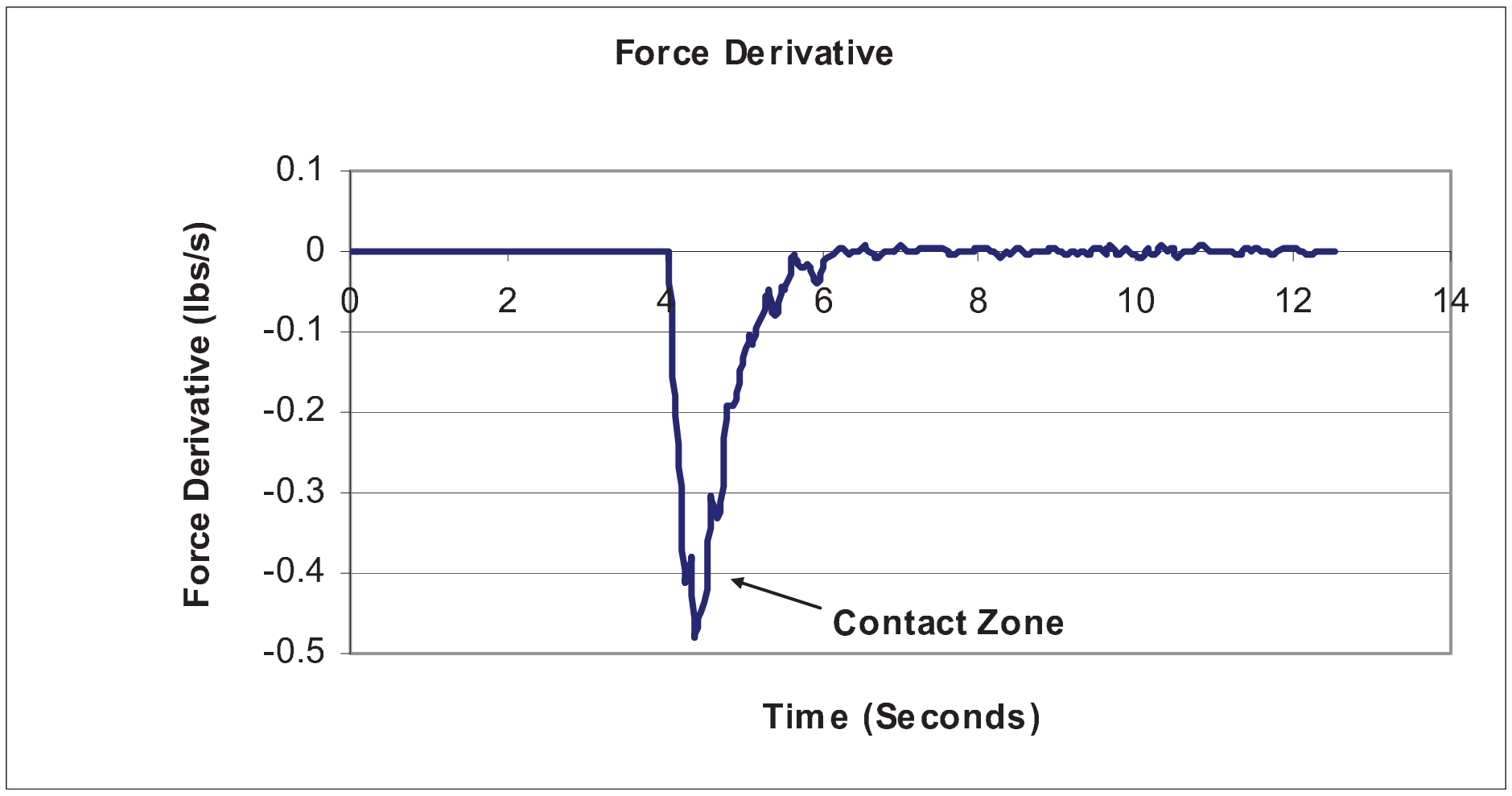

**Figure 4-10. Derivative of Force Signal during Contact Control**

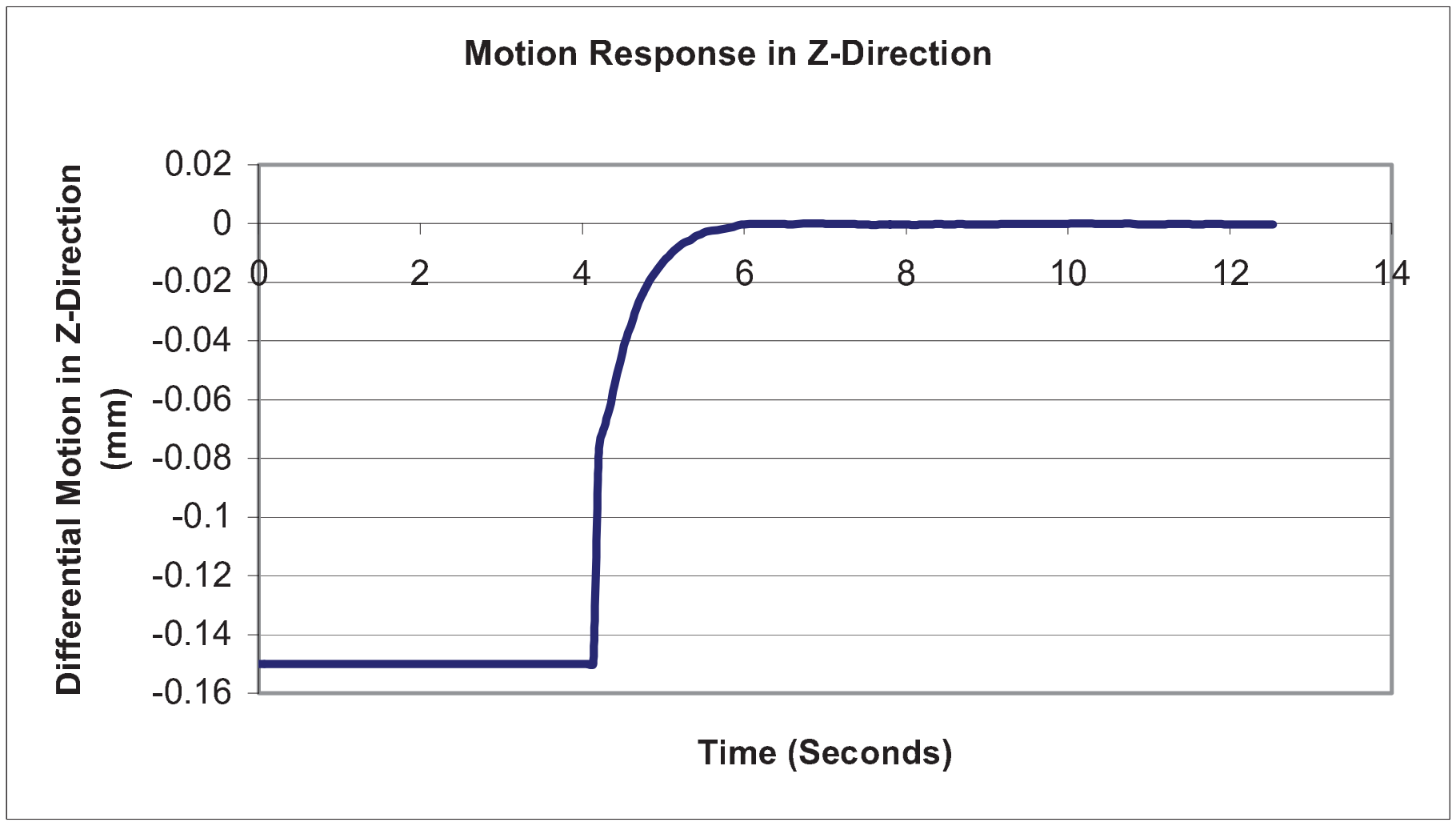

**Figure 4-11. Differential Motion Response**

### 4.3.2 Observations

A major demerit of the above mentioned pure force control scheme is that it requires the derivative of the force signal. The noise in the F/T sensor signal makes its derivative somewhat unacceptable. The approach velocity and environmental compliance used in this experiment are conservative yet typical. The experiments were carried out on a relatively rigid surface also. A vibration damper pad made of cork with hard rubber lining was used as a representative stiff surface. In this case, even a fairly moderate approach velocity (2.0mm/s) at a control bandwidth of 10Hz resulted in very large impulses ($|F.\Delta t| = (30.0\,\mathrm{N})(0.75\,\mathrm{s}) = 22.5\,\mathrm{N\text{-}s}$) as shown in Figure 4-12. This effort on stiff environments was not further pursued since such impulses were beyond the quasi-static force limits of the manipulator used for this work and thus detrimental.

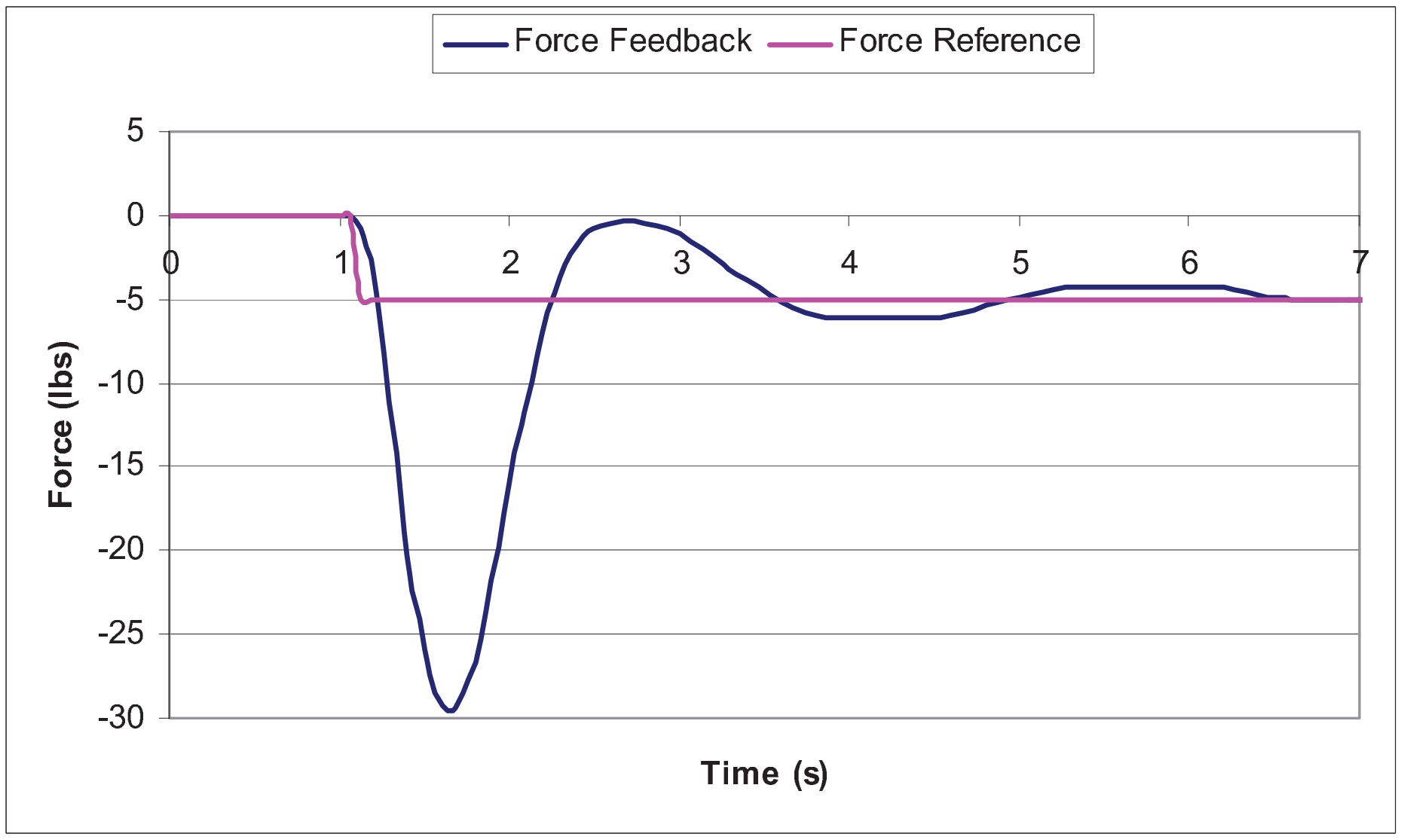

**Figure 4-12. Contact Force Response on a Stiff Surface**

Note that in this experiment there was effectively only one force constraint in the Z-direction. This is basic and was merely used as a test case for preliminary hands-on investigation rather than producing advanced results that could be used as a benchmark. A wrench constraint (in 6-DOF space) is more all-encompassing. **Force constraints in 3-DOF space (or more) are not only harder to deal with but also infinitely more demanding. We have yet to establish a foundation for this class of problem.**

## 4.4 Compliant Control

In this section we discuss results of experiments that were conducted on the PowerCube manipulator testbed using position-based compliant control. *Compliant Control* may be defined as a control scheme wherein the accommodative differential motion of the manipulator and the external force are proportional as given in Eq.(4-11).

$$\mathbf{\Delta\theta} = \left[\mathbf{G}_{\mathbf{\theta}}^{\mathbf{u}}\right]^{-1} \mathbf{K}_{\mathbf{P}}\mathbf{e}_{\mathbf{f}} \tag{4-11}$$

The above control algorithm is similar to the one developed for the assisted teleoperation application and was implemented on the PowerCube manipulator in 1-DOF (Z-direction in Cartesian space). The results of this implementation are presented in Figure 4-13 for various values of $K_P$. There is a trade-off between settling time and overshoot, as noticeable in the graphic below. For $K_P = 0.03$, the interaction force settles to the required value of 5 lbs in approximately 6 seconds but has an overshoot of approximately 14% of the reference force. This is not advisable for delicate operations. $K_P = 0.003$ facilitated safe achievement of the goal but at the cost of delayed settling time of more than 20 seconds. $K_P = 0.01$ seemed to work best for the application with no overshoot and settling time of 10 seconds. The above discussion shows that gain-scheduling in such algorithms is a tedious process.

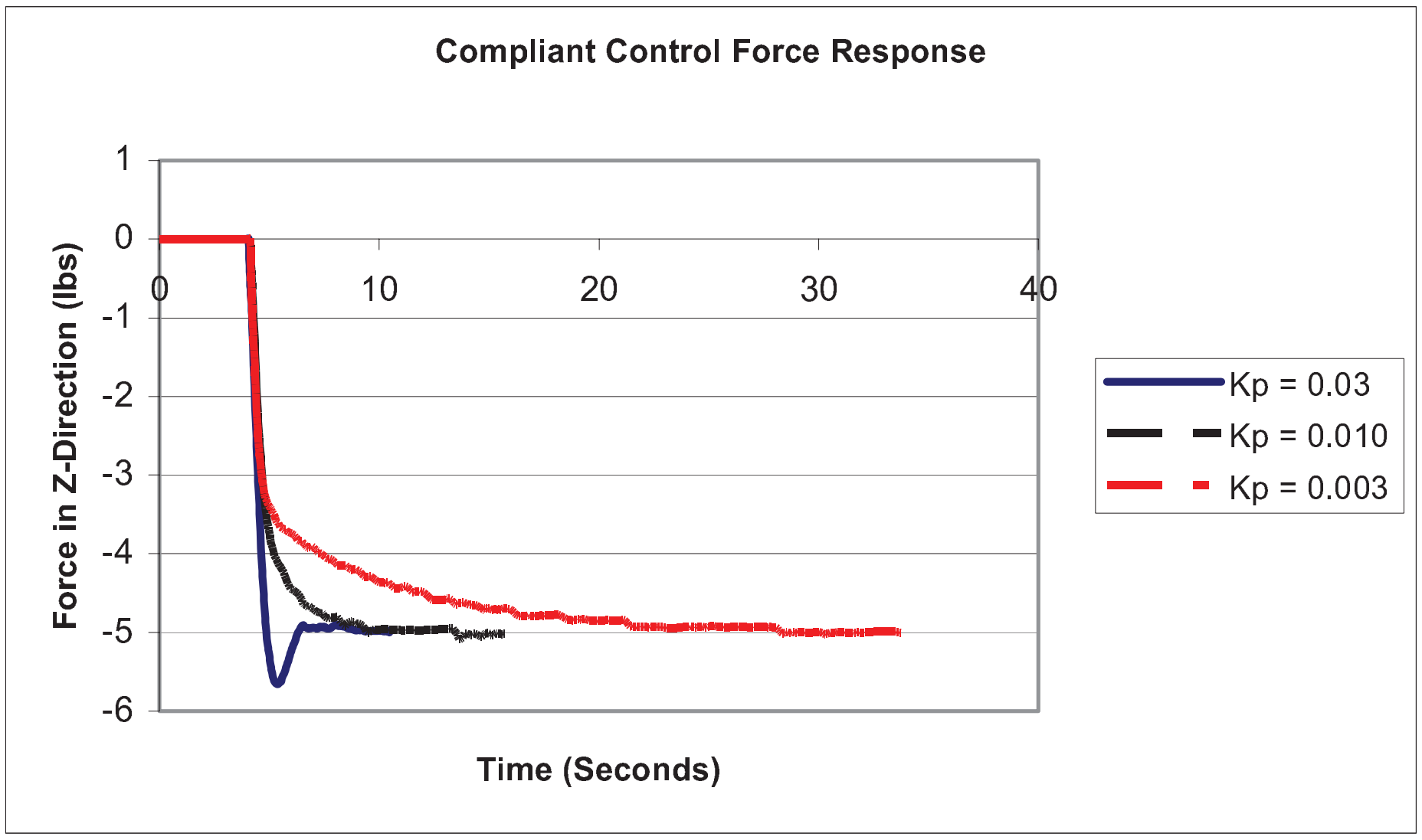

**Figure 4-13. Response using Compliance Control**

As can be noticed in the above discussion, tuning of the constants used in active control schemes is an arduous task before these systems are installed. An approach velocity of 2.5 mm/s was used for this compliant control application. Control bandwidth for this application was 15 Hz. The compliant (or stiffness) control algorithm described above is a very simplistic approach and thus results in very long settling times. The settling time can be improved with computational efficiency and tuning of the control parameters. However, this approach is unreliable and cannot be used as a basis for an expanded science.

## 4.5 Investigation of Force Tracking Performance

*Force regulation*, where the goal of the controlled system is to achieve a desired constant reference force, is a common and relatively simple application. On the contrary, *force tracking,* where the objective is to follow a pre-defined time-varying reference force profile, is hard to achieve. It is more so in the case of more than one (3DOF, 5DOF, or 6DOF) time-varying force constraint in task-space. This is especially true in the case when large uncertainties are present in the contact situation. In this section we investigate the performance of the above mentioned control laws, pure force control and compliant control, under force tracking.

The reference force trajectory used for judging tracking performance is the sinusoidal signal given in Eq.(4-12). $\mathrm{f}_{\max} = -3lbs$ and $\nu = \frac{1}{T} = \frac{1}{50} = 0.02Hz$ were, respectively, the amplitude and frequency of the sine wave. Note that the modulus is used to generate a purely negative sinusoidal signal. This was done deliberately because the F/T sensor measures only reactive forces and positive (active) forces cannot be measured when the manipulator is in contact with a surface which 'pushes' on it. Note that $\mathrm{t_c}$ is the cumulative time after contact is established.

$$\mathrm{f(t)} = \mathrm{f}_{\max} \left\| \sin\left( \frac{2\pi\, \mathrm{t_c}}{50} \right) \right\| \qquad \textbf{(4-12)}$$

The natural frequency of the contact interface is given in Eq.(4-13).

$$\hat{\nu} = \frac{1}{2\pi}\sqrt{\frac{k}{m}} = \frac{1}{2\pi}\sqrt{\frac{4391Nm^{-1}}{1.2kg}} = 9.623Hz \qquad \textbf{(4-13)}$$

Hence the ratio ($\mu_\nu$) of the sinusoidal signal frequency to the natural frequency of the contact interface is shown to be 0.207% in Eq.(4-14).

$$\mu_\nu = \frac{\nu}{\hat{\nu}} = \frac{0.02}{9.623} = 0.207\% \qquad \textbf{(4-14)}$$

Approach velocity and control bandwidth for this application were respectively 1.25 mm/s and 25 Hz. The performance of the pure force control strategy is shown in Figure 4-14.

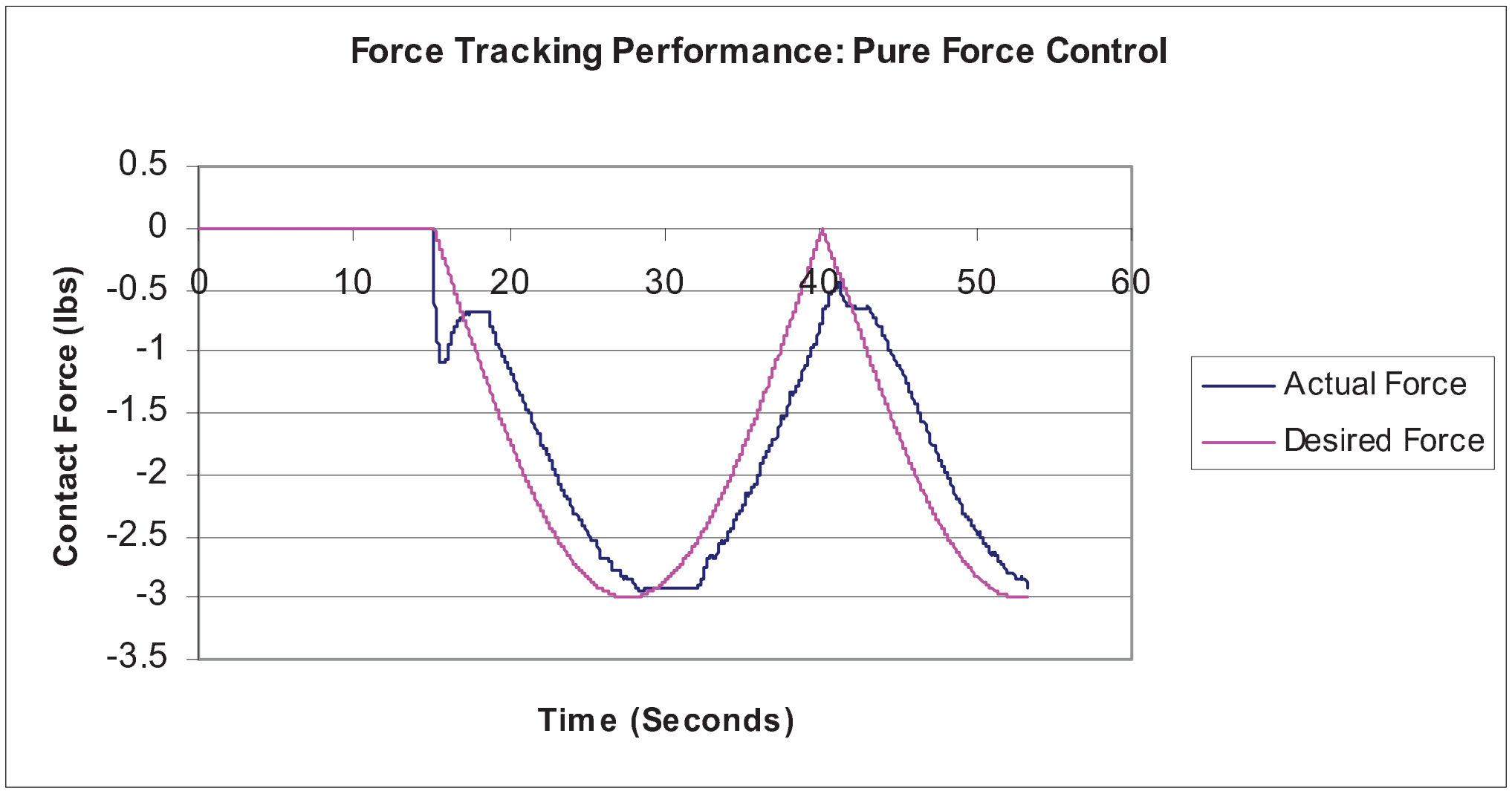

**Figure 4-14. Tracking Performance for Pure Force Control**

The experiment was recorded for around 50 seconds. Contact is established at the instant $t = 15.12\,s$. At this instant, the sinusoidal force reference is commanded. Impulse experienced by the EEF on contact is $|F.\Delta t| = (1.0\,\text{N})(2.0\,\text{s}) = 2.0\,\text{N-s}$ which is not detrimental. A constant time lag of 4% of the time-period is observed. Most force-controlled applications require that the manipulator stays in contact with the surface through the course of the operation. Hence a deadband width of 0.45 lbs was used to ensure that the manipulator does not lose contact with the surface inadvertently due to sensor noise. Hence the actual force signal is capped at -0.45 lbs as shown in Figure 4-14.

The tracking experiment described above was implemented using a compliant control scheme. The values of all parameters were maintained constant in the interest of standardization so that the two control algorithms can be fairly compared for tracking performance.

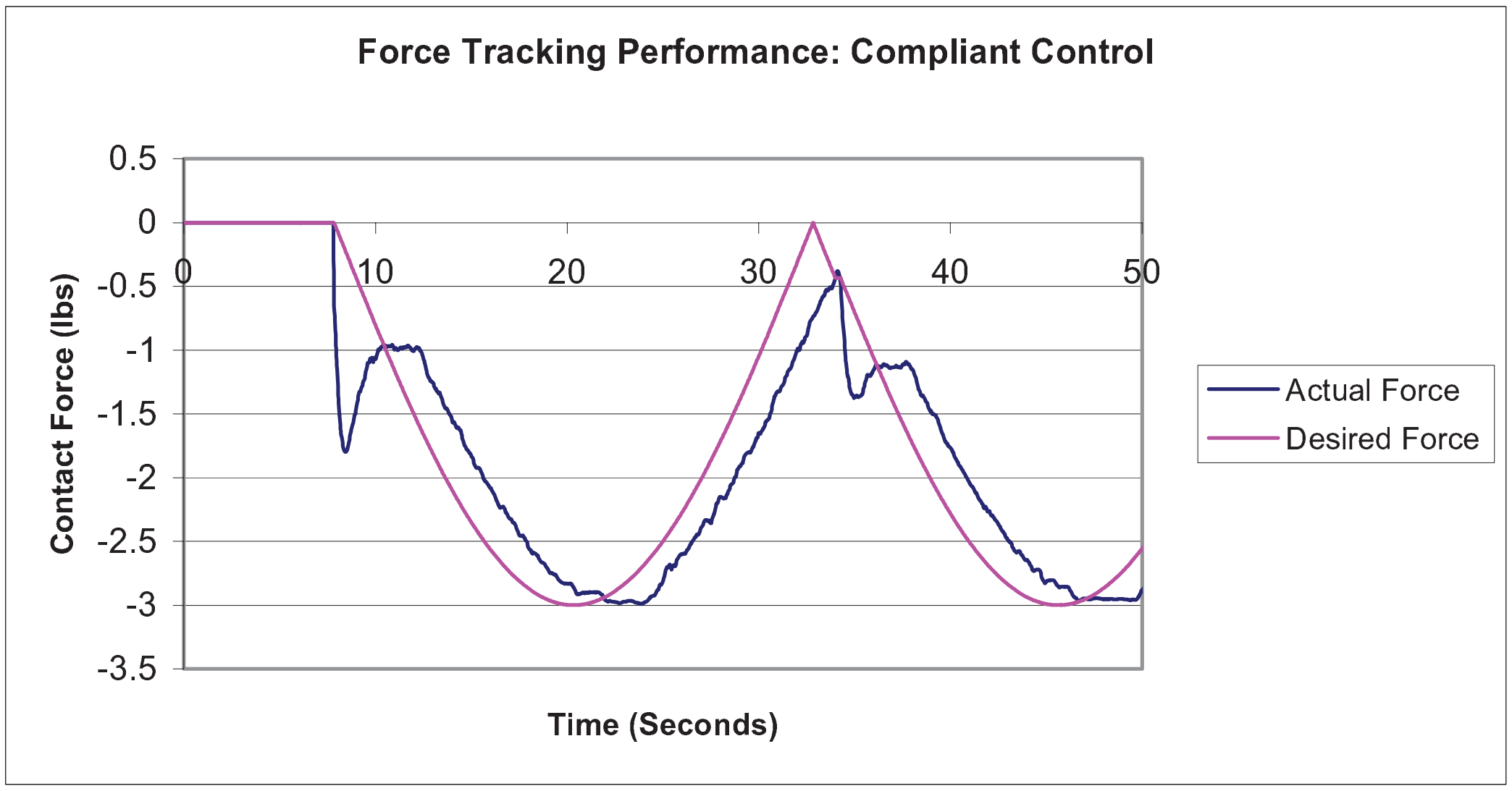

**Figure 4-15. Tracking Performance for Compliant Control**

The results from the force tracking experiment using compliant control are presented in Figure 4-15. Contact is established at $\mathrm{t} = 8.45\,\mathrm{s}$. On an average, the time lag during tracking was 4% of the time-period of the sinusoidal signal. The impulse experienced at contact, $|F.\Delta t| = (1.7\,\mathrm{N})(1.75\,\mathrm{s}) = 2.975\,\mathrm{N\text{-}s}$, is higher than that experienced under pure force control. This demonstration was conducted to illustrate the difficulty of force tracking applications and the relative ineptitude of the compliant control algorithm to successfully track a time-varying force command.

## 4.6 Effect of Actuator Parameters on Force Control

This section discusses the influence that actuator characteristics have on force and motion control performance in the Cartesian space. In geared transmissions, the reflected inertia at the robot system level is very high since the prime-mover inertias are scaled at the actuator output by the square of the gear ratios. Consequently in robots that use high gear transmission ratios, like the PowerCube manipulator that has harmonic drive transmissions, the operational inertia of the robot is extremely high and undesirable.

In the context of manipulator design, there has always been this argument about the advantages and disadvantages of geared transmission against direct-drive systems. For force control too, these transmissions have their own pros and cons. Cable driven systems are perceived to be direct drive arrangements. These mechanisms are clumsy and occupy a lot of space. Geared systems are relatively stiff and offer compactness which is a very desirable factor. Actuator transmission friction contributes additional dynamics and dissipates some energy. This is another argument in favor of direct drive mechanisms. In the current-torque mapping experiment for the PowerCube module (Section.3.1.1.1), a phenomenon called *stiction* (or static friction) was noticed. This is the minimum force/torque that an actuator should apply to produce useful motion at its output. Backlash or lost motion in geared transmissions is undesirable especially in space applications where gravity is negligible. Most of the manipulator compliance, or tendency to deflect under load, is contributed by its actuator transmissions. A force-controlled robot needs to be accurately calibrated to facilitate accurate motion control which leads to better force control performance. Hudgens and Tesar [1992] presented a broad treatment of the robot compliance parameter estimation problem and demonstrated its applicability to light machining robots. Experiments to test their improved linear quasi-static joint and link compliance model were conducted on a Cincinnati Milacron T3-776 industrial robot. They demonstrated reliable static compliance measurement and estimation through experimental data. Sklar and Tesar [1988] developed software in VAX APL v2.0 for robot metrology and calibration techniques. An effort to calibrate the PowerCube manipulator using an Indoor Global Position System (iGPS) is underway at the UTRRG metrology laboratory [Kang, Pryor and Tesar, 2004]. This research will be beneficial only if completely nonlinear models for bearing friction and compliance are determined [Brandlein et al., 1999].

Most of the issues outlined in this section are active topics of research within UTRRG. Currently, there are researchers working on developing nonlinear performance maps for gear trains, bearings, and prime-movers.

## 4.7 Force Control Implementation Issues

During the course of implementing the force control experiments described in this chapter, we came across some issues that are listed in this section.

- *Frame Selection at Contact Interface*: It is necessary to choose a comfortable frame of reference to implement control algorithms. For 6-DOF force control the task frame framework suggested by DeSchutter and Van Brussel [1988] is almost generic (there are some applications like incompatible seam following that cannot be represented by this formalism) and very helpful. The task frame is similar to the Frenet frame used for motion planning purposes [Wu and Jou, 1989].
- *Reference Force Profile*: Task planning for contact tasks involves both motion planning and interaction force planning. Together, these constraints may be referred to as *process planning*. Chang and Tesar [2004] have suggested a task based performance map that can be used for operational purposes (See Figure 4-16).

It is advisable to have an application independent process plan. However, currently, determination of the desired force trajectory is application dependent. Reference force trajectories are frequently based on experience. In Eq.(4-15), we present an example from [Chan and Liaw, 1996]. This is the desired force profile (in Newtons) for the insertion of Printed Circuit Boards (PCBs) into an edge-connector socket with spring contacts using a Selective Compliance Assembly Robot Arm (SCARA).

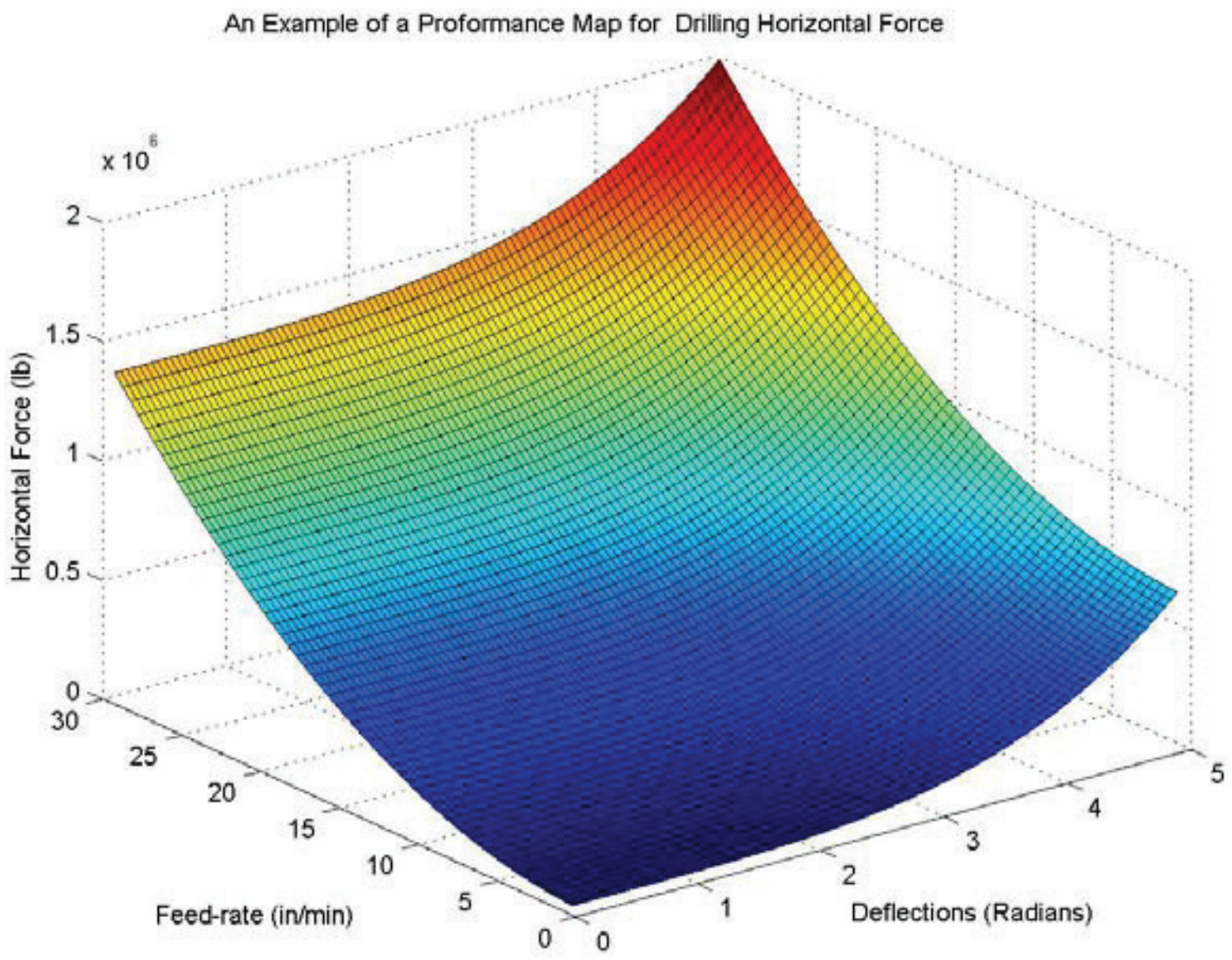

**Figure 4-16. Process Performance Map for Horizontal Force in Drilling**

$$f_d(t) = \begin{cases} 0.0 & : 0 \le t < 1.23s \\ 5789.63(t-1.23)^3 - 21994.91(t-1.23)^4 + 25041.65(t-1.23)^5 & : 1.23s \le t < 1.485s \\ 30 + 200(t-1.485) + 15644.23(t-1.485)^3 - 147313.65(t-1.485)^4 + 329844.99(t-1.485)^5 & : 1.485s \le t < 1.68s \\ 65 - 63443.07(t-1.68)^3 + 528692.27(t-1.68)^4 + 1174871.72(t-1.68)^5 & : 1.68s \le t < 1.86s \\ 28 & : 1.86s \le t \le 2.22s \end{cases} \quad \textbf{(4-15)}$$

The procedure they used for devising this trajectory was to jog the robot through three prescribed assembly phases, namely *approach, insert, and* *set-down*, record the forces in each phase, and introduce quintic polynomial segments to approximate the measured forces.

- *Configuration Space Selection*: A configuration space for a mechanism may be defined as a vector space of position co-ordinates that define the state of the mechanism. The two commonly used configuration spaces for a robot are the joint space $\mathbf{\Theta} \equiv \left\{ \begin{pmatrix} \theta_1 & \theta_2 & \ldots\ldots & \theta_N \end{pmatrix}^T : N = DOF \right\}$ and the EEF space $\mathbf{U} \equiv \left\{ \begin{pmatrix} x & y & z & \theta_x & \theta_y & \theta_z \end{pmatrix}^T \right\}$. The vector spaces $\mathbf{\Theta}$ and $\mathbf{U}$ are respectively the input and output spaces of the robot. The EEF space is also known as Cartesian Space (CS) or Operational Space (OS). In the experiments described in this chapter the EEF space was chosen for implementing the control algorithm. This is primarily because the force sensor information was transformed to this space. End-of-arm force control could still be used in joint space if the peripheral F/T sensory information is transformed to the joint space. Alternatively, joint space torque control can be implemented. In this method, the torque (or current) variation at an actuator is used for commanding accommodative motions in the joints. The disadvantage of this method is that end-point forces and torques cannot be accurately determined based only on the sensory information of the actuator. This is because of the many transformations and unmodeled effects (such as compliance) between joint space and Cartesian space. Hence, we recommend that for any force control implementation, end-point force sensing is necessary.
- *Gain Scheduling*: The tuning of gains/weights used in active force control is a time-consuming and undesirable process. In most cases, the control algorithms assume linearity which simplifies the system dynamics to unrealistic extents. Linear control and gain tuning for such methods are based on relatively ambiguous parameters and thus have their own limitations. Control and operation of systems should be based on a comprehensive approach such as intelligent control based on performance maps.

- *Contact Friction*: In real contact situations, there are different kinds of forces involved, viz. reaction force due to 'pushing' or 'pulling' on the environment, restoring force due to the compliance of the EEF-environment interface, and frictional forces due to interaction between the manipulator and the environment. Of the above forces, friction is hardest to model due to the inherent non-linearity of this phenomenon. Most force control algorithms simplify the interaction as a point contact on a plane surface. This simplifying assumption facilitates in partitioning the control space into orthogonal force-controlled and motion-controlled directions. However in real-life, contact situations are more complex and most control is a mixture of force and motion (or dynamic response in other words). We recommend here that a force control approach should include a task model (or map) that considers friction and other types of interaction forces in addition to the first order static reaction forces arising from contact.
- *Approach Velocity*: Interaction tasks are characterized by four phases: (i) unconstrained motion to approach the surface of interest (ii) contact transition (iii) controlled contact under force control to execute desired function (iv) transition to free space motion after task execution. This succession of events is depicted in the state diagram below (Figure 4-17). The velocity of approach in phase(i) influences the impulse experienced during phase(ii). This in turn affects the impulsive torques at the robot joints. Hence it is necessary to choose an approach velocity that is reasonably large and yet conservative to evade part/manipulator damage.

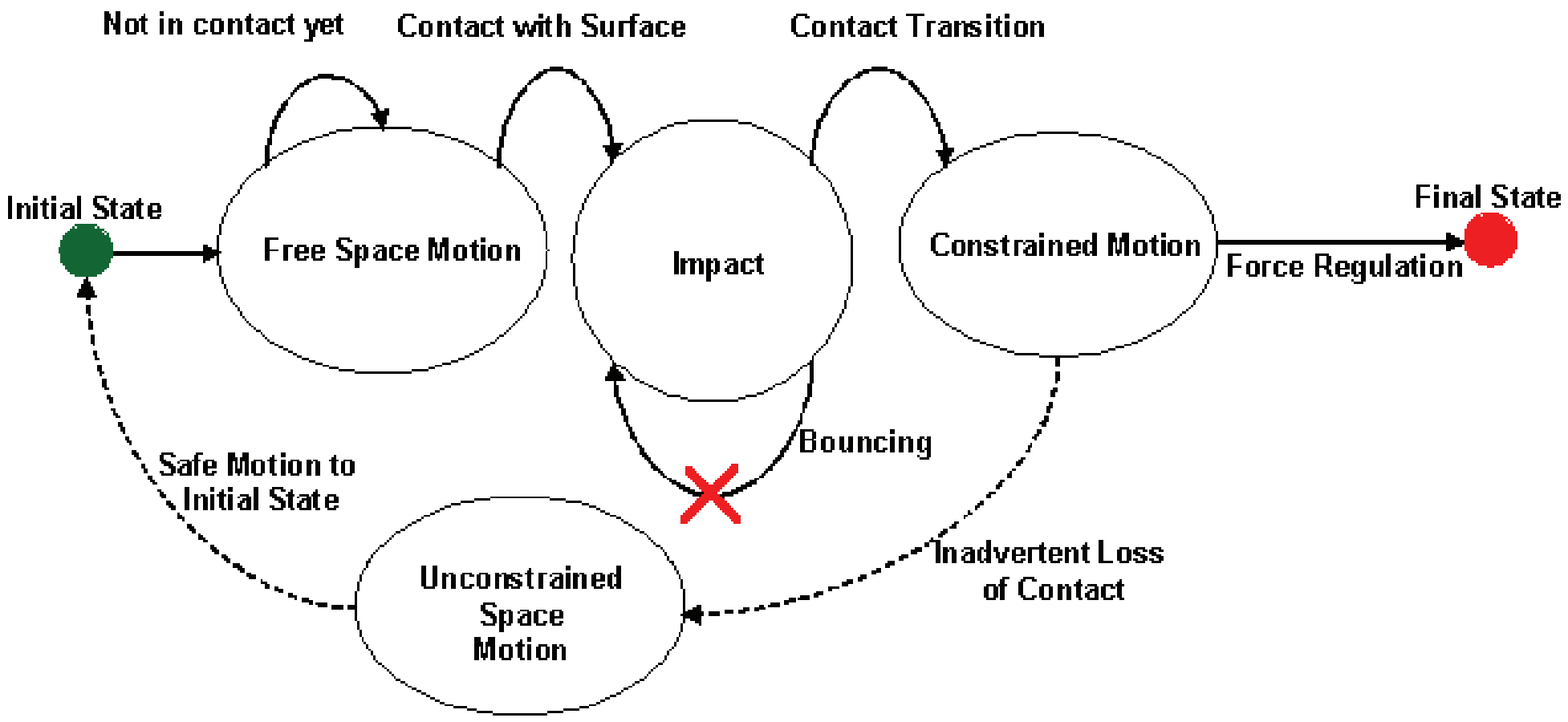

**Figure 4-17. State Diagram for Contact Task Execution**

This sensitivity of contact force response on approach velocity demonstrates the significance of motion planning for force control. At UTRRG, the approach we follow for producing smooth trajectories is based on the theory of algebraic curves and their geometric and dynamic properties [March and Tesar, 2004].

- *Environment/Manipulator Compliance*: This is a factor that directly affects the restoring force that acts on the EEF on contact. The safety of the force-controlled operation improves with greater interface compliance, but at the cost of positional accuracy. We recommend that the operator should have complete control over the interface compliance irrespective of the mechanical design of the system. The virtual spring experiment in this chapter demonstrates that it is feasible to program the stiffness of a manipulator in different directions of the task space.

- *Human Input Tremors*: In the case where the manipulator is used as a receptive input device, operator tremors might induce vibrations of the EEF. Although tremors are invisibly small, these jitters may be amplified by the closed loop gains to create noticeable vibrations of the manipulator EEF. This may be minimized by implementing a filter on the force/torque signal. This reduces the effect of both sensor noise and input tremors. In addition, it also makes the manipulator less sensitive to impulses (which contain many frequencies). Kim and Tesar [1990] showed that the operator tremor is more intense in the case of hard grip. For a 1-DOF force control application using a force feedback manual controller, they used a low-pass filter with a cut-off frequency of 5 Hz to stabilize human jittering.
- *System Natural Frequency*: In a force control application, especially compliant control using a passive compliant element or active control, an effort has to be made to keep the frequency of the active control dynamics as far away as possible from the natural frequency of the manipulator-environment contact interface. This issue is frequently not a major concern however it becomes more relevant in the case of very compliant contact surfaces. If the contact interface can be modeled as a *spring-mass-damper* system ($M\ddot{x} + C\dot{x} + Kx = F(t)$), then $M\ddot{x} \cong Kx$ is an undesirable condition leading to an exchange of equal amounts of kinetic and potential energy in an uncontrolled oscillation. Tesar and Tosunoglu [1992] showed that to judge this condition, the ratio $\frac{F^i}{K}$ may be used, where $F^i$ is the inertia force and $K$ is the interface stiffness. Behi and Tesar [1991] determined the dynamic and compliance properties of an industrial manipulator based on experimental modal analysis with the goal of predicting the lowest natural frequency in real-time.

## 4.8 Chapter Summary

This chapter presented the results from some elementary force control experiments conducted on a modular robot testbed at the UTRRG robotics lab. As this manipulator cannot yet be controlled in dynamic mode, we used position-based control methods. The demonstrations in this chapter, although simplistic, raised some issues related to force control implementations and effect of actuator characteristics on force control performance. In addition to this, they contributed to the capability of the lab. Two cases were considered. In the first case (Case A) the manipulator was used as a receptive device that reacts to an input end-point force according to a pre-defined control algorithm. In the second case (Case B) the manipulator was used as an active device to execute a force-controlled task on a surface of arbitrary stiffness.

In Case A, the behaviors of spring and inertia-damper elements were emulated at the EEF using a delta controller. The motivation for such applications was demonstrated through assisted motion and virtual fixture implementations. The possibility of using a powered manipulator as a teleoperation input device was also addressed under this section. In Case B, the active force control algorithms considered were pure force control (which uses the error signal between desired force and actual force to command an accommodative motion) and compliant control (which defines the interaction between the EEF and the contact surface as a compliance). The force-tracking performance of the above active control schemes was also investigated. A description of the effect of actuator parameters on force control performance of the manipulator was then presented. The chapter concludes with an analysis of implementation issues of force control applications.

## Chapter 5

# Investigation of Multi-Input Control At the Actuator Level

Actuators that populate a complex system are usually taken for granted by system-level control developers. However, significant performance improvements at the component level can enhance the capabilities of the overall system. The approach at UT Austin has been to maximize the number of choices within the actuator to enhance its intelligence. This includes layered control (multiple physical scales), dual-level control for fault tolerance (force or velocity summing), and a mixture of force and motion control. This chapter primarily explores Force/Motion Control (FMC) as a means of establishing this performance improvement in the force/motion domain. The goal of FMC is to facilitate enhanced capability for complex functional processes at the output of the system level like die-finishing, deburring, force-fit assembly, and fixturing. We propose here that this can be accomplished by expanding the performance envelope of the actuator by providing two distinct inputs, one in the force domain and another in the motion (velocity) domain. These inputs should be in parallel and have a relative scale change of (10:1) - (15:1) so that their influence on each other is minimal (i.e, they should remain as independent as possible). The concept of FMC is woven around the Force/Motion Actuator (FMA), conceptually described and simulated in this chapter. To achieve an improvement in overall performance of this actuator system, it should be able to use the force and motion sub-systems "intelligently" in real-time and in the presence of external disturbances.

The chapter is structured such that the background and motivation for multi-domain inputs are presented at the outset. The concept of FMC is described. The conceptual design and analysis of the FMA is presented. In this analysis, a dynamic model of the FMA is developed and results of some preliminary simulations are presented. The chapter concludes with a note on two issues that need to be investigated about the FMA, viz., the performance envelope and performance criteria.

## 5.1 Background and Motivation

The overarching philosophy of multi-input actuation is to let different actuator properties from more than one input have independent pathways to the output. The presence of these redundant inputs results in the freedom to mix these properties, based on a criteria-based decision making system, to achieve complex process control at the output.

### 5.1.1 Pure Force Control and Pure Motion Control

Pure force control may be construed as the management of force levels in the system without the intent of creating motion (over very short time periods). In such an application, the transitional elements between the force input and output should be rigid so that the force flow is achieved without position pose definition losses due to deformation. Rigidity of the transfer medium is also important to ensure the quickest possible response.

Pure motion control, on the other hand, is a control scheme wherein the output motion is most important and the forces are secondary. Considering representative application scenarios, a catapult system requires pure velocity (motion) control whereas a fixturing system (where you are trying to hold a position against a force) needs to be purely force controlled. A window cleaning application involves both force and motion control (in this case, these are orthogonal).

### 5.1.2 Multi-Input Intelligent Actuation

Embedding equal or distinct prime-movers inside the same actuator has many advantages. Such actuators are called *multi-input actuators*. If the component prime-movers in these actuators have influences in the same domain (motion or force), they can be used to assemble fault-tolerant or layered control systems. With subsystems in different domains (force and motion), a physical embodiment for a combined force/motion actuator can be achieved. The challenge in such multi-input actuators is to prioritize inputs based on output requirements and a set of criteria related to the mixing of performance levels from the constituent sub-systems.

#### 5.1.2.1 Fault Tolerance

Fault-Tolerant systems have more inputs (dual and equal) than outputs so that there is no single-point failure in the system during operation. By effectively using condition-based maintenance, health margins of the dual sides of these actuators may be monitored in real-time and the output requirements may be partitioned among the healthy actuators in the event of partial or complete failure.

*Fault-tolerant* actuators have two prime-movers in series or in parallel. The two equal sub-systems are in the same domain. They can be either force summing or velocity (motion) summing. There is a complete duality between equal force systems and equal motion systems. Velocity summing (in series) means the forces are incidental to the motion (See Figure 5-1). Force summing (in parallel) means that the velocities are secondary (See Figure 5-2).

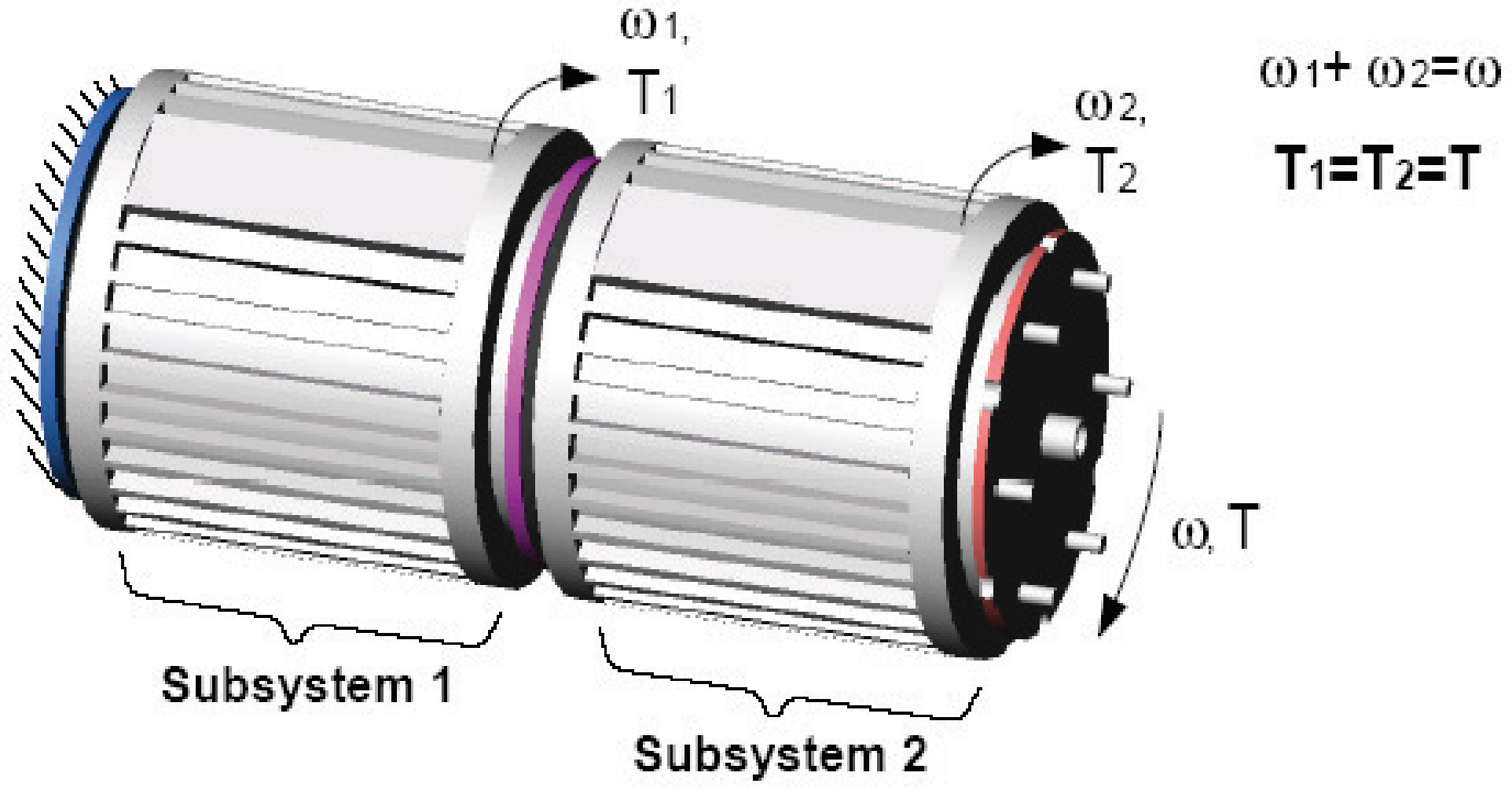

**Figure 5-1. Concept of a Velocity Summing Fault-Tolerant Rotary Actuator**

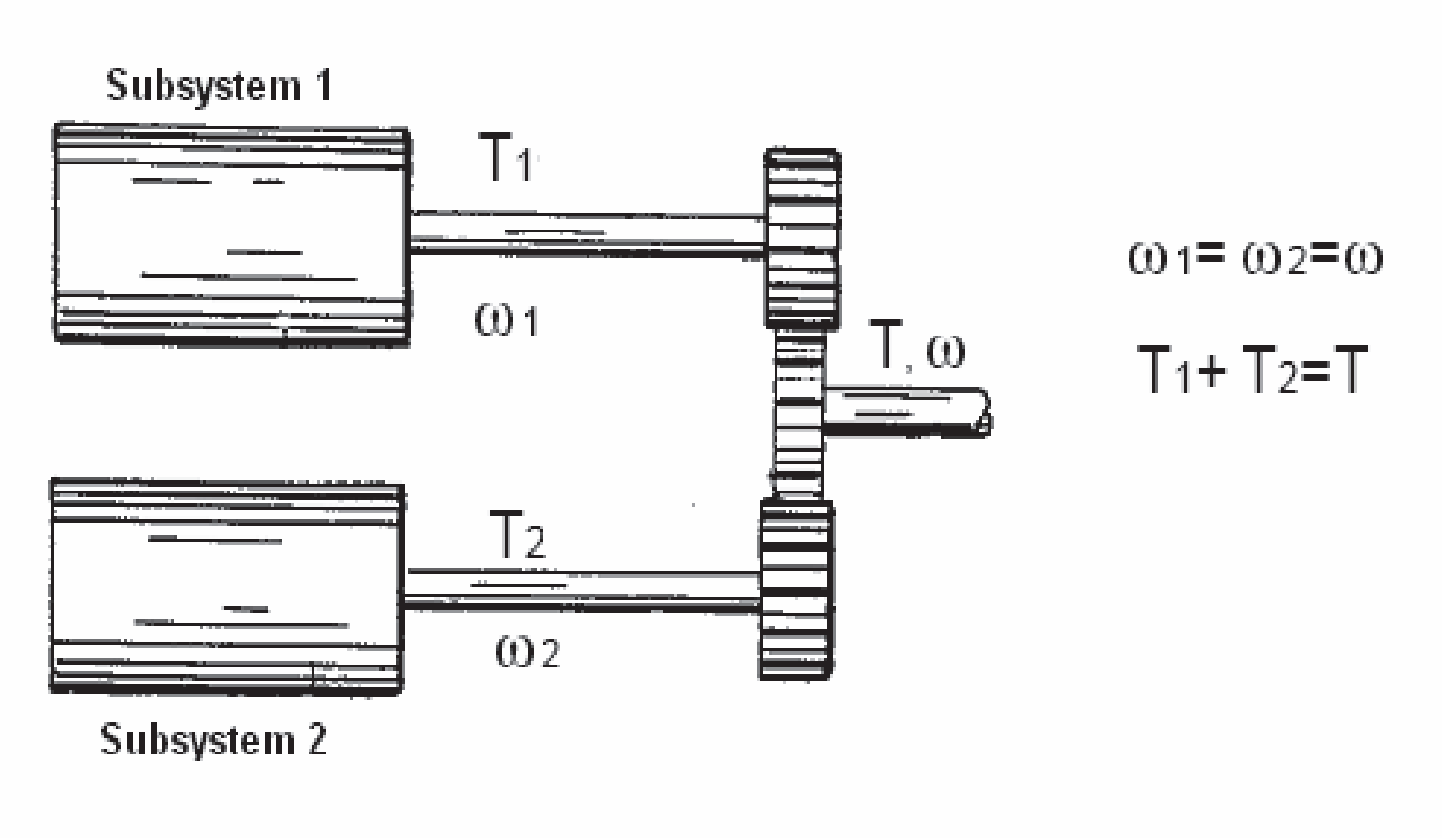

**Figure 5-2. Concept of a Force Summing Fault Tolerant Rotary Actuator**

#### 5.1.2.2 Layered Control Using Uni-Domain Hybrid Actuators

*Layered control* is achieved when you have a combination of two motion control actuators having a mixture of motion scales in series on the same output. The concept of layered control is illustrated in Figure 5-3. The slowly moving curve is the desired output motion of the actuator system. This motion is usually at a relatively lower frequency (<10 Hz) and contains predictable or relatively easily measured operational forces. A large scale electromechanical actuator subsystem configured in a compact design using exceptional component technologies is used to meet the requirements of this relatively slow motion.

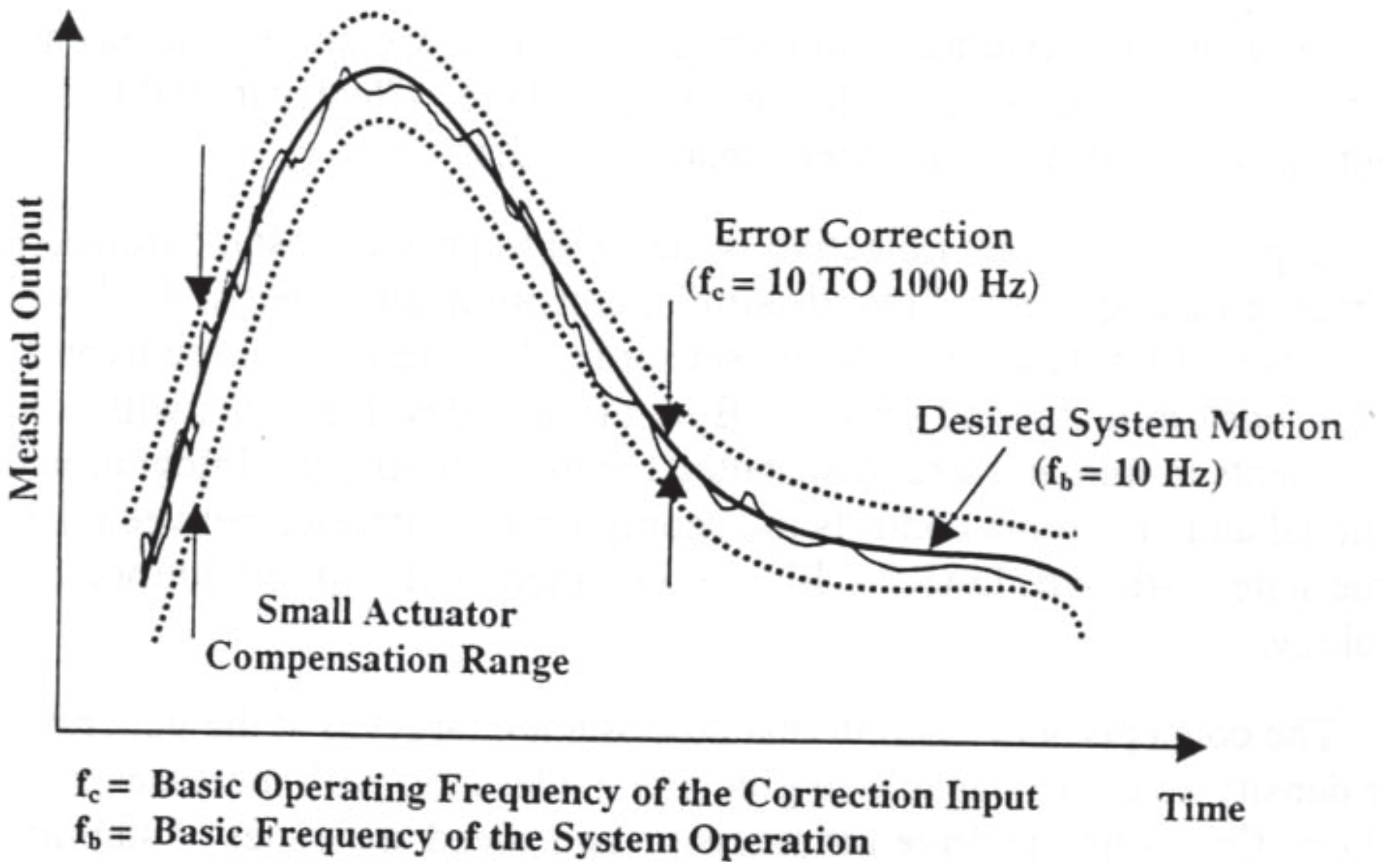

**Figure 5-3. Conceptual Diagram for Layered Control [Tesar, 1999b]**

Superimposed on this desired motion is a higher frequency (10-1000 Hz), relatively small error generated by self-induced and external disturbances that are difficult to measure. Since the small scale motion is typically less than one percent of the magnitude of the larger desired motion and occurs at relatively high speed, an inherently nonlinear and sluggish large scale system cannot adequately eliminate the external disturbances. Consequently, another small-scale subsystem, created using smart materials, may be used to remove these higher frequency external disturbances. In doing so, resorting to linear control methods for the small actuator controller is warranted. The basis for linear control-in-the-small is that the system first-order kinematic influence coefficients (or g-functions) are effectively constant because the small-scale system operates in a very small time scale while the large system has changed very little. In other words, if the controlling equation of this system was temporarily expressible in the form $M\ddot{x} + C\dot{x} + Kx = F(t)$, then $M, C, K$ may momentarily be considered constants.

The advantages of using such a layered pair of actuators over a single-scale actuator system are presented below [Tesar, 1999b]:

- Superior precision and accuracy over the full operating range
- Compactness and light weight for increased power density
- Increased overall operational speed and frequency response
- Greatly enhanced stability due to high frequency disturbance rejection
- Expanded range of motion along with increased stiffness and load capacity

With simultaneous integration of performance-criteria, sensing, accurately modeled parameters, and decision-making that is integrated over both scales, a number of application requirements can be met. These applications involve a wide parameter space in speed, load capacity, accuracy, and disturbance rejection. Some representative application domains for this concept are the airborne laser system, strategic missile nozzle actuation, the "more electric" aircraft, and active control of launch systems. There are also several high value manufacturing processes, like airframe manufacturing and auto panel stamping die-finishing, that can be impacted by multi-scale intelligent hybrid actuation [Tesar, 1999b].

#### 5.1.2.3 Force/Motion Control Using Multi-Domain Hybrid Actuators

Most forms of control are combinations of output parameters such as position, velocity, force, friction, and energy/power [Tesar, 2003b]. Force/Motion control maybe defined as the management of force and motion on a given output independently based on a set of force and motion criteria. The objective of Force/Motion Control (FMC) is to expand the choice of force and motion modifications at the output of a nonlinear mechanical system by embedding independent force and motion generators at the actuator level.

A force generator has to have a fast response and thus change force-levels quickly. This force is generated independent of the position. A force-actuator is backdrivable (mechanically soft) and requires a force/torque sensor for feedback on its force/torque levels. Its location in space is of lesser importance and thus it does not need an accurate position or velocity sensor. A torque sensor will however be useful for knowing the dynamic/inertia force-level.

On the contrary, a motion generator produces motion independent of force-level and is relatively stiff since it is not backdrivable. It requires accurate position/velocity sensing and does not need a torque sensor. Force/Torque sensing for such an actuator is only useful for disturbance rejection. Table 5-1 summarizes the characteristics of force and motion sub-systems in a multi-domain hybrid actuator.

**Table 5-1. General Characteristics of Force and Motion Sub-Systems**

| | Primary Domain | Representative Gear Ratio | Backdriveability | Sensors Required |
|---|---|---|---|---|
| **Ideal Force Generator** | Force | 15:1 | Backdrivable (Mechanically Soft) | Torque Sensor, Acceleration Sensor |
| **Ideal Velocity Generator** | Motion | 200:1 | Non-Backdrivable (Mechanically Rigid) | Position and Velocity Sensors |

## 5.2 FMA: Goal and Concept

The objective of designing a Force/Motion Actuator (FMA) is to permit independent properties from constituent sub-systems to flow to the output with minimal mutual interactions or disturbances. The principle of the FMA is based on summing the force and motion inputs such that they preserve their distinct function at the output. This may be accomplished by means of an appropriate two-input-one-output gear summer in a rotary configuration as described in a following section. In the linear arrangement, this goal may be established by using spindle screws with suitable transfer functions.

It is relatively easier to design the force/motion actuator in the rotary case than in the linear case. Figure 5-4 shows a conceptual sketch. The idea is to combine a force prime-mover and motion prime-mover using a 2-DOF gear train to embed these choices in the actuator.

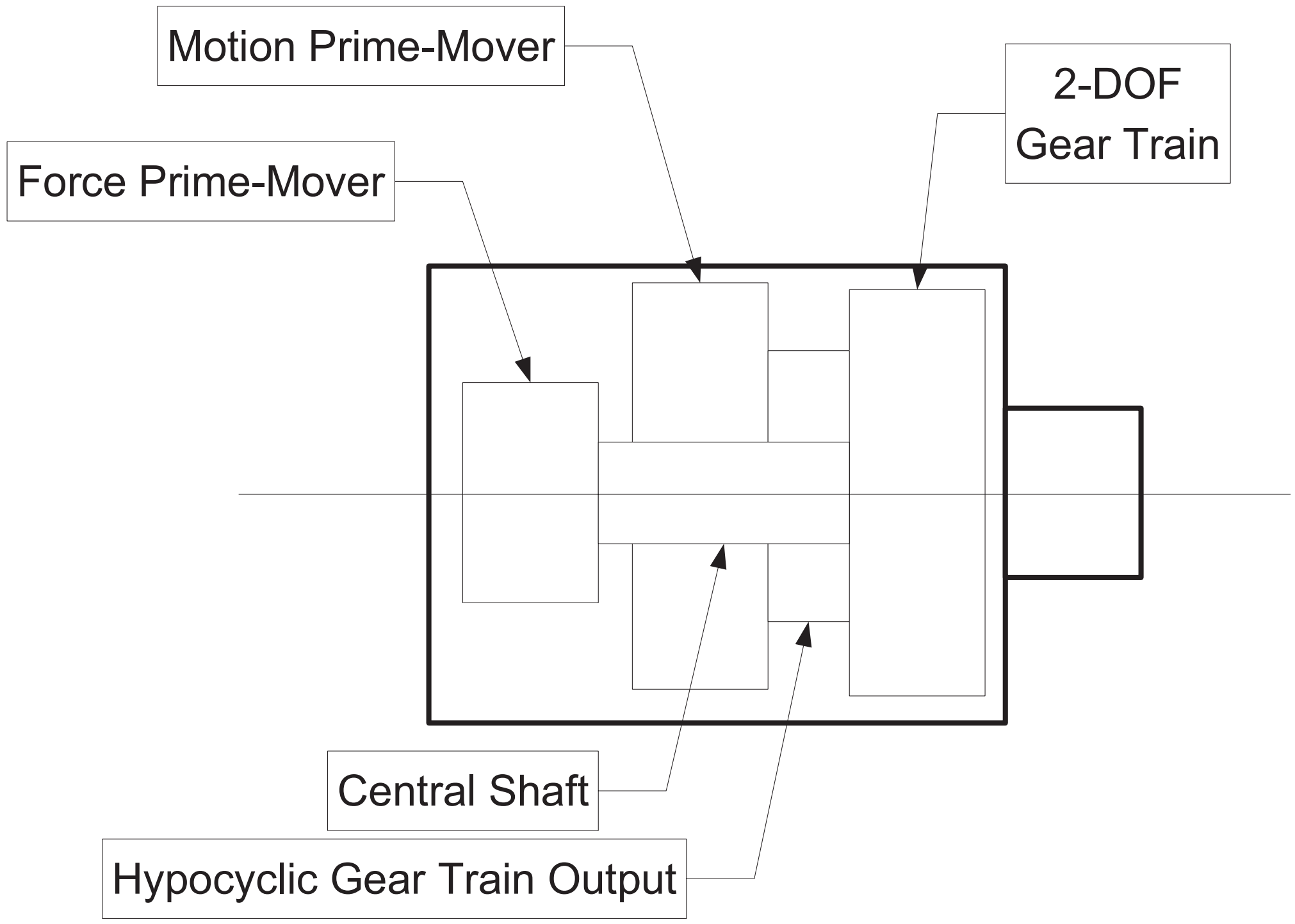

**Figure 5-4. Conceptual Layout of the Force/Motion Actuator**

## 5.3 FMA: Analysis

In this section, we will analyze the kinematic transformation between the prime mover and the output shaft for both the force and motion sub-systems. Also, we present an actuator dynamic model based on kinematic influence coefficients. This model will be used to simulate the system and, consequently, to generate the performance maps of the component stages.

### 5.3.1 Gear Train Kinematics

This section analyzes the gear ratios for the two pathways from both inputs (force and motion prime-movers) to the output. Shown in Figure 5-5 is the front view of a 2-dof star compound gear train.

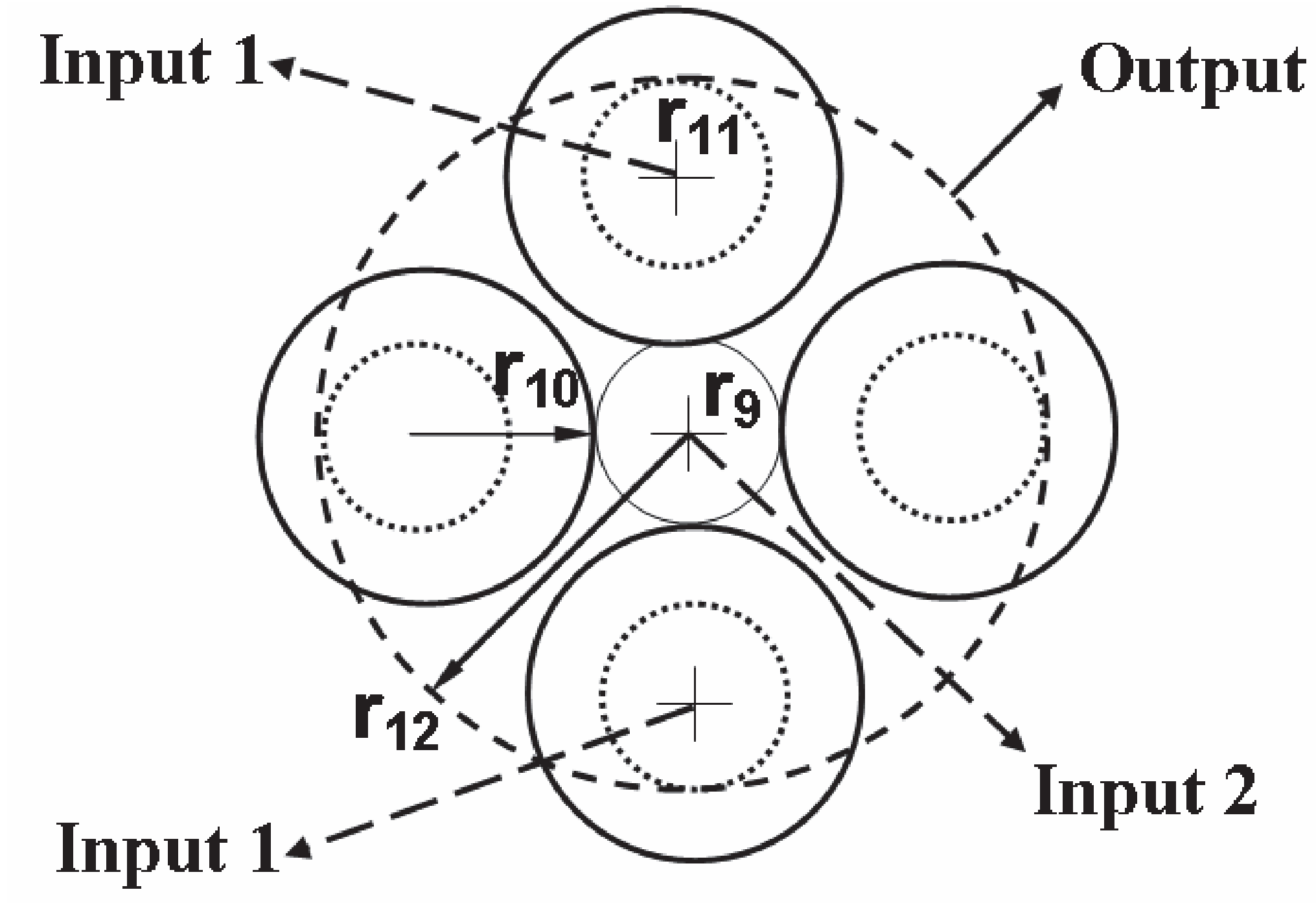

**Figure 5-5. Front View Schematic of a 2-DOF Star-Compound Gear Train**

Input 1 is driven by the velocity prime-mover and Input 2 is driven by the force prime-mover. The radii of the sun gear, planet gear (stage 1), planet gear (stage 2) and the ring gear are respectively $r_9$, $r_{10}$, $r_{11}$ and $r_{12}$.

From the ratios calculated in Table 5-2, we will now determine the kinematic transformations for the force and motion sub-systems.

Let $\omega_f$ be the angular velocity of the sun gear (driven by the force prime-mover), $\omega_v$ be that of the carrier (driven by the output from the hypocyclic gear train) and $\omega_o$ be that of the ring gear (actuator output).

**Table 5-2. Table for Gear Ratio Calculation**

| Action | Carrier (rpm) | Sun (rpm) | Planet (rpm) | Ring (rpm) |
|---|---|---|---|---|
| With carrier locked, sun is given a positive rotation of $\omega$ (CCW) | 0 | $\omega$ | $-\omega \frac{r_9}{r_{10}}$ | $-\omega \frac{r_9 r_{11}}{r_{10} r_{12}}$ |
| With all moving parts locked, the whole assembly is given a positive rotation of $\omega_v$ (CCW) | $\omega_v$ | $\omega_v$ | $\omega_v$ | $\omega_v$ |
| ***Resultant Rotation*** *(Adding Rows 2 and 3)* | $\omega_v$ | $\omega + \omega_v$ | $\omega_v - \omega \frac{r_9}{r_{10}}$ | $\omega_v - \omega \frac{r_9 r_{11}}{r_{10} r_{12}}$ |

From Table 5-2, it is clear that

$$\omega_f = \omega_v + \omega \tag{5-1}$$

$$\omega_o = \omega_v - \omega \frac{r_9 r_{11}}{r_{10} r_{12}} \tag{5-2}$$

Substituting Eq.(5-1) in Eq.(5-2), we get

$$\omega_o = \omega_v - (\omega_f - \omega_v) \frac{r_9 r_{11}}{r_{10} r_{12}} \tag{5-3}$$

$$\omega_o = \omega_v \left( 1 + \frac{r_9 r_{11}}{r_{10} r_{12}} \right) - \omega_f \left( \frac{r_9 r_{11}}{r_{10} r_{12}} \right) \tag{5-4}$$

The carrier is driven by the output from the hypocyclic gear train and the sun is driven directly by the force prime-mover (Refer Figure 5-5). Consider the reduction of the hypocyclic gear train to be $g_{hypo}$. Hence the overall kinematic transformation from velocity motor to the output is $g_{hypo}\left(1+\frac{r_9 r_{11}}{r_{10} r_{12}}\right)$ and that from the force motor to the output is $\left(-\frac{r_9 r_{11}}{r_{10} r_{12}}\right)$. Hence the following expression may be used for evaluating the output angular velocity ($\omega_o$) of the actuator.

$$\omega_o = [\mathbf{G}_p^o]\boldsymbol{\omega}_\mathbf{p} \tag{5-5}$$

$$[\mathbf{G}_p^o] = \begin{bmatrix} g_1 & g_2 \end{bmatrix} \qquad g_1, g_2 < 1 \tag{5-6}$$

$$g_1 = g_{hypo}\left(1+\frac{r_9 r_{11}}{r_{10} r_{12}}\right) \tag{5-7}$$

$$g_2 = -\left(\frac{r_9 r_{11}}{r_{10} r_{12}}\right) \tag{5-8}$$

$$\boldsymbol{\omega}_\mathbf{p} = \begin{bmatrix} \omega_{vp} \\ \omega_{fp} \end{bmatrix} \tag{5-9}$$

$[\mathbf{G}_p^o]$ is the kinematic influence coefficient matrix from the prime-movers to the actuator output. Let us choose the following numbers for gear radii. $r_9 = r_{11} = 1$ unit, $r_{10} = 2.3$ units. Consequently $r_{12} = r_9 + r_{10} + r_{11} = 4.3$ units. If the gear ratio of the hypocyclic stage is chosen to be 150:1, then the following are the transformation ratios for the force and motion subsystems.

$$g_1 = g_{hypo}\left(1+\frac{r_9 r_{11}}{r_{10} r_{12}}\right) = \frac{1}{150}\left(1+\frac{1}{9.89}\right) = 0.007341 \tag{5-10}$$

$$g_2 = -\left(\frac{r_9 r_{11}}{r_{10} r_{12}}\right) = -\left(\frac{1}{9.89}\right) = -0.10111 \quad \textbf{(5-11)}$$

The relative scale of change $\rho$ between force and motion sub-systems is:

$$\rho = \left|\frac{g_2}{g_1}\right| = \frac{\left(\frac{r_9 r_{11}}{r_{10} r_{12}}\right)}{g_{hypo}\left(1 + \frac{r_9 r_{11}}{r_{10} r_{12}}\right)} = \frac{0.10111}{(0.007341)} = 13.773 \quad \textbf{(5-12)}$$

### 5.3.2 Dual Actuator Dynamic Model with Load and External Disturbance

In this section, we develop a second-order dynamic model of the dual actuator system with a load under external force disturbance. The system considered is a dual-actuator with a rigid link attached to its output.

$$I(q)\ddot{q} + V(q,\dot{q}) + F(\dot{q}) + G(q) = (\tau + \tau_{ext}) \quad \textbf{(5-13)}$$

The variables used have the following meaning:

- $I(q)$ is the inertia of the output link about the actuator output shaft (in kg-m$^2$)
- $V(q,\dot{q})$ is the centrifugal/Coriolis torque, which in the case of a single-link is non-existent. However the term is retained to maintain generality
- $F(\dot{q})$ is the joint friction torque at the actuator output
- $G(q)$ is the gravitational torque due to change in link angle
- $\tau$ is the actuator output torque
- $\tau_{ext}$ is the external torque from the environment measured by the force-sensor

Lewis, Abdallah and Dawson [1993] presented the dynamics of armature controlled motors. Extending their model to a multi-input actuator, the dynamics of the force and velocity prime-movers may be organized into a matrix equation as represented in Eq.(5-14). Note that the armature inductance is neglected here.

$$\mathbf{I_M}\ddot{\mathbf{q}}_\mathbf{M} + \mathbf{B_M}\dot{\mathbf{q}}_\mathbf{M} + \mathbf{F_M} + [\mathbf{G_p^o}]^T(\tau + \tau_{ext}) = \mathbf{K_M}\mathbf{v} \quad \textbf{(5-14)}$$

The variables used have the following meaning:

- $\mathbf{I_M} \in \mathcal{R}^{2x2}$ is a diagonal matrix of motor inertias ( in kg-m$^2$)
- $\mathbf{B_M} \in \mathcal{R}^{2x2}$ is a diagonal matrix such that $B_{M_{i,i}} = B_{M_i} + \frac{K_{b_i} K_{m_i}}{R_{a_i}}$
  - $B_{M_i}$ is the motor damping constant of the i$^{th}$ motor (in N-s)
  - $K_{m_i}$ is the torque constant of the i$^{th}$ motor (in N-m/A)
  - $K_{b_i}$ is the back e.m.f constant of the i$^{th}$ motor (in V/rad/s)
  - $R_{a_i}$ is the armature resistance of the i$^{th}$ motor (in Ohms)
- $\mathbf{K_M} \in \mathcal{R}^{2x2}$ is a diagonal matrix of motor torque constants $K_{m_i}$
- $\mathbf{F_M} \in \mathcal{R}^2$ is the vector of motor friction torques (in N-m)
- $\tau$ is the torque at the actuator output (in N-m)
- $[\mathbf{G_p^o}]$ is the kinematic influence coefficient matrix of the g-functions (dimensionless) from the prime-movers to the output of the actuator (See Eq.(5-5) to (5-9))
- $\mathbf{v} \in \mathcal{R}^2$ is the vector of motor inputs (in Volts)
- $\mathbf{q_M} \in \mathcal{R}^2$ is the vector of motor shaft displacement (in rads)

Please note that all the physical parameters used in the dynamic model should be based on the "as-built" numbers and not the "as-designed" or "as-modeled" ones.

From Eq.(5-5) we know that

$$\dot{q} = [\mathbf{G}_p^o]\dot{\mathbf{q}}_\mathbf{M} \tag{5-15}$$

Since $[\mathbf{G}_p^o]$ is non-square, the system in Eq.(5-15) is under-constrained and not readily invertible. Hence we have to resort to a generalized inverse based on a cost function. Consider the cost function (or performance criterion) $\phi(\dot{\mathbf{q}}_\mathbf{M})$ evaluated as follows:

$$\phi(\dot{\mathbf{q}}_\mathbf{M}) = \frac{\mathbf{1}}{\mathbf{2}}\dot{\mathbf{q}}_\mathbf{M}{}^T\mathbf{W}\dot{\mathbf{q}}_\mathbf{M} \tag{5-16}$$

where $\mathbf{W} \in \mathcal{R}^{2x2}$ is a suitable symmetric positive definite weighting matrix and $\dot{\mathbf{q}}_\mathbf{M} \in \mathcal{R}^2$ is the vector of prime-mover velocities. If we minimize the performance criterion in Eq.(5-16) based on the constraint in Eq.(5-15), the generalized inverse of $[\mathbf{G}_p^o]$ may be computed as in Eq.(5-17) and Eq.(5-18) and is called the *right pseudo-inverse solution*.

$$\dot{\mathbf{q}}_\mathbf{M} = [\mathbf{G}_p^o]^+\dot{q} \tag{5-17}$$

$$[\mathbf{G}_p^o]^+ = \mathbf{W}^{-1}[\mathbf{G}_p^o]^T\left([\mathbf{G}_p^o]\mathbf{W}^{-1}[\mathbf{G}_p^o]^T\right)^{-1} \tag{5-18}$$

From Eq.(5-16), in a particular simplified case where $\mathbf{W}$ is the identity matrix $\mathbf{I}_{2x2}$, the generalized inverse takes the form as in Eq.(5-19). In this case the generalized inverse solution locally minimizes the norm of the vector $\dot{\mathbf{q}}_\mathbf{M}$. Performance criteria will be discussed in more detail in a later section.

$$[\mathbf{G}_p^o]^+ = [\mathbf{G}_p^o]^T\left([\mathbf{G}_p^o][\mathbf{G}_p^o]^T\right)^{-1} \tag{5-19}$$

Substituting Eq.(5-17) in Eq.(5-14)

$$\mathbf{I}_\mathbf{M}[\mathbf{G}_p^o]^+\ddot{q} + \mathbf{B}_\mathbf{M}[\mathbf{G}_p^o]^+\dot{q} + \mathbf{F}_\mathbf{M} + [\mathbf{G}_\mathbf{p}^\mathbf{o}]^T(\tau + \tau_{ext}) = \mathbf{K}_\mathbf{M}\mathbf{v} \tag{5-20}$$

Finding an expression for $\tau$ from Eq.(5-20)

$$\tau = [\mathbf{G_p^o}]^{T+}\left[\mathbf{K_M v} - \left(\mathbf{I_M}[\mathbf{G}_p^o]^+\ddot{q} + \mathbf{B_M}[\mathbf{G}_p^o]^+\dot{q} + \mathbf{F_M}\right)\right] - \tau_{ext} \qquad \textbf{(5-21)}$$

Substituting Eq.(5-21) in Eq.(5-13)

$$\begin{aligned} &I(q)\ddot{q} + V(q,\dot{q}) + F(\dot{q}) + G(q) = \\ &[\mathbf{G_p^o}]^{T+}\left[\mathbf{K_M v} - \left(\mathbf{I_M}[\mathbf{G}_p^o]^+\ddot{q} + \mathbf{B_M}[\mathbf{G}_p^o]^+\dot{q} + \mathbf{F_M}\right)\right] - \tau_{ext} \end{aligned} \qquad \textbf{(5-22)}$$

This may be reduced to the following form

$$I'(q)\ddot{q} + V'(q,\dot{q}) + F'(\dot{q}) + G(q) = \mathbf{K'_M\, v} - \tau_{ext} \qquad \textbf{(5-23)}$$

where

- $I'(q) = I(q) + [\mathbf{G_p^o}]^{T+}\mathbf{I_M}[\mathbf{G}_p^o]^+$
- $V'(q,\dot{q}) = V(q,\dot{q}) + [\mathbf{G_p^o}]^{T+}\mathbf{B_M}[\mathbf{G}_p^o]^+\dot{q}$
- $F'(\dot{q}) = F(\dot{q}) + [\mathbf{G_p^o}]^{T+}\mathbf{F_M}$
- $\mathbf{K'_M} = [\mathbf{G_p^o}]^{T+}\mathbf{K_M}$

Eq.(5-23) represents the complete second-order dual actuator dynamic model considering load and external force disturbance.

## 5.4 FMA: Preliminary Simulations

This section outlines the results of preliminary simulations carried out to investigate force/motion control. The objective of this simulation effort was to assess the dynamic model of the dual actuator and run it through representative reference trajectories to evaluate its response. Since the dual actuator is a Dual Input Single Output (DISO) by design, energy was used for optimizing the performance of the force and motion subsystems. A computed torque controller was used for controlling the system. The results expected from this effort are the response of the actuator to force disturbances, its performance envelope, partitioning of the total kinetic energy between the force and motion subsystems so we can monitor the energy transfer between the two sub-systems.

### 5.4.1 System Description

The system considered is the dual force/motion actuator with a link at the output (Figure 5-6). The link is assumed to be rigid with a point mass at its center of mass.

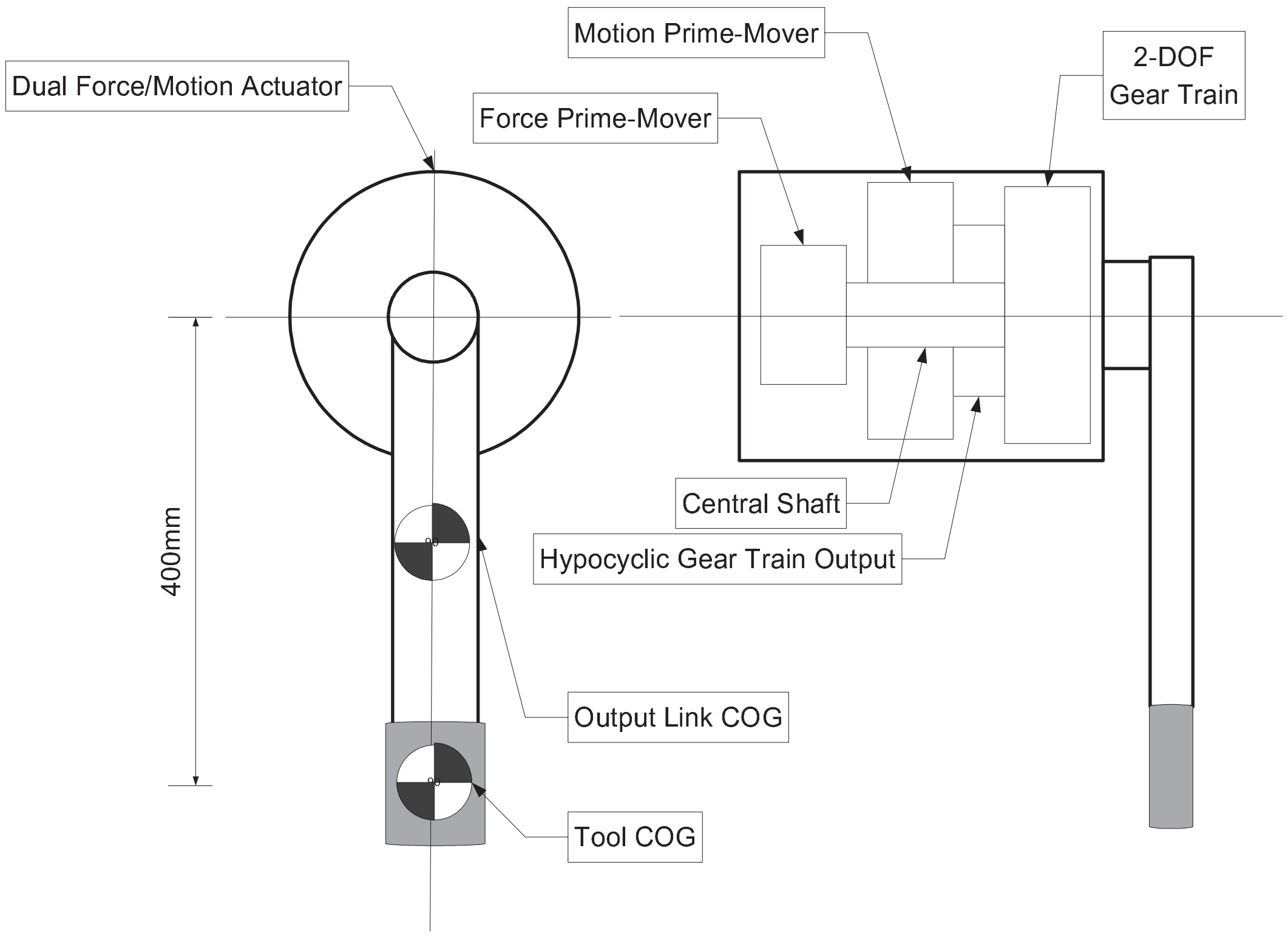

**Figure 5-6. Dual Actuator and Output System Schematic**

The system kinematic model and dynamic model have been described in full-length in Sections 5.3.1 and 5.3.2 respectively. The motion and force prime-movers were chosen from the BH02300 and HT02305 brushless DC motor series respectively from Emoteq Inc [Emoteq Website]. The following tables list the system parameters:

**Table 5-3. Output Link Properties**

| Property | Link | Tool |
|---|---|---|
| "As-Designed" Mass (kg) | 10 | 5 |
| "As-Designed" Length (kg) | 0.40 | Negligible |
| "As-Built" Mass (m) | 13 | 4.9 |
| "As-Built" Length (m) | 0.40 | Negligible |

**Table 5-4. Prime Mover Properties**

| Property | Motion Prime-Mover[6] | Force-Prime-Mover[7] |
|---|---|---|
| Inertia ($kg\text{-}m^2$) | $5.4x10^{-6}$ | $8.9x10^{-5}$ |
| Motor Damping Constant (Nm/RPM) | $2.3x10^{-7}$ | $3.1x10^{-5}$ |
| Torque Constant (Nm/A) | 0.039 | 0.36 |
| Back e.m.f Constant (V/rad/s) | 0.04 | 0.36 |
| Armature Resistance (Ohms) | 2.23 | 2.23 |
| Transmission Ratio | 136 | 9.89 |

---

[6] Values based on BH02300 high-speed brushless DC motor from Emoteq, Inc

[7] Values based on HT02305 high-torque brushless DC motor from Emoteq, Inc

### 5.4.2 Task Planning

The task plan chosen for this simulation is motivated by deburring. Deburring illustrates most of the challenges for force/motion control since force and motion have to be managed in the same direction. The deburring tool pressure should be managed in the direction normal to the feed. The feed rate should be maintained constant at a relatively high value while rejecting the force/torque disturbances from the environment. A trapezoidal motion plan was created for the output link velocity as described in Eq.(5-24). $T = 10\,\sec$ is the total time for which the simulation is run. To simulate the "bumping into a burr" scenario, a contact force model based on viscous friction was developed as described in Eq.(5-27).

$$\dot{q}_{desired} = \begin{cases} \dfrac{\omega_o^{peak}}{\left(\dfrac{T}{4}\right)} t & if \quad t \le \dfrac{T}{4} \\ \omega_o^{peak} & if \quad \dfrac{T}{4} < t < \dfrac{3T}{4} \\ \dfrac{\omega_o^{peak}}{\left(\dfrac{T}{4}\right)} (T - t) & if \quad t \ge \dfrac{3T}{4} \end{cases} \tag{5-24}$$

$$\begin{aligned} T &= 10\,\sec \\ \omega_o^{peak} &= \frac{2\pi}{T} \end{aligned} \tag{5-25}$$

$$B_{friction} = \begin{cases} 5.0 & if \quad 1.0^c < q < 2.0^c \\ 25.0 & if \quad 3.0^c < q < 4.0^c \\ 0.0 & otherwise \end{cases} \tag{5-26}$$

$$\begin{aligned} \tau_{ext} &= B_{friction}\dot{q} + \eta \\ \eta &= \mathrm{N}(0, 2) \end{aligned} \tag{5-27}$$

The force sensor signal usually has some noise. To simulate this an additive white noise (standard normally distributed) with mean 0 and standard deviation 2 N-m, $\eta = \mathrm{N}(0,2)$, is used in Eq.(5-27).

### 5.4.3 FMA: Decision Making and Control

For control purposes, a computed torque control system is used. The control signal may be expressed in terms of the reference trajectory and the dynamic parameters as follows:

$$\mathbf{v} = \mathbf{K}_{\mathbf{M}}^{-1}[\mathbf{G}_p^o]^T \left[ \hat{I}'(q)\ddot{q} + \hat{V}'(q,\dot{q}) + \hat{F}'(\dot{q}) + \hat{G}(q) \right] \quad \textbf{(5-28)}$$

where $\hat{I}'(q)$, $\hat{V}'(q,\dot{q})$, $\hat{F}'(\dot{q})$, $\hat{G}(q)$ have the usual meanings described in Section 5.3.2 but are calculated based on the ‘as-designed’ parameters rather than the “as-built” ones.

The joint friction, mainly due to friction in the transmission, is designed based on the Stribeck friction model. Kennedy and Desai [2003] estimated the harmonic drive friction for a Mitsubishi PA-10 Robot using this model.

$$F(\dot{q}) = 0.20 + 1.506\dot{q} - 0.9602(1 - e^{-0.0047\dot{q}}) \quad \textbf{(5-29)}$$

According to Eq.(5-16) a positive definite matrix $\mathbf{W}$ maybe used for partitioning the control horse-power used for the force and motion sub-systems. Rather than using the inertia matrix, a dynamic weighting matrix was used to govern this partitioning based on the measured external force disturbance as given in Eq.(5-30) and Eq.(5-31).

$$\mathbf{W} = \begin{bmatrix} 1 & 0 \\ 0 & w \end{bmatrix} \quad \textbf{(5-30)}$$

$$w = \begin{cases} 164.5 & if \quad \tau_{ext} < 4.0Nm \\ 16.45 & otherwise \end{cases} \quad \textbf{(5-31)}$$

The weight 16.45, for operation under presence of disturbance, is the ratio between the prime-mover rotor inertias $\frac{I_{M_2}}{I_{M_1}}$. The weight for the case of absence of force disturbance is 10x the weight for the other case. These numbers seemed to work well for the simulation.

### 5.4.4 Simulation Results and Interpretation

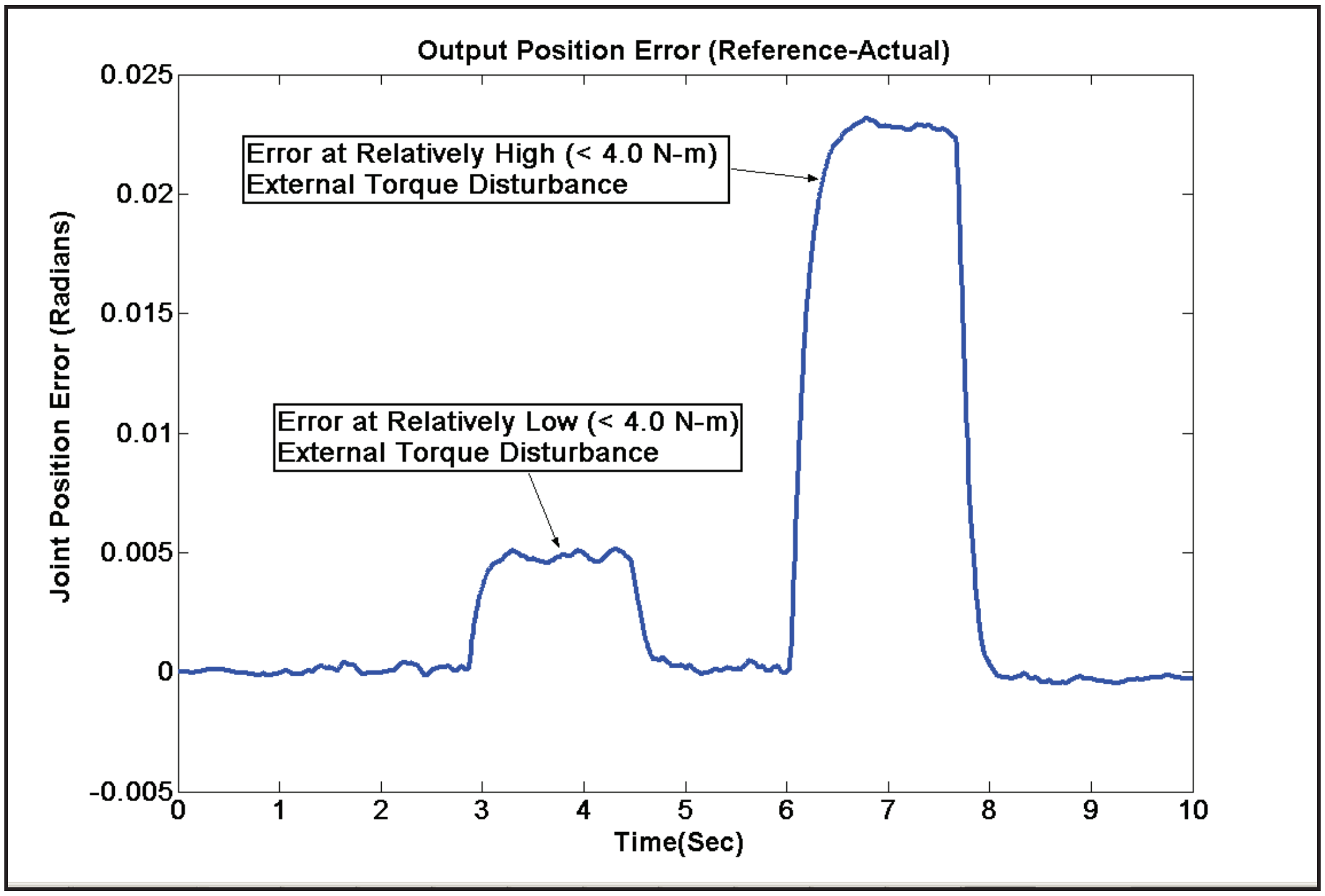

**Figure 5-7. Position Tracking Performance**

This section presents the FMA simulation results. Please note that the two external force disturbances according to Eq.(5-26) occur between 3.0-4.7 sec and 6.0-8.1 sec respectively as shown by the relatively high position tracking errors in these time intervals (See Figure 5-7). In these two time-intervals, a kink in the velocity tracking (Figure 5-8) is noticeable (not as much in the first interval as that in the latter one). This is when the output link slows down for a brief instant (0.7 seconds) and then follows the reference even while the disturbance is still existent.

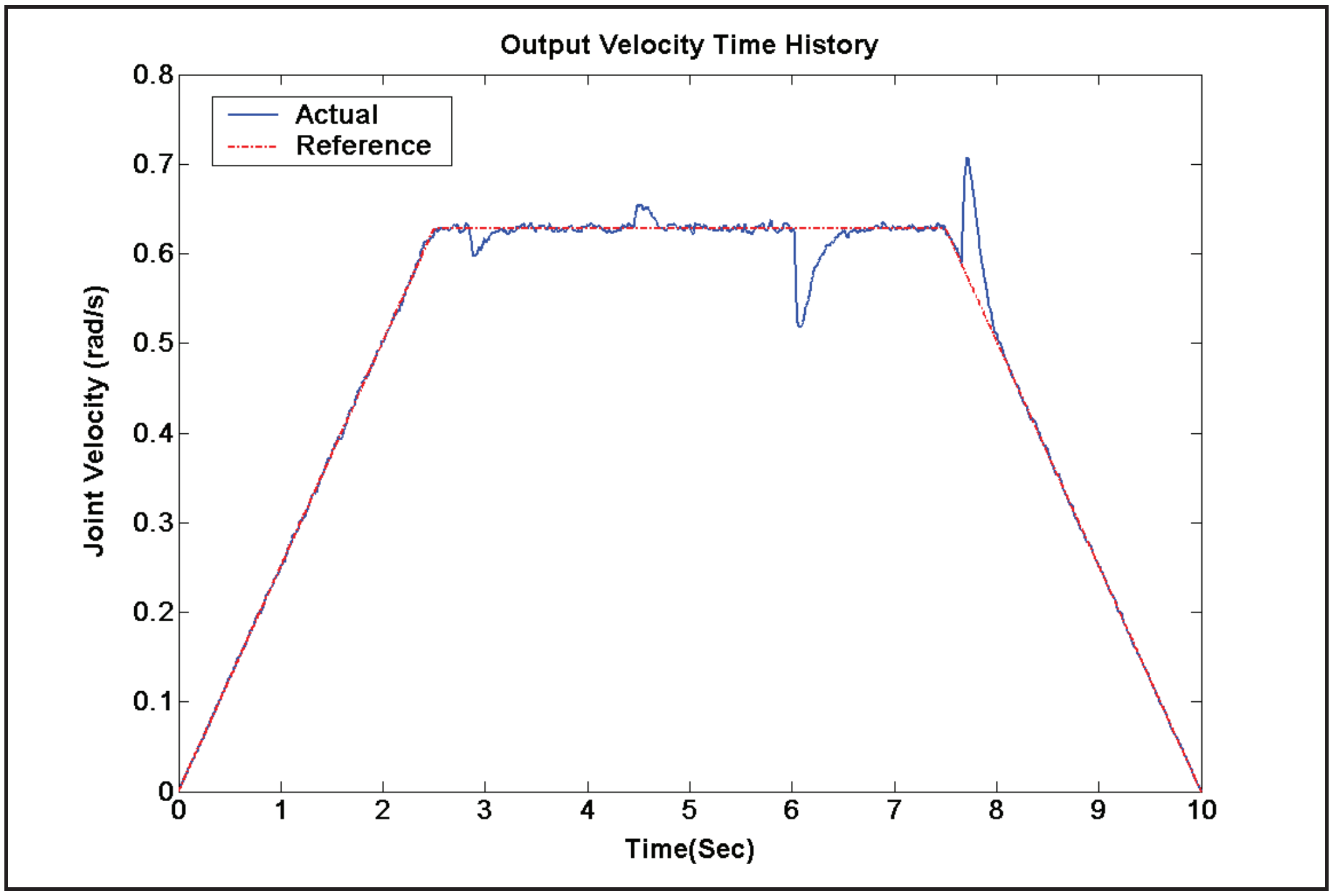

**Figure 5-8. Velocity Tracking Performance**

Also notice the brief increase in velocity (See Figure 5-8) when the external force disturbance is removed. This simulates a scenario wherein a tool makes and breaks contact with a burr.

Figure 5-9 shows the voltage history of the prime-movers of the motion and force sub-systems. Notice that the input is non-zero after execution of the whole trajectory since the final absolute position of the output link is 4.71 radians and thus there are static gravity and frictional torques acting on the link in this final configuration.

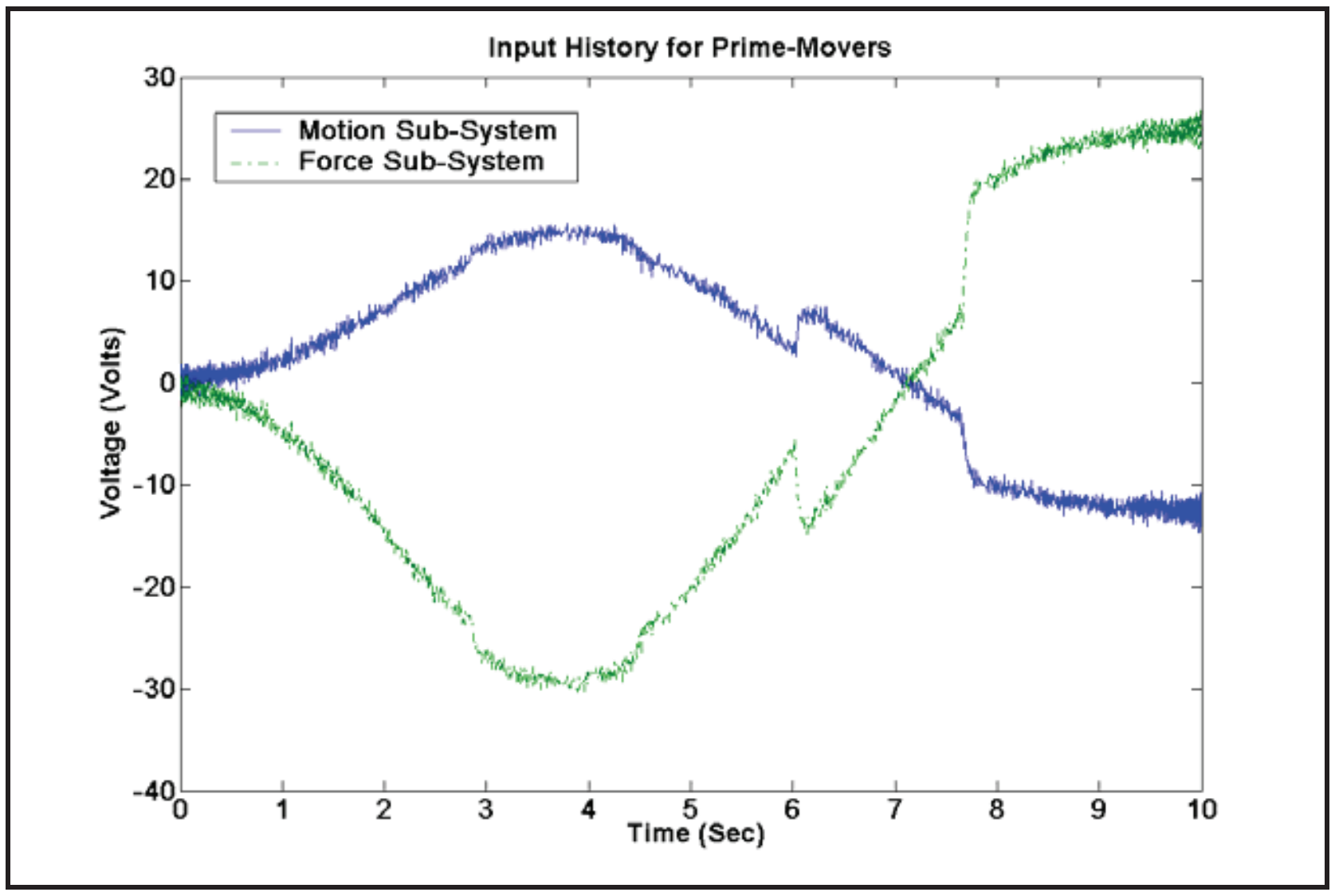

**Figure 5-9. Prime-Mover Input (Voltage)[8] History**

Figure 5-10 shows the contributions of the prime-movers in the velocity domain and also the kinetic energies of the motion and force sub-systems during the task. Note here that average velocity of the motion prime-mover is around 7.5 times that of the force prime-mover. This shows that the major contributor in the velocity domain is the motion (i.e., velocity) sub-system.

---

[8] For this simulation, the dynamic model of the motor was second order because the armature inductance was assumed to be negligible. If this is not assumed, then the motor dynamics would be third-order and the input would be current instead of voltage

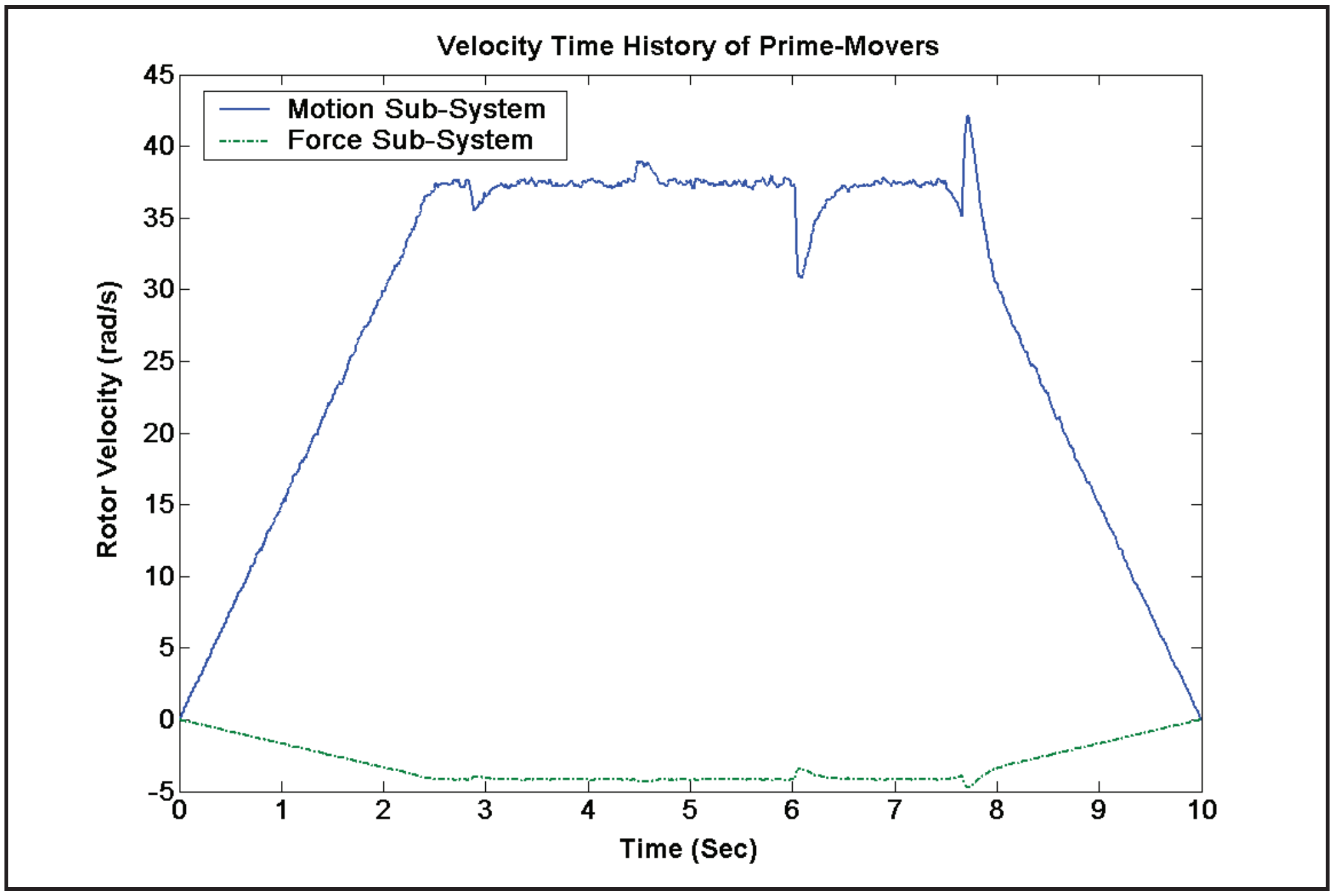

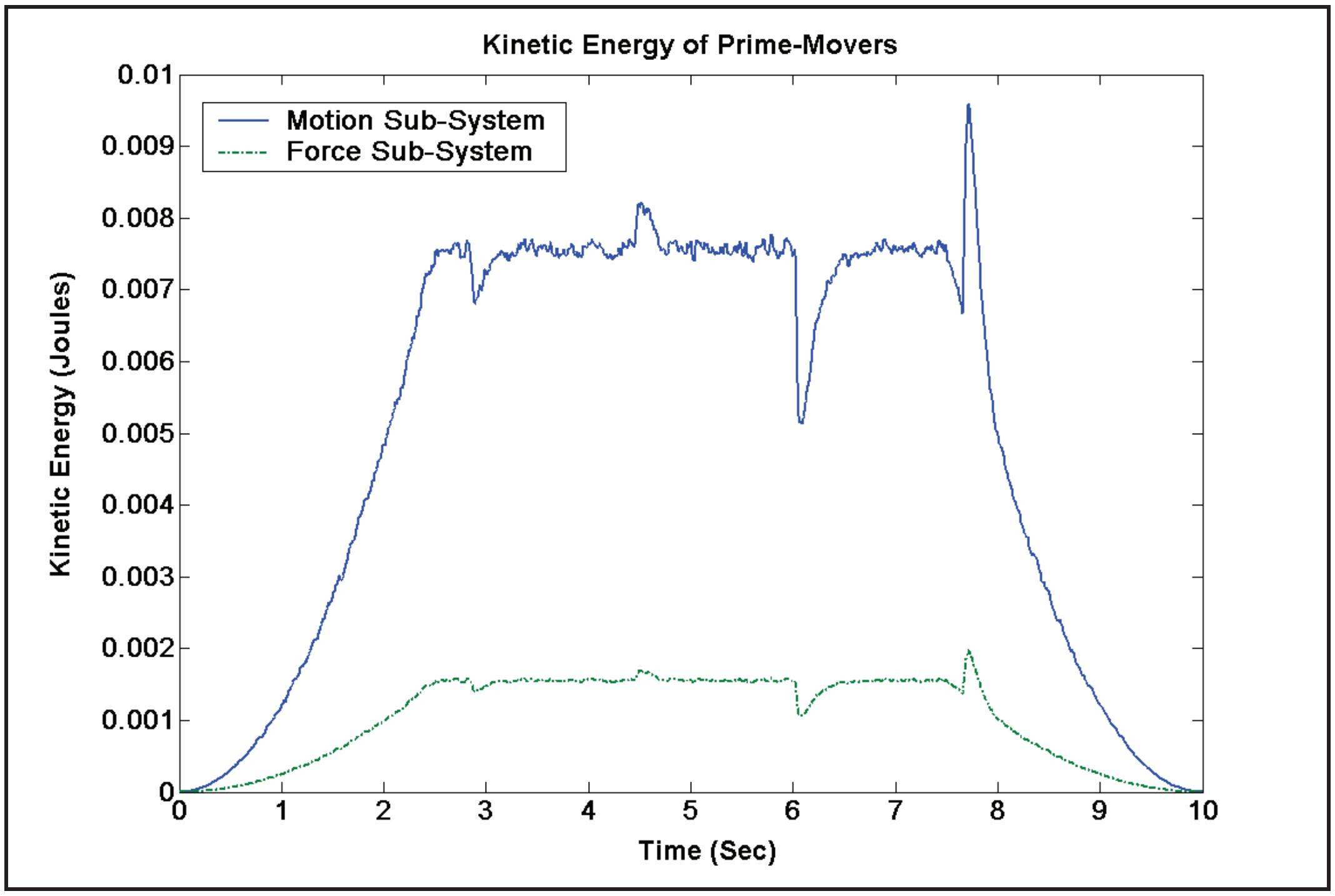

**Figure 5-10. Prime-Mover Contributions in Motion Domain**

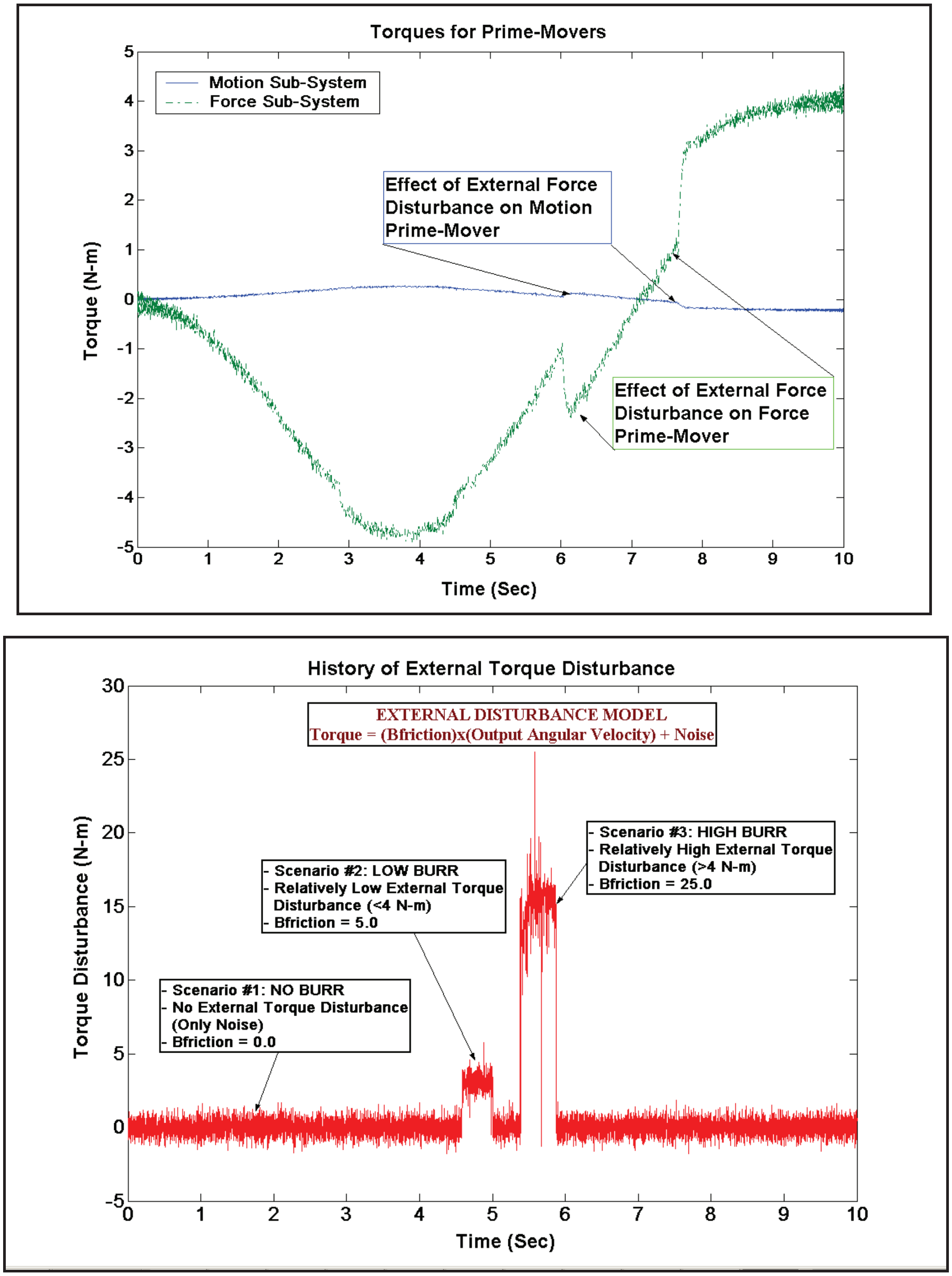

**Figure 5-11. Prime-Mover Contributions in Force Domain Under Disturbance**

The Partition Value of Kinetic Energies (PVKE) was monitored and a constant partitioning of 83% to 17% between the motion and force prime-movers was observed (See Figure 5-10). This is a direct result of minimizing the weighted norm of the prime-mover velocities using the weighting matrix dependent on the external force disturbance formulated in Eq.(5-30).

Figure 5-11 shows the contributions of the motion and force sub-systems in the force domain. The force sub-system produces about 20 times as much torque as that of the motion sub-system on an average. This shows that the principal contributor of torque during task execution is the force prime-mover. Also notice that the velocity prime-mover is virtually unaffected by the external torque disturbance whereas the force prime-mover is significantly affected (Refer prime-mover torque history in Figure 5-11). This is primarily due to the kinematic scaling of approximately 14:1 between the force and motion sub-systems shown in Eq.(5-12). This reaction to external disturbance is dependent on the backdriveability of the force sub-system. Notice that the motion sub-system is virtually non-backdriveable (and thus mechanically rigid) while the force sub-system is backdriveable (and thus mechanically soft) (Refer Table 5-1). Figure 5-11 also shows the external torque disturbance that simulates three characteristic scenarios in a deburring application (viz. *No Burr*, *Low Burr*, and *High Burr* respectively referred to as Scenarios# 1, 2, and 3 in the following discussion).

### 5.4.5 Simulation Summary

To summarize the presented results, Figure 5-7 shows that a tracking performance of over 95% (approximately 0.005 radians positional error) is achievable for small disturbances (< 4 N-m) and this performance deteriorates to approximately 84% (approximately 0.025 radians positional error) for a time-interval of 0.7 seconds under large external disturbances (> 4 N-m). Figure 5-10 shows that the force sub-system has minimal motion relative to the velocity sub-system, the latter responding (on an average) at about 7.5 times the rate at which the former responds. At the same time, Figure 5-11 shows that the velocity sub-system has minimal average torque relative to the force sub-system, the latter producing (on an average) at least 20 times the torque produced by the former.

Another observation from the simulation (See Figure 5-11) is that under the presence of external torque disturbance, the force sub-system is more responsive to output demands in the force level in comparison to the velocity sub-system. This is because the former is relatively mechanically soft than the latter.

Minimizing the weighted norm of the prime-mover rotor velocities forces the use of the motion prime-mover to contribute mainly toward motion-tracking (Scenario# 1 described above in Figure 5-11) and the force prime-mover toward external force disturbance rejection (Scenario# 2 and Scenario# 3 in Figure 5-11).

This simulation points to performance criteria as major vehicles to capitalize on the independent force and motion inputs. The performance criterion used in this simulation was the weighted norm of the prime-mover velocities. Many more useful and physically significant criteria could be used. Minimizing the power input (voltage x current) to the prime-movers is another feasible criterion.

Another issue not addressed by this simulation is the dynamic coupling between the two prime-movers. The dynamic model of the prime-movers presented in Eq.(5-14) assumes that the dynamics of the two sub-systems are mutually exclusive. A more accurate model could be developed which considers this dynamic coupling (which will be significant). This will be given consideration in future research.

## 5.5 FMA: Performance Envelope

A performance envelope for an actuator may be defined as the closed surface describing the range of attainable outputs of a system (actuator) in terms of all relevant performance criteria as parametric functions of system states, controlled inputs, uncontrolled inputs, and parameters, and constraints on the states, inputs, and parameters Eq.(5-32) [Hvass and Tesar, 2004]. A performance envelope bounds the set of all attainable operating points for a specific set of performance criteria using the same reference operating parameters.

$$PE_j = range(f_1(x,u_c,u_d,\theta), f_2(x,u_c,u_d,\theta)...f_n(x,u_c,u_d,\theta)) \quad \textbf{(5-32)}$$

A performance envelope for an actuator may be characterized as a set of data points which serves as a guide to motor performance capability. It may be thought of as a look-up table or a point cloud that can be used in real-time for control and decision making. Torque-speed curves are often used to assess the capability of an actuator. A torque-speed performance envelope is the locus of all operating points (or ordered pairs of torque and speed) that can be achieved by the actuator system. Figure 5-12 shows the schematic of such a torque-speed performance envelope. The schematic suggests three envelopes:

- Conservative performance region (C), which is the capability of actuator necessary for nominal operation
- Enhanced performance region (E), which may be achieved by using redundant resources and criteria based decision-making
- Performance surge to include reduced reserve (R), which is the actuator capability demanded under worst-case operating conditions
- From the above definitions, $C \subset E \subset R$

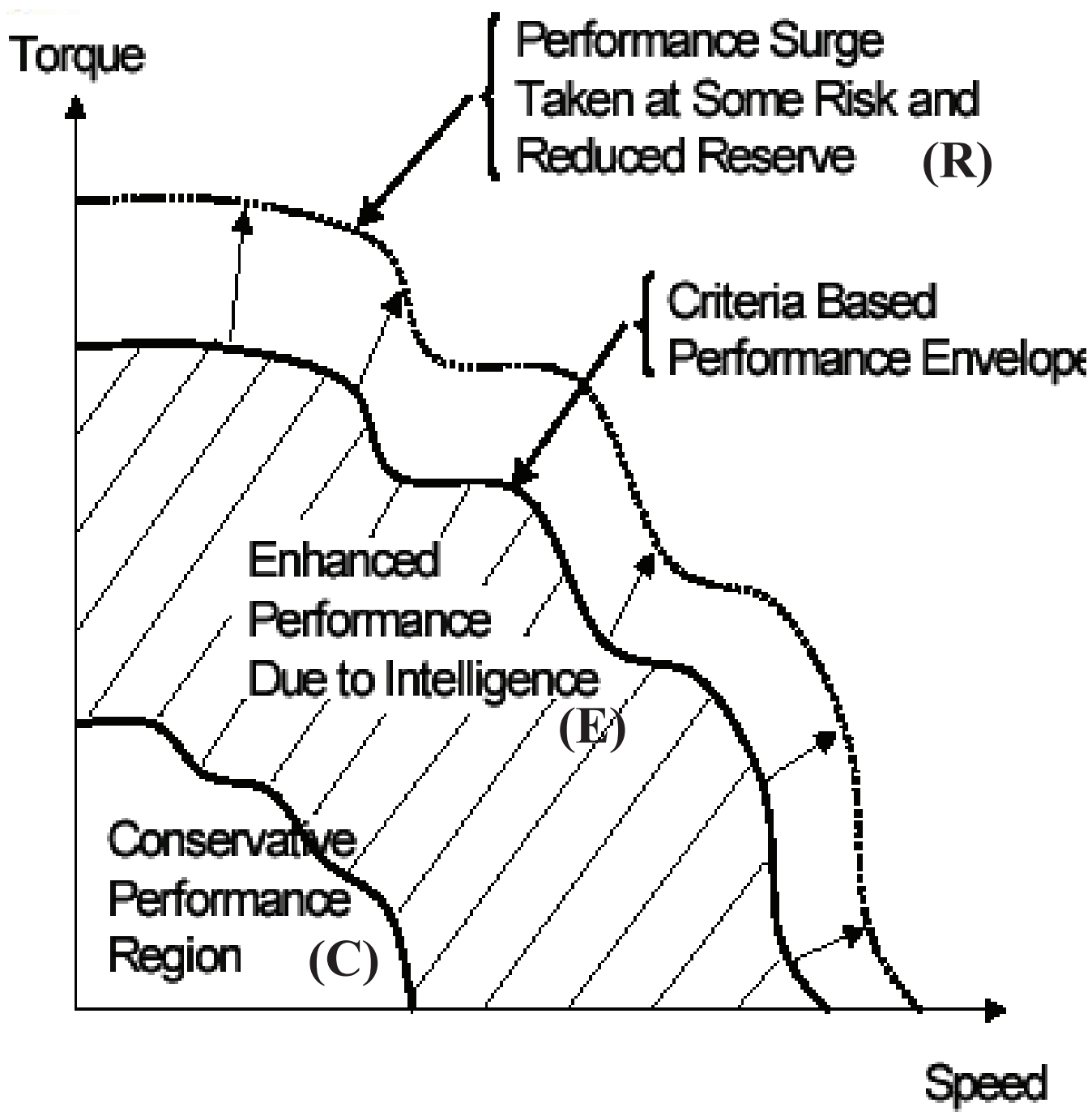

**Figure 5-12. Actuator Performance Envelope Concept**

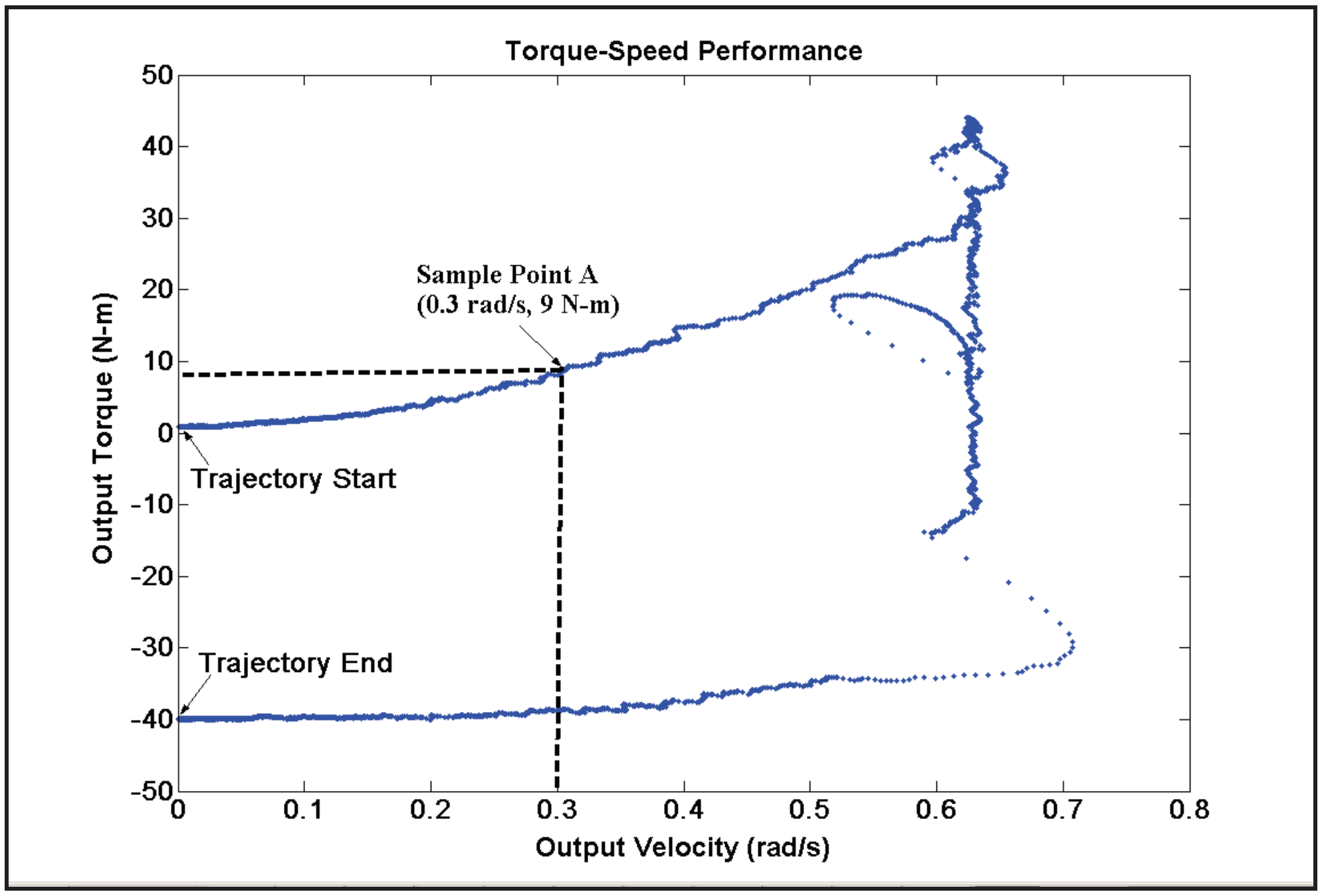

**Figure 5-13. Torque-Speed Performance of FMA for Trapezoidal Motion Plan**

Performance envelopes may represent different quantities such as response, efficiency, noise etc. A dedicated study of performance envelopes for the FMA is required to know its complete capability. A starting step toward that goal is to evaluate its torque-speed capability for various output quantities.

Figure 5-13 shows the torque-speed performance of the FMA for the trapezoidal velocity tracking task with force disturbances. This figure shows the torque and speed demands on the force/motion actuator *output* to complete the trapezoidal motion planning task under the conditions of the simulation described in Section 5.4. This is an illustration of how a torque-speed performance envelope can be constructed using a desired operational specification (trapezoidal motion plan in the simulation) and a specific set of performance criteria (weighted norm of the prime-mover velocities in the simulation). In future work, many more representative reference signals and performance criteria for robotic actuators should be used to create an exhaustive torque-speed performance envelope for the FMA. Such an envelope would tell us the performance capability of the actuator in terms of achievable torques and speeds at the output. A map such as this can be used to compare the capability of the FMA with other actuators (such as the pure velocity generator or pure force generator). Figure 5-13 shows a simple example (for trapezoidal reference velocity trajectory) of this exhaustive envelope that needs to be developed in future research. Note that there are fewer data points on this plot than is actually needed to make it useful. The graphic merely demonstrates how such a performance map can be developed.

## 5.6 FMA: Performance Criteria

Performance criteria may be defined as functions of inputs, outputs and/or properties of the system which mathematically represent secondary task goals. In the literature, criteria are often referred to as cost functions which need to be maximized or minimized to achieve a specific task goal. Criteria are used for redundancy resolution in redundant systems, i.e, systems with more inputs than outputs.

The force/motion actuator, by design, is a redundant system with two inputs, namely the force and motion prime-movers, governing a common output. Due to this redundancy, it is possible to satisfy secondary task goals by optimizing performance criteria. Like the criterion in Eq.(5-16), let us specify another cost function as given below:

$$\phi(\dot{\mathbf{q}}_{\mathbf{M}}) = \frac{\mathbf{1}}{\mathbf{2}}\left(\dot{\mathbf{q}}_{\mathbf{M}} - \dot{\mathbf{q}}_{\mathbf{M0}}\right)^T \mathbf{W}\left(\dot{\mathbf{q}}_{\mathbf{M}} - \dot{\mathbf{q}}_{\mathbf{M0}}\right) \tag{5-33}$$

where $\mathbf{W} \in \mathcal{R}^{2x2}$ is a suitable symmetric positive definite weighting matrix and $\dot{\mathbf{q}}_{\mathbf{M0}} \in \mathcal{R}^2$ is a vector of arbitrary prime-mover velocities. If we minimize the performance criterion in Eq.(5-33) based on the constraint in Eq.(5-15), the generalized inverse of $[\mathbf{G}_p^o]$ may be computed as follows.

$$\dot{\mathbf{q}}_{\mathbf{M}} = [\mathbf{G}_p^o]^+ \dot{q} + \left(\mathbf{I} - [\mathbf{G}_p^o]^+ [\mathbf{G}_p^o]\right)\dot{\mathbf{q}}_{\mathbf{M}0} \tag{5-34}$$

$$[\mathbf{G}_p^o]^+ = \mathbf{W}^{-1}[\mathbf{G}_p^o]^T \left([\mathbf{G}_p^o]\mathbf{W}^{-1}[\mathbf{G}_p^o]^T\right)^{-1} \tag{5-35}$$

In the case where $\mathbf{W} = \mathbf{I}_{\mathbf{2x2}}$ the generalized inverse solution locally minimizes the norm of the vector $\left(\dot{\mathbf{q}}_{\mathbf{M}} - \dot{\mathbf{q}}_{\mathbf{M0}}\right)$, i.e, $\dot{\mathbf{q}}_{\mathbf{M}}$ is chosen as close as possible to $\dot{\mathbf{q}}_{\mathbf{M0}}$. The vector $\dot{\mathbf{q}}_{\mathbf{M0}} \in \mathcal{R}^2$ is arbitrary and hence facilitates satisfying a secondary objective. The second term in Eq.(5-34), known as the *homogenous solution*, contains the matrix $\left(\mathbf{I} - [\mathbf{G}_p^o]^+[\mathbf{G}_p^o]\right)$ which is a projector matrix that projects any arbitrary $\dot{\mathbf{q}}_{\mathbf{M}0}$ to the null-space of $[\mathbf{G}_p^o]$. Figure 5-14 shows the null-space in the prime-mover velocity space. Any point on this null-line results in a zero velocity at the output link.

The vector $\dot{\mathbf{q}}_{\mathbf{M0}}$ is usually calculated based on a performance criterion as shown below:

$$\dot{\mathbf{q}}_{\mathbf{M0}} = k(\nabla h) \tag{5-36}$$

where h is a performance criterion and k is a positive (or negative) constant if we wish to locally maximize (or minimize) $h$. Table 5-5 lists various criteria that could be used for actuators in the force and motion domains.

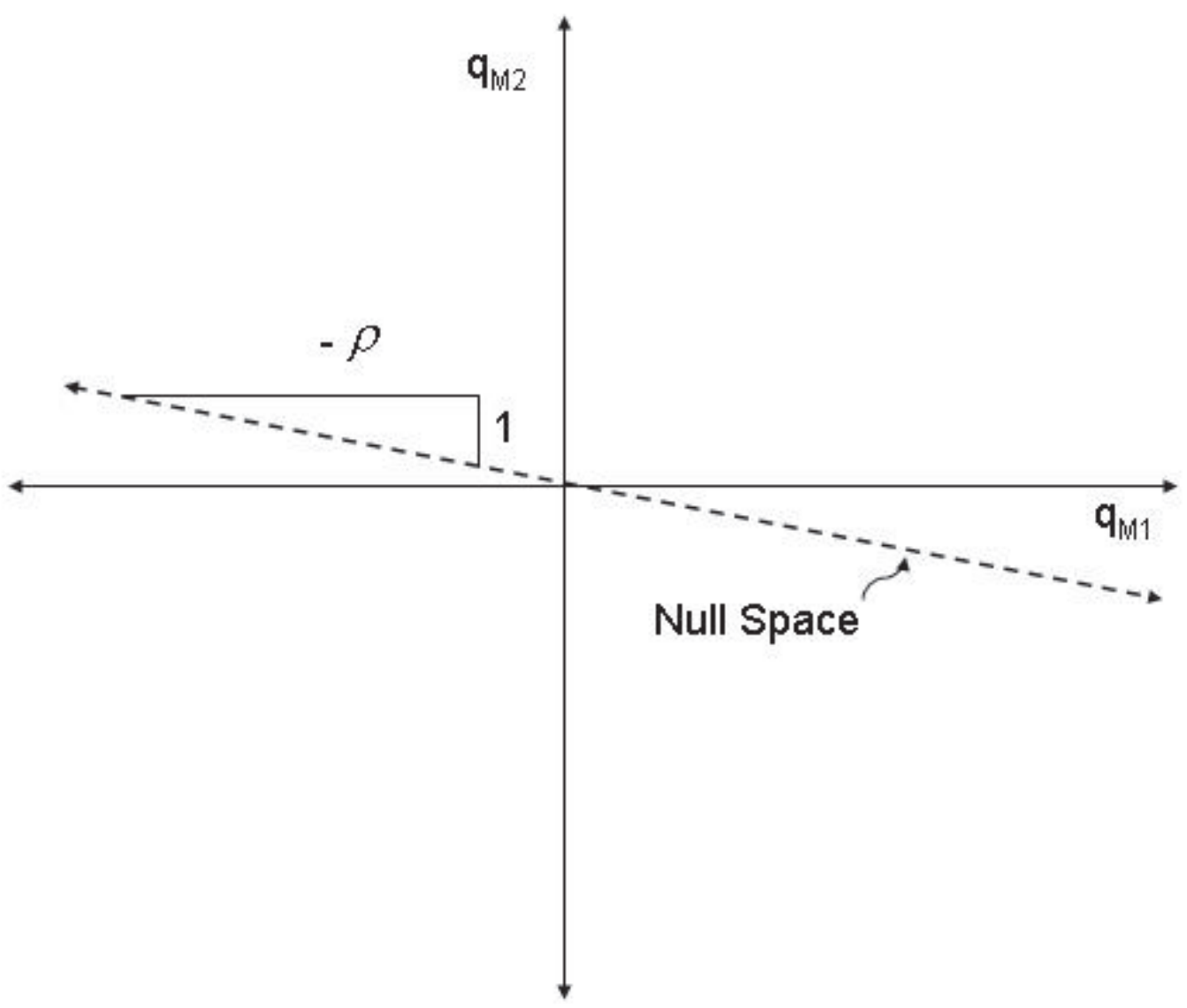

**Figure 5-14. Null-Space in the Prime-Mover Velocity Space**

**Table 5-5. Criteria Corresponding to Different Actuation Domains**

| Force Criteria | Motion Criteria | Mixed Criteria (Force/Motion) |
|---|---|---|
| ○ Force Tracking Error<br>○ Force Ellipsoids<br>○ Load Capacity | ○ Motion Tracking Error<br>○ Velocity Ellipsoids<br>○ Motion Range | ○ Precision<br>○ Accuracy<br>○ Disturbance Rejection<br>○ Power/Energy<br>○ Kinetic Energy Partition Value |

Some desirable characteristics of performance criteria are as follows [Tisius and Tesar, 2004]:

- Physically significant – must display mathematical or experimental improvement
- Multiple physical meanings – this increases effectiveness without additional computational expense
- Varies over the workspace – allows decision making
- Single valued – allows deterministic solutions
- Continuous – allows integral and differential calculations
- Computationally efficient – important for a system to run in real-time
- Mathematically independent – keeps criteria from overlapping effects
- Bounded in magnitude – makes normalization possible
- Task independent – one formulation regardless of task being performed

## 5.7 Chapter Summary

The principal aim of this chapter was to initiate an investigation to determine the capability of the Force/Motion Actuator (FMA). The literature pertaining to multi-domain inputs within the Robotics Research Group and their application areas were covered to demonstrate the motivation for this study. A thorough kinematic and dynamic model of the FMA was presented. Based on these analytics, preliminary simulations were carried out to explore the capabilities of this actuator. A trapezoidal output motion plan was considered as a reference operational specification for this numerical simulation. It was observed that for a motion tracking application with external force disturbances, the motion sub-system contributes predominantly to the velocity tracking, while the force sub-system is the principal contributor to the output torque. We also observed that with a kinematic scaling of approximately 14:1 between the force and motion sub-systems, the effect of external force disturbances is minimal on the velocity side while it is significant on the force side. This is because the velocity sub-system is virtually non-backdriveable while the force sub-system is relatively mechanically soft (due the kinematic scaling 14:1). This shows that the design choice of 14:1 scaling for the FMA results in the selective flow of force and motion sub-system attributes to the output. Embedding these multiple behaviors in force and motion domains in the same actuator is hence desirable.

Two areas, not completely researched in this chapter, which need to be addressed, are the performance envelope for the FMA and meaningful performance criteria to base decision-making and control on. Initial thoughts toward research in this direction are presented. A more accurate dynamic model for the FMA that considers the coupling between the motion and force sub-systems needs to be developed. Before that, a thorough static analysis using FEM should be conducted. We suggest that extensive experimentation and prototype development will be the definitive way to prove the capability of this multi-domain hybrid actuator system.

# Chapter 6

# Summary and Roadmap

Interaction control, which involves the management of both contact force and motion, enhances the capability of an intelligent machine (such as a serial-chain manipulator) to perform high-value manufacturing processes like deburring, grinding, assembly, force-fitting, stamping, and die finishing. In this report, a detailed review of the literature (Chapter 2) pertinent to this operational issue (at the system and component levels) was compiled. Experiments were implemented on a modular robot testbed (Chapter 3 and Chapter 4) to identify the main issues in implementation of force control applications on real robot systems and also to study the effect of actuator characteristics on system level force control performance. The report presents *Multi-Domain Inputs* at the actuator level as means of expanding the choices in the force and motion domains. The Force/Motion Actuator (FMA) is an example of such an actuator, consisting of distinct force and motion subsystems combined in parallel. Conceptual design, modeling, and control of the FMA were presented in Chapter 5.

This work was broad to encompass system level experimental force control and investigation of the FMA. However, for future work it is necessary to emphasize multi-domain inputs, interaction between their constituent sub-systems, and their effect on the operational performance of the overall system.

**Table 6-1: Literature Summary**

| Area | Contribution | Credit(s) | Institution(s) | Year(s) |
|---|---|---|---|---|
| **Task Programming** | Task Frame Formalism (TFF)** | Hendrik Van Brussel, Joris De Schutter Bruyninckx | KU Leuven | 1980-1996 |
| | Compliance Frame Formalism (CFF)* | Matthew Mason | AI Lab, MIT | 1981 |
| **Operational Software** | OSCAR** | Kapoor and Tesar | UTRRG | 1996 |
| | OROCOS** | Bruyninckx | KU Leuven | 2001 |
| **Component Technology** | Multi-Input Actuators, Layered Control, CITS and EMAA*** | Tesar | UTRRG | 1978-2003 |
| | Parallel Coupled Micro-Macro Actuator** (PaCMMA) | Morell and Salisbury | MIT | 1996 |
| | Distributed Macro-Mini ($DM^2$) Actuation** | Zinn, Khatib and Roth | Stanford | 2000-2004 |
| **Algorithms** | Hybrid Control Research Thread* | Raibert and Craig | MIT | 1981 |
| | | Lipkin and Duffy | UFL | 1988 |
| | Compliant Control** | Hendrik Van Brussel | KU Leuven | 1980 |

*** Most relevant to this report

** Intermediate relevance to this report

* Some relevance to this report

## 6.1 Literature Summary

Past works that were deemed relevant to the research and presented in this report are listed in Table 6-1. The goal of multi-input actuation is to use distinct inputs that are scaled with respect to each other such that one set is responsible for attaining task goals (containing low frequency content) and the other set is used for disturbance rejection (containing high frequency content). If the inputs are all in the velocity domain, then this approach is called *Control-In-The-Small (CITS)* and was patented by Tesar [1985]. Tesar showed that CITS can be used to control the output at various scales of motion in the CHAMP proposal [1999b]. Parallel Coupled Micro-Macro Actuator (PaCMMA) developed by Morell and Salisbury [1996] at MIT, and the Distributed Macro-Mini ($DM^2$) actuation method for human-centric robotics proposed by Zinn, Khatib and Roth [2004] at Stanford draw on this same concept (CITS). The Electromechanical Actuator Architecture (EMAA) [Tesar, 2003a] is the result of a two decade actuator research history at UT Austin. It conceptually describes in considerable detail dual fault-tolerant, layered control, and force/motion actuators as part of 10 basic classes of actuator designs. Apart from above mentioned works, the treatment of component level technology for force/motion control in the literature is inadequate. For important literature on algorithms, task programming and software architectures for force/motion management, please refer Table 6-1.

## 6.2 Result Summary and Interpretation

Conceptual design, modeling and decision making for the FMA were presented in Chapter 5. The system (See Figure 6-1) was simulated with the results presented in Table 6-2. A trapezoidal trajectory (total time of 10s with peak of 0.6rad/s from t=2.5s to t=7.5s) in the motion domain with disturbances in the force domain (0.4N-m) was used as the task plan.

**Table 6-2. Summary of FMA Simulation Results**

| | Force/Motion Actuator |
|---|---|
| Kinematic Scaling | 13.773 (Force:Motion) |
| Partition Value of Kinetic Energies (PVKE) | 83:17 (Force:Motion) |
| Performance Criterion Used | Minimizing Weighted Norm of Kinetic Energies |
| Tracking Performance (Small Disturbance < 4N-m) | 95% |
| Tracking Performance (Large Disturbance > 4N-m) | 84% |

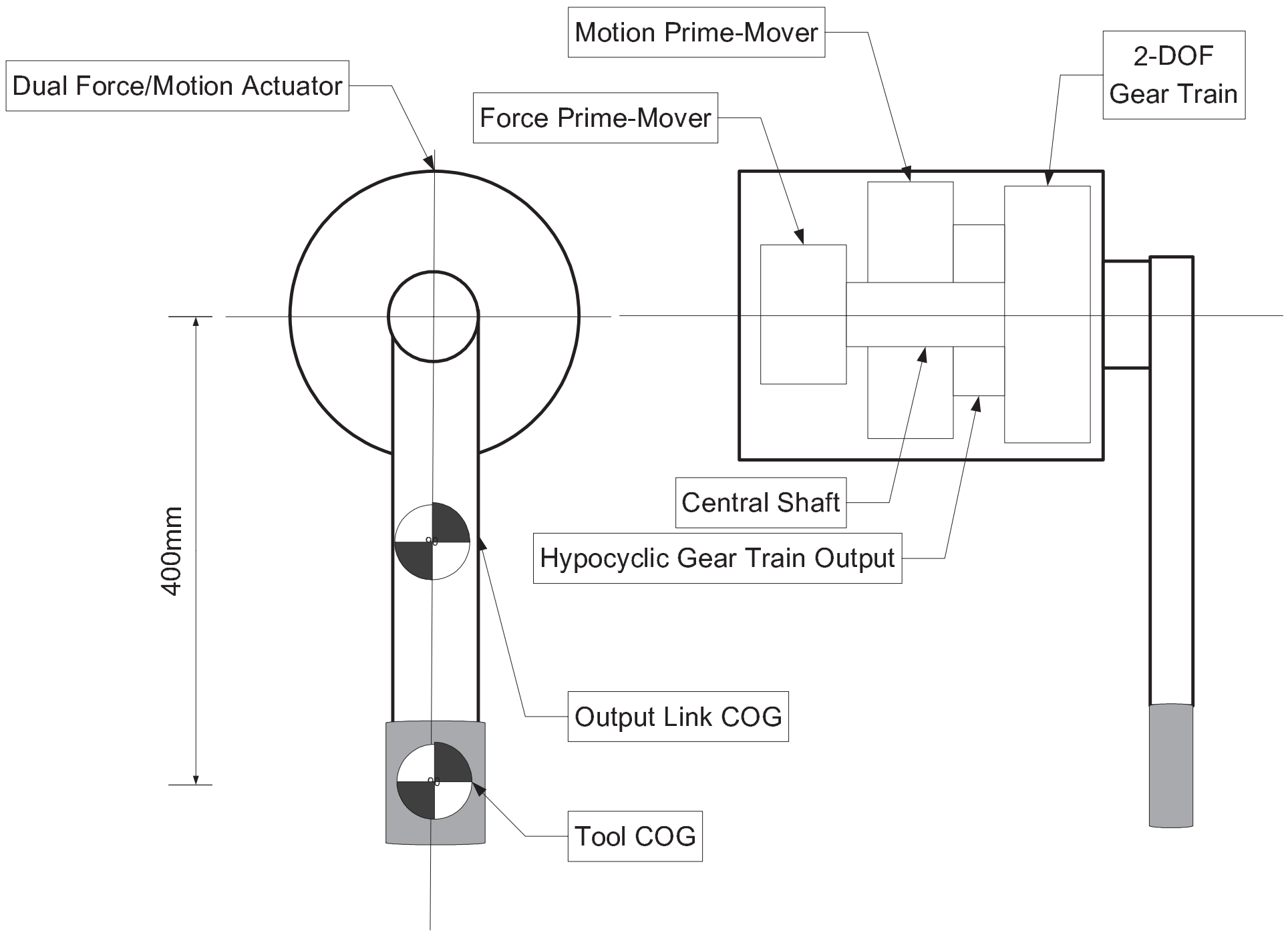

**Figure 6-1. Force/Motion Actuator Concept**

The Stribeck model was used to simulate actuator transmission friction. The decision making technique used was Computed Torque Control (CTC).

**Table 6-3. Force Control Experiment Results on Compliant Surface[9]**

| Application | Parameters | Pure Force Control | Compliant Control |
|---|---|---|---|
| *Force Regulation* | *Best Settling Time (Seconds)* | **1.3** | 6 |
| | *Impulse (N-s)* | **0.375** | ~10 |
| | *Approach Velocity (mm/s)* | 2.25 | 2.5 |
| | ***(Kp,Kv,Ki)** OR **K*** | (0.1,0.1,0.01) | 0.03 |
| | *Control Bandwidth (Hz)* | 15 | 15 |
| *Force Tracking* | *Time Lag (% Time Period)* | **4%** | **4%** |
| | *Impulse (N-s)* | **2.0** | 2.975 |
| | *Approach Velocity (mm/s)* | 1.25 | 1.25 |
| | ***(Kp,Kv,Ki)** OR **K*** | (0.1,0.1,0.01) | 0.003 |
| | *Control Bandwidth (Hz)* | 25 | 25 |

[9] The best values for each metric are in bold

In Chapter 4, some elementary force control experiments on a modular robot system were reported. The virtual spring and virtual inertia-damper demonstrations confirmed the spatial force transformation from the local sensor frame to the robot world frame. Spin-off application domains of these demonstrations (Human Augmentation, Virtual Fixtures, and Teleoperation) were discussed. In this section, implementation of a plane virtual fixture was presented. Subsequently, implementations of pure force control and compliant control algorithms in a 1-DOF force control task on a compliant environment were reported. These algorithms were implemented on a modular manipulator testbed for both force regulation and force tracking cases. The results of these experiments on a compliant surface are summarized in Table 6-3.

The modular robot testbed used for the above experimental activity was described in Chapter 3. This includes specifications/descriptions of the manipulator modules, force/torque sensor, and test contact surfaces. Chapter 3 also documents experiments conducted to determine the inertia parameters and current-torque mapping of the manipulator modules to facilitate dynamic control. The results of these experiments and specifications of the manipulator are listed in the appendix.

## 6.3 Questions Raised

This section summarizes some major research issues encountered during the work presented in this report. These issues have been tabulated (in Table 6-4) as a series of questions that serve as thought starters and thus pave the way for future research in this area. They have also been ranked in order of relevance (on a scale of 1, lowest priority, to 5, highest priority) to the main research focus, viz. Multi-Domain Inputs (MDI) at the actuator level to support Multi-Domain Control (MDC) at the system level.

**Table 6-4. Summary of Research Issues**

| No | Issue | Relative Priority | Domain |
|---|---|---|---|
| 1 | What different combinations of force and motion generators are possible in MDI actuators? | 3 | *Design* |
| 2 | How are the constituent subsystems of MDI actuators coupled in terms of kinematic influence coefficients and dynamic characteristics? | 4 | *Modeling* |
| 3 | How do we prioritize inputs of the subsystems based on output requirements? What sets of force, motion, and mixed force/motion criteria can be used for mixing performance levels of constituent subsystems? | 5 | *Operation* |
| 4 | Under what operational conditions are constituent subsystems essentially independent? | 4 | *Operation* |
| 5 | How do we determine reference wrench profiles for a general 6-DOF contact task? Can task performance envelopes be developed for these? | 2 | *Modeling* |
| 6 | How does the Performance Envelope (PE) expansion of the MDI actuator affect the PE of the task? Given a set of MDI actuators, what is the analytical transformation that relates input parameters to output force and motion parameters for system operation? | 3 | *Modeling* |
| 7 | How do we manage real-time decision-making and resource allocation for MDI actuators? | 2 | *Operation* |

## 6.4 Recommendations for Future Work

This section lays out recommendations for future work in the area of multi-domain actuation, which is the generic concept of which the force/motion actuator is an example. The approach and objective at UT Austin has been to maximize the number of choices within the actuator to enhance its intelligence, and in turn improve the performance of the overall system. It means, primarily, that we can control 12 separate task parameters at the end-effector of the robot dramatically expanding the functional capacity of all robot systems. Multi-Domain Input (MDI) is embedded at the very core of this premise. Before a detailed study of multi-domain actuation systems is undertaken, some fundamental questions have to be raised and issues relating to the different domains of a mechanical system have to be addressed. The following sections attempt to present these issues and outline a course for future work. The following is the structure we will follow in this section.

1. Mechanical Systems as Energy Transformers
2. How Do Force and Motion Transform?
3. Intelligent actuators that facilitate this transformation
4. Real-time control of these actuators and the system as a whole
5. How do we assemble components into systems?
6. Milestones and 3-year research plan

### 6.4.1 Mechanical Systems as Energy Transformers

A high-valued function, such as deburring or drilling, involves energy transfer. The main intention of a mechanical system, like a linkage or manipulator for example, is conversion of energy from an input source to an output to perform useful work while maintaining precision between the tool and the workpiece. A mechanical system, like any other, is not conservative. In other words, the energy transfer is not 100% efficient and there is some energy lost due to various non-conservative systems of forces acting on the system during the task execution. The two distinct physical quantities that define mechanical power product are force (sometimes called *effort*) and velocity (sometimes called *flow*), i.e, power = (force)x(velocity). Hence most tasks (or functions) required of mechanical systems may be considered as interactions between force and motion domains. Force and motion variables share a dual relationship and define the state of a system.

For an interaction task, in a strictly kinetostatic sense, under ideal conditions of contact, the end-effector differential motion and reactive forces are "reciprocal" to each other [Lipkin and Duffy, 1988], i.e, the instantaneous work vanishes at the point of contact. However, in reality, interaction forces could be of several types. Static (or reactive) force is first order and transforms to the output using g-functions. Inertia force is second order and relates to the output using h-functions. Losses due to frictional force, if modeled as fluid friction, are dependent on velocity. In the non-ideal world where there is no perfectly rigid mechanical component, every force causes a corresponding motion (or more restrictively, *differential motion*). In the case of static structural force, this differential motion is the deflection of the material on which it acts, and such a force is called a *restoring force* because it stores deflection energy (a form of potential energy) that can be reused. Impulse forces, caused by hammering, are characterized by high frequency content and incite vibrations and the system dissipates this energy due to various hysteresis losses.

### 6.4.2 How Do Force and Motion Transform?

From the power product, note that force and motion are dual variables. Consider $n$ sources of forces ($f_i$) and corresponding differential motions ($v_i$) contributing to a common output represented by the ordered pair $(f_o, v_o)$. The power output may be represented as in Eq.(6-1) where $\eta_i$ is the efficiency of energy transformation associated with the $i^{th}$ source.

$$\sum_{i=1}^{i=n} \eta_i f_i v_i = f_o v_o \tag{6-1}$$

For discussion purposes let us assume *n=2* and also that $\eta_i = 1 \forall i$. If we combine these input sources in series (Figure 6-2), output displacement is the series sum of displacements of the two input stages as are the power and velocity as shown in Eq.(6-2) and Eq.(6-3) respectively.

$$f_1 v_1 + f_2 v_2 = f_o v_o \tag{6-2}$$

$$v_1 + v_2 = v_o \tag{6-3}$$

From these relations follows the fact that the forces associated with each of the subsystems is equal to that in the output as shown in Eq.(6-4)

$$f_1 = f_2 = f_o \tag{6-4}$$

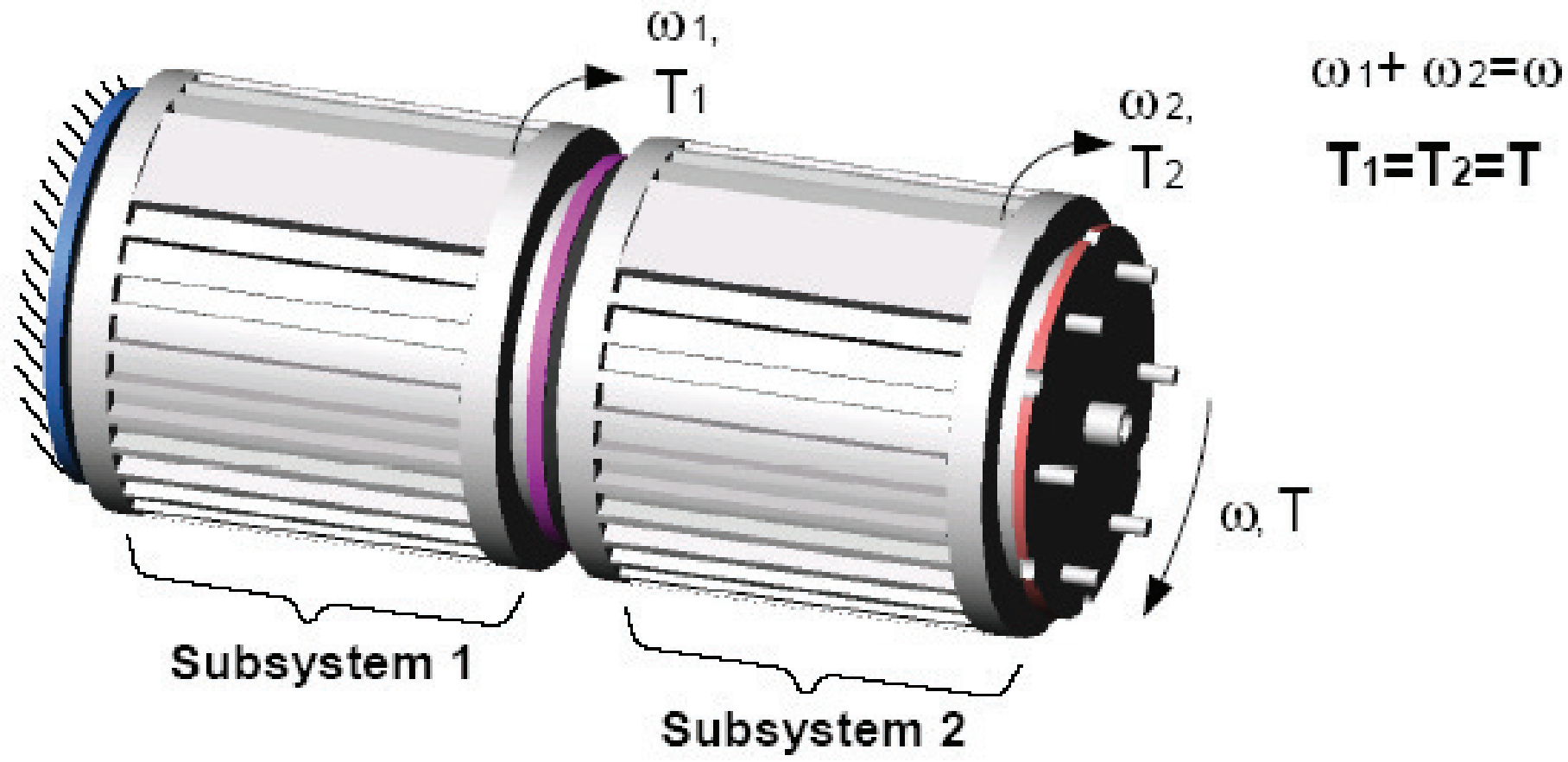

**Figure 6-2. Concept of Velocity Summing Fault Tolerant Rotary Actuator**

Now consider a torque-summing actuator by combining the input sources in parallel (See Figure 6-3). As the input forces have independent paths to the output, they add up to contribute to the output force as in Eq.(6-5).

$$f_1 + f_2 = f_o \qquad \textbf{(6-5)}$$

Nevertheless, power transformation is as given in Eq.(6-2). This information may be used to arrive at Eq.(6-6). From this follows the fact that each of the subsystem velocities is equal to the output velocity as in Eq.(6-7), considering that the input forces are non-zero.

$$f_1(v_1 - v_o) + f_2(v_2 - v_o) = 0 \qquad \textbf{(6-6)}$$

$$v_1 = v_2 = v_o \qquad \textbf{(6-7)}$$

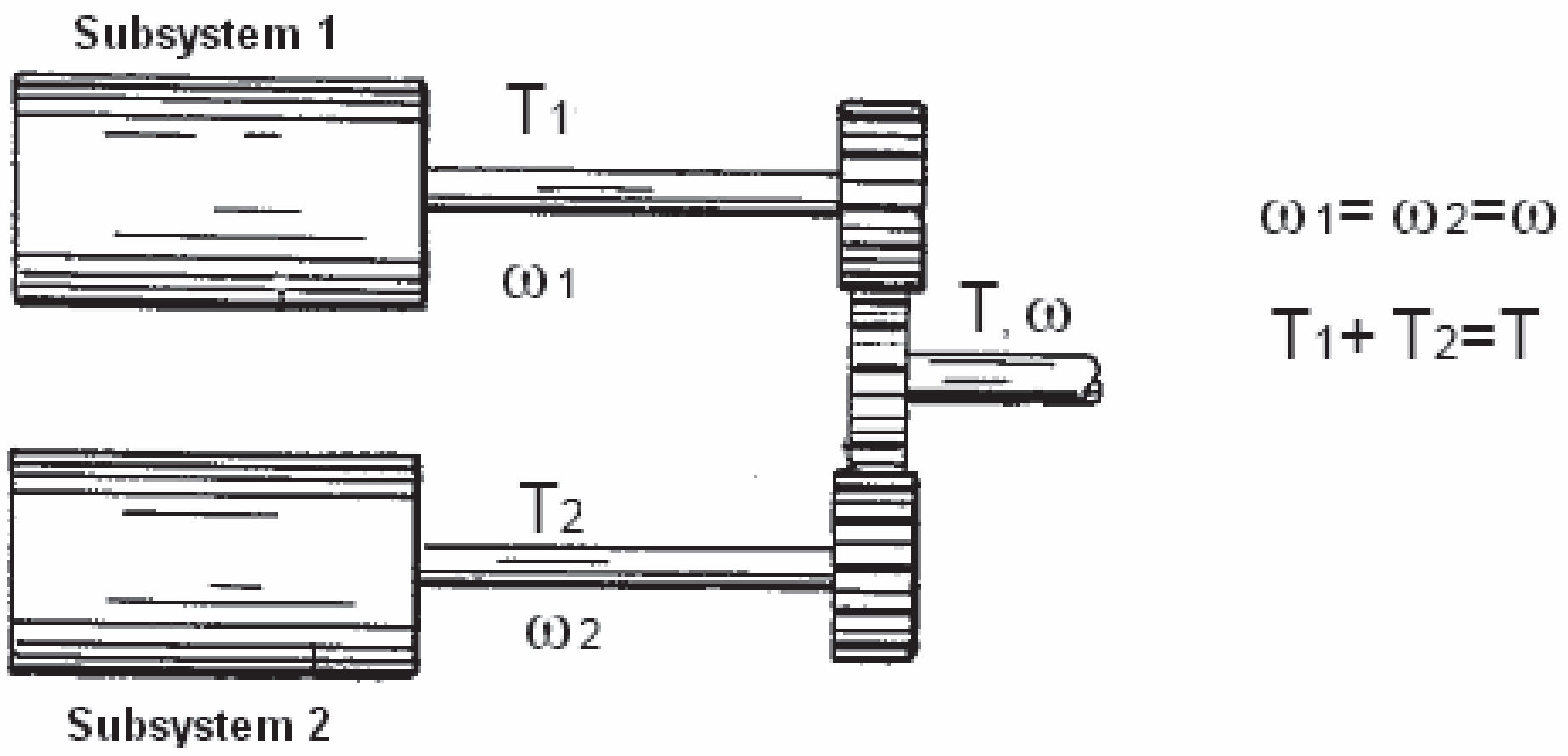

**Figure 6-3. Concept of Force Summing Fault Tolerant Rotary Actuator**

Also it can be shown that the transformation from input to output (in terms of g-functions) for static forces and differential motions are inverses of each other. G-functions or Kinematic Influence Coefficients (KICs) are geometry-based functions that relate input and output of the system. Relatively constant g-functions imply linearity as in the case of a small motion micromanipulator (See Figure 6-4) where the fine-motion stage has a small movement and thus approximately constant g-function over this small motion range.

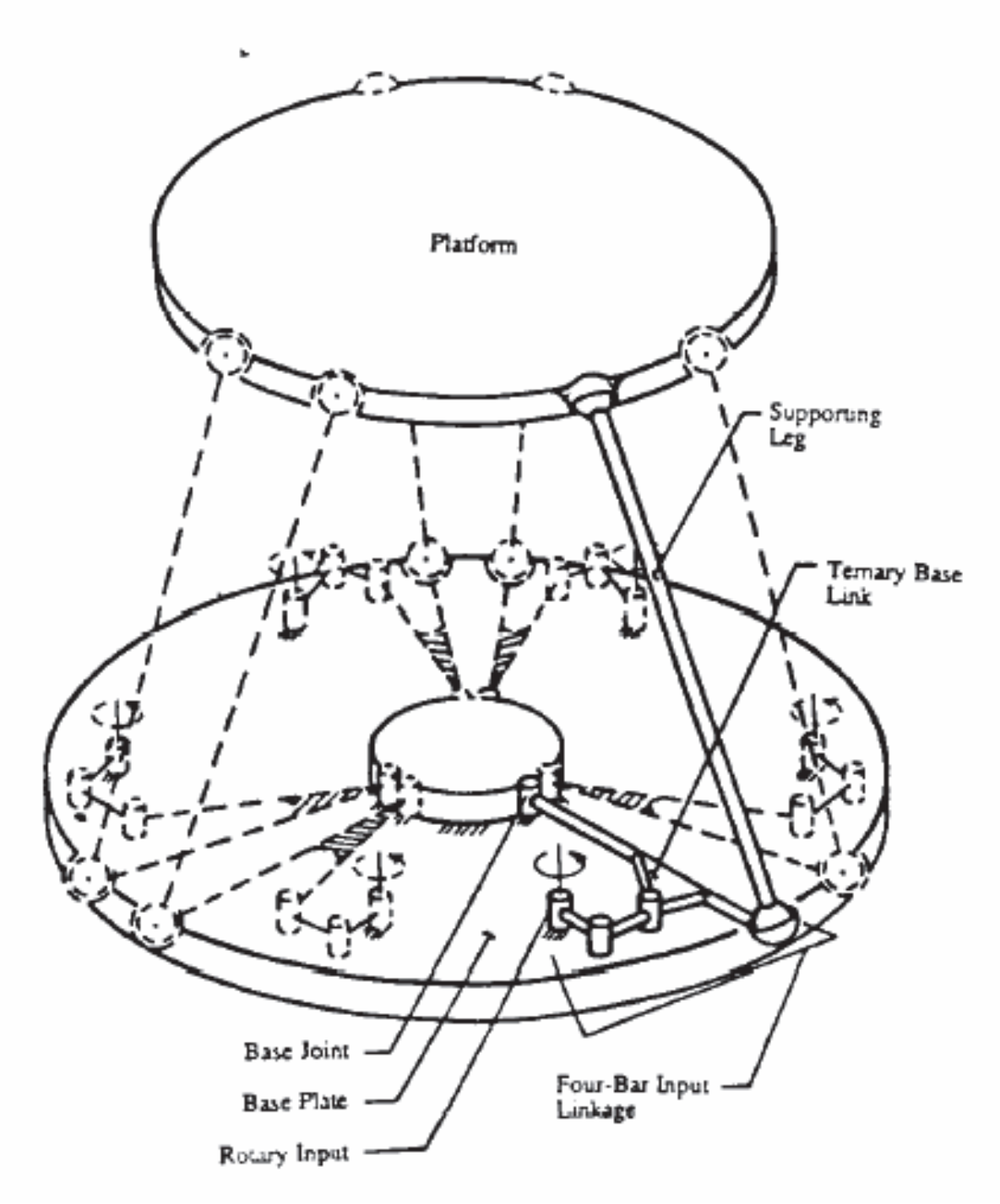

**Figure 6-4. Parallel Geometry Micromanipulator [Hudgens and Tesar, 1991]**

### 6.4.3 Multi-Input Intelligent Actuators

Having shown how force and motion are transformed between input and output in the previous section, we now present classes of intelligent actuators that facilitate various combinations of force and motion domains. Considering dual-inputs, there are three possible factors that govern the characteristic of the combination:

- Constituent domains, viz. force/force, motion/motion, and force/motion

- Type of arrangement, viz. series and parallel
- Relative scale of change between the constituent subsystems

All classes of Multi-Domain Input (MDI) actuators can be constructed using various combinations of the above listed factors. Note that all such combinations are not necessarily successful. In other words, we need not get good physical embodiments in the actuator for all cases. For example, summing torque and motion in series does not give any significant advantage over existing simpler designs (series velocity summing or parallel force summing). In the above discussion, by force and motion inputs we mean ideal force and motion generators. In Chapter 5 we presented the dynamic model for the force/motion actuator. In this the mass, damping, and stiffness matrices were all diagonal. This is based on the idealistic premise that there is no coupling between the force and motion stages. However there could be cross-coupling terms. Also note that the Force/Motion Actuator (FMA) is a special case of the more general concept of Multi-Domain Inputs. However we will gain significant insight into MDI by investigating the FMA.

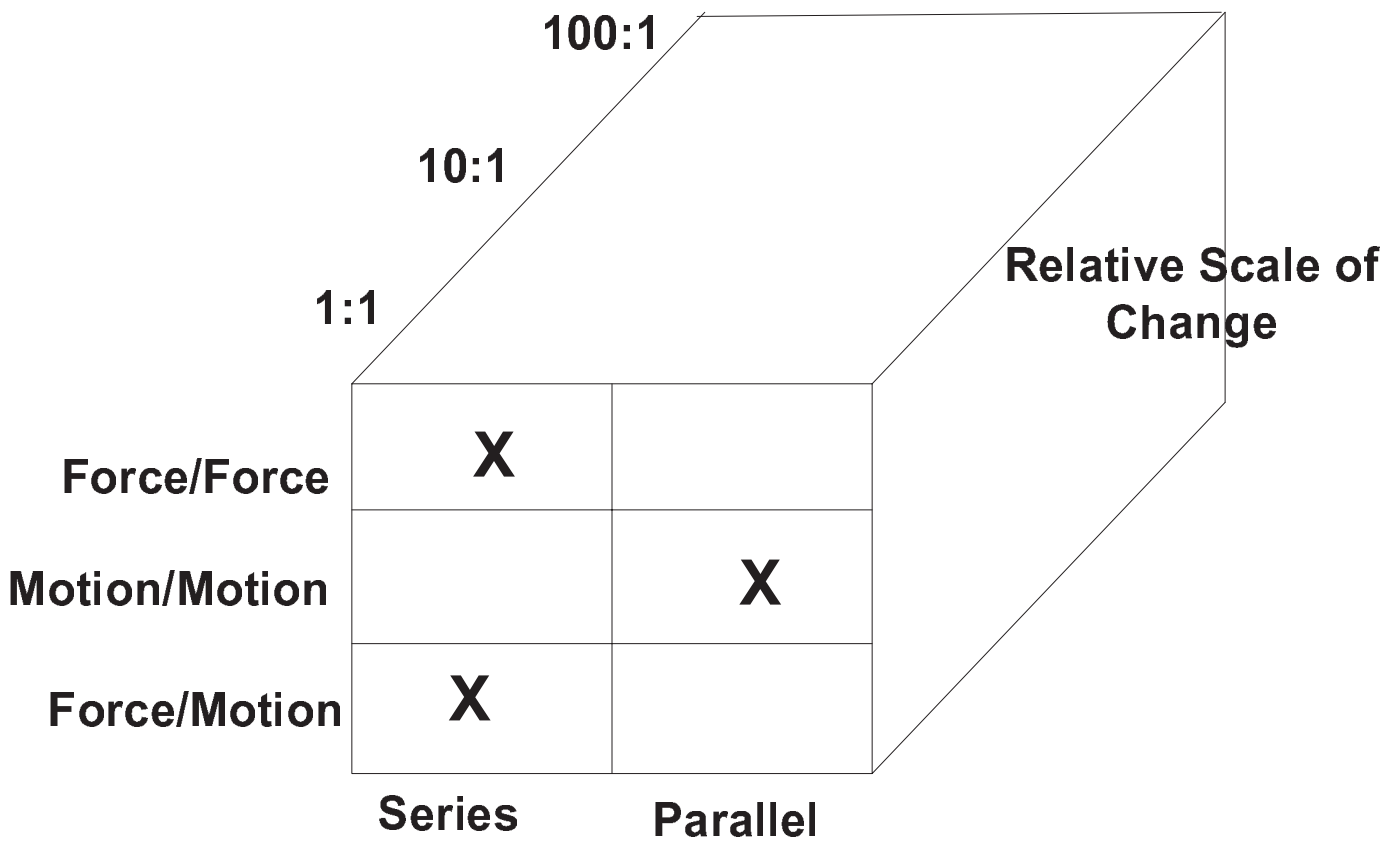

**Figure 6-5. Classes of Multi-Domain Input (MDI) Actuators**

**Table 6-5. Combinations of Force and Motion Domains**

| Actuator Type | Actuator Sub-Systems *(Notation)* | Combination | Jacobian Transform | Output Domains | Reference |
|---|---|---|---|---|---|
| *Dual Fault-Tolerant* | Force/Force *(FFA)* | Parallel | $\left[G_{\phi}^{u}\right]$ | Force/Force | [Tesar, 2003a] |
| | Motion/Motion *(VVA)* | Serial | $\left[G_{\phi}^{u}\right]$ | Motion/Motion | [Tesar, 2003a] |
| *Mixed Scales (Layered Control)* | Force/Force *(FfA)* | Serial | $\left[G_{\phi}^{u}\right]$ | Force/Force | [Tesar, 1999b] |
| | Motion/Motion *(VvA)* | Serial | $\left[G_{\phi}^{u}\right]$ | Motion/Motion | [Tesar, 1985] [Tesar, 1999b] |
| *Mixed Domains (Force/Motion Control)* | Force/Motion *(FVA)* | Parallel | $\left[G_{\phi}^{u}\right]$ | Force/Motion | [Tesar, 2003b] |

To understand the significance of these actuation systems within the context of the future of robotics technologies, please refer to Tesar's [1989] thirty-year forecast. See Figure 6-6 for a listing of functional regimes of MDI actuators.

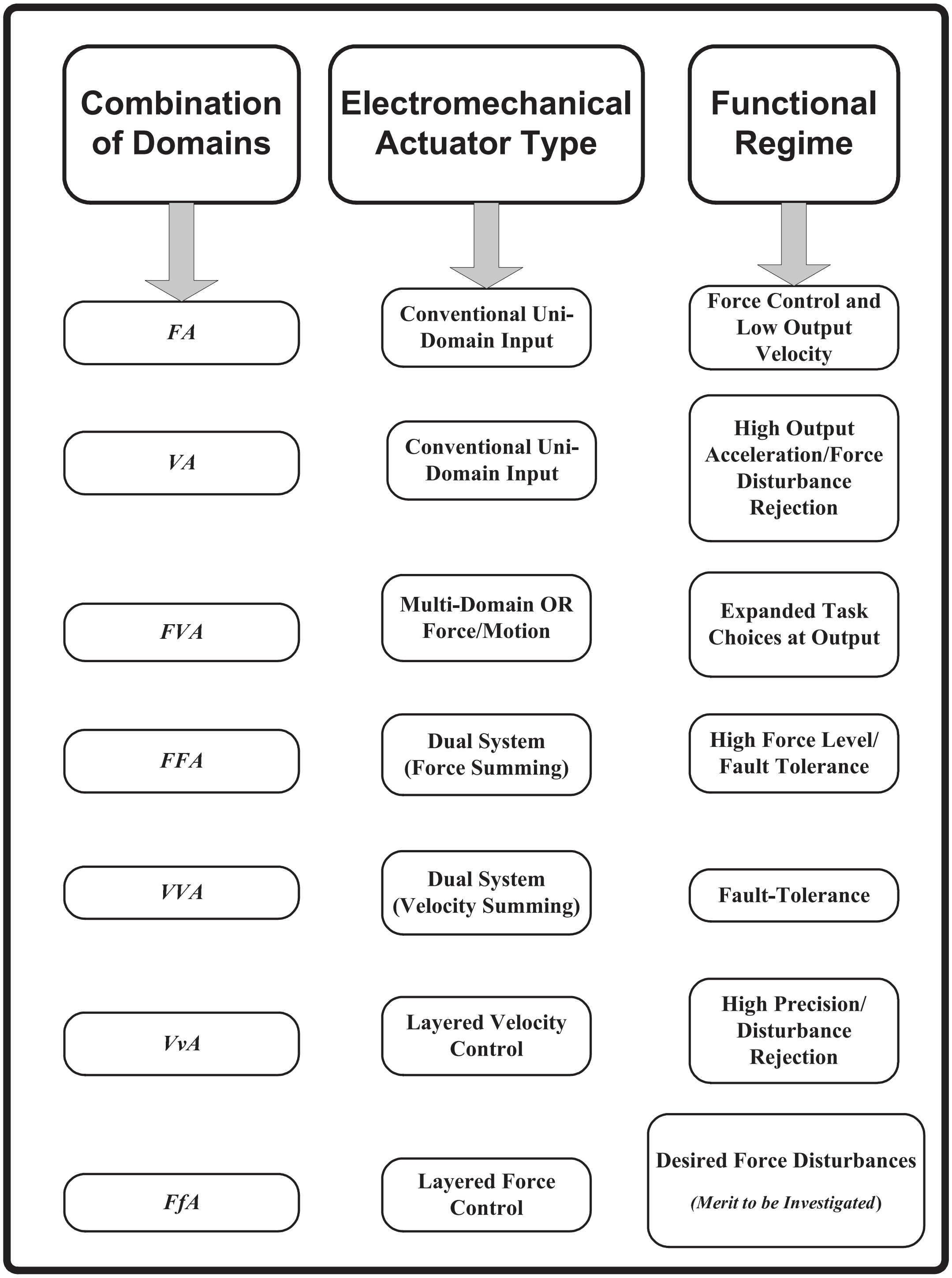

**Figure 6-6. Functional Regimes of Multi-Domain Actuators**

### 6.4.4 System Operation and Control

If successful embodiments of various MDI actuators are achieved, then the next issue that we are faced with is the real-time management of input commands. Consider a manipulator with generalized topology. The requirements of the task at hand have to be translated into force and/or motion requirements in the input space. We can think of three configuration spaces for such a system, namely Actuator Space (AS), defined by prime-mover torques/velocities, Joint Space (JS), defined by joint torques/velocities, and Operational Space (OS), characterized by output torques/velocities (See Figure 6-7). The transformation from AS to JS is through the structure matrix of the transmission ([A]) which contains the gear-ratios in the case of a geared transmission. This is frequently a constant matrix. Bearing compliance, rotor inertias, and lost motion are important performance metrics in AS. The transformation from JS to OS is established through the first order kinematic influence coefficients or g-function matrix ([G]) which is called the *Jacobian* in robotics literature. The operational goal of the decision and control system is to achieve task requirements in OS by commanding appropriate inputs in AS, at the same time allocating resources efficiently if the system is redundant.

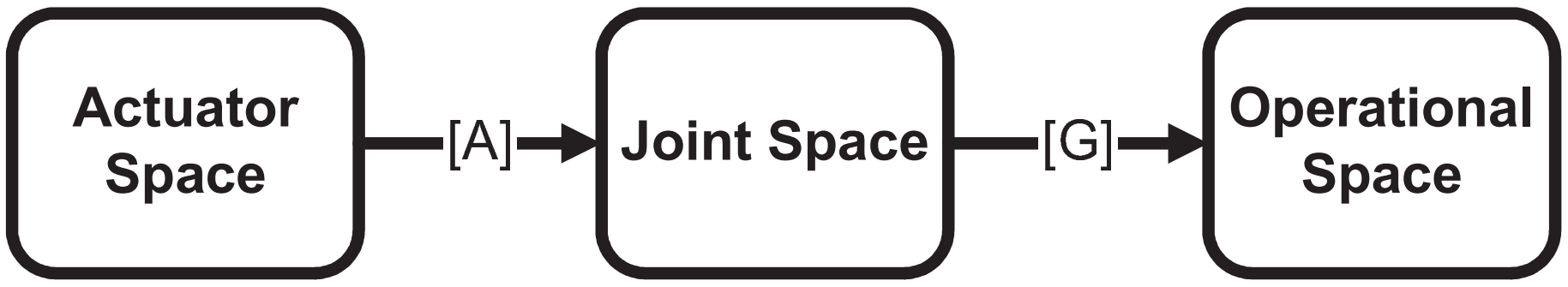

**Figure 6-7. Configuration Spaces of a Manipulator**

#### 6.4.4.1 Criteria-Based Process Management System

Most control is a mixture of force and motion. The task requirements for a contact task are usually expressed as a desired motion plan and a target force profile. Together these requirements may be termed as a *process plan*. In a more general sense we could think of process performance envelopes which describe the complex process plan in terms of maps. This proposition is being studied by Chang and Tesar [2004]. A conceptual example of such a process performance map is shown in Figure 6-8. It shows the relationship between the lateral/horizontal force (in drilling), drill feed-rate, and the lateral deflection of the drill bit. Note that this graphic shows a conceptual performance map and that the numerical values are not necessarily realistic.

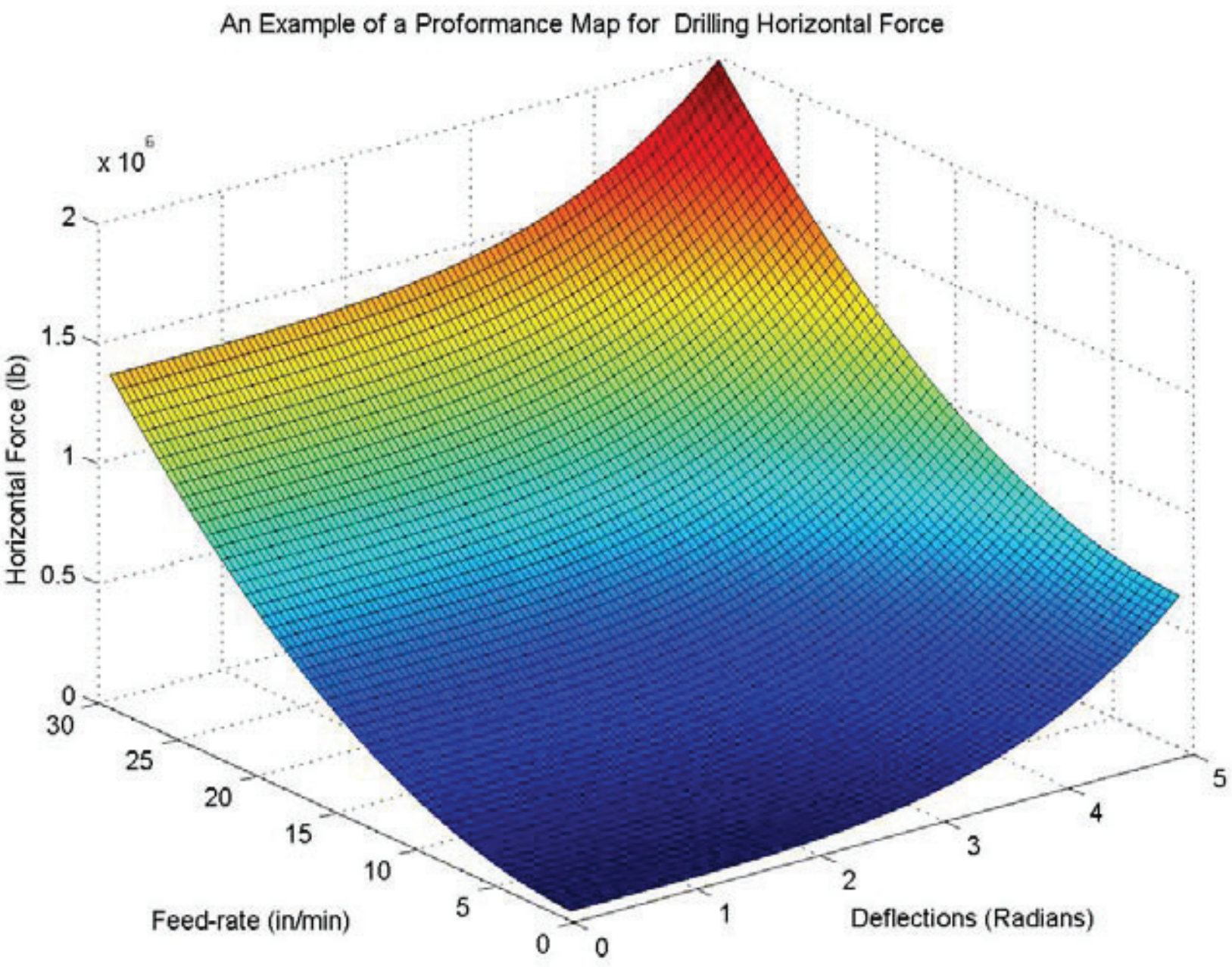

**Figure 6-8. Process Performance Map for Horizontal Force in Drilling [Chang and Tesar, 2004]**

Based on the task requirements in OS at any instant, the resource allocation to component subsystems in AS for MDI actuators should done through a criteria based decision making scheme.

### 6.4.5 Components to Systems

This section introduces the operational issues at the system level while using MDI actuators to assemble systems on demand.

#### 6.4.5.1 Multi-Domain Control in the Context of Process Control

Intelligent Automation can be defined as the science that would lead to the fully integrated manufacturing cell made of 30 to 50 standard modules assembled on demand. "Intelligent Automation" concentrates on market driven production systems operated from a database with the maximum integration of all modern technologies within a full architecture (mechanical, electronic and software) based on an enhanced science of design [Tesar, 1997].

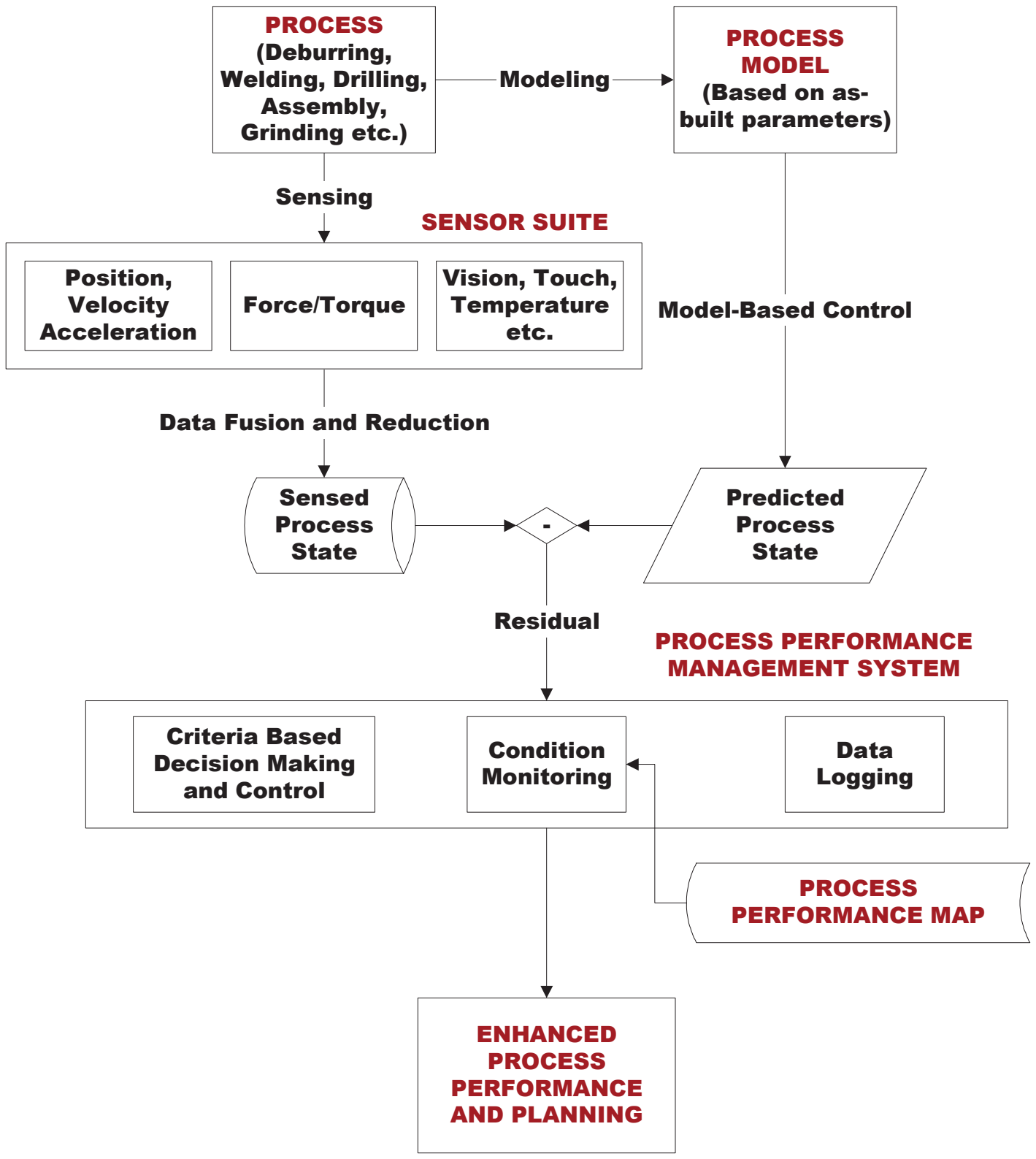

**Figure 6-9. Schematic of Process Control**

*Sensor Based Manufacturing Process Control* is a principal component of intelligent automation. In this approach the residual between a parametrically accurate model reference and a sensor reference is used as a basis for decision making to enhance process performance and planning (See Figure 6-9). The term *process* should be interpreted as the embodiment of all progressive events that lead to the execution of a task with acceptable performance accompanied by optimal demands on the system resources (or control inputs). This interpretation illustrates that this paradigm is more appropriate for a system with redundant resources (such as a redundant serial chain manipulator). Redundancy can include multiple ways of completing the task and also redundant resources within the actuator.

## 6.5 Contribution to RRG's Vision

This section shows the association between other research threads within RRG and Multi-Domain Inputs (MDI).

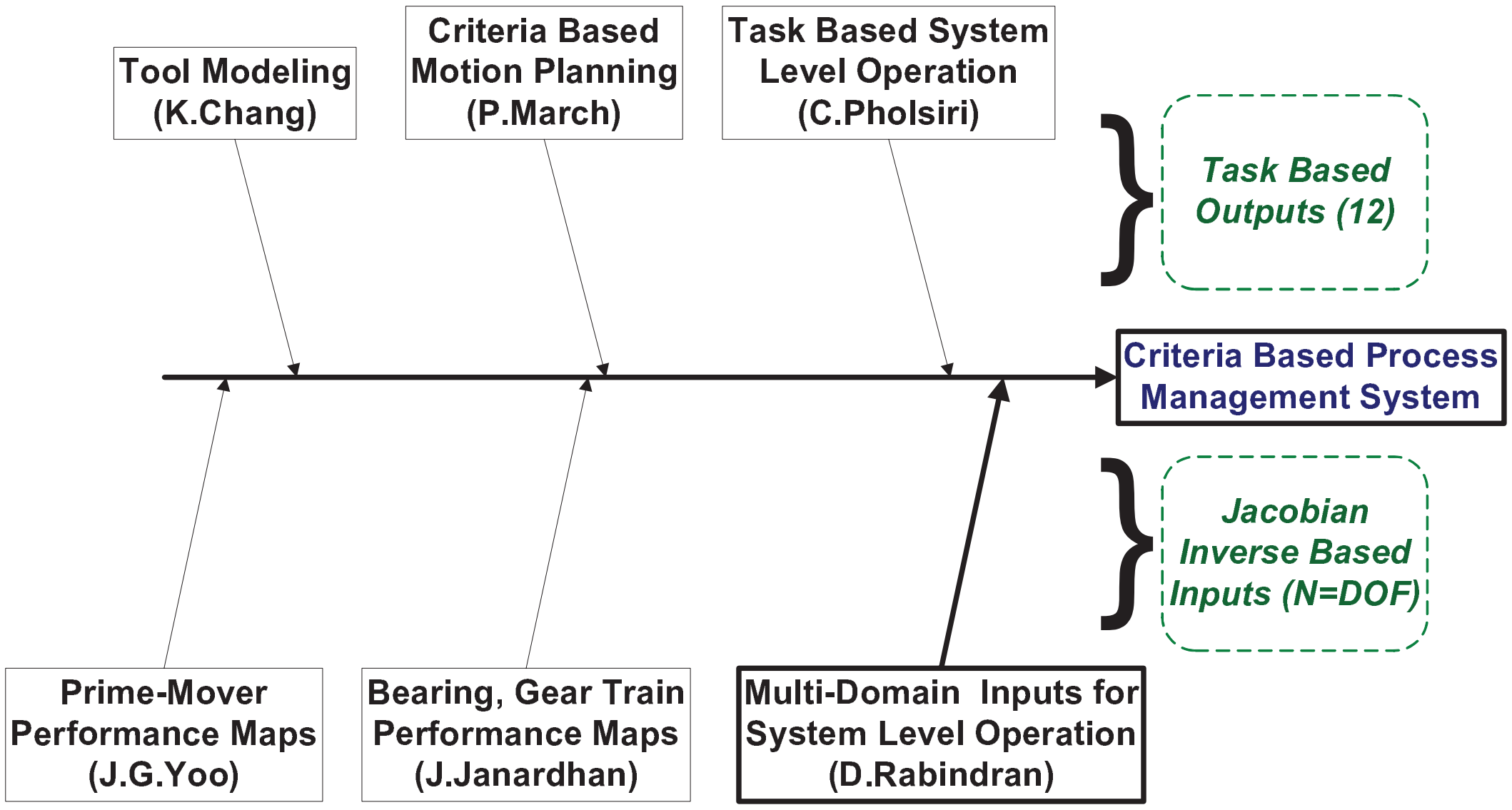

**Figure 6-10. RRG Research Threads for Process Control**

Figure 6-10 shows the various research topics and associated researchers.

Given a contact task, the contribution of various research threads to successful completion of the task would be as follows:

- *Tool modeling* and task performance envelopes for some representative classes of tools and processes [Chang and Tesar, 2004] will provide the necessary information for determining system demands (Tool norms, ellipsoids, tool performance maps etc.). These are but force/motion priorities in the operational space.
- *Motion planning* based on curve-based criteria [March and Tesar, 2004] will help in determining the motion planning component of the process plan. This research component can be used to determine a smooth geometric path that circumvents shocks.
- *Task based system operation and control* [Pholsiri and Tesar, 2004] provides new approaches for configuration management at the system level for constrained and unconstrained physical tasks.
- *Performance maps for prime-movers* [Yoo and Tesar, 2002] will provide an understanding of the motor parameters. Yoo and Tesar are currently developing motor performance envelopes based on experimental data from a non-linear actuator testbed.
- *Performance maps for gear trains* [Janardhan and Tesar, In Process] increases our understanding of transmission parameters. Bearing friction and compliance will be a major part of this study.
- *Multi-domain inputs* for system operation and control will provide the framework for criteria based resource allocation at the actuator level based on system level demands for the general geometric system case. This entails formalizing the transformation of force/motion priorities from Operational Space to Actuator Space (See Figure 6-7).

## 6.6 Deliverables for Future Research

This section describes the future course of work in this research thread. The aim here is to embed multiple inputs within the same actuator to expand force and velocity choices at the output of the system. In other words, Multi-Domain Input (MDI) actuation systems will be used for Multi-Domain Control (MDC) (force and velocity domains) of high-value robotic processes that have a variety of performance parameters in the force and velocity domains (like high precision/low force, high force/low precision, low force/low precision, etc). The hypothesis to be tested may be stated as:

"*Multi-Domain Input (MDI) actuators with scaled or distinct sub-systems (layered control and force/motion) dramatically improve the choice on process parameters (force and velocity) at the system output, thus resulting in the expansion of the functional capacity (in terms of force and velocity) of open- and closed-chain robotic systems*"

The approach adopted for testing this hypothesis would be as follows:

- Prove performance enhancement of MDI at the actuator level by comparing with uni-domain inputs. This requires modeling, simulation, and experimentation
- Present MDI in the analytical framework [Thomas and Tesar, 1982] for system operation
- Conduct numerical simulations with representative robotic processes to test the above mentioned analytical framework.

Table 6-6 presents a detailed three-year program of work in this area.

**Table 6-6. Three Year Program of Future Work**

| Time | Tasks |
|---|---|
| **Year I** | 1. Comparison of three basic classes of Multi-Domain Input actuators (dual fault-tolerant, layered control, and force/motion) with each other and also with uni-domain force and velocity inputs<br>a. Consider basic behavior of component system (qualitative comparison)<br>b. Consider performance for precision and force regulation tasks (quantitative comparison) |
| **Year I** | 2. Simulation of Force/Motion Actuator (FMA) performance (ACTUATOR LEVEL)<br>a. Kinematic analysis of actuator ***(In Progress)***<br>b. Dynamic model of actuator ***(In Progress)***<br>c. Static Analysis (FEM) using output torques from above simulation |
| **Year I-II** | 3. Investigation of dynamic coupling between component subsystems in FMA (ACTUATOR LEVEL)<br>a. Study of criteria that can be used for resource allocation at the actuator level (force, velocity, and mixture criteria)<br>b. Model the coupling between the component sub-systems and identify when they are relatively independent. Consider various scales of change |
| **Year II** | 4. Experimental Evaluation of Force/Motion actuator with a desktop prototype (ACTUATOR LEVEL)<br>a. Build a desk top set-up with near-perfect velocity generator, near-perfect force generator, and a 2-DOF gear train, with a link attached to the output<br>b. Evaluate input performance parameters for a set of output reference signals – trapezoidal, sinusoidal, step input, etc<br>c. Compare the results with that from a perfect motion generator and perfect force generator. Quantify performance improvement through data analysis |
| **Year II** | 5. Extension of [Thomas and Tesar, 1982] to relate MDI actuator input parameters to task/process parameters – Analytical Work (SYSTEM LEVEL)<br>a. Deep understanding of force and motion domains at system level<br>b. Consider both *process specification parameters* and *process control parameters* (for precision and force tasks) |
| **Year III** | 6. Proof of Concept Numerical Simulation (SYSTEM LEVEL)<br>a. Consider 3-4 representative real-world robotic processes/high-value functions that span precision to force demands<br>b. Use the extended analytical formulation developed in the previous task<br>c. Use serial chain and parallel topologies (to generalize the formulation)<br>d. Develop process simulations |

# Appendix

## A. Parameters for PowerCube Modular Manipulator

The PowerCube modular manipulator used in this research is based on standard industrial manipulator geometry with a Roll-Pitch-Pitch-Roll-Pitch-Roll configuration. It weighs $78.8 \pm 2.0$ lbs, has a reach of approximately $3.5$ ft, and a payload rating of $8$ lbs at $3.0$ ft/sec.

The following D-H parameters are based on the frame assignment described in "Introduction to Robotics: Mechanics and Control" authored by John. J. Craig. Note that all $\theta_i$ are variables since all joints are rotary.

| i | $\alpha_{i-1}$ (Degrees) | $a_{i-1}$ (mm) | $d_i$ (mm) | $\theta_i$ | Offsets (Degrees) |
|---|---|---|---|---|---|
| 1 | 0 | 0 | 0 | $\theta_1$ | 0 |
| 2 | 90 | 0 | 0 | $\theta_2$ | -90 |
| 3 | 0 | 265.0 | 0 | $\theta_3$ | 90 |
| 4 | -90 | 0 | 412.5 | $\theta_4$ | 0 |
|  | 90 | 0 | 0 | $\theta_5$ | 0 |
| 6 | 90 | 0 | 0 | $\theta_6$ | 0 |

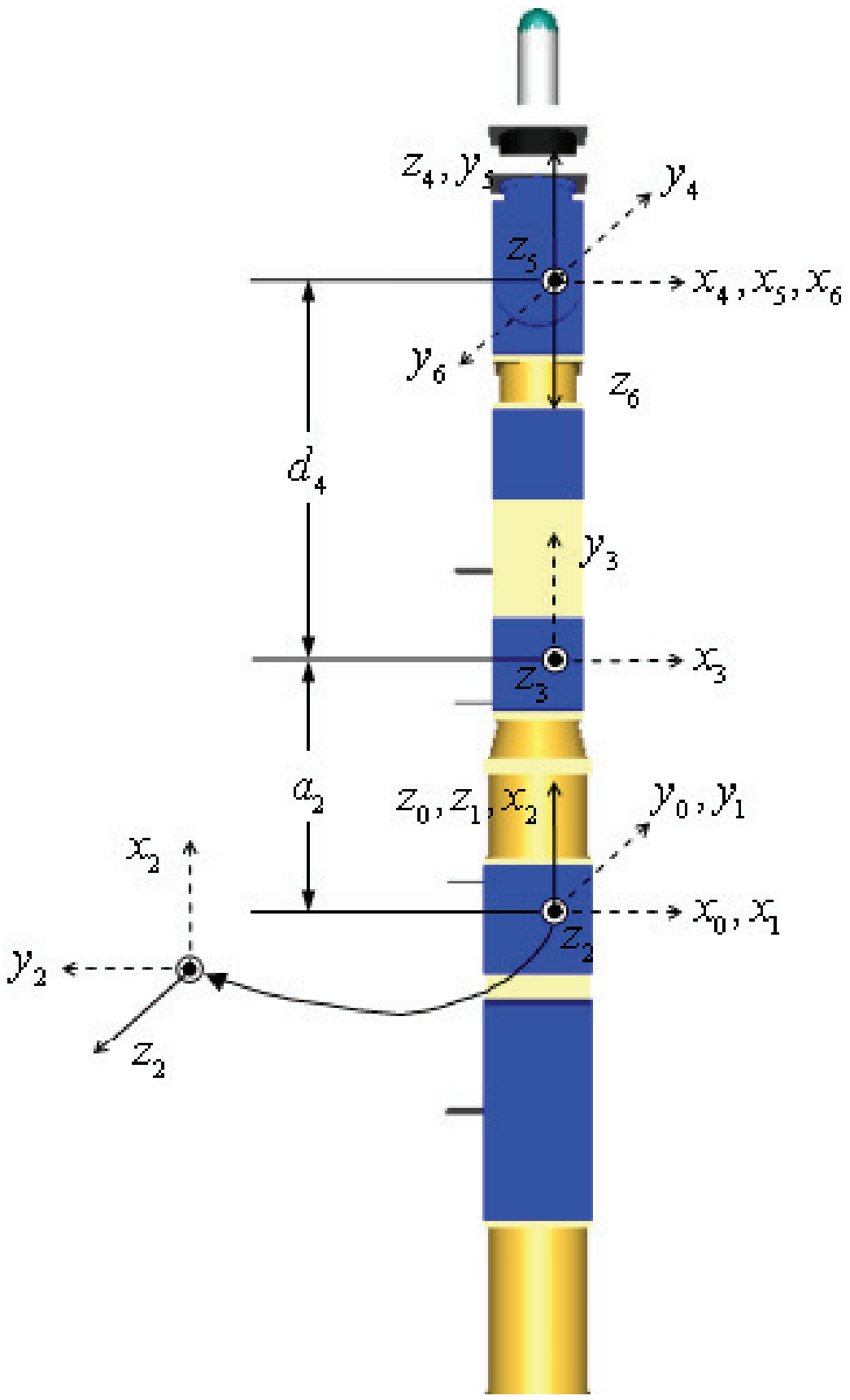

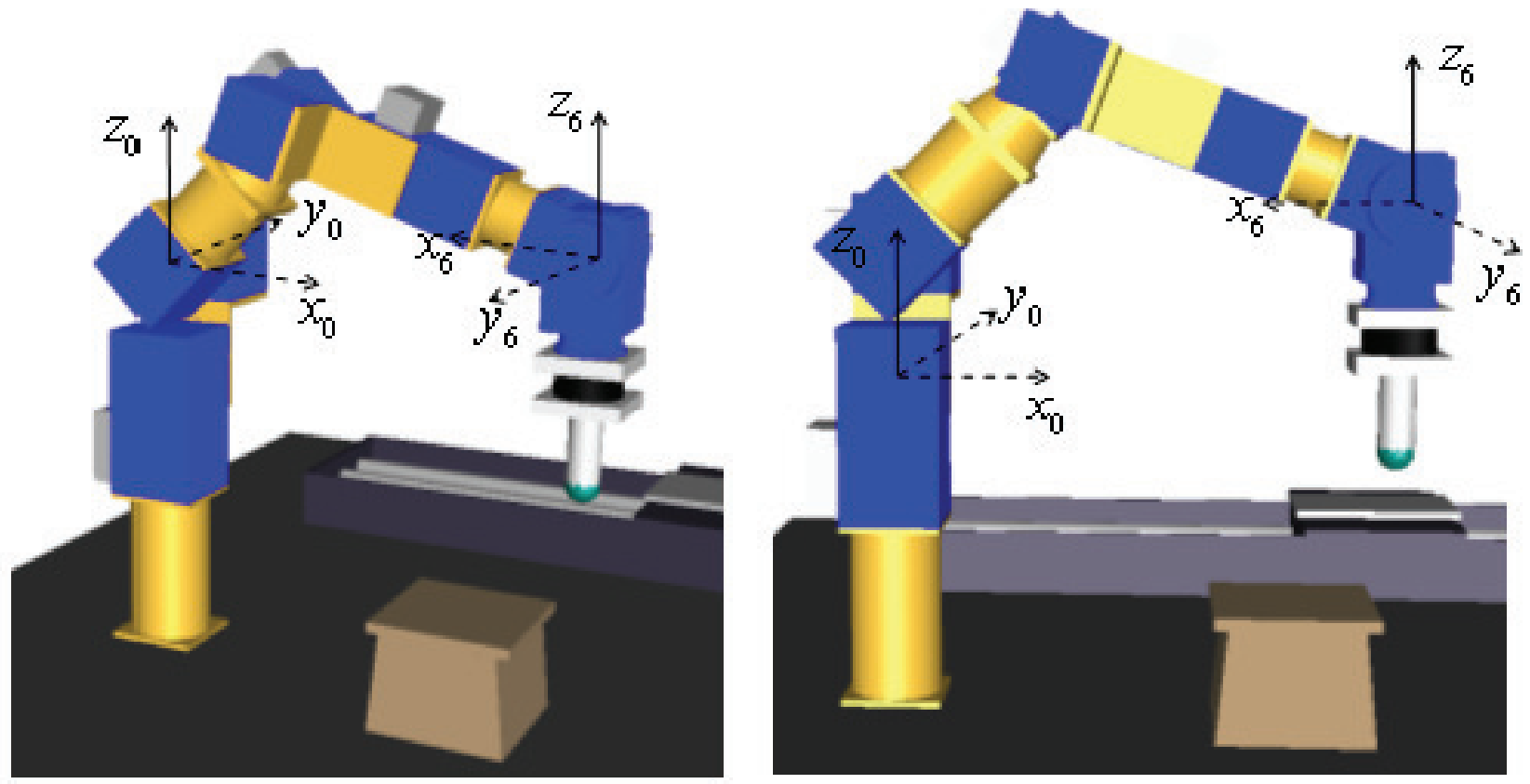

| | $z_0, z_1(x_0)$ *link twist* | $z_0, z_1(x_0)$ *link length* | $x_0, x_1(z_1)$ *link offset* | $x_0, x_1(z_1)$ *joint angle* |
|---|---|---|---|---|
| $i$ | $\alpha_{i-1}$ | $a_{i-1}$ | $d_i$ | $\theta_i$ |
| 1 | 0 | 0 | 0 | $\theta_1$ |
| 2 | $90^0$ | 0 | 0 | $\theta_2 - 90^0$ |
| 3 | 0 | $a_2$ | 0 | $\theta_3 + 90^0$ |
| 4 | $-90^0$ | 0 | $d_4$ | $\theta_4$ |
| 5 | $90^0$ | 0 | 0 | $\theta_5$ |
| 6 | $90^0$ | 0 | 0 | $\theta_6$ |

## B. Parameters of the Trifilar Pendulum

| | |
|---|---|
| Length of Suspension String (m) | 1.8288 |
| Mass of the Plates (kg) | 3.671 |
| Diameter of the Plates (m) | 0.012954 |
| Measured Moment of Inertia of Bottom Plate (kg-m$^2$) | 0.299744 |
| Theoretical Moment of Inertia of Bottom Plate (kg-m$^2$) | 0.268634 |

## C. Inertia Parameters of the PowerCube Modules

| Module | $J_X$ (kg-m$^2$) | $J_Y$ (kg-m$^2$) | $J_Z$ (kg-m$^2$) |
|---|---|---|---|
| 110x110x40 Link | 0.001058611 | 0.000661663 | 0.000661663 |
| 110x110x95 Link | 0.001559981 | 0.001641261 | 0.001641261 |
| 90x90x40 Link | 0.000498446 | 0.000343699 | 0.000343699 |
| 90x90x95 Link | 0.000856721 | 0.000993569 | 0.000993569 |
| PR 110 Actuator | 0.03080969 | 0.016473629 | 0.016473629 |
| PG 070 Gripper | 0.008453821 | 0.009485337 | 0.007628006 |
| PR 090 Actuator | 0.014914532 | 0.009269184 | 0.009269184 |
| PW 090 Wrist | 0.016535598 | 0.011403202 | 0.013613901 |
| 90x90 Angle Link *X* | 0.001119517 | $7.751 \times 10^{-8}$ | $4.06 \times 10^{-9}$ |
| *Y* | $7.751 \times 10^{-8}$ | 0.001092373 | 0.000311758 |
| *Z* | $4.06 \times 10^{-9}$ | 0.000311758 | 0.000702996 |
| 110x110 Angle Link *X* | 0.002526201 | 0.000000935 | 0.000003727 |
| *Y* | 0.000000935 | 0.002389596 | 0.000695662 |
| *Z* | 0.000003727 | 0.000695662 | 0.001711138 |

## D. Error Analysis of the Inertia Values

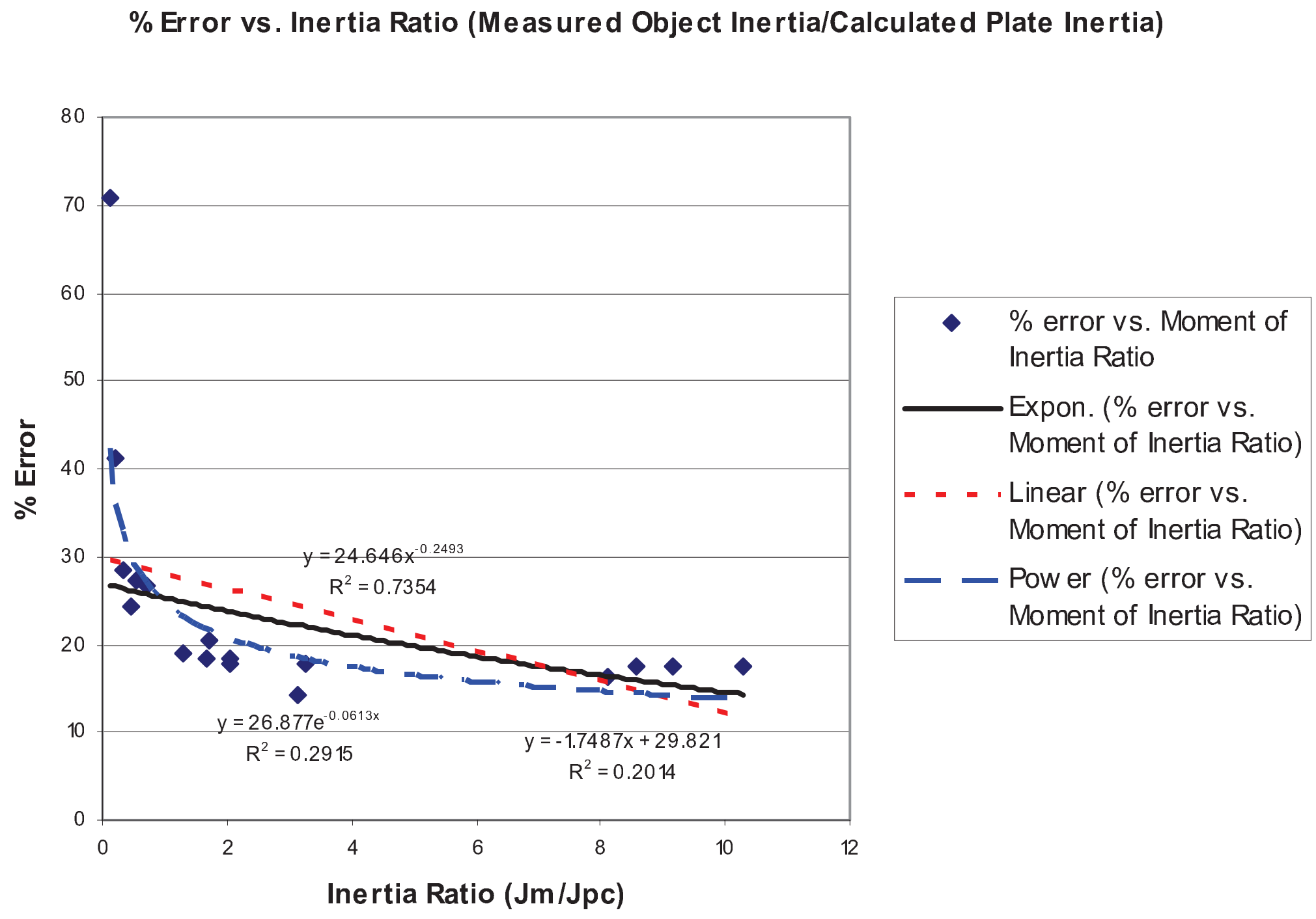

## E. Frame Transformation for the Force/Torque Sensor

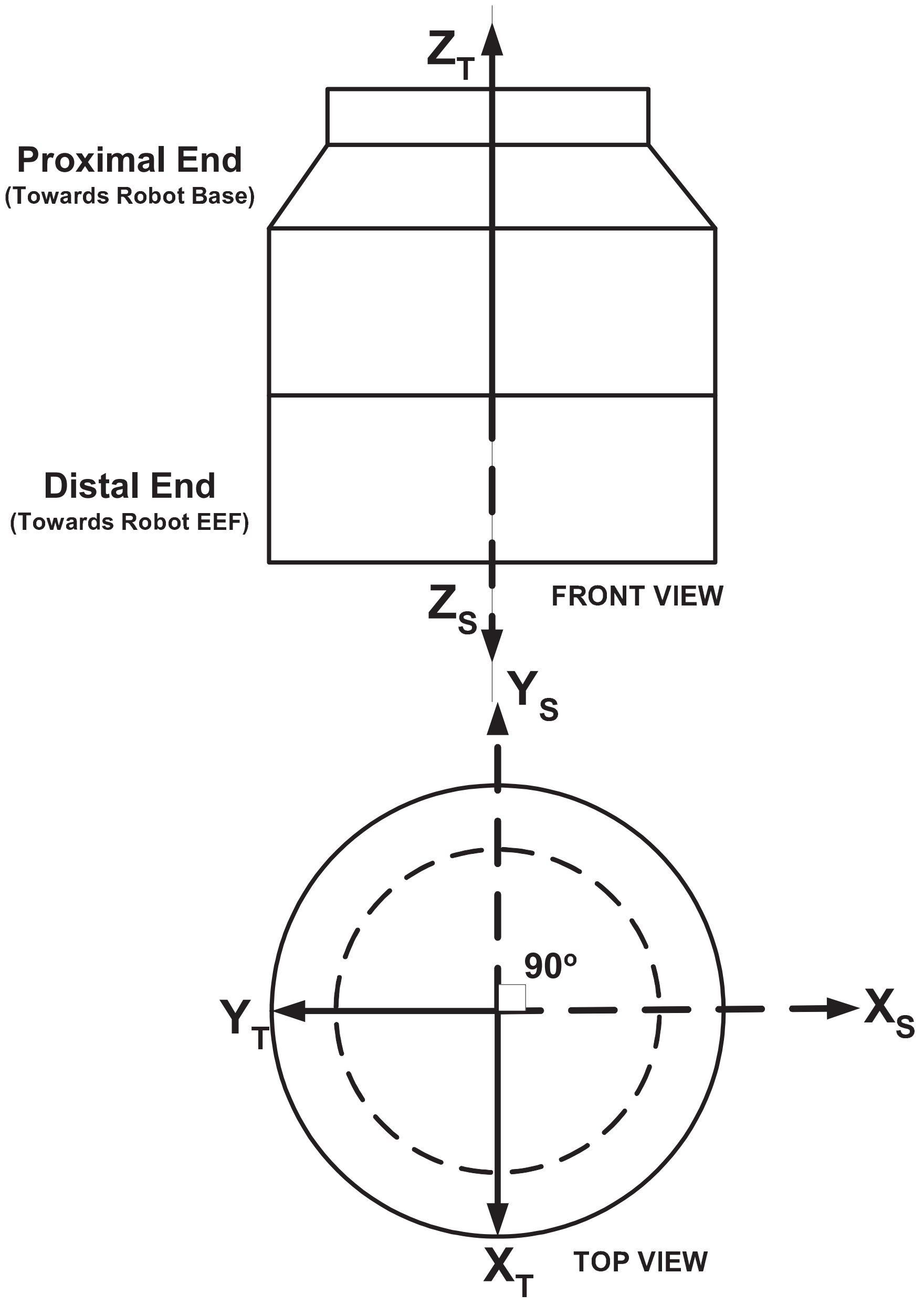

$\Pi_S \equiv X_S Y_S Z_S$ is the local sensor reference frame and $\Pi_T \equiv X_T Y_T Z_T$ is the EEF frame of reference. The relative orientation of $\Pi_S$ w.r.t $\Pi_T$ is shown in the figure above. The rotation matrix $\mathbf{R}_S^T$ is given by:

$$\mathbf{R}_S^T = \left[\mathbf{R}_{Z_T,\frac{\pi}{2}}\right]\left[\mathbf{R}_{X_T,\pi}\right] = \begin{bmatrix} 0 & 1 & 0 \\ -1 & 0 & 0 \\ 0 & 0 & 1 \end{bmatrix}\begin{bmatrix} 1 & 0 & 0 \\ 0 & -1 & 0 \\ 0 & 0 & -1 \end{bmatrix} = \begin{bmatrix} 0 & -1 & 0 \\ -1 & 0 & 0 \\ 0 & 0 & -1 \end{bmatrix}$$

Hence the spatial transformation matrix between $\Pi_S$ and $\Pi_T$ may be represented as:

$$\mathbf{X}_S^T = \begin{bmatrix} \left(\mathbf{R}_S^T\right)_{3\times3} & (\mathbf{0})_{3\times3} \\ (\mathbf{0})_{3\times3} & \left(\mathbf{R}_S^T\right)_{3\times3} \end{bmatrix} = \begin{bmatrix} 0 & -1 & 0 & 0 & 0 & 0 \\ -1 & 0 & 0 & 0 & 0 & 0 \\ 0 & 0 & -1 & 0 & 0 & 0 \\ 0 & 0 & 0 & 0 & -1 & 0 \\ 0 & 0 & 0 & -1 & 0 & 0 \\ 0 & 0 & 0 & 0 & 0 & -1 \end{bmatrix}$$